\documentclass[pdflatex,sn-mathphys]{sn-jnl}% Math and Physical Sciences Reference Style
\usepackage{subfigure}

\usepackage{color}
\usepackage{multirow, makecell}
\usepackage{threeparttable}
\usepackage{threeparttablex}
\usepackage{longtable}
\usepackage{caption}
\usepackage{hhline}

\usepackage{pdflscape}
\usepackage{afterpage}
\usepackage{rotating}

\usepackage{pdfpages}

\usepackage{censor} % blackout text
\newcommand\review[1]{\textcolor{black}{#1}}
\newcommand\reviewII[1]{\textcolor{black}{#1}}

\makeatletter
\newcommand{\textsupersubscript}[2]{%
  \begingroup
    \settowidth{\@tempdima}{\textsuperscript{#1}}%
    \settowidth{\@tempdimb}{\textsubscript{#2}}%
    \ifdim\@tempdima<\@tempdimb
      \setlength{\@tempdima}{\@tempdimb}%
    \fi
    \makebox[\@tempdima][l]{%
      \rlap{\textsuperscript{#1}}\textsubscript{#2}}%
  \endgroup}
\makeatother

\newlength\MAX  
\definecolor{majority_class}{RGB}{31, 119, 180}
\definecolor{minority_class}{RGB}{255, 127, 14}
\newcommand*\IRchart[1]{\rlap{\textcolor{minority_class}{\rule{\MAX}{2ex}}}\textcolor{majority_class}{\rule{#1\MAX}{2ex}}}
\jyear{2022}%

\theoremstyle{thmstyleone}%
\theoremstyle{thmstyletwo}%

\theoremstyle{thmstylethree}%

\usepackage{framed}
\newenvironment{shadequote}
{\begin{snugshade}\begin{quote}}
{\hfill\end{quote}\end{snugshade}}
\definecolor{shadecolor}{rgb}{0.9,0.9,0.9}

\begin{document}

\title[SMOClust]{SMOClust: Synthetic Minority Oversampling based on Stream Clustering for Evolving Data Streams}


\author*[1]{\fnm{Chun Wai} \sur{Chiu}}\email{c.chiu@keele.ac.uk}

\author*[2]{\fnm{Leandro L.} \sur{Minku}}\email{l.l.minku@bham.ac.uk}
%\equalcont{This author contributed equally to this work.}

\affil*[1]{\orgdiv{School of Computer Science and Mathematics}, \orgname{Keele University}, \orgaddress{\street{Keele}, \city{Staffordshire}, \postcode{ST5 5BG}, \country{United Kingdom}}}

\affil*[2]{\orgdiv{School of Computer Science}, \orgname{University of Birmingham}, \orgaddress{\street{Edgbaston}, \city{Birmingham}, \postcode{B15 2TT}, \country{United Kingdom}}}

% \affil[2]{\orgdiv{Department}, \orgname{Organization}, \orgaddress{\street{Street}, \city{City}, \postcode{10587}, \state{State}, \country{Country}}}


% \abstract{The abstract serves both as a general introduction to the topic and as a brief, non-technical summary of the main results and their implications. Authors are advised to check the author instructions for the journal they are submitting to for word limits and if structural elements like subheadings, citations, or equations are permitted.}
\abstract{Many real-world data stream applications not only suffer from concept drift but also class imbalance. Yet, very few existing studies investigated this joint challenge. Data difficulty factors, which have been shown to be key challenges in class imbalanced data streams, are not taken into account by existing approaches when learning class imbalanced data streams. In this work, we propose a drift adaptable oversampling strategy to synthesise minority class examples based on stream clustering. The motivation is that stream clustering methods continuously update themselves to reflect the characteristics of the current underlying concept, including data difficulty factors. This nature can potentially be used to compress past information without caching data in the memory explicitly. Based on the compressed information, synthetic examples can be created within the region that recently generated new minority class examples. Experiments with artificial and real-world data streams show that the proposed approach can handle concept drift involving different minority class decomposition better than existing approaches, especially when the data stream is severely class imbalanced and presenting high proportions of safe and borderline minority class examples.}




\maketitle

\section{Introduction} \label{section:introduction}

In the past years, the volume and incoming speed of data have increased tremendously. Data frequently arrive continuously in the form of data stream rather than forming a single static data set. Therefore, \textit{data stream learning}, which is able to learn incoming data upon arrival, becomes an increasingly important approach to extract knowledge from data. It has been widely used in real-world applications, such as credit card fraud detection \cite{CreditCardFraud}, software defect prediction \cite{tabassum2020} and spam filtering \cite{SpamFiltering}. \review{There are many types of problems / tasks in data stream learning, for examples, classification, regression, clustering, anomaly detection etc. This work focuses on binary classification.} %aims to address several challenges in data stream classification, specifically binary classification, by leveraging stream clustering techniques.}

\review{Concept drift is a common challenge in data streams. It is a change in the underlying distribution of the problem. Such a change can deteriorate the predictive performance of the data stream learning algorithm because the predictive model built previously may not be valid anymore. To deal with concept drift, data stream learning algorithms can be categorised to as explicit and implicit approaches \cite{nonStationary,LearningUnderConceptDriftOverview}. Explicit approaches employ a concept drift detection method to detect if there is a concept drift, and then adopt strategies to update predictive model to cope with such drift  \cite{nonStationary,LearningUnderConceptDriftOverview}. Implicit approaches do not employ any concept drift detection method but continuously evolve themselves to reflect the current underlying concept, thus adapting to concept drifts \cite{nonStationary,LearningUnderConceptDriftOverview}.}

\review{Data stream learning algorithms can also be categorised by their mode of operation: batch-based (chunk-based) learning and online learning \cite{ConceptDriftAdaptationSurvey,nonStationary}. Batch-based learning refers to as learning the data stream by batches of new training data. It has the advantage of having more data to learn at a given time step, thus the learning approach can better capture the current underlying concept \cite{ConceptDriftAdaptationSurvey,nonStationary}. In contrast, online learning has a stricter requirement which only allows the data stream learning approach to process each training example \reviewII{separately and then discard it \cite{ConceptDriftAdaptationSurvey,nonStationary},rendering it applicable to problems with stricter time and memory requirements}. To deal with concept drifts in a timely fashion, online learning usually is more preferable than batch-based learning which needs to wait for whole batches of training examples to arrive. Moreover, batch-based learning assumes that all training data within the same batch are drawn from the same underlying concept, which may not always be the case in most real-world applications. Thus, this work focuses on \textit{online learning}.}

\review{Another challenge frequently present in real world data stream applications is that their class distribution is often skewed, an issue that is} commonly referred to as \textit{class imbalance} \cite{OnlinceCILSurvey}. For example, in credit card fraud detection, there are always more genuine transactions than fraudulent transactions. In software defect prediction, there are typically more clean than defective components. When class imbalance exists, the data stream learning algorithms are likely to get biased towards the majority class, being likely to misclassify minority class examples. Yet, the minority class is usually the class of interest in the classification task, meaning that misclassifying minority class examples could lead to a high cost. This work focuses on binary classification, thus, there is a majority class and a minority class when the data stream is class imbalanced.

% data-level strategy is the most common approach in the literature because it is not specific to a particular type of learning algorithm. It addresses class imbalance by balancing the number of examples from each class to train the learning algorithm. To achieve this, one can oversample the minority class and / or undersampling the minority class. In particular,

To deal with class imbalance, a category of oversampling strategies has shown to be successful in data set learning (offline learning). They create synthetic examples to enrich the minority class, which causes less overfitting than reusing existing minority class examples \cite{SMOTE,BorderlineSMOTE,GaussianbasedSMOTE}. Some recent work attempted to bring such a successful idea into the field of data stream learning \cite{OnlineBagging_Boosting_CIL,CSMOTE}. However, they usually cache all the minority class examples seen so far into the memory which is impractical for data stream learning. Moreover, reusing all past minority class examples may prevent these approaches from dealing with \textit{concept drifts} (changes in the underlying probability distribution, a.k.a., \textit{concept} \cite{EnsembleSurvey}) affecting the minority class.

% Data streams typically present concept drift, which deteriorates the predictive performance of the data stream learning algorithm, as the predictive model built previously is not valid anymore. So, strategies are required to adapt to such changes. To react to concept drifts in a timely fashion, data stream learning approaches preferably operate in an online way, where each incoming training example is processed separately and sequentially rather than in batches \cite{OnlineBagging}. Thus, this work focuses on \textit{online learning}.

\review{Dealing with the joint issue of concept drift and class imbalance is a challenging task. In particular, the relatively small number of minority class examples mean that it may be more difficult to detect or adapt to concept drifts affecting the minority class.} Many studies have been proposed to deal with either class imbalance or concept drift. However, existing work to deal with their joint challenge remains little. Although a recent survey work \cite{OnlinceCILSurvey} showed that class imbalance is a more dominant factor than concept drift in affecting the predictive performance, the effectiveness of the existing class imbalance techniques for data stream learning could potentially be compromised by concept drift as most of them are not prepared to deal with drifts. Recent work addresses this challenge by finding relevant past minority class examples for oversampling \cite{HUWRS.IP} or performing synthetic minority class oversampling based on the statistics of the minority class after drift detection \cite{VFC_SMOTE}. These methods are not always ideal because relevant past minority class examples might not exist while relying on drift detection to reset minority class statistics could be detrimental, especially when the drift is gradual. 

\review{In addition, the method of storing past examples as proposed in \cite{HUWRS.IP} may be impractical for data stream learning when there are strict space constraints. Similarly, synthesising new examples based on simple statistics of past examples as proposed in \cite{VFC_SMOTE} overlooks important data difficulty factors within the class. Specifically, this method does not consider the location of past examples in the feature space. These data difficulty factors include concept drifts involving different movements of the minority class sub-clusters, changing class imbalance ratio, and changing the ratios of different types of minority class examples. Existing work has shown that these factors are critical in learning from drifting class imbalanced data streams \cite{DataDifficulty}.}

%
% A more recent work from the field of software effort estimation proposed to create synthetic minority class data by adding Gaussian noise to existing examples \cite{GauNoise}. Such method is memory efficient but it may cause overfitting to the most recent decision area of the minority class.
%
%Moreover, another study \cite{DataDifficulty} recently showed that \textit{data difficulty factors} are also crucial in learning drifting class imbalanced data streams. These factors include concept drifts involving different movements of the minority class sub-clusters, changing class imbalance ratio, and changing the ratios of different types of minority class examples. Yet, existing approaches do not take data difficulty factors into account to improve concept drift handling. For instance, synthesising examples based on simple statistics of past examples \cite{VFC_SMOTE} ignores the data difficulty factors within the class, as it does not take location in the feature space into the account. 
Therefore, new approaches are needed to better address concept drifts with multiple data difficulty factors in class imbalanced data streams. To fill this research gap, this paper aims to answer the following research questions:

% The RQs below are copied from the thesis.
\begin{itemize}
    \item \textbf{RQ1)} How to produce minority class synthetic examples for oversampling so that we could explore the decision areas of the minority class to better consider data difficulty factors while adapting to concept drift?
    \item \textbf{RQ2)} How does the proposed approach perform compared to existing approaches in different types of concept drift affecting the minority class? For which types does it perform the best and worst? Why?
    \item \textbf{RQ3)} How does the proposed approach perform compared to existing approaches when applied to real-world data streams?
\end{itemize}

To answer RQ1, we propose a novel method to create synthetic minority class examples for oversampling based on stream clustering. The motivation is that stream clustering methods use a set of micro-clusters as the abstraction/compression of the examples they have seen so far. They usually retain micro-clusters by temporal order, which means old micro-clusters are forgotten. Therefore, the information they hold reflects the characteristics of the current underlying concept. Our novel method exploits this nature of stream clustering methods to track the current decision areas of the minority class. \review{It then generates synthetic minority class samples for oversampling within the region where real minority class examples have been recently observed.} With this strategy, the proposed method is less likely to produce noisy synthetic examples while being able to explore the decision areas of the minority class, better considering data difficulty factors when adapting to concept drift (RQ1).

% within the region that recently generated new minority class examples

The proposed approach is compared against five existing approaches through experiments on artificial data streams considering different data difficulty factors and class imbalance ratios, and real-world data streams (RQ2, RQ3). The results show that SMOClust handled concept drifts of different minority class sub-clusters movements better than existing approaches (RQ2, RQ3). It also performed better than existing approaches when the data stream is severely class imbalanced and presents high proportions of safe and borderline \cite{TypesOfMinorityClass} minority class examples (RQ2, RQ3).
% Its major weakness is to handle rare and outlier minority class examples(RQ2, RQ3).
\review{Its major weakness is to handle data streams presenting large proportions of rare and outlier \cite{TypesOfMinorityClass} minority class examples (RQ2, RQ3).}

The rest of this paper is organised as follows. Section \ref{section:related work} discusses related work on synthetic minority class oversampling techniques and state-of-the-art approaches in dealing with class imbalance and concept drift in data stream learning. Section \ref{section:proposed approach} presents the proposed approach. Section \ref{section:experiments} presents the experimental study and discusses the results. Section \ref{section:conclusion} concludes this study and discusses the future work.

\section{Related Work} \label{section:related work}

This section first introduces class imbalance and existing resampling methods for class imbalanced learning in Section \ref{section:resampling methods}. Section \ref{section:drifting class imbalanced data stream learning approaches} then discusses the state-of-the-art approaches to deal with class imbalance and concept drift in data stream learning. \reviewII{Table \ref{table:SMOClust vs others - Novelty} summarises the main characteristics of the approaches discussed in this section. At the end of this table, we contrast these with SMOClust, the approach that we propose in Section \ref{section:proposed approach} of this paper.}

\afterpage{
\begin{landscape}
\begin{table}[!h]
\centering
\caption{\reviewII{Comparison of Principal Characteristics Between Related Works and SMOClust}}
\label{table:SMOClust vs others - Novelty}
\renewcommand\tabcolsep{1.7pt}
\begin{threeparttable}
\begin{tabular}{l|c|c|c|c}
\hline
\makecell[c]{Approach} & Classifier Type & Approach Type & Concept Drift Adaptation & Class Imbalance Strategy \\
\hline
\hline
OnlineUnderOverBagging \cite{OnlineBagging_Boosting_CIL} & \makecell{Online \\ Ensemble} & N/A & N/A & \makecell{Undersampling +\\ Oversampling} \\
\hline
OnlineSMOTEBagging \cite{OnlineBagging_Boosting_CIL} & \makecell{Not Strictly Online\tnote{a} \\ Ensemble} & N/A & N/A & SMOTE \\
\hline
% OnlineAdaC2 \cite{OnlineBagging_Boosting_CIL} & \makecell{Online \\ Ensemble} & N/A & N/A & Cost Weight \\
% \hline
% OnlineCSB2 \cite{OnlineBagging_Boosting_CIL} & \makecell{Online \\ Ensemble} & N/A & N/A & Cost Weight \\
% \hline
% OnlineRUSBoost \cite{OnlineBagging_Boosting_CIL} & \makecell{Online \\ Ensemble} & N/A & N/A & Undersampling \\
% \hline
% OnlineSMOTEBoost \cite{OnlineBagging_Boosting_CIL} & \makecell{Partly Online\tnote{a} \\ Ensemble} & N/A & N/A & SMOTE \\
% \hline
Learn++.CDS \cite{Learn++CDS/Learn++NIE} & \makecell{Batch-based \\ Ensemble} & Implicit & Weighted Majority Ensemble & SMOTE \\
\hline
Learn++.NIE \cite{Learn++CDS/Learn++NIE} & \makecell{Batch-based \\ Ensemble} & Implicit & Weighted Majority Ensemble & Sub-ensemble Method \\
\hline
DWMIL \cite{DWMIL} & \makecell{Batch-based \\ Ensemble} & Implicit & Weighted Majority Ensemble & Undersampling \\
\hline
HUWRS.IP \cite{HUWRS.IP} & \makecell{Batch-based \\ Ensemble} & Implicit & \makecell{Hellinger Distance +\\ Random Subspace Method} & Instance Propagation \\
\hline
OOB \cite{OOB-UOB} & \makecell{Online \\ Ensemble} & Implicit\tnote{b} & Fading Factor & Oversampling \\
\hline
UOB \cite{OOB-UOB} & \makecell{Online \\ Ensemble} & Implicit\tnote{b} & Fading Factor & Undersampling \\
\hline
ESOS-ELM \cite{ESOS_ELM} & \makecell{Online \\ Ensemble} & Explicit & \makecell{Hypothesis testing +\\ Weighted Majority Ensemble} & Sub-ensemble Method \\
\hline
C-SMOTE \cite{C_SMOTE} & \makecell{Not Strictly Online\tnote{a} \\ Meta-strategy} & Explicit & ADWIN & SMOTE + ADWIN \\
\hline
VFC-SMOTE \cite{VFC_SMOTE} & \makecell{Online \\ Meta-strategy} & Explicit & Employ Drift Detector & \makecell{Synthetic Minority Oversampling by Beta \\ Distribution Sampling + Histogram-based Sketch}  \\
\hline
SMOTE-OB \cite{SMOTE_OB} & \makecell{Online \\ Ensemble} & Explicit & Employ Drift Detector & \makecell{Synthetic Minority Oversampling by Beta \\ Distribution Sampling + Histogram-based Sketch} \\
\hline
CSARF \cite{CSARF} & \makecell{Online \\ Ensemble} & Explicit & \makecell{Employ Drift Detector +\\ Weighted Majority Ensemble} & \makecell{Cost-sensitive Weighting +\\ Sub-ensemble Method} \\
\hline
ROSE \cite{ROSE} & \makecell{Not Strictly Online\tnote{a} \\ Ensemble} & Explicit & \makecell{ADWIN +\\ Weighted Majority Ensemble} & \makecell{Cost-sensitive Weighting +\\ Undersampling} \\
\hline
\hline
\makecell[l]{SMOClust \\ (Proposed approach)} & \makecell{Online \\ Meta-strategy} & Explicit & Employ Drift Detector & \makecell{Synthetic Minority Oversampling by Multivariate \\ Skewed Gaussian Sampling + Stream Clustering} \\
% \makecell[l]{Generate synthetic data by multivariate \\ skewed Gaussian sampling method.} 
\hline
\end{tabular}
\begin{tablenotes}
\begin{footnotesize}
\item[a] Not Strictly Online: Whilst these approaches process training examples upon arrival, they store training examples for later use.
\item[b] Can only deal with P(Y) drift.
\end{footnotesize}
\end{tablenotes}
\end{threeparttable}
\end{table}
\end{landscape}
}

\subsection{Resampling Methods for Class Imbalance} \label{section:resampling methods}

Class imbalance refers to the data set or data stream having at least one under-represented class (minority class). In this situation, the machine learning model tends to misclassify minority class examples more frequently than the majority class because there exists very little information about the minority class.

Approaches to address class imbalance are mainly divided into three categories: algorithm-level approach, \review{ensemble approach, and data-level approach} \cite{OnlinceCILSurvey}. Algorithm-level approaches are often called cost-sensitive approaches, as they place a higher cost when misclassifying minority class examples than majority class examples. Ensemble approaches create different class balanced training subsets to train each ensemble member. Data-level approaches modify the class distribution using a resampling method, such that standard machine learning models can learn from both classes with the same amount of information. They can be applied during the data pre-processing phase. Due to this generic nature, this work focuses on data-level approaches.

% SMOTE, Borderline-SMOTE, G-SMOTE
\textit{Undersampling} and \textit{Oversampling} are two main types of data-level approaches \cite{OnlinceCILSurvey}. Undersampling methods reduce the number of majority class examples for training, usually removing noisy examples or examples that are deemed to have a low impact on the decision boundary. Yet, it has the chance to cause important information loss. On the other hand, oversampling methods increase the number of minority class examples, by replication or synthesis. They will not cause any information loss but they cause longer training time and are likely to cause overfitting when training on the same examples multiple times.

Synthetic Minority Oversampling Technique (SMOTE) \cite{SMOTE} is a very renowned oversampling technique in offline learning, which synthesises minority class examples for oversampling, thus balancing the data set. In practice, SMOTE first randomly chooses an existing minority class example from the data set, denoted as $x_i$. It then randomly chooses one of the k-nearest neighbours of $x_i$ in the minority class, denoted as $x'_i$. After that, a difference vector between $x_i$ and $x'_i$ is calculated. To create a point along the line between $x_i$ and $x'_i$, each dimension of the difference vector is multiplied by a random number $\theta$ ($0 < \theta < 1$), then the resulting difference vector is added to $x_i$ dimensionwisely. SMOTE performs this procedure until the target oversampling rate $M$ is reached. This oversampling rate $M$ and the $k$ for k-nearest neighbours are the hyper-parameters of the algorithm.

Many variants of SMOTE have been proposed in the last two decades. For example, Borderline-SMOTE \cite{BorderlineSMOTE} considers that examples close to the decision boundary are more difficult to learn, thus it synthesises minority class examples around this area. Gaussian-based SMOTE (G-SMOTE) \cite{GaussianbasedSMOTE} is a more recent approach which tend to synthesise examples very close to the existing minority class examples. Other well-known methods of synthetic minority oversampling include ADASYN \cite{ADASYN}, DBSMOTE \cite{DBSMOTE}, SWIM \cite{SWIM} etc.

On top of the class imbalance ratio, it has been pointed out that data difficultly factors also greatly impact the classification performance \cite{TypesOfMinorityClass} . These factors describe the characteristic of a given example (usually the minority class example) in the feature space:

\begin{itemize}
    \item Safe: Surrounded by examples from the same class.
    \item Borderline: Located near the decision boundary.
    \item Rare: Located deep inside the decision region of the opposite class, together with handful examples from the same class.
    \item Outlier: Isolated and located deep inside the decision region of the opposite class.
\end{itemize}

The aforementioned methods can also be considered as taking the data difficulty factors into the account. For example, Borderline-SMOTE synthesises minority class examples around the borderline region, while G-SMOTE can be considered as synthesising minority class examples in the safe region.

However, these synthetic minority oversampling methods could not be applied to class imbalanced data stream learning directly as they cache the entire data set into memory, which is impractical in data stream learning. For example, OnlineSMOTEBagging \cite{OnlineBagging_Boosting_CIL} \review{is one of this kind. It} replaces simple oversampling with SMOTE in OnlineUnderOverbagging. \review{In our preliminary experiments with the data streams used in this work, we attempted to run OnlineSMOTEBagging. However, OnlineSMOTEBagging consumed all the memory we had access to (8GB), resulting in failure to complete the run. Furthermore,} the underlying concept of the data stream may change over time (concept drift). The cached examples may be from different concepts. Thus, synthesising minority class examples based on them may not always follow the current underlying concept.

Additionally, one recent work in the field of software effort estimation is also quite inspiring \cite{GauNoise}. They enlarge the software project data set by adding Gaussian noise to the existing examples. This method could be particularly related to synthetic minority oversampling for class imbalanced data stream learning as it is memory efficient and fast to perform. The potential risk is that, if we apply it to the most recent minority class examples, it might cause overfitting to such a recent area.

\subsection{Approaches for Class Imbalanced Data Stream Learning in the Presence of Concept Drift} \label{section:drifting class imbalanced data stream learning approaches}

% How existing work deal with concept drift in class imbalanced data stream learning?
% - Explicit approaches:
%   - The most basic form is to use a class imbalance technique to reduce the base learner's bias towards the majority class, while using a drift detector to actively estimate whether a concept drift has happened.
%   - DDM-OCI, LFR, PAUC-PH
%   Other explicit approaches:
%   - OOB, UOB, C-SMOTE, ESOS-ELM
%   (Need justification on why SMOClust was not compared with C-SMOTE, Shuo asked this during the viva / to the thesis.)

% - Implicit approaches:
%   - Learn++.CDS, Learn++.NIE, DWMIL, HUWRS.IP

\reviewII{This section discusses approaches that are closely related to the proposed approach. For a comprehensive survey on class imbalanced data stream learning, please refer to \cite{aguiar2022survey}.}

\reviewII{Broadly speaking,} existing approaches to deal with class imbalance and concept drift have two main categories: explicit approach and implicit approach.

\paragraph{Explicit Approaches}
Explicit approaches estimate whether a concept drift has happened, usually by employing a drift detector to monitor the predictive performance of the base learner / main ensemble. This drift detector can be any from the literature, ideally using a class imbalance insensitive metric, such as DDM-OCI \cite{DDM-OCI}, LFR \cite{LFR}, PAUC-PH \cite{PAUC_PH} etc.
% Examples of explicit approaches include Oversampling-based and Undersampling-based Online Bagging (OOB and UOB) \cite{OOB-UOB}, Continuous-SMOTE (C-SMOTE) \cite{CSMOTE}, and ESOS-ELM \cite{ESOS_ELM}.

Continuous-SMOTE (C-SMOTE) \cite{CSMOTE} is \reviewII{one of the pioneers who bring SMOTE to drifting class imbalanced data stream learning}. It uses an Adaptive Window (ADWIN) \cite{ADWIN} to store the most recent examples and applies SMOTE to the minority class examples inside the ADWIN for oversampling. Upon drift detection, the old window of ADWIN is dropped as it is deemed to belong to the old concept.
% The ADWIN also monitors the error rate of the base learner, as a strategy to detect concept drift. Upon drift detection, the older sub-window is discarded. However, this strategy is not responsible to adapt the base learner to concept drift. It only makes sure that the examples in the ADWIN belong to the current concept. This means that another concept drift detector is needed if the base learner is an explicit learner. 
\reviewII{However, when there is no concept drift detection, C-SMOTE keeps storing minority class examples which can cause memory issues. Besides, SMOTE does not take decision boundaries and data difficulty factors into consideration, thus noisy examples may be generated.}

\reviewII{Very Fast Continuous-SMOTE (VFC-SMOTE) \cite{VFC_SMOTE} was proposed to solve the issues faced by C-SMOTE. It uses a dynamic summary data structure, called ``sketch'', to summarise the statistics of past examples. It generates synthetic examples by Beta distribution sampling from a set of summaries in the sketch, where each summary has the information of one input feature of past examples. When generating synthetic minority class examples, VFC-SMOTE tends to choose summaries that represent more past examples, which means it tends to generate synthetic minority class examples in the dense area of minority class. Nevertheless, this method may generate considerably noisy synthetic examples because it samples each input feature individually and does not adopt mechanisms to try to respect decision boundaries.}

% SMOTE-OB
\reviewII{SMOTE with Online Bagging (SMOTE-OB) \cite{SMOTE_OB} is another approach that is similar to VFC-SMOTE. It incorporates the strategy of generating synthetic minority class examples from VFC-SMOTE into OnlineUnderOverBagging \cite{OnlineBagging_Boosting_CIL}. With this design, SMOTE-OB combines three data-level re-balancing methods to combat class imbalance while training the base learners diversely \cite{SMOTE_OB}. However, as SMOTE-OB uses the same synthetic minority class examples generating strategy as VFC-SMOTE, it faces the same disadvantages in terms of potentially generating considerably noisy synthetic examples.}

Ensemble of Subset Online Sequential Extreme Learning Machine (ESOS-ELM) \cite{ESOS_ELM} is another notable explicit approach for drifting class imbalanced data stream learning. It uses a sub-ensemble method to train each base learner with an approximately equal number of majority and minority class examples, thus dealing with class imbalance. To deal with concept drift, it uses a threshold-based strategy with hypothesis testing to detect any significant change in the predictive performance of the main ensemble, thus reporting concept drift. Meanwhile, it also uses a weighted majority vote system, based on G-Mean, to adapt to any potential concept drift that could not be detected by the aforementioned method. ESOS-ELM's sub-ensemble method is time efficient in dealing with class imbalance as it does not replicate or synthesise any examples. However, it does not provide additional information to explore the decision areas of minority class. Besides, ESOS-ELM is restrictive in terms of base learner type. It only allows to use ELMs.

% CSARF
\reviewII{Cost-sensitive Adaptive Random Forest (CSARF) \cite{CSARF} is an online, cost-sensitive sub-ensemble method designed to address the challenges of drifting class imbalanced data streams. It is a variant of the Adaptive Random Forest (ARF) \cite{ARF} algorithm. It incorporates a drift detector and a weighted majority ensemble to handle concept drift. To deal with class imbalance, CSARF utilises the Matthews Correlation Coefficient (MCC), a class imbalance insensitive metric, to assign weights to internal decision trees and ensure that all trees are trained with examples from the minority class \cite{CSARF}. While CSARF offers speed and memory efficiency due to its cost-sensitive approach, it fails to consider factors related to data difficulty. Additionally, CSARF is limited to using only the Hoeffding Tree \cite{HTNB} as base learners.}

% ROSE
\reviewII{Robust Online Self-Adjusting Ensemble (ROSE) \cite{ROSE} is a cost-sensitive ensemble method designed specifically for learning from drifting class imbalanced data streams. It employs ADWIN as a drift detector and uses a weighted majority ensemble to handle concept drift. To address class imbalance, ROSE employs self-adjusting $\lambda$ bagging (where $\lambda$ is determined based on estimated class sizes), and enforces the Hoeffding bound to improve predictive performance in the minority class. Furthermore, ROSE maintains sliding windows per class to store the most recent examples and to create a class balanced data set through undersampling. This class balanced data set is used to build new background base learners. However, similar to CSARF, ROSE does not consider data difficulty factors in its class imbalance adaptation strategy. Additionally, ROSE's strategy for building new background base learners may be prone to more extreme levels of class imbalance in non-stationary data streams because such a scenario would require using very old minority class examples to build new base learners, besides the sliding window initially taking time to get filled with minority class examples.}

In short summary, most existing explicit approaches to deal with class imbalance and concept drift do not explore the decision areas of the minority class. Whilst a few recent work \reviewII{\cite{CSMOTE,VFC_SMOTE,SMOTE_OB}} attempted to fill this research gap, they did not \reviewII{strictly take decision boundaries} and data difficulty factors, which are crucial in data stream learning, into consideration.

\paragraph{Implicit Approaches}
Implicit approaches are usually ensemble learners. They do not actively detect concept drift but continuously update the voting weights of the base learners, thus adapting to any potential changes in the underlying concept. However, in class imbalanced data stream learning, the weighting strategy also needs to consider that the base learner may bias toward the majority class. To address this issue, one can place a higher penalty on the weight of the base learners performing poorly in the minority class (cost-sensitive approach). Another method is to employ a resampling method to reduce the learning bias (data-level approach).
% Examples of implicit approaches include Learn++.CDS \cite{Learn++CDS/Learn++NIE}, Learn++.NIE \cite{Learn++CDS/Learn++NIE}, \acrfull{DWMIL} \cite{DWMIL}, and \acrfull{HUWRS.IP} \cite{HUWRS.IP}.

\reviewII{Oversampling-based and Undersampling-based Online Bagging (OOB and UOB) \cite{OOB-UOB} are two pioneers of data-level approach for class imbalanced data streams. Their idea is to incorporate random oversampling or random undersampling with Online Bagging (OB) \cite{OnlineBagging}. They estimate the current class size based on an exponential smoothing function with a fading factor $\theta$. Whenever a new example $s_t$ with a class label $y_t$ arrives, it is first used to calculate the class imbalance ratio of class $y_t$ to the majority class (OOB) or the minority class (UOB). This ratio is used as the parameter $\lambda$ of Poisson distribution in OB, thus deciding the number of times to train each ensemble member on $s_t$. While OOB and UOB are effective in addressing class imbalance with simple resampling methods, they can only deal with concept drifts that affect the posterior probability of the classes ($P(Y)$).}

Learn++ for Concept Drift with SMOTE (Learn++.CDS) and Learn++ for Non-stationary and Imbalanced Environments (Learn++.NIE) \cite{Learn++CDS/Learn++NIE} are two pioneer \reviewII{batch-based} approaches in this category. They were both based on the well-known approach, Learn++ for Non-Stationary Environment (Lean++.NSE) \cite{Learn++NSE/NOAA}. Learn++.CDS uses SMOTE to balance the most recent batch of training data, while Learn++.NIE is a sub-ensemble method which bootstraps the majority class in the most recent batch of training examples to create different class balanced training sets. They both use weighted majority vote as a strategy to deal with concept drift where ensemble members performing well in the minority class have a higher weight. While they are both great methods to deal with class imbalance, they could struggle when the data stream is severely class imbalanced because there could exist some training batches which has no minority class examples.

Dynamic Weighted Majority for Imbalance Learning (DWMIL) \cite{DWMIL} brought the renowned Dynamic Weighted Majority (DWM) into class imbalanced data stream learning. In general, it changes the weighting metric from accuracy to a class imbalance insensitive metric, such as G-Mean, while adopting UnderBagging \cite{UnderOverBagging&BatchSMOTEBagging}, which is an offline learning approach, as the base learner to deal with class imbalance.

Heuristic Updatable Weighted Random Subspaces with Instance Propagation (HUWRS.IP) \cite{HUWRS.IP} is a batch-based learning approach to deal with drifting class imbalanced data streams. It is based on the approach of HUWRS \cite{HUWRS} which was proposed to learn class balanced data streams. The main novelty of HUWRS.IP is the example selection mechanism, called Instance Propagation (IP), which selects relevant past minority class examples for oversampling the most recent train batch. However, these examples may not exist in the memory.

Shortly summarising, existing implicit approaches to deal with class imbalance and data stream learning either rely on sub-ensemble methods or reusing relevant past examples. These methods do not explore the decision areas of the minority class. They do not take data difficulty factors into account either. Besides, these approaches are batch-based approaches, thus they are unlikely to react to concept drift swiftly due to the need to wait for whole batches to arrive.

\section{Proposed Approach} \label{section:proposed approach}

To answer the RQ1 posed in Section \ref{section:introduction}, we proposed a novel approach called Synthetic Minority Oversampling based on stream Clustering (SMOClust). The main novelty of this approach is to produce synthetic minority class examples for oversampling based on the information compressed by the stream clustering method. Most stream clustering methods represent this information in the form of micro-clusters, which summarise the statistics of past examples that are close together in the feature space. These statistics usually include the vectors of the dimensional-wise cumulative sum and squared sum. Thus, they do not need to cache all the past examples in the memory. Most importantly, this strategy could potentially deal with gradual drift involving different data difficulty factors because stream clustering methods continuously update themselves to reflect the characteristics of the current underlying concept.
% \reviewII{To highlight the novelty of SMOClust, Table \ref{table:SMOClust vs others - Novelty} briefly summarises the key idea of SMOClust and compares it against a few state-of-the-art SMOTE-based approaches}

SMOClust also employs a concept drift detector to monitor the predictive performance of the base learner, as a strategy to deal with abrupt drift. Thus, it is an explicit concept drift adaptation approach. Upon drift detection, the base learner will be reset.
% SMOClust does not store any past base learner into its memory.
Although this strategy may not always be ideal \cite{DP,CDCMS}, this work focuses on investigating the effectiveness of the novel stream clustering based synthetic minority oversampling strategy in learning class imbalanced data streams with concept drift. So, it is intended to keep other components of SMOClust simple to analyse the characteristics of the proposed strategy.

Algorithm \ref{alg:SMOClust-Overview} presents the pseudo-code over-viewing SMOClust. The details of its working mechanism are described and explained as follows.

% Alg. 1
\begin{algorithm}[!ht]
\footnotesize
% \small
\caption[]{Synthetic Minority Oversampling based on Stream Clustering - SMOClust} \label{alg:SMOClust-Overview}
{\textbf{Hyper-parameters}: Base Learner($b$), Stream Clustering Method($sc$), Class Size Fading Factor($\theta$), Gaussian Noise Variance($v$), Categorical Change Probability($P_c$), $k$-Nearest neighbour($k$), Drift Detector($d$), Data Stream($S$)} \\
{\textbf{Variables}: Base Learner($B$), Stream Clustering Methods array($SC[]$)}
% \vspace{-0.2cm}
\begin{algorithmic}[1]
\State Create a array of stream clustering methods ($SC[]$). Each of them corresponds to a class of the classification task. \label{SMOClust-Alg1:create-stream-clustering-methods}
\For{each new example $s_t$ from the data stream $S$}
    \State Update the drift detector by the prediction made by base learner $B$ to $s_t$
    \If {drift detection alarm is issued} \label{SMOClust-Alg1:drift-detection}
    \State Reinitialise the base learner $B$ \label{SMOClust-Alg1:reset-base-learner}
    \EndIf
    \State Train the base learner $B$ and update its estimated class size using $s_t$  \label{SMOClust-Alg1:update-base-learner-with-s_t}
    \State Store the latest example of each class \label{SMOClust-Alg1:store-s_t}
    \State $class_{maj}, class_{min} \gets$ Determine the current majority and minority class based on the\par
    \hskip\algorithmicindent estimated class size by $B$
    \While{\textbf{(}the minority class estimated by $B$ is smaller than that of majority class \textbf{AND}\par
        \hskip\algorithmicindent all stream clustering methods can provide micro-clustering results\textbf{)} \textbf{OR}\par
        \hskip\algorithmicindent SMOClust has observed any minority class example} \label{SMOClust-Alg1:while-loop-start}
        \If{all stream clustering methods are ready to provide micro-clustering results}
            \State $mCluster_{anchor} \gets$ Randomly pick a frequently updated micro-cluster of $class_{min}$ \label{SMOClust-Alg1:weighed-random-draw-anchor}
            \If{$mCluster_{anchor}$ is surrounded by micro-clusters of $class_{min}$} \label{SMOClust-Alg1:Gen-synth-by-mCluster-start}
                \State $synthInst^{Bin} \gets$ create a synthetic example using Alg. \ref{alg:Gen-Synth-kNN} \label{SMOClust-Alg1:Gen-synth-by-kNN}
            \Else
                \State $synthInst^{Bin} \gets$ create a synthetic example by Gaussian\par
                \hskip\algorithmicindent sampling $mCluster_{anchor}$ \label{SMOClust-Alg1:Gen-synth-by-anchor}
            \EndIf \label{SMOClust-Alg1:Gen-synth-by-mCluster-end}
            % \State $synthInst \gets binaryToNominal(synthInst^{Bin}.copy())$
            \State Train $SC[class_{min}]$ using $synthInst^{Bin}$ (without class attribute) \label{SMOClust-Alg1:Gen-synth-by-mCluster-train-SC}
            \State Train the base learner $B$ and update its estimated class size using $synthInst^{Bin}$ \label{SMOClust-Alg1:Gen-synth-by-mCluster-train-B}
        \Else
            \State $synthInst \gets$ create a synthetic example by adding Gaussian noise to latest\par
            \hskip\algorithmicindent minority class example \cite{GauNoise} \label{SMOClust-Alg1:Gen-synth-by-GauNoise}
            % \State $synthInst^{Bin} \gets nominalToBinary(synthInst.copy())$
            \State Train $SC[class_{min}]$ using $synthInst$ (without class attribute) \label{SMOClust-Alg1:Gen-synth-by-GauNoise-train-SC}
            \State Train the base learner $B$ and update its estimated class size using $synthInst$ \label{SMOClust-Alg1:Gen-synth-by-GauNoise-train-B}
        \EndIf
    % \algstore{SMOClust-overview}
    \EndWhile
    % \State $s^{Bin,noClass}_{t} \gets nominalToBinary(s_t.copy())$ \label{SMOClust-Alg1:convert-s_t}
    \State Use $s_t$ (without class attribute) to train the stream clustering method that corresponds\par
    \hskip\algorithmicindent to the class value of $s_t$ \label{SMOClust-Alg1:train-SC-with-s_t}
\EndFor
\end{algorithmic}
\end{algorithm}

SMOClust \review{is a data stream learning algorithm that uses} a base learner $B$ to learn from and make predictions to new examples. This base learner $B$ could be any single learner, such as Hoeffding Tree \cite{HTNB}, or an ensemble learner, such as Online Bagging \cite{OnlineBagging}. SMOClust does not store past models. It uses stream clustering methods $SC[]$ to manage sets of micro-clusters that compress the information of past examples. There is one stream clustering method for each class of the problem (line \ref{SMOClust-Alg1:create-stream-clustering-methods}, Algorithm \ref{alg:SMOClust-Overview}). The stream clustering method can be arbitrary from the literature, \review{such as Clustream \cite{Clustream}, StreamKM++ \cite{StreamKM++}, DenStream \cite{DenStream}, Clustree \cite{Clustree} etc. In this work, Clustream was chosen because it is largely invariant for different types of concept drifts, meaning that it can effectively adapt to concept drift without compromising the quality of its clustering results \cite{ClusteringUnderConceptDrift}.} The strategy of synthesising minority class examples for oversampling based on micro-clusters is explained in Section \ref{section:SMOClust-Generate Syntheitc Minority Class example}.

The most recent example $s_t$ will be first used for concept drift detection (line \ref{SMOClust-Alg1:drift-detection}, Algorithm \ref{alg:SMOClust-Overview}). This concept drift detection method can be arbitrary from the literature, such as DDM \cite{DDM}, DDM-OCI \cite{DDM-OCI}, PAUC-PH \cite{PAUC_PH}, ADWIN \cite{ADWIN} etc. Upon drift detection, the base learner $B$ and time decay class sizes are reset but not the stream clustering methods $SC[]$ because they are prepared to adapt to concept drifts (line \ref{SMOClust-Alg1:reset-base-learner}, Algorithm \ref{alg:SMOClust-Overview}). That said, after concept drift detection, the stream clustering methods will still retain some knowledge belonging to the previous concept. This has two advantages: 1) In the case of false positive drift detection, SMOClust can exploit the knowledge stored in the micro-clusters to train the base learner. 2) Knowledge of the pre-drift concept could help to learn the post-drift concept, especially when the drift has low severity \cite{DDD}.

After that, SMOClust uses $s_t$ to train $B$ and to update the time decay class sizes (line \ref{SMOClust-Alg1:update-base-learner-with-s_t}, Algorithm \ref{alg:SMOClust-Overview}). The time decay class sizes estimate the current minority class and thus determine the oversampling rate. Equation \ref{eq:time decay class size} presents the calculation of the normalised class size of class $c_m$ at time step $t$ \cite{OOB-UOB}:

\begin{equation} \label{eq:time decay class size}
classSize(c_m)^{(t)} = \begin{cases}\frac{1}{|M|}, & \text{if}\ t=f \\
\frac{[c_{s_t} = c_m]+\theta\times classSize(c_m)^{(t-1)} \times(t-f)}{t-f+1}, & \text{otherwise}\end{cases}
\end{equation}

\noindent where $m \in M$ and $M = \{0,1\}$, considering binary classification tasks and $\theta$ $(0<\theta<1)$ is a predefined time decay factor. $c_{s_t}$ is the true class of $s_t$. Thus, $[c_{s_t} = c_m]=1$ if the true class of $s_t$ is $c_m$, otherwise 0. $f$ is the first time step used in the calculation. Note that, unlike OOB and UOB \cite{OOB-UOB} which estimate the current class sizes of the data stream, SMOClust estimates the class imbalance degree of the information seen by the base learner rather than the class imbalance degree of the data stream. Thus, synthetic examples are also used to update the class sizes. The reason behind this design is discussed together with the strategy of training the base learner with synthetic examples.

SMOClust \review{first records} the most recent examples from each class (line \ref{SMOClust-Alg1:store-s_t}, Algorithm \ref{alg:SMOClust-Overview})\review{, then} checks if the base learner has learnt from both classes equally (line \ref{SMOClust-Alg1:while-loop-start}, Algorithm \ref{alg:SMOClust-Overview}). If not, SMOClust will generate synthetic minority class examples for oversampling based on the micro-clusters \review{of the minority class} (line \ref{SMOClust-Alg1:Gen-synth-by-mCluster-start}-\ref{SMOClust-Alg1:Gen-synth-by-mCluster-end}, Algorithm \ref{alg:SMOClust-Overview}), which is detailed in Section \ref{section:SMOClust-Generate Syntheitc Minority Class example}. 

In the case that \review{not all stream clustering methods can provide micro-clustering results and SMOClust has observed and recorded} the most recent ``real'' example of the minority class (denoted as $s_{last\_minority}$), SMOClust will generate a synthetic minority class by adding Gaussian noise to $s_{last\_minority}$ for oversampling (line \ref{SMOClust-Alg1:Gen-synth-by-GauNoise}, Algorithm \ref{alg:SMOClust-Overview}). This strategy follows the strategy proposed by \cite{GauNoise}, except SMOClust treats ordinal attributes as categorical attributes due to the limitation in MOA \cite{MOA}.
% Generally speaking, SMOClust adds Gaussian noise to a numeric attribute of $s_{last\_minority}$ by applying a Gaussian distribution centred at the current value of $s_{last\_minority}$ at that numeric attribute with a predefined variance $v$. For nominal attributes (including categorical and ordinal attributes), SMOClust changes them to any other possible value with a predefined probability $P_c$. 

No matter the synthetic minority class example is generated based on micro-clusters or Gaussian noise, SMOClust will use it to train the base learner and the corresponding stream clustering method, and to update the class size immediately (line \ref{SMOClust-Alg1:Gen-synth-by-mCluster-train-SC}-\ref{SMOClust-Alg1:Gen-synth-by-mCluster-train-B}, \ref{SMOClust-Alg1:Gen-synth-by-GauNoise-train-SC}-\ref{SMOClust-Alg1:Gen-synth-by-GauNoise-train-B}, Algorithm \ref{alg:SMOClust-Overview}). \review{This strategy can prevent the base learner from biasing towards the majority class when there are no ``real'' minority class examples arrive for a long period, which is likely to happen when the data stream is extremely class imbalanced. Also, updating the class sizes with both ``real'' and synthetic examples allows us to estimate if the base learner has learnt from both classes equally. If not, SMOClust will then create synthetic minority class examples to train the base learner immediately.}

In the case of none of the above conditions being satisfied\review{, i.e., none of the conditions of the while-loop are satisfied (line \ref{SMOClust-Alg1:while-loop-start}, Algorithm \ref{alg:SMOClust-Overview})}, SMOClust will not perform any oversampling because this means either oversampling is not needed or there is no information about the minority class for SMOClust to generate synthetic examples. Lastly, a copy of the most recent example $s_t$ is converted to a suitable format to train the stream clustering method, corresponding to the class value of $s_t$ (line \ref{SMOClust-Alg1:train-SC-with-s_t}, Algorithm \ref{alg:SMOClust-Overview}).

\subsection{Generating a Synthetic Minority Class Example for Oversampling using Micro-clusters} \label{section:SMOClust-Generate Syntheitc Minority Class example}

This section presents the overview of generating a synthetic minority class example for oversampling based on micro-clusters. The general idea is to create synthetic minority class examples in one of the dense areas of the minority class. In this way, we can consolidate the knowledge learnt in the existing minority class areas without being greatly affected by noise. In the case where a dense area does not exist, SMOClust will pick one of the past minority class areas to explore the decision boundary around it. 

Algorithm \ref{alg:Gen-Synth-kNN} presents the pseudo-code of this method. The details of generating a synthetic minority class example using micro-clusters can be described as follows.

% Alg. 2
\begin{algorithm}[!ht]
\footnotesize
\caption[]{Generate Synthetic Instance with k-NN Micro-Clusters} \label{alg:Gen-Synth-kNN}
\begin{algorithmic}[1]
\Function{genSynthInstBykNN}{$SC[class_{min}]$, $mCluster_{anchor}$, $class_{min}$, $k$}
    % \State $kNNmClusters \gets$ get $k$ nearest micro-clusters around $mCluster_{anchor}$ \label{SMOClust-Alg2:get-kNN-mClusters}
    \State $sphere\_cluster \gets$ combine $mCluster_{anchor}$ with its $k$ nearest micro-clusters, using Alg. \ref{alg:Combine-mClusters} \label{SMOClust-Alg2:combine-anchor-with-kNN-mClusters}
    \State $synthInst \gets$ create a synthetic example by sampling $sphere\_cluster$, using Alg. \ref{alg:Sample-Skewed-Gaussian} \label{SMOClust-Alg2:Gen-synth-from-skewed-Gaussian}
    % \State $synthInst.setClassValue(s_t.classValue())$
    \State \Return $synthInst$
\EndFunction
\end{algorithmic}
\end{algorithm}

First of all, SMOClust randomly takes one of the micro-clusters from the minority class as an anchor (denoted as $mc_{anchor}^{minority}$) (line \ref{SMOClust-Alg1:weighed-random-draw-anchor}, Algorithm \ref{alg:SMOClust-Overview}). Micro-clusters that are created recently or are updated frequently and recently have higher chance to be chosen as this anchor. After that, SMOClust checks if $mc_{anchor}^{minority}$ is surrounded by the micro-clusters from the same class (line \ref{SMOClust-Alg1:Gen-synth-by-mCluster-start}, Algorithm \ref{alg:SMOClust-Overview}). If this condition is satisfied, SMOClust can consider such area is dense enough to create synthetic minority examples for oversampling. It will then make a copy of $mc_{anchor}^{minority}$ and then combine it with its $k$-Nearest micro-clusters (based on hull distance) in class $class_{min}$ to form a temporary micro-cluster $mc_{temp}$ (line \ref{SMOClust-Alg2:combine-anchor-with-kNN-mClusters}, Algorithm \ref{alg:Gen-Synth-kNN}). We denote such set of $k$-Nearest micro-clusters as $MC^{kNN,minority}$, thus, $\lvert MC^{kNN,minority}\rvert =k$ and each $k$-Nearest micro-clusters is denoted as $mc_i^{kNN,minority}\in MC^{kNN,minority}$. The details of how to combine a set of micro-clusters into one are presented in Algorithm \ref{alg:Combine-mClusters}.

\begin{algorithm}[!ht]
\footnotesize
\caption[]{Combining a set of micro-clusters into one} \label{alg:Combine-mClusters}
\begin{algorithmic}[1]
\Function{combine}{$mClusters[]$}
    \State $c_{new} \gets$ compute the weighted average of the centres of micro-clusters in $mClusters[]$ \label{SMOClust-Alg3:calculate centre}
    \For {each micro-cluster $mClust \in mClusters[]$} \label{SMOClust-Alg3:calculate radius start}
        \State $d \gets$ compute the distance between $c_{new}$ and the centre of $mClust$
        \State $r_n \gets r_n \cup$ (the radius of $mClust$ + $d$)
    \EndFor
    \State $r_{new} \gets$ find the largest value in $r_n$ \label{SMOClust-Alg3:calculate radius end}
    \State \Return a new micro-cluster with centre $c_{new}$ and radius $r_{new}$
\EndFunction
\end{algorithmic}
\end{algorithm}

\begin{figure}[!ht]
\centering
% \includepdf[pages=-]{CombineMC.pdf}
\includegraphics[width=\textwidth]{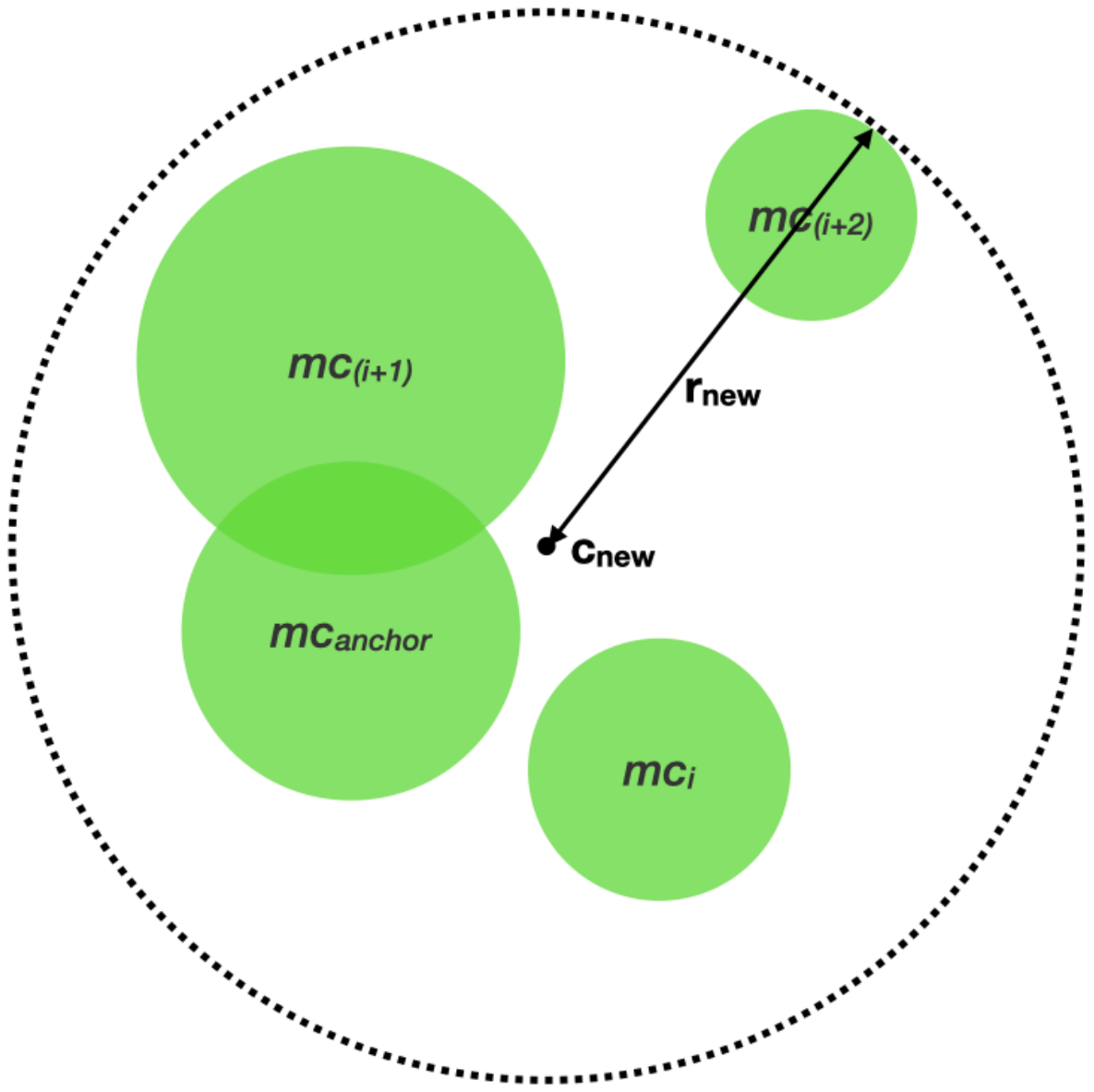}
\caption{Illustration of Combining $mc_{anchor}^{minority}$ with 3-nearest neighbours into one micro-cluster}
\label{figure:combineMC}
\end{figure}

To combine a set of micro-clusters into one, we first need to calculate the new centre $c_{new}$ of the resulting micro-cluster $mc_{temp}$. This can be achieved by getting the weighted average of the centres of the original set of micro-clusters, dimensionwisely (line \ref{SMOClust-Alg3:calculate centre}, Algorithm \ref{alg:Combine-mClusters}). After that, we set the radius $r_{new}$ of the resulting micro-cluster $mc_{temp}$ to as the distance between the new centre to the farthest hull (boundary) among all the original micro-clusters (line \ref{SMOClust-Alg3:calculate radius start}-\ref{SMOClust-Alg3:calculate radius end}, Algorithm \ref{alg:Combine-mClusters}). Figure \ref{figure:combineMC} illustrates an example of combining $mc_{anchor}^{minority}$ with its 3-nearest neighbours into one micro-cluster.

\begin{figure}[!ht]
\centering
\includegraphics[width=\textwidth]{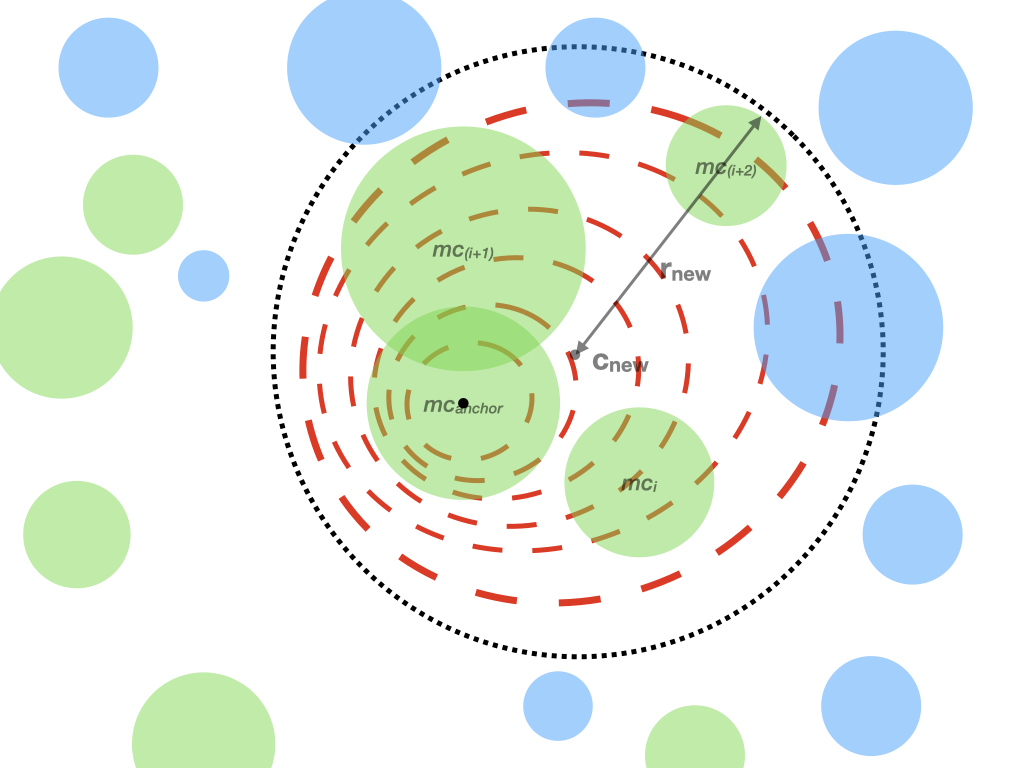}
\caption{Illustration of Sampling a Synthetic Minority Class Example from $mc_{temp}$}
\label{figure:SampleSkewGau}
\end{figure}

A synthetic minority class example will then be generated by sampling from this resulting micro-cluster with the highest chance near the centre of $mc_{anchor}^{minority}$ (line \ref{SMOClust-Alg2:Gen-synth-from-skewed-Gaussian}, Algorithm \ref{alg:Gen-Synth-kNN}). Figure \ref{figure:SampleSkewGau} illustrates an example of sampling from a synthetic minority class example from $mc_{temp}$.

In Figure \ref{figure:SampleSkewGau}, the green circles are the micro-clusters belonging to the minority class while the blue circles are the micro-clusters belonging to the majority class.\footnote{Note that, the size and the number of micro-clusters in Figure \ref{figure:SampleSkewGau} do not necessarily reflect the number of examples in each class. This figure just focuses on a particular region in the feature for explanation purposes.} The black circle line represents $mc_{temp}$ and the red dashed lines are the contour of the probability density function to sampling a point. The closer to the centre of $mc_{anchor}^{minority}$, the higher the probability.

The reason for sampling a new synthetic minority class example close to $mc_{anchor}^{minority}$ is that this $mc_{temp}$ could overlap with the micro-cluster from the other class. If we just sample from $mc_{temp}$ randomly or by a multivariate Gaussian distribution with a mean at $c_{new}$, we will have a high chance to sample a point that is close to the region or the majority class. Therefore, sampling points as synthetic minority class examples from $mc_{temp}$ but close to the centre of $mc_{anchor}^{minority}$ can reduce the risk of generating noisy examples while maintaining the ability to explore this dense region of the minority class.

Although Figure \ref{figure:SampleSkewGau} only illustrates an example in two-dimensional feature space, this idea can be applied to any multi-dimensional space.
% Besides, if we consider the $mc_{temp}$ as a hyper-sphere, this sampling strategy is indirectly using a skewed multivariate Gaussian distribution into $mc_{temp}$ with the maximum of the probability density function at the centre of $mc_{anchor}^{minority}$.
This sampling strategy is further detailed in Section \ref{section:SMOClust-Skewed Gaussian Sampling}.

In the case that $mc_{anchor}^{minority}$ is not surrounded by the micro-clusters belonging to the same class, SMOClust will generate a synthetic minority class example by performing multivariate Gaussian sampling inside $mc_{anchor}^{minority}$ (line \ref{SMOClust-Alg1:Gen-synth-by-anchor}, Algorithm \ref{alg:SMOClust-Overview}). For example, this will be the case when when $mc_{(i+2)}$ (top right green circle in Figure \ref{figure:SampleSkewGau}) is chosen to be the $mc_{anchor}^{minority}$. The the mean of the multivariate Gaussian distribution is the centre of $mc_{anchor}^{minority}$ and the standard deviation is set as a third of the radius of $mc_{anchor}^{minority}$ ($radius / 3$). In other words, the boundary of $mc_{anchor}^{minority}$ is set at three units standard deviations (or standard score = 3) from the centre. Therefore, we have 99.9\% of chance to sample a point within $mc_{anchor}^{minority}$. Gaussian distribution was chosen rather than uniform distribution in sampling $mc_{anchor}^{minority}$ because $mc_{anchor}^{minority}$ could partly overlap with the majority class region. Therefore, sampling a new point as synthetic minority class example close to the centre of $mc_{anchor}^{minority}$ is a safe strategy.

% The reason for sampling from multivariate distribution rather than univariate distribution in each dimension is explained in Section \ref{section:SMOClust-Uniform Sampling}.

% Note that the simple strategy of dimensionwise uniform sampling cannot be used to pick this random direction uniformly because such strategy is sampling from a hyper-cube where all the distances from the centre to the hull are different. This means that the direction with a long distance has a higher chance to be chosen. That is why we need to use Muller's method \cite{Muller1959} to pick this random direction uniformly.

\subsection{Sampling from a Micro-cluster with the Highest Probability at a Designated Location} \label{section:SMOClust-Skewed Gaussian Sampling}

\begin{figure}[!ht]
\centering
\subfigure[Step 1]{\includegraphics[width=0.45\textwidth]{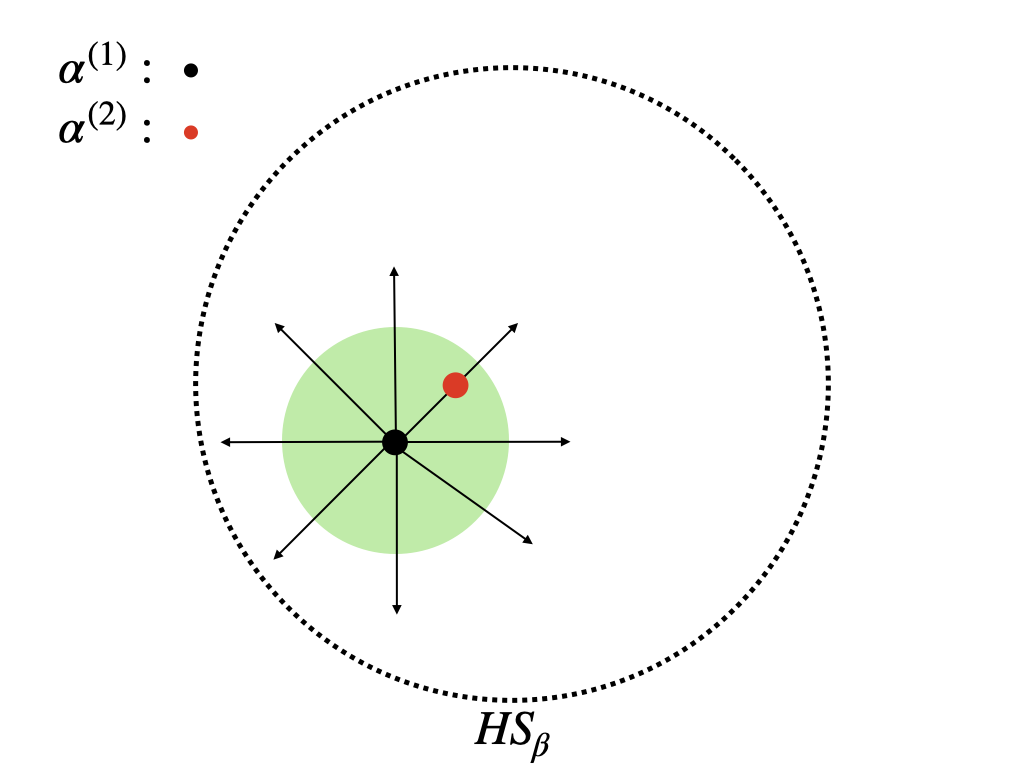} \label{figure:sampling from skew Gau - step 1}}
\subfigure[Step 2]{\includegraphics[width=0.45\textwidth]{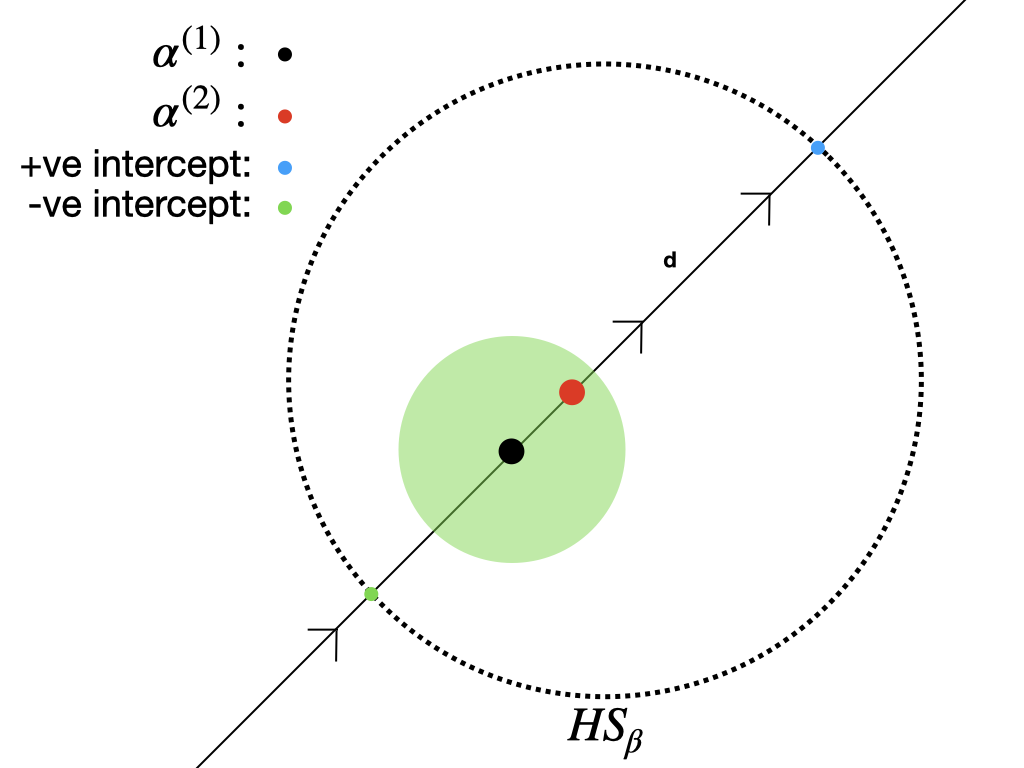} \label{figure:sampling from skew Gau - step 2}}
\subfigure[Step 3]{\includegraphics[width=0.45\textwidth]{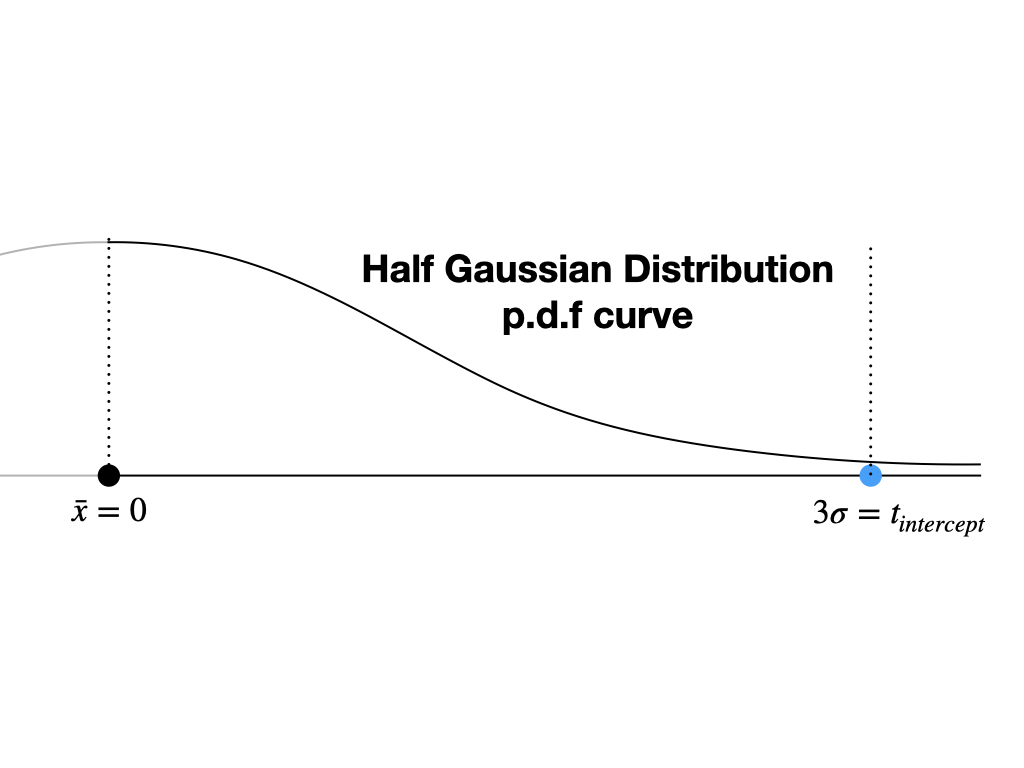} \label{figure:sampling from skew Gau - step 3}}
\subfigure[Two-Dimensional Example of the Sampling Result]{\includegraphics[width=0.45\textwidth]{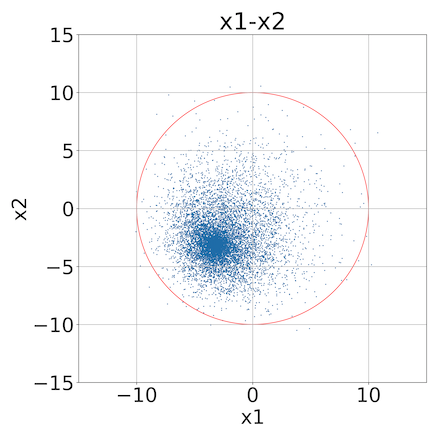} \label{figure:sampling from skew Gau - example}}
\caption{Illustration of Sampling from $mc_{temp}$}
\label{figure:sampling from skew Gau}
\end{figure}

This section present the strategy to sampling points from the temporary micro-cluster $mc_{temp}$ which is formed by combining $mc_{anchor}^{minority}$ and $mc_i^{kNN,minority}\in MC^{kNN,minority}$ with the highest probability at the centre of $mc_{anchor}^{minority}$. The general idea is to sample random points that are inside $mc_{temp}$ and these points are likely to be close to the centre of $mc_{anchor}^{minority}$. The pseudocode of this sampling strategy is presented in Algorithm \ref{alg:Sample-Skewed-Gaussian}. Figure \ref{figure:sampling from skew Gau} illustrates the steps of this sampling strategy and it can be explained as follows.

% Alg. 4
\begin{algorithm}[!ht]
\footnotesize
\caption[]{Sampling from a Hyper-Sphere by Skewed Gaussian with the Maximum of the Probability Density Function at a Designated Location}
\label{alg:Sample-Skewed-Gaussian}
\begin{algorithmic}[1]
\Function{sample\_around\_target}{$\alpha^{(1)}$, $sphere\_cluster$}
    \State $\beta \gets sphere\_cluster.getCentre()$
    \State $r \gets sphere\_cluster.getRadius()$
    \State $dimensions \gets \beta.numOfDimensions()$
    \State $\delta \gets createArrayWithSize(dimensions)$
    \State $\gamma \gets createArrayWithSize(dimensions)$
    \State $\alpha^{(2)} \gets sample\_random\_from\_hypersphere(\alpha^{(1)}, 1)$ \Comment{By Muller's Method \cite{Muller1959}} \label{SMOClust-Alg4:sample-alpha2}
    \State $A \gets 0; B \gets 0; C \gets 0$
    \For {$i \in range(0..dimensions)$}
        \State $\delta[i] \gets \alpha^{(2)}[i] - \alpha^{(1)}[i]$ \label{SMOClust-Alg4:denote-delta}
        \State $\gamma[i] \gets \beta[i] - \alpha^{(1)}[i]$ \label{SMOClust-Alg4:denote-gamma}
        \State $A \gets A + (\delta[i] * \delta[i])$ \Comment{$A=\sum^n_{i=0}\delta^2_i$} \label{SMOClust-Alg4:denote-A}
        \State $B \gets B + (\delta[i] * \gamma[i])$ \Comment{$\sum^n_{i=0}\delta_i\gamma_i$} \label{SMOClust-Alg4:denote-B-sum}
        \State $C \gets C + (\gamma[i] * \gamma[i])$ \Comment{$\sum^n_{i=0}\gamma^2_i$} \label{SMOClust-Alg4:denote-C-sum}
    \EndFor
    \State $B \gets B * -2$ \Comment{$B=-2(\sum^n_{i=0}\delta_i\gamma_i)$} \label{SMOClust-Alg4:denote-B}
    \State $C \gets C - (r * r)$ \Comment{$C=(\sum^n_{i=0}\gamma^2_i)-r^2$} \label{SMOClust-Alg4:denote-C}
    \State \Return $(-B + sqrt(B * B - 4 * A * C)) / (2 * A)$ \Comment{$\frac{-B+\sqrt{B^2-4AC}}{2A}$}
\EndFunction
\end{algorithmic}
\end{algorithm}

Let us first denote the micro-cluster $mc_{temp}$ as $HS_\beta$ which is a hyper-sphere with radius $r$ and centred at $\beta=(\beta_1,\beta_2,\beta_3,...,\beta_n)$, where $n$ is the number of dimensions of the input space of the problem, the equation of this hyper-sphere is:

\begin{equation} \label{eq:hyper-sphere}
\sum^n_{i=0}(x_i-\beta_i)^2=r^2
\end{equation}

\noindent Let us also denote the centre of $mc_{anchor}^{minority}$ to as  $\alpha^{(1)}=(\alpha^{(1)}_1,\alpha^{(1)}_2,\alpha^{(1)}_3,...,\alpha^{(1)}_n)$ (the black dot in Figure \ref{figure:sampling from skew Gau - step 1}), which should always be inside $HS_\beta$. First of all, we need to pick a random direction from $\alpha^{(1)}$ (Figure \ref{figure:sampling from skew Gau - step 1}). This can be achieved by randomly and uniformly picking a point from a unit hyper-sphere centred at $\alpha^{(1)}$, using the Muller's method \cite{Muller1959}.
% (Algorithm \ref{alg:Mullers-method}).
% The detail of the Muller's method \cite{Muller1959} is explained in Section \ref{section:SMOClust-Uniform Sampling}.
We then denote this point to as $\alpha^{(2)}=(\alpha^{(2)}_1,\alpha^{(2)}_2,\alpha^{(2)}_3,...,\alpha^{(2)}_n)$ (the red dot in Figure \ref{figure:sampling from skew Gau - step 1}) (line \ref{SMOClust-Alg4:sample-alpha2}, Algorithm \ref{alg:Sample-Skewed-Gaussian}). Points $\alpha^{(1)}$ and $\alpha^{(2)}$ form an $n$-dimensional infinite long straight line (the line $d$ in Figure \ref{figure:sampling from skew Gau - step 2}), whose parameterised equation is:

\begin{equation} \label{eq:parameterised line}
x_i=\alpha^{(1)}_i+t(\alpha^{(2)}_i-\alpha^{(1)}_i)
\end{equation}

\noindent where $t$ is a scalar and $(\alpha^{(2)}_i-\alpha^{(1)}_i)$ is the direction vector. To find the intercepts of this infinite long line to the hull of $HS_\beta$ (the blue and green dots in Figure \ref{figure:sampling from skew Gau - step 2}), we can substitute Equation \ref{eq:parameterised line} into Equation \ref{eq:hyper-sphere}\footnote{The idea is inspired by the discussion on https://math.stackexchange.com/questions/151064/calculating-line-intersection-with-hypersphere-surface-in-mathbbrn?rq=1}:

\begin{equation} \label{eq:sub parameterised line into hyper-sphere}
\sum^n_{i=0}((\alpha^{(2)}_i-\alpha^{(1)}_i)t+(\alpha^{(1)}_i-\beta_i))^2=r^2
\end{equation}

\noindent Let us denote $\delta_i =\alpha^{(2)}_i-\alpha^{(1)}_i$ and $\gamma_i=\beta_i-\alpha^{(1)}_i$ (line \ref{SMOClust-Alg4:denote-delta} and \ref{SMOClust-Alg4:denote-gamma}, Algorithm \ref{alg:Sample-Skewed-Gaussian}), then Equation \ref{eq:sub parameterised line into hyper-sphere} becomes:

\[ \sum^n_{i=0}(\delta_it-\gamma_i)^2=r^2 \]

\begin{equation} \label{eq:expaned sub}
(\sum^n_{i=0}\delta^2_i)t^2-2(\sum^n_{i=0}\delta_i\gamma_i)t+(\sum^n_{i=0}\gamma^2_i)-r^2=0
\end{equation}

\noindent Let us denote $A=\sum^n_{i=0}\delta^2_i$ (line \ref{SMOClust-Alg4:denote-A}, Algorithm \ref{alg:Sample-Skewed-Gaussian}), $B=-2(\sum^n_{i=0}\delta_i\gamma_i)$ (line \ref{SMOClust-Alg4:denote-B-sum} and \ref{SMOClust-Alg4:denote-B}, Algorithm \ref{alg:Sample-Skewed-Gaussian}) and $C=(\sum^n_{i=0}\gamma^2_i)-r^2$ (line \ref{SMOClust-Alg4:denote-C-sum} and \ref{SMOClust-Alg4:denote-C}, Algorithm \ref{alg:Sample-Skewed-Gaussian}) to solve Equation \ref{eq:expaned sub} based on Bhaskara's equation:

\[ t=\frac{-B\pm\sqrt{B^2-4AC}}{2A} \]

\noindent Here, we just take the positive root of $t$ because it ``follows'' the direction vector, while the negative root ``oppositely follows'' the direction vector (the direction is denoted by the arrows on line $d$ in Figure \ref{figure:sampling from skew Gau - step 2}). i.e.

\begin{equation} \label{eq:positive intercept}
t_{intercept}=\frac{-B+\sqrt{B^2-4AC}}{2A}
\end{equation}

\noindent Substituting $t_{intercept}$ into Equation \ref{eq:parameterised line} will obtain the intercept of the line and the hyper-sphere, following the direction vector (the blue dot in Figure \ref{figure:sampling from skew Gau - step 2}). Thus, to sample points within the $HS_\beta$, we can simply sample a scalar $t_{sample}$ between 0 and $t_{intercept}$ (Figure \ref{figure:sampling from skew Gau - step 3}) and substitute it into Equation \ref{eq:parameterised line} to obtain the sampled point. As we want to sample this point with the highest chance at the target point $\alpha^{(1)}$, we can sample $t_{sample}$ using Gaussian distribution with the mean = 0 and standard deviation = $\frac{t_{intercept}}{3}$. i.e. 

\[ g\sim N(0,(\frac{t_{intercept}}{3})^2) \]
\[ t_{sample}=|g| \]

\noindent At last, we substitute $t_{sample}$ into Equation \ref{eq:parameterised line} to obtain the sample point.

The reason for setting the standard deviation to be $\frac{t_{intercept}}{3}$ is that we want the sampled point to be within the micro-cluster. Yet, the probability density function of the Gaussian distribution has no bounds. Thus, we set the $t_{intercept}$ at 3 standard score (z-score = 3), such that 99.9\% area under the probability density function curve of the Gaussian distribution is between $-t_{intercept}$ and $+t_{intercept}$. Also, we want $t_{sample}$ to ``follow'' the direction vector (i.e. we only interested in line segment between the black and the blue dots on $d$ in Figure \ref{figure:sampling from skew Gau - step 2}), thus, we only accept the positive value of $t_{sample}$.

Figure \ref{figure:sampling from skew Gau - example} presents a two-dimensional example of using the aforementioned strategy to sample points in a hyper-sphere centred at (0,0) with a radius of 10. The points have the highest probability to be sampled at (-7,0)

\section{Experiments to Evaluate the Predictive Performance of SMOClust} \label{section:experiments}

This section presents the design of the experiments to evaluate SMOClust. The predictive performance of SMOClust was first compared against five existing approaches from the literature on artificial data streams of different types of drifts. This is to investigate for which types of drift SMOClust will be advantageous and disadvantageous, answering RQ2. SMOClust was then compared against the same set of existing approaches on real-world data streams to obtain a general idea of its performance in practical situations, answering RQ3. Massive Online Analysis (MOA) \cite{MOA} was chosen to be the experimentation platform. Section \ref{section:SMOClust-data streams} presents the details of artificial and real-world data streams used in the experiments. Section \ref{section:SMOClust-experment setup} presents the detailed setup of the experiments, including the procedure of hyper-parameter tuning and the evaluation method used in the experiments.

\subsection{Data Streams} \label{section:SMOClust-data streams}

\review{As discussed in Sections \ref{section:introduction} and \ref{section:related work}, data difficulty factors play a crucial role in class imbalanced data stream learning with concept drift. Therefore, it is important to evaluate class imbalance data stream learning approaches based on data streams with different data difficulty factors. In line with that, the  artificial data stream generator} proposed by \cite{DataDifficulty} was adopted \review{because it} allows us to simulate concept drifts that affect different data difficulty factors, including the class imbalance ratio, movement of the clusters in the minority class, and the proportion of safe, borderline and rare minority class examples. \review{We have generated a large variety of artificial data streams to avoid any bias in the evaluation and enable us to understand the conditions under which SMOClust performs well and the conditions under which it fails, as well as the reason for such behaviour.}

%\review{It is also worth noting that while both the data stream generator and SMOClust take data difficulty factors into account in data stream learning, it does not necessarily mean that the generator is biased towards SMOClust. In some cases, existing methods that were not specifically designed to handle data difficulty factors may be adequate. As demonstrated in the experiment results presented in Section \ref{section:SMOClust-analysis-artificial data streams}, this is indeed the case. Moreover, real-world data streams, which will be discussed later in this section, have also been included in the experiments to benchmark SMOClust.}

Table \ref{table:SMOClust-artificial stream details} presents a summary of artificial data streams used in the experiments. Each of them has five numerical input attributes $\{x_i \in (-1,1)\}^5_{i=1}$ and a class label $y_i \in \{0,1\}$. They all consist of 200k examples where concept drift happens gradually from 70k to 100k time steps. \reviewII{The continuous movement of minority class sub-clusters in gradual drift scenarios creates a complex and dynamic environment for evaluation.} We created thirty artificial data streams of each type with different random seeds.
% thirty different sets of random seeds. Each set of random seeds consists of three random seeds to control the behaviour of the clusters in the minority class, generating training examples and testing examples respectively.
Each of the thirty streams is used to evaluate the data stream learning approaches in a single run. The evaluation method is detailed in Section \ref{section:SMOClust-experment setup}

Following the default setting by \cite{DataDifficulty}, when the artificial data stream has no drift or no modifier specified, it is: 1) class balanced, 2) composed of a single cluster representing class 1, uniformly surrounded by the examples of class 0, and 3) examples only appear in safe regions. When the data stream is class imbalanced, class 1 is the minority class while class 0 is the majority class.

\begin{table}[!ht]
% \footnotesize
\centering
\caption{Summary of Artificial Data Streams}
\label{table:SMOClust-artificial stream details}
\renewcommand\tabcolsep{20pt}
\begin{threeparttable}
\begin{tabular}{c|c}
\hline
\rowcolor[rgb]{0.9,0.9,0.9} Imbalance Ratio Drift & Single Factor Drift with Static Imbalance Ratio \\
\hline
\makecell{StaticIm10\_Im1, \\ StaticIm1\_Im10, \\ Im1, \\ StaticIm1\_Im50} & \makecell{StaticIm\{30/10/1\}\_Split\{3/7\}, \\ StaticIm\{30/10/1\}\_Move\{3/7\}, \\ StaticIm\{30/10/1\}\_Merge\{3/7\}, \\ StaticIm\{30/10/1\}\_Borderline\{20/100\} \\ StaticIm\{30/10/1\}\_Rare\{20/100\}} \\
\hline
\hline
\rowcolor[rgb]{0.9,0.9,0.9} Double Factor Drift & Complex Factor Drift \\
\hline
\makecell{Im1+Rare100, \\ Im10+Rare60, \\ Split5+Im10, \\ Im1+Borderline100, \\ Im10+Borderline20} & \makecell{StaticIm10\_Split5+Im1+Rare100, \\ StaticIm10\_Split5+Im1+Borderline100, \\ Split5+Im10+Borderline40+Rare40, \\ Split5+Im10+Borderline80, \\ Im10+Borderline20+Rare20} \\
\hline
\end{tabular}
\begin{tablenotes}
\begin{footnotesize}
\item[-] All artificial data streams have 200k examples, where a single concept drift occurs from 70k-th time step to 100k-th time step.
\item[-] ``+'' refers to the factors occurring simultaneously during the concept drift. 
\item[-] StaticIm\{$N$\} refers to a static minority class ratio of $N\%$ throughout the entire stream.
\item[-] Im\{$N$\} refers to the minority class ratio of $N\%$ after the concept drift.
\item[-] Split\{$N$\}, Move\{$N$\}, Merge\{$N$\} refer to drifts which split, move and merge $N$ clusters in the minority class respectively.
\item[-] Borderline\{$N$\}, Rare\{$N$\} refer to drifts changing $N\%$ of the minority class examples from appearing in a safe region of the clusters to being borderline region and rare cases respectively.
\end{footnotesize}
\end{tablenotes}
\end{threeparttable}
\end{table}

As shown in Table \ref{table:SMOClust-artificial stream details}, we considered four groups of drift from \cite{DataDifficulty}'s work in this study. The first group (Imbalance ratio drift) considers concept drift affecting the class imbalance ratio only. The second group (Single factor drift with static imbalance ratio) considers data streams with a static class imbalance ratio while the concept drift happens in the form of five factors, which were discussed by \cite{DataDifficulty}: splitting, moving, merging clusters and decreasing the ratio of safe examples while increasing the ratio of borderline or rare examples. In the third (Double factor drift) and the fourth (Complex factor drift) groups, we have chosen ten artificial data streams (five for each group) with concept drift affecting two factors and a group of factors, respectively. These artificial data streams were chosen evenly across the lists of data streams from \cite{DataDifficulty}'s work with double factor drift and complex factor drift in \cite{DataDifficulty}'s work respectively. These lists were sorted by the average performance of the compared data stream learning approaches in their work. Thus, picking data streams evenly from these lists means that we are taking scenarios with different difficulty levels.

As the analysis which is presented in Section \ref{section:SMOClust-analysis-artificial data streams} shows that SMOClust performed well in severely imbalanced data streams, we performed additional experiments with the aforementioned single factor drift streams with more severe static class imbalance ratio to further evaluate SMOClust in extreme cases. These additional severely class imbalanced artificial data streams are summarised in Table \ref{table:SMOClust-severe imbalance artificial stream details}. Note that, although we reused the static imbalance ratio of 1\% minority class examples, we used another set of random seeds when performing these additional experiments. 

\begin{table}[!t]
\centering
\caption{Summary of Single Factor Drift Artificial Data Streams with Severe Imbalance Ratio}
\label{table:SMOClust-severe imbalance artificial stream details}
\renewcommand\tabcolsep{65pt}
\begin{threeparttable}
\begin{tabular}{c}
\hline
\rowcolor[rgb]{0.9,0.9,0.9} Single Factor Drift with Severe Static Imbalance Ratio \\
\hline
\hline
\makecell{StaticIm\{5/3/1/07/05/03\}\_Split\{3/7\}, \\ StaticIm\{5/3/1/07/05/03\}\_Move\{3/7\}, \\ StaticIm\{5/3/1/07/05/03\}\_Merge\{3/7\}, \\ StaticIm\{5/3/1/07/05/03\}\_Borderline\{20/100\}, \\ StaticIm\{5/3/1/07/05/03\}\_Rare\{20/100\} } \\
\hline
\end{tabular}
\begin{tablenotes}
\begin{footnotesize}
\item[-] All artificial data streams have 200k examples, where a single concept drift from 70k-th time step to 100k-th time step.
\item[-] StaticIm\{$N$\} refers to a static minority class ratio to be $N\%$ throughout the entire stream. StaticIm\{$0N$\} refers to a static minority class ratio to be $0.N\%$ throughout the entire stream.
\item[-] Split\{$N$\}, Move\{$N$\}, Merge\{$N$\} refer to drifts which split, move and merge $N$ clusters in the minority class respectively.
\item[-] Borderline\{$N$\}, Rare\{$N$\} refer to drifts changing $N\%$ of the minority class examples from appearing in a safe region of the clusters to being borderline region and rare cases respectively.
\end{footnotesize}
\end{tablenotes}
\end{threeparttable}
\end{table}

Apart from experiments with artificial data streams, we also performed experiments with different real-world data streams to evaluate SMOClust in practical applications. These real-world data streams are summarised in Table \ref{table:SMOClust-real-world stream details} and their details are as follows.

\begin{table}[!t]
% \footnotesize
\centering
\caption{Summary of Real-World Data Streams}
\label{table:SMOClust-real-world stream details}
\renewcommand\tabcolsep{1pt}
\begin{threeparttable}
\begin{tabular}{c|c|c|c|c|l|l}
\hline
\makecell{Stream} & \makecell{\#Examples \\ (Pre)} & \makecell{\#Examples \\ (Actual)} & \makecell{\#Nom. \\ Attr.} & \makecell{\#Num. \\ Attr.} & \multicolumn{1}{c|}{\makecell{Imbalance \\ Ratio (Pre)}} & \multicolumn{1}{c}{\makecell{Imbalance \\ Ratio (Actual)}} \\
\hline
\hline
%  Airlines & 53,938 & 485,445 & 4 & 3 & 0.646:0.354 \IRchart{0.646} & 0.544:0.456 \IRchart{0.544} \\
Luxembourg & 190 & 1711 & 15 & 16 & 0.532:0.468 \IRchart{0.532} & 0.512:0.488 \IRchart{0.512} \\
NOAA & 1,815 & 16,344 & 0 & 8 & 0.698:0.303 \IRchart{0.698} & 0.685:0.315 \IRchart{0.685} \\
Ozone & 253 & 2,281 & 0 & 72 & 0.893:0.107 \IRchart{0.893} & 0.942:0.058 \IRchart{0.942}\\
PAKDD2009 & 4999 & 44998& 13 & 14 & 0.798:0.202 \IRchart{0.798} & 0.803:0.197 \IRchart{0.803}\\
Covtype\textsubscript{(c\textsubscript{1}=\{1-6\})} & 58,101 & 522,911 & 2 & 10 & 0.785:0.215 \IRchart{0.785} & 0.619:0.381 \IRchart{0.619} \\
Covtype\textsubscript{(c\textsubscript{1}=1)} & 58,101 & 522,911 & 2 & 10 & 0.595:0.405 \IRchart{0.595} & 0.524:0.476 \IRchart{0.524} \\
Covtype\textsubscript{(c\textsubscript{1}=2)} & 58,101 & 522,911 & 2 & 10 & 0.963:0.037 \IRchart{0.963} & 0.936:0.064 \IRchart{0.936} \\
Covtype\textsubscript{(c\textsubscript{1}=3)} & 58,101 & 522,911 & 2 & 10 & 0.963:0.037 \IRchart{0.963} & 0.999:0.001 \IRchart{0.999} \\
Covtype\textsubscript{(c\textsubscript{1}=4)} & 58,101 & 522,911 & 2 & 10 & 0.958:0.042 \IRchart{0.958} & 0.987:0.014 \IRchart{0.987} \\
Covtype\textsubscript{(c\textsubscript{1}=5)} & 58,101 & 522,911 & 2 & 10 & 0.963:0.037 \IRchart{0.963} & 0.971:0.029 \IRchart{0.971} \\
Covtype\textsubscript{(c\textsubscript{1}=6)} & 58,101 & 522,911 & 2 & 10 & 0.963:0.037 \IRchart{0.963} & 0.965:0.035 \IRchart{0.965}\\
INSECTS\textsuperscript{inc.} & 45,204 & 406,840 & 0 & 33 & 0.899:0.101 \IRchart{0.899} & 0.905:0.095 \IRchart{0.905} \\
INSECTS\textsubscript{abr.} & 35,527 & 319,748 & 0 & 33 & 0.912:0.088 \IRchart{0.912} & 0.907:0.093 \IRchart{0.907} \\
INSECTS\textsupersubscript{inc.}{grad.} & 14,342 & 128,981 & 0 & 33 & 0.921:0.079 \IRchart{0.921} & 0.899:0.101 \IRchart{0.899} \\
INSECTS\textsupersubscript{inc.}{abr. re.} & 45,204 & 406,840 & 0 & 33 & 0.895:0.105 \IRchart{0.895} & 0.905:0.095 \IRchart{0.905} \\
INSECTS\textsupersubscript{inc.}{re.} & 45,204 & 406,840 & 0 & 33 & 0.895:0.105 \IRchart{0.895} & 0.905:0.095 \IRchart{0.905} \\
\reviewII{Amazon} & 800 & 7,200 & 0 & 30 & 0.728:0.272 \IRchart{0.782} & 0.875:0.125 \IRchart{0.875} \\
% Elec & 4,531 & 40,781 & 1 & 7 & 0.605:0.395 \IRchart{0.605} & 0.572:0.428 \IRchart{0.572} \\
\reviewII{Twitter} & 909 & 8,181 & 0 & 30 & 0.814:0.186 \IRchart{0.814} & 0.846:0.154 \IRchart{0.846} \\
\hline
\end{tabular}
\begin{tablenotes}
\begin{footnotesize}
\item[-] Total number of attributes = \#Nominal attributes + \#Numeric attributes + Class attribute.
 \item[-] ``Pre'' refers to hyper-parameter tuning sets (i.e. the first 10\% of the original data set). ``Actual'' refers to actual experiment sets (i.e. the remaining 90\% of the original data set.
 \item[-] Covtype\textsubscript{(c\textsubscript{1}=\textit{x})}: ``c\textsubscript{1}=$x$'' refers to the class 1 is the class $x$ in the original data set while the rest of the classes are combined to be the class 0 in the ``Actual'' experiment stream. ``c\textsubscript{1}=\{$x_0-x_n$\}'' refers to the class 1 is the class $x_0$-$x_n$ in the original data set combined while the rest of the classes are combined to be the class 0 in the ``Actual'' experiment stream.
 \item[-] For all INSECTS data streams, ``ae-albopictus'' is the class 1. ``inc.'' refers to incremental, ``abr.'' refers to abrupt, ``grad'' refers to gradual, and ``re.'' refers to recurring.
\end{footnotesize}
\end{tablenotes}
\end{threeparttable}
\end{table}

The Luxembourg stream \cite{Luxembourg} was constructed from the European Social Survey from 2002 to 2007. The classification task is to predict whether internet usage is high or low. The NOAA stream \cite{Learn++NSE/NOAA} contains weather records collected over five decades (1949-1999). These records include temperature, pressure, wind speed, precipitation and other weather-related events. The classification task is to predict whether the next day will rain. The Ozone stream \cite{Ozone} consists of air measurements collected from 1998 to 2004. The task is to predict the ozone level eight hours ahead of time. The PAKDD2009 stream \cite{PAKDD2009} consists of private label credit card application records and the task is to decide whether a given application should be approved. Forest Covertype (Covtype) stream \cite{Covtype} contains the cartographic information about the forest of 30 $\times$ 30-meter cells and the task is to predict the cover type for a given cell. Covtype stream originally is a multi-class classification problem with seven forest cover types. To make it suitable for this study, it has been converted into seven binary classification streams. Each of them takes one of the forest cover types as one class while combining other forest cover types to be the other class. INSECTS streams \cite{Real-World/INSECTS} were constructed using a smart trap with optical sensors to collect the flying data of three different species of insects in a non-stationary environment for around three months. The temperature of the data collection environment was controlled to simulate concept drifts. INSECTS streams originally have six classes: three species of mosquitoes with two genders. We converted them into binary classification tasks by combining classes belonging to the species of ae-albopictus as the minority class while combining the rest of the classes as the majority class. Also, it has to note that \cite{Real-World/INSECTS} originally proposed seven INSECTS streams but we only adopted six of them which contain concept drifts and left the INSECT-out-of-control stream unused as it does not contain any concept drift. \reviewII{The Amazon stream \cite{Amazon} comprises reviews of books, DVDs, electronics, and kitchen appliances. Reviews with a rating greater than 3 were labelled as positive. The objective is to discern whether a review has a rating above 3. The Twitter stream \cite{Twitter} consists of labelled tweets about popular topics. The goal is to predict whether the sentiment of a given tweet is positive or negative.}

To facilitate analysing the predictive performance of SMOClust, we also analysed the characteristics of the minority class of the real-world data streams, including the potential number of clusters, and the ratios of safe, borderline, rare and outlier examples. Note that we only analysed the portion of the real-world data streams used in the actual experiments, which excludes the first 10\% of each original real-world data stream that was used for hyper-parameter tuning (see Section \ref{section:SMOClust-experment setup} for details the hyper-parameter tuning procedure). The procedure of this analysis follows the methodology proposed by \cite{DataDifficulty} and is described as follows.

The characteristics of each real-world data stream are estimated in successive batched of examples. We followed \cite{DataDifficulty} to use a batch size of 2000 examples for all data streams except for \reviewII{Luxembourg, NOAA, Amazon, and Twitter}, where a batch size of 200 was used as these data streams have less than 10,000 examples. The class imbalance ratio and the ratios of each minority class type are estimated for each batch. It is worth noting that we only focused on analysing the class 1 because it is the global minority class of all the real-world data streams (see Table \ref{table:SMOClust-real-world stream details}), even though this class could potentially become a majority during certain periods of the data stream, e.g., when there is potential concept drift affecting $P(Y)$, changing the roles of majority and minority classes temporarily. As for types of minority class examples, they were estimated using the method proposed by \cite{TypesOfMinorityClass}. This method first finds the $k$-Nearest neighbours of each minority class example. Based on the class ratios among these $k$-Nearest neighbours, it then categorises each minority class example as safe, borderline, rare, or outlier. Here, we followed \cite{TypesOfMinorityClass} to adopt $k=5$.

Following \cite{DataDifficulty}'s procedure, we also estimated the number of minority class clusters in each batch, using the affinity propagation algorithm \cite{AffinityPropagation} and removing clusters with less than six minority class examples \cite{DataDifficulty}. The affinity propagation algorithm was run thirty times with different random seeds for each batch. The average estimated number of minority class clusters is then recorded. 

Lastly, we reported the ranges of the aforementioned characteristics across the different batched and their medians in Table \ref{table:SMOClust-real world stream characteristics}. Note that we only performed analysis about types of minority class examples and the potential number of clusters on batched that contain at least six ($k+1$) minority class (class 1) examples. This is to prevent always categorising the minority class examples as rare cases or outliers when the total number of minority class examples in the batch is extremely low. The number of batches with less than size minority class examples is reported in brackets in the third column of Table \ref{table:SMOClust-real world stream characteristics}.

% \caption{Characteristics of Real-World Data Streams Used in Experiments. (Values in the brackets are the median)}
\afterpage{
\begin{landscape}
\begin{table}[!h]
\footnotesize
\centering
\caption{Characteristics of Real-World Data Streams (Values in the brackets are the median)}
\label{table:SMOClust-real world stream characteristics}
\renewcommand\tabcolsep{4pt}
\begin{threeparttable}
\begin{tabular}{c|c|c|c|c|c|c|c|c}
\hline
\makecell{Stream} & \#Examples & \makecell{\#Batches \\ (Uncounted)} & \makecell{Est. \#Clust. \\ (Median)} & \makecell{Minority \\ Ratio} & Safe & Borderline & Rare & Outlier \\
\hline
\hline
Luxembourg & 1711 & 9 (0) & 4-8 (6) & 46\%-53\% (48\%) & 47\%-62\% (51\%) & 32\%-48\% (42\%) & 3\%-9\% (4\%) & 0\%-3\% (1\%) \\
NOAA & 16344 & 9 (0) & 9-33 (28) & 25\%-37\% (31\%) & 29\%-45\% (33\%) & 33\%-50\% (43\%) & 11\%-18\% (15\%) & 6\%-13\% (9\%) \\
Ozone & 2281 & 12 (5) & 0-2 (0) & 0\%-18\% (4\%) & 0\%-22\% (0\%) & 0\%-61\% (17\%) & 14\%-58\% (30\%) & 5\%-70\% (33\%) \\
PAKDD2009 & 44998 & 23 (0) & 19-32 (28) & 17\%-22\% (20\%) & 0\%-4\% (2\%) & 27\%-38\% (32\%) & 33\%-41\% (36\%) & 22\%-34\% (30\%) \\
Covtype\textsubscript{(c\textsubscript{1}=\{1-6\})} & 522911 & 262 (8) & 1-26 (1) & 0\%-76\% (36\%) & 66\%-99\% (92\%) & 0\%-26\% (6\%) & 0\%-8\% (1\%) & 0\%-5\% (1\%) \\
Covtype\textsubscript{(c\textsubscript{1}=1)} & 522911 & 262 (2) & 1-36 (1) & 0\%-91\% (46\%) & 69\%-100\% (93\%) & 0\%-25\% (5\%) & 0\%-5\% (1\%) & 0\%-4\% (0\%) \\
Covtype\textsubscript{(c\textsubscript{1}=2)} & 522911 & 262 (149) & 0-33 (5) & 0\%-89\% (0\%) & 0\%-100\% (90\%) & 0\%-78\% (7\%) & 0\%-14\% (2\%) & 0\%-11\% (1\%) \\
Covtype\textsubscript{(c\textsubscript{1}=3)} & 522911 & 262 (240) & 0-6 (1) & 0\%-3\% (0\%) & 0\%-62\% (44\%) & 0\%-67\% (19\%) & 0\%-38\% (16\%) & 9\%-67\% (20\%) \\
Covtype\textsubscript{(c\textsubscript{1}=4)} & 522911 & 262 (108) & 0-9 (3) & 0\%-10\% (1\%) & 0\%-100\% (68\%) & 0\%-79\% (23\%) & 0\%-56\% (6\%) & 0\%-50\% (2\%) \\
Covtype\textsubscript{(c\textsubscript{1}=5)} & 522911 & 262 (159) & 0-14 (3) & 0\%-28\% (0\%) & 0\%-99\% (81\%) & 0\%-61\% (13\%) & 0\%-38\% (3\%) & 0\%-29\% (2\%) \\
Covtype\textsubscript{(c\textsubscript{1}=6)} & 522911 & 262 (110) & 0-17 (2) & 0\%-21\% (1\%) & 0\%-100\% (91\%) & 0\%-93\% (6\%) & 0\%-33\% (1\%) & 0\%-22\% (1\%) \\
INSECTS\textsuperscript{inc.} & 406840 & 204 (1) & 0-27 (15) & 0\%-19\% (9\%) & 0\%-24\% (5\%) & 0\%-46\% (32\%) & 0\%-42\% (29\%) & 14\%-100\% (33\%) \\
INSECTS\textsubscript{abr.} & 319748 & 160 (0) & 0-49 (13) & 0\%-39\% (9\%) & 0\%-50\% (5\%) & 0\%-53\% (32\%) & 0\%-46\% (27\%) & 4\%-100\% (32\%) \\
INSECTS\textsupersubscript{inc.}{grad.} & 128981 & 65 (0) & 7-27 (16) & 5\%-19\% (10\%) & 0\%-43\% (8\%) & 17\%-46\% (33\%) & 5\%-42\% (27\%) & 8\%-47\% (28\%) \\
INSECTS\textsupersubscript{inc.}{abr. re.} & 406840 & 204 (1) & 0-26 (15) & 0\%-18\% (10\%) & 0\%-29\% (5\%) & 0\%-57\% (32\%) & 7\%-41\% (28\%) & 6\%-93\% (32\%) \\
INSECTS\textsupersubscript{inc.}{re.} & 406840 & 204 (1) & 0-26 (15) & 0\%-17\% (10\%) & 0\%-27\% (4\%) & 0\%-58\% (32\%) & 0\%-41\% (28\%) & 5\%-100\% (32\%) \\
\reviewII{Amazon} & 7200 & 36 (20) & 1-5 (4) & 0\%-34\% (0\%) & 0\%-13\% (4\%) & 27\%-55\% (40\%) & 25\%-46\% (36\%) & 10\%-38\% (17\%) \\
% Elec & 40781 & 21 (0) & 1-24 (17) & 32\%-52\% (42\%) & 44\%-74\% (64\%) & 19\%-40\% (27\%) & 3\%-12\% (7\%) & 2\%-5\% (3\%) \\
\reviewII{Twitter} & 8181 & 41 (0) & 0-7 (2) & 5\%-41\% (13\%) & 0\%-22\% (0\%) & 0\%-52\% (20\%) & 0\%-59\% (31\%) & 9\%-90\% (41\%) \\
\hline
\end{tabular}
\begin{tablenotes}
\begin{footnotesize}
\item[-] Covtype\textsubscript{(c\textsubscript{1}=\textit{x})}: ``c\textsubscript{1}=$x$'' refers to the class 1 is the class $x$ in the original data set while the rest of the classes are combined to be the class 0 in the ``Actual'' experiment stream. ``c\textsubscript{1}=\{$x_0$-$x_n$\}'' refers to the class 1 is the class $x_0$-$x_n$ in the original data set combined while the rest of the classes are combined to be the class 0 in the ``Actual'' experiment stream.
\item[-] For all INSECTS data streams, ``ae-albopictus'' is the class 1. ``inc.'' refers to incremental, ``abr.'' refers to abrupt, ``grad'' refers to gradual, and ``re.'' refers to recurring.
\end{footnotesize}
\end{tablenotes}
\end{threeparttable}
\end{table}
\end{landscape}
}

As shown in Table \ref{table:SMOClust-real world stream characteristics}, PAKDD2009 and NOAA streams usually present the most number of clusters of minority class examples, with medians of twenty-eight clusters, meaning that the minority class is split into several clusters in this data stream. INSECTS streams usually present fewer clusters of the minority class than PAKDD2009 and NOAA streams, which have medians ranging from thirteen to sixteen clusters. Luxembourg, Ozone\reviewII{, Covtype, Amazon and Twitter streams} usually present the least number of clusters of the minority class, having medians ranging from zero to six.

As for the types of minority class examples, Table \ref{table:SMOClust-real world stream characteristics} shows that the Ozone, PAKDD2009\reviewII{, INSECTS, Amazon, and Twitter} streams mainly consist of borderline, rare, and outlier minority class examples. Luxembourg and NOAA streams mainly consist of safe and borderline minority class examples. Most Covtype streams mainly consist of safe minority class examples. Regarding the minority ratios, most of them have a small range, indicating that the potential concept drifts only affect $P(Y)$ with mild severity. In contrast, Covtype\textsubscript{(c\textsubscript{1}=\{1-6\})}, Covtype\textsubscript{(c\textsubscript{1}=1)} and Covtype\textsubscript{(c\textsubscript{1}=2)} streams have a very large range, indicating that that they potentially present severe concept drifts affecting $P(Y)$. In particular, Covtype\textsubscript{(c\textsubscript{1}=2)} presents a large range of minority class ratio with a very small median (1\%). This may indicate that the severe concept drifts affecting $P(Y)$ could potentially be abrupt.

\subsection{Experiment Setup} \label{section:SMOClust-experment setup}

This section presents the procedure of hyper-parameter tuning and experiments. The following are the approaches from the literature that were considered in this study and the reason behind the choice. All of these approaches are strict online approaches, which do not require storage of any past data, so that the comparisons are fair.

\begin{itemize}
    \item OOB\textsubscript{(d)} and UOB\textsubscript{(d)} \cite{OOB-UOB}: Baseline approaches that use simple oversampling or undersampling to deal with class imbalance in data stream learning.
    \item OnlineUnderOverBagging\textsubscript{(d)} (oUnderOverB\textsubscript{(d)}) \cite{OnlineBagging_Boosting_CIL}: A simple existing approach which combines simple undersampling and oversampling for class imbalance data stream learning. \reviewII{We slightly modified it to use time decay class sizes with the ``oversampling'' equation from OOB to controlling the resampling rate. We chose to adopt the ``oversampling" equation from OOB because the research paper \cite{OnlineBagging_Boosting_CIL} explicitly states that the resampling rate for OnlineUnderOverBagging should be greater than 1. On the other hand, the ``undersampling" equation from UOB produces a fractional number, which is not suitable in this context.}
    \item \reviewII{VFC-SMOTE \cite{VFC_SMOTE}: An existing approach which addresses class imbalance by generating synthetic minority class examples using histogram-based summaries of past examples.}
    \item \reviewII{SMOTE-OB \cite{SMOTE_OB}: An existing approach which incorporates the class imbalance adaptation strategy of VFC-SMOTE into OnlineUnderOverBagging\cite{OnlineBagging_Boosting_CIL}.}
    \item OnlineOversampling\textsubscript{(d)} (oOS\textsubscript{(d)}): A variant of the proposed approach which always uses the most recently seen minority class example for oversampling. This approach is used as a baseline to support the investigation of when the proposed strategy of creating synthetic minority class examples for oversampling is advantageous / disadvantageous.
    \item SMOGauNoise: A variant of the proposed approach inspired by \cite{GauNoise}, which proposed a data augmentation method for software effort estimation. SMOGauNoise has the same learning and making prediction strategies as the proposed approach but it always creates synthetic minority class examples for oversampling by adding Gaussian noise to the most recent minority class example. Note that this is the first time to investigate \cite{GauNoise}'s data augmentation method in the context of classification problems.
\end{itemize}

Approaches followed by ``(d)'' refers to these approaches that were not designed to handle concept drift originally\footnote{Except OOB and UOB can handle concept drift affecting $P(Y)$.}. We used a wrapper to enable them to use a concept drift detector. Their system reset upon concept drift detection.
% Besides, OnlineSMOTEBagging \cite{OnlineBagging_Boosting_CIL} was also included in our preliminary experiments. However, it consumed too much memory and took over fourteen hours to complete a single run of hyper-parameter tuning. Thus, we decided to exclude it from our comparison because of such run-time and memory consumption are not feasible in real-world application, regardless its predictive performance.

For the evaluation method, we modified the periodic holdout test for the experiments with artificial data streams. This modified periodic holdout test takes the data difficulty factors into the consideration, which includes the position of the minority class clusters, class imbalance ratio, and the proportions of borderline and rare examples. During a single run, the data stream learning approach was tested on a holdout test set $B^{test}_t$ of $m$ examples after training on every $n$ example. Its predictive performance in G-Mean was then recorded. The holdout test sets are class balanced and they follows the same underlying joint probability distribution (concept) at the evaluation time step $t$, where $t \mod n = 0$, i.e., $B^{test}_t \sim P_t(X,Y)$. At the end of the run, we summarised their performance across the stream by taking an average of their G-Mean performance on the test sets.

%%% Mention the options for the hyper-parameters here. We should mention the chosen values as well. But I think it maybe okay to not posting everything in the form of table as in the thesis.

For hyper-parameter tuning purposes, an additional artificial data stream was created. It also consists of 200k examples where the concept drift happens from 70k to 100k time steps but the class imbalance ratio and the drift behaviour were randomly selected from the set of all combinations of drift factors used in \cite{DataDifficulty}. We denote this data stream as the ``hyper-parameter tuning stream''. The set of hyper-parameter values of each approach that leads to the best ten runs average of G-Mean across this stream was then used in the experiments. In the experiments, we adopted thirty runs rather than ten runs to reduce the effect of randomness on the results.

Experiments with real-world data streams have a similar procedure. The first 10\% or each real-world data stream was used for the hyper-parameter tuning purposes. The prequential evaluation was used because the underlying concepts are unknown in advance. The set of hyper-parameter values of each approach that leads to the best ten runs average of G-Mean across the first 10\% of each real-world data stream was then chosen to be used in the experiment of the corresponding data stream which consists of the remaining 90\% of examples. The time decayed G-Mean performance was sampled at every 500 time steps, \reviewII{except they were sampled at every fifty time steps for NOAA, Ozone, Amazon, Twitter streams and every ten time steps for Luxembourg stream} \review{due to the fact that these streams are a lot shorter than other streams (i.e., they have a lot fewer examples than other data streams). Thus, sampling at shorter intervals allows us to see how the performance of the approaches changes throughout these relatively short data streams.} We adopted a time decay factor of 0.999 to make their past predictive performance less important to the current time step. We recorded their thirty runs average G-Mean performance across each stream for evaluation and comparative analysis.

At the end of the experiments, the predictive performance of the approaches was compared by different concept drift data difficulty factors. The corresponding rankings in the groups were then presented. Friedman test with a level of significance of 0.05 was applied to each group, confirming if there is any statistical significance between the predictive performance of different approaches. If there is, Nememyi post-hoc test was used to determine which approaches performed significantly different from the top-ranked approach. In the statistical tests, each group corresponds to a data stream learning approach while each observation within a group corresponds to the average predictive performance across a given data stream in a single run. The thirty runs average predictive performance of the approaches are also reported to facilitate us in analysing the margin of the performance difference. 
% A12 effect size \cite{A12} is used to see how likely the proposed approach performed better / worse than the existing approaches with such a margin of the performance difference.

\subsection{Results with Artificial Data Streams} \label{section:SMOClust-analysis-artificial data streams}

This section presents the analysis done to compare the predictive performance of SMOClust against existing approaches on artificial data streams which consider different drift difficulties in the minority class.
% This analysis allows us to understand when SMOClust performed better and worse than existing approaches.
General comparisons are first given based on the Friedman rankings of average G-Mean of the approaches grouped by different drift difficulty factors, presented in Table \ref{table:SMOClust-Friedman Ranks-GMean-artificial}. It is then followed by a detailed analysis of the behaviour of SMOClust in representative cases where it performed better and worse than existing approaches in Sections \ref{section:SMOClust-analysis-artificial data streams-better} and \ref{section:SMOClust-analysis-artificial data streams-worse} respectively.

% \caption[]{Statistical (Friedman) Ranking of G-Mean on Artificial Streams Grouped by Factors}
\begin{table}[!ht]
\footnotesize
\centering
\caption[]{Statistical (Friedman) Ranking of G-Mean on Artificial Streams Grouped by Factors}
\label{table:SMOClust-Friedman Ranks-GMean-artificial}
\renewcommand\tabcolsep{1.1pt}
\begin{threeparttable}
\begin{tabular}{c|ccccccccccc|c}
\hline
Groups & OOB & UOB & oOS & \makecell{oUnder- \\ OverB} & OOB\textsubscript{d} & UOB\textsubscript{d} & oOS\textsubscript{d} & \makecell{oUnder- \\ OverB\textsubscript{d}} & \makecell{SMO- \\ Gau- \\ Noise} & \makecell{\reviewII{VFC-} \\ \reviewII{SMO-} \\ \reviewII{TE}} & \makecell{\reviewII{SMO-} \\ \reviewII{TE-} \\ \reviewII{OB}} & \makecell{SMO- \\ Clust} \\
\hline
\hline
\makecell{Imbalance \\ Ratio \\ Drift} & \cellcolor{lime} 3.77 & \cellcolor{lime}\underline{4.44} & \cellcolor{lime}\underline{4.72} & \cellcolor{lime} 3.27 & 7.35 & 7.59 & 8.37 & \underline{5.05} & \underline{6.93} & 11.65 & 9.1 & \underline{5.76} \\
\makecell{Double \\ Factor} & \underline{6.03} & 7.24 & \underline{5.71} & \underline{6.85} & 4.2 & 7.39 & \underline{6.05} & \underline{5.47} & \cellcolor{lime} 2.67 & 11.38 & 9.21 & \underline{5.80} \\
\makecell{Complex \\ Factor} & \underline{5.16} & 8.2 & \underline{5.27} & \underline{6.4} & \cellcolor{lime} 4.27 & \underline{6.09} & \underline{6.88} & \underline{4.99} & \cellcolor{lime} 2.94 & 11.71 & 10.4 & \underline{5.68} \\
\hline
\multicolumn{13}{c}{\cellcolor[rgb]{0.95,0.95,0.95} Single Factor Drift with Static Imbalance Ratio} \\
\hline
\makecell{StaticIm\{*\}\tnote{a} \\ \_Split} & 7.04 & 6.93 & 7.69 & \underline{5.63} & \underline{4.96} & 6.93 & \underline{6.09} & \cellcolor{lime} 3.72 & \cellcolor{lime} 2.97 & 11.6 & 9.48 & \underline{4.97} \\
\makecell{StaticIm\{*\}\tnote{a} \\ \_Move} & \underline{5.13} & 6.97 & \underline{4.75} & \underline{4.65} & \underline{5.93} & 8.04 & \underline{6.66} & \underline{5.34} & \cellcolor{lime} 3.19 & 11.84 & 9.96 & \underline{5.53} \\
\makecell{StaticIm\{*\}\tnote{a} \\ \_Merge} & \underline{5.40} & \underline{6.84} & 4.70 & \cellcolor{lime} 3.84 & \underline{6.38} & 8.01 & \underline{6.60} & 4.74 & \cellcolor{lime} 3.30 & 11.77 & 10.01 & \underline{6.41} \\
\makecell{StaticIm\{*\}\tnote{a} \\ \_Borderline} & \underline{5.89} & \underline{6.17} & \underline{5.91} & \cellcolor{lime} 4.26 & \underline{5.46} & \underline{7.65} & 7.91 & \cellcolor{lime} 4.08 & \cellcolor{lime} 3.47 & 11.61 & 9.09 & \underline{6.51} \\
\makecell{StaticIm\{*\}\tnote{a} \\ \_Rare} & \cellcolor{lime} 2.52 & \underline{6.98} & \cellcolor{lime} 3.04 & 4.34 & 6.6 & \underline{7.56} & \underline{8.58} & 5.72 & 4.06 & 11.3 & \underline{9.26} & \underline{8.04} \\
StaticIm30\_\{*\}\tnote{b} & 5.84 & 9.02 & 3.40 & \underline{6.94} & 3.81 & \underline{7.93} & 3.75 & 5.54 & \cellcolor{lime} 2.19 & 11.27 & 11.12 & \underline{7.20} \\
StaticIm10\_\{*\}\tnote{b} & 4.89 & \underline{8.98} & 5.03 & 3.87 & 4.14 & \underline{9.17} & 7.49 & 3.56 & \cellcolor{lime} 2.23 & 11.91 & 7.77 & \underline{8.96} \\
StaticIm1\_\{*\}\tnote{b} & 4.86 & \cellcolor{lime}\underline{2.34} & 7.23 & \cellcolor{lime}\underline{2.83} & 9.64 & 5.81 & 10.27 & 5.06 & 5.78 & 11.69 & 9.78 & \cellcolor{lime}\underline{2.72} \\
\hline
\hline
All & 5.16 & 6.78 & 5.23 & 4.90 & 5.63 & 7.43 & 7.12 & 4.87 & \cellcolor{lime} 3.58 & 11.61 & 9.57 & \underline{6.12} \\
\hline
\end{tabular}
\begin{tablenotes}
\begin{footnotesize}
\item[a] ``StaticIm\{*\}\tnote{a}'' refers to StaticIm\{30/10/1\}, which means the group includes all artificial data streams of that type in static minority class ratio of 30\%, 10\%, and 1\% respectively.
\item[b] ``\{*\}'' refers to {Split/Move/Merge/Borderline/Rare}, which means the group includes all artificial data streams of the above five types of drifts with the same static minority class ratio.
\item[-] Smaller values for the rankings are better values.
\item[-] The p-values of Friedman tests are all $\leq$2.2E-16.
\item[-] Highlighted ranks denote significant superior performance.
\item[-] Underlined ranks denote the corresponding approach's performance have no statistical significance with SMOClust.
\end{footnotesize}
\end{tablenotes}
\end{threeparttable}
\end{table}

Table \ref{table:SMOClust-Friedman Ranks-GMean-artificial} shows that SMOClust was one of the top-ranked approaches when the data stream is extremely class imbalanced (minority class ratio: 1\%), indicating that SMOClust handled extremely class imbalanced data stream better than most existing approaches, while it performed similarly to UOB and OnlineUnderOverBagging. However, SMOClust was one of the low-ranked approaches in the group of rare cases, indicating that it could not handle rare cases very well. For other groups, although SMOClust was not one of the top-ranked approaches, it usually performed similarly to mid-ranked approaches.

\begin{figure}[p]
\centering
\includegraphics[width=\textwidth]{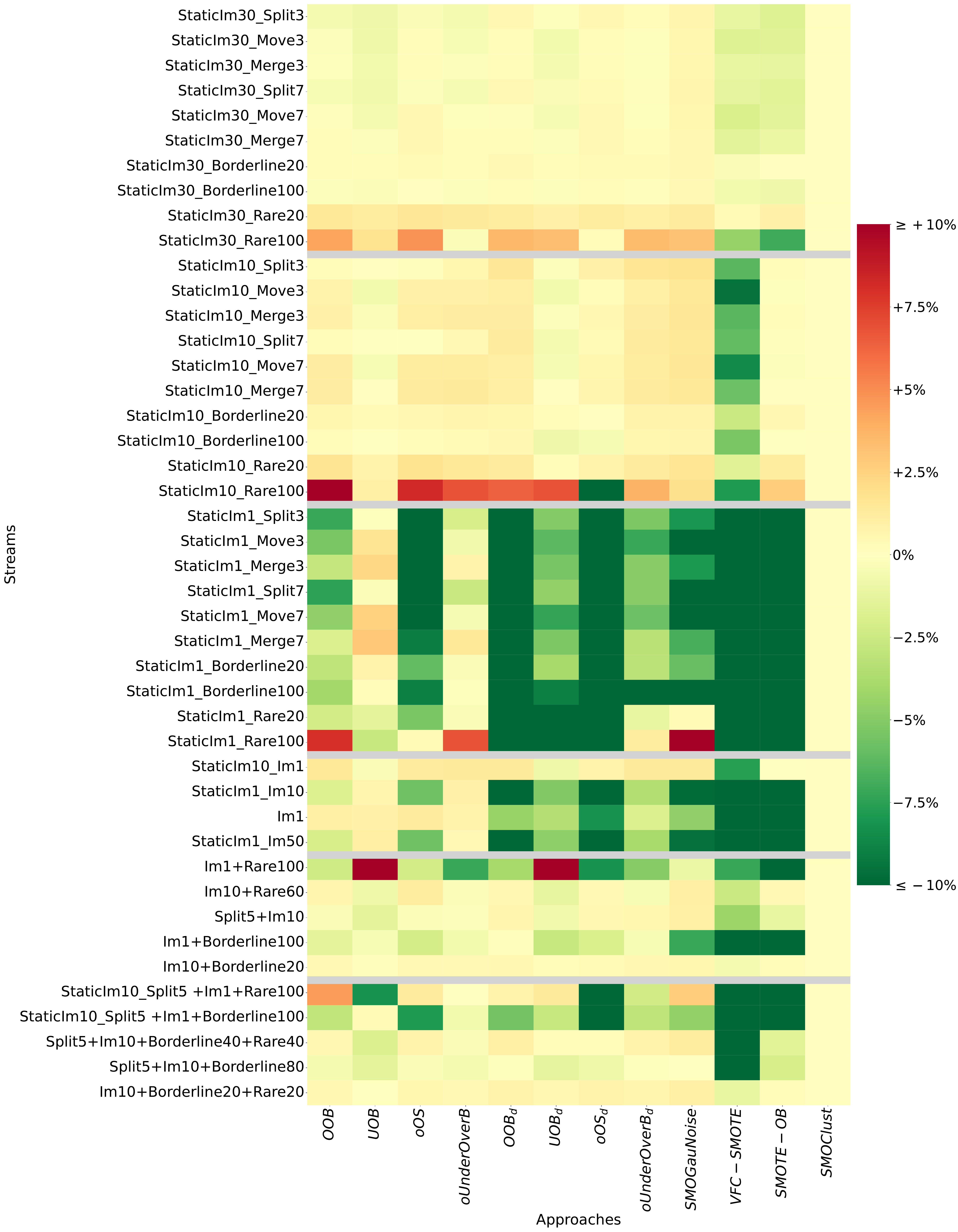}
\caption{Difference in Average G-Mean Against SMOClust on Class Imbalanced Artificial Data Streams Based on 30 Runs (Green cells indicate SMOClust performed better; Red cells indicate SMOClust performed worse; Grey horizontal lines separate different groups of data streams, i.e., StaticIm\{30/10/1\}, Imbalance Ratio Drift, Double Factor, and Complex Factor)}
\label{figure:SMOClust-Avg GMean Artificial}
\end{figure}

As Friedman rankings only show the relative position of approaches' predictive performance but they do not provide any information about the margin of difference. To investigate how much did SMOClust performed better in severely class imbalanced streams and worse in other groups of factors, we further compared their thirty runs average G-Mean on each artificial data stream. \review{The results of their difference in average G-Mean are presented in the form of a heat-map in Figure \ref{figure:SMOClust-Avg GMean Artificial}. Green cells indicate results favourable to SMOClust, whereas red cells indicate results favourable to the compared approach. For a comprehensive table of the predictive performance of the approaches, please refer to the supplementary document.}%\footnote{The supplementary document is available at: https://github.com/michaelchiucw/SMOClust}.} 
% A12 effect size \cite{A12} was adopted to compare how likely does SMOClust perform better or worse than existing approaches on each artificial data stream with the corresponding margin.

% \captionof{table}{30 Runs Average G-Mean on Artificial Data Streams (A12 SMOClust vs Others)}

Table \ref{table:SMOClust-Friedman Ranks-GMean-artificial} shows that SMOClust usually obtained lower rankings than other approaches in less severe class imbalanced data streams. However, Figure \ref{figure:SMOClust-Avg GMean Artificial} reveals that the margin of the under-performance was usually small as we can rarely see saturated red cells in the table. In contrast, the high ranking achieved by SMOClust in the group of StaticIm1\_\{*\} was supported by a lot of saturated green cells in the sector StaticIm1 of Figure \ref{figure:SMOClust-Avg GMean Artificial}, meaning that SMOClust performed a lot better than existing approaches in cases with severe class imbalanced ratio. Besides, Figure \ref{figure:SMOClust-Avg GMean Artificial} further confirms that SMOClust could not handle rare minority class examples very well as we can see that cases involving Rare100 drift have lots of saturated red cells. In particular, OOB and OnlineUnderOverBagging handled rare minority class examples better than SMOClust.

\review{One potential reason why SMOClust did not perform well in handling data streams with a large proportion of rare minority class examples is the conservative nature of the proposed synthetic example generation method, where most synthetic examples are generated in the dense area of the minority class. To address this, it might be helpful to generate synthetic examples in a more diverse manner. However, generating synthetic examples diversely can also introduce a significant amount of noise or even create artificial concept drifts. Moreover, it can be challenging to ensure that a certain area belongs to the minority class if there are no real minority class examples in that area. The proposed method is less prone to these risks and uncertainties, while overcoming the problems of existing work, which ignore data difficulty factors and rely on caching all (minority class) examples for synthetic minority class oversampling.}

Comparing the predictive performance of SMOClust against UOB and OnlineUnderOverBagging in the group of StaticIm1\_\{*\}, Table \ref{table:SMOClust-Friedman Ranks-GMean-artificial} shows that they performed similarly. Yet, the sector of StaticIm1 in Figure \ref{figure:SMOClust-Avg GMean Artificial} reveals that SMOClust performed better than UOB by small margins (around 1-2\% G-Mean\review{, light green cell}) in cases presenting concept drift of increasing rare minority class ratio, yet, it performed worse than UOB by medium-small margins (around 3\% G-Mean\review{, light red cells}) in cases presenting concept drift of moving and merging minority class clusters. SMOClust performed better than OnlineUnderOverBagging by medium-small margins (around 2-3\% G-Mean\review{, light green cells}) in cases presenting a concept drift of splitting minority class clusters. However, it performed slightly worse than OnlineUnderOverBagging (around 1\% G-Mean\review{, light red cell}) in cases presenting concept drift of merging minority class clusters. It also performed worse than OnlineUnderOverBagging by a large margin (around 7\% G-Mean\review{, saturated red cell}) in StaticIm1\_Rare100 case. In short, SMOClust performed similarly to both UOB and OnlineUnderOverBagging in most StaticIm1 cases, except OnlineUnderOverBagging performed a lot better in StaticIm1\_Rare100 case.

\reviewII{When comparing the predictive performance of SMOClust against two approaches that also summarise past knowledge to support the generation of synthetic examples (VFC-SMOTE and SMOTE-OB), Table \ref{table:SMOClust-Friedman Ranks-GMean-artificial} and Figure \ref{figure:SMOClust-Avg GMean Artificial} show that SMOClust performed better in most cases, especially in StaticIm1 cases. This indicates that the proposed synthetic minority oversampling strategy in SMOClust is superior.}

Based on the aforementioned results, additional experiments were performed with the same set of single factor drift artificial data streams but enforced with extremely severe class imbalance ratios (minority class ratio $0.3\%$ to $5\%$, summarised in Table \ref{table:SMOClust-severe imbalance artificial stream details}) to further evaluate if SMOClust can usually perform better than existing approaches in extremely class imbalanced data streams.

\begin{figure}[p]
\centering
\includegraphics[width=\textwidth]{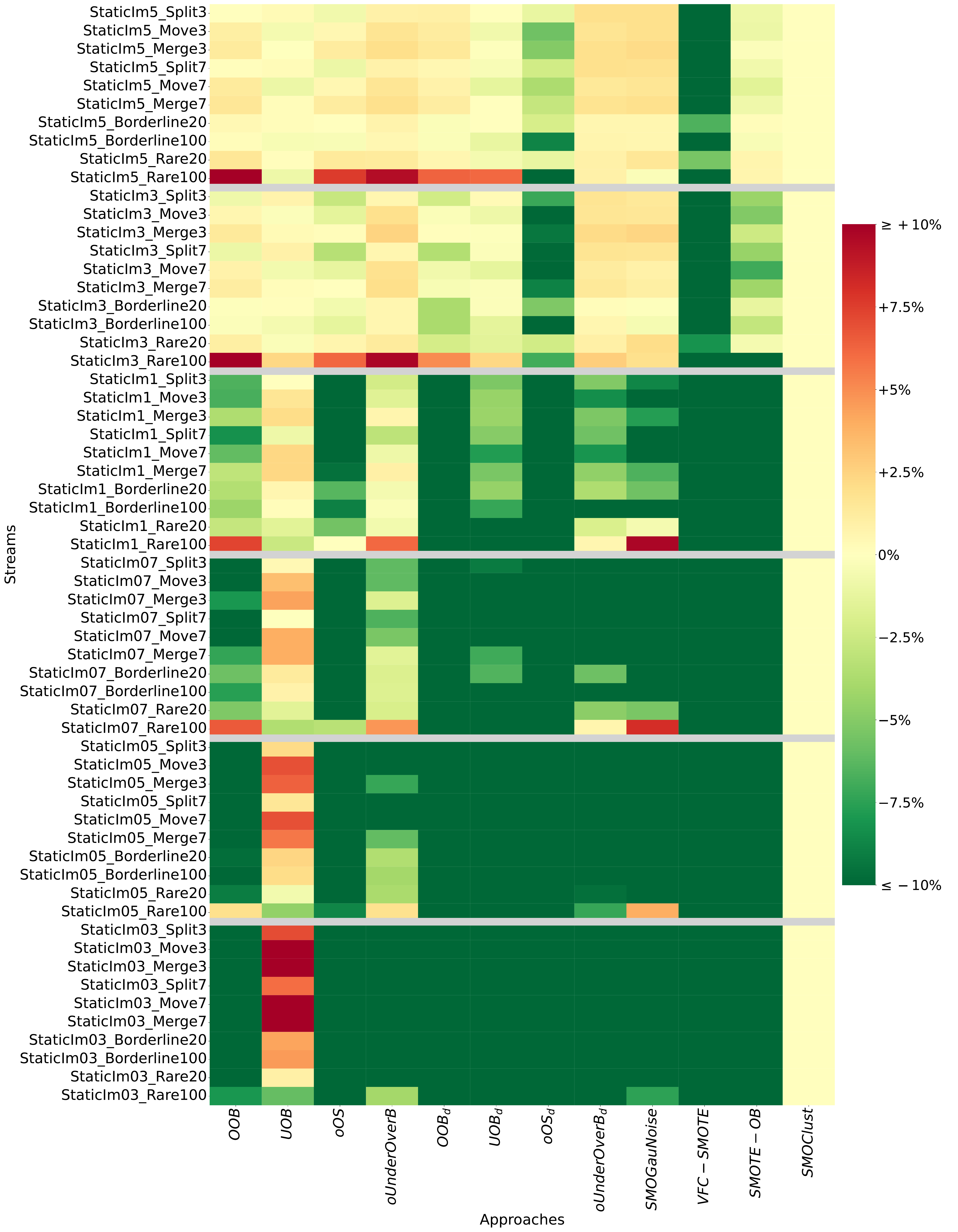}
\caption{Difference in Average G-Mean Against SMOClust on Severely Class Imbalanced Artificial Data Streams Based on 30 Runs (Green cells indicate SMOClust performed better; Red cells indicate SMOClust performed worse; Grey horizontal lines separate different groups of data streams, i.e., StaticIm\{5/3/1/07/05/03\})}
\label{figure:SMOClust-Avg GMean Artificial Severe}
\end{figure}

Table \ref{table:SMOClust-Friedman Ranks-GMean-artificial severe} presents the Friedman rankings of average G-Mean by groups of different drift difficulty factors on the severely class imbalanced artificial data streams. It shows that SMOClust can indeed achieve higher rankings when the class imbalance ratio is very severe (minority class ratio $\leq 1\%$). Figure \ref{figure:SMOClust-Avg GMean Artificial Severe} \review{presents the difference in average G-Mean (based on thirty runs) between the compared approaches and SMOClust on severely class imbalanced artificial data streams in the form of a heat-map with the same colour scheme as Figure \ref{figure:SMOClust-Avg GMean Artificial}. Similarly, please refer to the supplementary document for a comprehensive table of the predictive performance of the approaches.}
% and the A12 effect size comparison results.
It supports the aforementioned deduction with a lot of saturated green cells in the cases of minority class ratio $\leq1\%$, indicating the superior performance of SMOClust. The exception here is the comparison against UOB, with the margin of under-performance increasing as the severity of the class imbalance ratio increases by case. When compared against OnlineUnderOverBagging, SMOClust generally performed better in cases other than Rare100 drift, with the margin of superior performance increasing as the severity of the class imbalance ratio increases by case.

% \caption[]{Statistical (Friedman) Ranking of G-Mean on Severely Class Imbalanced Artificial Streams Grouped by Factors}
\begin{table}[!ht]
\footnotesize
\centering
\caption[]{Statistical (Friedman) Ranking of G-Mean on Severely Class Imbalanced Artificial Streams Grouped by Factors}
\label{table:SMOClust-Friedman Ranks-GMean-artificial severe}
\renewcommand\tabcolsep{1.1pt}
\begin{threeparttable}
\begin{tabular}{c|ccccccccccc|c}
\hline
Groups & OOB & UOB & oOS & \makecell{oUnder- \\ OverB} & OOB\textsubscript{d} & UOB\textsubscript{d} & oOS\textsubscript{d} & \makecell{oUnder- \\ OverB\textsubscript{d}} & \makecell{SMO- \\ Gau- \\ Noise} & \makecell{\reviewII{VFC-} \\ \reviewII{SMO-} \\ \reviewII{TE}} & \makecell{\reviewII{SMO-} \\ \reviewII{TE-} \\ \reviewII{OB}} & \makecell{SMO- \\ Clust} \\
\hline
\hline
\makecell{StaticIm\{*\}\tnote{a} \\ \_Split} & 5.78 & \cellcolor{lime}\underline{2.61} & 8.18 & \underline{3.62} & 8.58 & 5.30 & 9.90 & \underline{4.17} & 4.84 & 11.92 & 9.81 & \cellcolor{lime}\underline{3.30} \\
\makecell{StaticIm\{*\}\tnote{a} \\ \_Move} & 4.80 & \cellcolor{lime}\underline{3.20} & 7.53 & \cellcolor{lime}\underline{2.82} & 7.62 & 6.25 & 10.25 & 4.76 & 5.13 & 11.96 & 10.12 & \cellcolor{lime}\underline{3.57} \\
\makecell{StaticIm\{*\}\tnote{a} \\ \_Merge} & \underline{4.78} & \cellcolor{lime} 3.09 & 7.19 & \cellcolor{lime} 2.64 & 8.15 & 6.06 & 10.26 & \underline{4.66} & 5.14 & 11.80 & 10.03 & \underline{4.20} \\
\makecell{StaticIm\{*\}\tnote{a} \\ \_Borderline} & 4.68 & \cellcolor{lime}\underline{2.73} & 6.72 & \cellcolor{lime}\underline{2.63} & 9.52 & 6.39 & 10.52 & 4.92 & 6.12 & 11.32 & 9.20 & \cellcolor{lime}\underline{3.25} \\
\makecell{StaticIm\{*\}\tnote{a} \\ \_Rare} & \cellcolor{lime} 3.16 & 5.33 & 5.64 & \cellcolor{lime} 2.81 & 8.81 & 7.78 & 9.84 & \underline{5.29} & \underline{3.84} & 11.38 & 9.68 & \underline{4.44} \\
StaticIm5\_\{*\}\tnote{b} & 4.29 & \underline{7.56} & 6.44 & \cellcolor{lime} 2.53 & 5.39 & \underline{8.23} & 10.05 & \cellcolor{lime} 3.17 & \cellcolor{lime} 2.57 & 11.94 & \underline{8.25} & \underline{7.60} \\
StaticIm3\_\{*\}\tnote{b} & 4.48 & \underline{5.60} & 7.34 & \cellcolor{lime} 2.26 & 7.54 & 6.89 & 10.41 & \cellcolor{lime} 2.87 & 3.30 & 11.95 & 9.48 & \underline{5.88} \\
StaticIm1\_\{*\}\tnote{b} & 4.99 & \cellcolor{lime}\underline{2.36} & 7.32 & \cellcolor{lime}\underline{2.92} & 9.05 & 5.60 & 10.42 & 5.31 & 5.77 & 11.71 & 9.89 & \cellcolor{lime}\underline{2.65} \\
StaticIm07\_\{*\}\tnote{b} & 4.80 & \cellcolor{lime}\underline{1.93} & 7.21 & \underline{3.25} & 9.76 & 5.38 & 10.44 & 5.49 & 6.15 & 11.48 & 9.81 & \cellcolor{lime}\underline{2.29} \\
StaticIm05\_\{*\}\tnote{b} & 4.62 & \cellcolor{lime}\underline{1.61} & 6.97 & 3.27 & 10.10 & 5.88 & 10.05 & 5.76 & 6.11 & 11.36 & 10.17 & \cellcolor{lime}\underline{2.11} \\
StaticIm03\_\{*\}\tnote{b} & 4.65 & \cellcolor{lime}\underline{1.31} & 7.03 & 3.18 & 9.37 & 6.17 & 9.56 & 5.97 & 6.16 & 11.61 & 11.00 & \cellcolor{lime}\underline{1.99} \\
\hline
\hline
All & 4.64 & \underline{3.39} & 7.05 & \cellcolor{lime} 2.90 & 8.54 & 6.36 & 10.15 & 4.76 & 5.01 & 11.68 & 9.77 & \underline{3.75} \\
\hline
\end{tabular}
\begin{tablenotes}
\begin{footnotesize}
\item[a] ``StaticIm\{*\}\tnote{a}'' refers to StaticIm\{5/3/1/07/05/03\}, which means the group includes all artificial data streams of that type in static minority class ratio of 5\%, 3\%, 1\%, 0.7\%, 0.5\%, and 0.3\% respectively.
\item[b] ``\{*\}'' refers to {Split/Move/Merge/Borderline/Rare}, which means the group includes all artificial data streams of the above five types of drifts with the same static minority class ratio.
\item[-] Smaller values for the rankings are better values.
\item[-] The p-values of Friedman tests are all $\leq$2.2E-16.
\item[-] Highlighted ranks denote significant superior performance.
\item[-] Underlined ranks denote the corresponding approach's performance have no statistical significance with SMOClust.
\end{footnotesize}
\end{tablenotes}
\end{threeparttable}
\end{table}

Figure \ref{figure:SMOClust-Avg GMean Artificial Severe} also confirms that SMOClust usually does not handle rare minority class examples very well, especially when compared against OOB, OnlineUnderOverBagging and SMOGauNoise. However, an extremely severe class imbalance ratio may give advantage to SMOClust in dealing with Rare100 drift as cases involving Rare100 present less saturated red cells when the class imbalance ratio is $\leq1\%$. In particular, the case of StaticIm03\_Rare100 presents a row of saturated green cells. Anyhow, these results are consistent with previous results of experiments with less severe class imbalanced artificial data streams.

% \captionof{table}{30 Runs Average G-Mean on Severely Class Imbalanced Artificial Data Streams (A12 SMOClust vs Others)}

Besides, Table \ref{table:SMOClust-Friedman Ranks-GMean-artificial severe} also shows that SMOClust could not achieve high rankings in the groups concerning minority class ratio of 5\% and 3\%. This may due to the fact that the artificial data streams are long enough to have quite a lot of minority class examples, despite the minority class ratios were low. Therefore, the advantage of SMOClust was not manifested. The sectors of StaticIm5 and StaticIm3 on Figure \ref{figure:SMOClust-Avg GMean Artificial Severe} show that SMOClust usually performed slightly worse than most existing approaches but it performed better than OnlineOversampling\textsubscript{d}\reviewII{, VFC-SMOTE and SMOTE-OB}.

Considering all cases in Figure \ref{figure:SMOClust-Avg GMean Artificial Severe}, we can see that, when the minority class ratio decreases, SMOClust usually had a smaller margin of performance reduction than other approaches, except UOB. This shows that the aggressive nature of undersampling may be generally more advantageous than oversampling when the number of minority class examples in the data stream is extremely low. Yet, we can still see from Figure \ref{figure:SMOClust-Avg GMean Artificial Severe} that SMOClust performed better than UOB in most cases of Rare100 drift. This means that, when the minority class has extreme low number of examples and is difficult to learn, SMOClust still has more advantage than undersampling.
% In comparison, UOB performed similarly to SMOClust in cases of Rare100 drift when minority class ratios $> 1\%$. It then start to performed worse than SMOClust in cases of Rare100 when the the minority class ratio throughout the stream decreases, case by case. Meanwhile, OnlineUnderOverBagging performed better than SMOClust with a large margin in cases of Rare100 drift when the minority class ratios $> 1\%$. Yet, the margin decreases gradually as the minority class ratio throughout the stream decreases, case by case, until minority class ratio reaches 0.3\%.
One reason could be the fact that the compared approaches focus on learning the most recent decision areas of both classes, whereas SMOClust was designed to reinforce its knowledge in past minority class decision areas. This means that SMOClust is likely to have a better generalisation on the sub-areas of the minority class than existing approaches.

In the following sections, representative cases were chosen to discuss why SMOClust performed better and worse than existing approaches respectively, providing a more detailed understanding of the results.

% Besides, we also discussed that UOB tends to perform better than SMOClust in most cases as the minority class ratio throughout the stream decreases, case by case. This potentially because UOB is an aggressive approach which usually tries to learn and to perform well in the minority class but sacrifices its predictive performance in the majority class. Such strategy may be advantageous especially when the data stream only presents concept drift of minority class clusters movements and the class imbalance ratio is extremely severe. However, it is likely to sacrifice some predictive performance in the majority class because it usually considers majority class areas in between minority class areas as minority class. In other words, the improvements in the recall on the minority class may occur in higher detriment of the majority class than for the proposed approach. In the detailed analysis (Section \ref{section:SMOClust-analysis-artificial data streams-better} and \ref{section:SMOClust-analysis-artificial data streams-worse}), we further look into this characteristic of UOB by inspecting its learnt decision area at different time steps of the chosen representative cases.

% In Sections \ref{section:SMOClust-analysis-artificial data streams-better} and \ref{section:SMOClust-analysis-artificial data streams-worse}, representative cases were chosen to discuss why SMOClust performed better and worse than existing approaches respectively, providing a more detailed understanding of the results.

\subsubsection{Cases where SMOClust performed better} \label{section:SMOClust-analysis-artificial data streams-better}

This section discusses why SMOClust performed better than most other approaches in artificial data streams with severe class imbalance ratio when the class imbalance ratio is extremely severe (minority class ratio $\leq 1\%$ throughout the stream). StaticIm1-Move7 stream was chosen from Figure \ref{figure:SMOClust-Avg GMean Artificial} as the representative case to discuss the behaviour of SMOClust in details.

As mentioned in Section \ref{section:SMOClust-data streams}, the artificial data streams have five input attributes and a class label. Therefore, it is difficult to visualise the learnt decision areas of the approaches and understand their behaviour in details. Because of this, we created a version of the representative streams with two input attributes and a class label while preserving the characteristics which include the class imbalance ratio and the drift difficulty factors etc.
% The two-dimensional version of representative streams allows us to visualise the learnt decision areas of the approaches, thus, to analyse their behaviours further. 
Note that we only created a single copy of each two-dimensional representative stream, such that we can compare the data stream learning approaches with their median predictive performance in thirty runs on the same data stream. Also, the hyper-parameters of the approaches were tuned based on a separated random two-dimensional artificial data stream, following the procedure explained in Section \ref{section:SMOClust-experment setup}.

Table \ref{table:SMOClust-Average GMean 2D better} presents the their thirty runs average G-Mean on the two-dimensional version of StaticIm1\_Move7 stream. It shows that SMOClust performed the best. These results are slightly inconsistent with the results of the corresponding five dimensional stream in Figure \ref{figure:SMOClust-Avg GMean Artificial}, where SMOClust performed slightly worse than UOB but similarly to OnlineUnderOverBagging. Yet, in general, SMOClust still performed better than other approaches in both two-dimensional and five dimensional versions of StaticIm1\_Move7 stream. This may indicate that SMOClust tends to perform better in low-dimensional data stream. Anyhow, the detailed analysis presented in the following paragraphs can still explain the characteristics of SMOClust and why it performed better than most other approaches in this representative case.

\begin{table}
% \footnotesize
\centering
% \caption{30 Runs Average G-Mean on Two-Dimensional Version of Representative Artificial Data Streams where SMOClust Performed Better (A12 SMOClust vs Others)}
\caption{30 Runs Average G-Mean on Two-Dimensional Version of Representative Artificial Data Streams where SMOClust Performed Better}
\label{table:SMOClust-Average GMean 2D better}
\renewcommand\tabcolsep{3pt}
\begin{threeparttable}
\begin{tabular}{c|cccccc}
\hline
Stream & OOB & UOB & oOS & \makecell{oUnder- \\ OverB} & OOB\textsubscript{d} & UOB\textsubscript{d} \\
\hline
\hline
% \multicolumn{11}{c}{\cellcolor[hsb]{0.95,0.95,0.95} Set 3'} \\
% \hline
StaticIm1\_Move7 & \cellcolor[hsb]{0.25,0.912,1} 82.11\% & \cellcolor[hsb]{0.25,1.0,1} 76.3\% & \cellcolor[hsb]{0.25,1.0,1} 79.46\% & \cellcolor[hsb]{0.25,0.597,1} 85.26\% & \cellcolor[hsb]{0.25,1.0,1} 53.45\% & \cellcolor[hsb]{0.25,1.0,1} 56.94\% \\
% StaticIm03\_Rare100 & \cellcolor[hsb]{0.25,1.0,1} 52.61\% & \cellcolor[hsb]{0.25,0.213,1} 64.18\% & \cellcolor[hsb]{0.25,1.0,1} 51.04\% & \cellcolor[hsb]{0.25,0.879,1} 57.52\% & \cellcolor[hsb]{0.25,1.0,1} 21.21\% \\
\hline
\hline
Stream & oOS\textsubscript{d} & \makecell{oUnder- \\ OverB\textsubscript{d}} & \makecell{SMO- \\ GauNoise} & \makecell{VFC- \\ SMOTE} & \makecell{SMOTE- \\ OB} & SMOClust \\
\hline
\hline
StaticIm1\_Move7 & \cellcolor[hsb]{0.25,1.0,1} 76.88\% & \cellcolor[hsb]{0.25,1.0,1} 45.12\% & \cellcolor[hsb]{0.25,0.829,1} 82.94\% & \cellcolor[hsb]{0.25,1.0,1} 1.04\% & \cellcolor[hsb]{0.25,1.0,1} 33.09\% & 91.23\% \\
% StaticIm03\_Rare100 & \cellcolor[hsb]{0.25,1.0,1} 28.33\% & \cellcolor[hsb]{0.25,1.0,1} 41.9\% & \cellcolor[hsb]{0.25,1.0,1} 43.45\% & \cellcolor[hsb]{0.25,0.935,1} 56.96\% & 66.31\% \\
\hline
\end{tabular}
\begin{tablenotes}
\begin{footnotesize}
\item[-] Based on the average G-Mean, cells are highlighted in lime / light grey when SMOClust performed better than the corresponding approach and cells are highlighted in orange / dark grey cells when SMOClust performed worse than the corresponding approach. The colour intensity scales with the absolute difference of average G-Mean between the SMOClust and the approach of the column and the intensity reaches the maximum when such difference is $\geq 10\%$.
% \item[-] Symbols [*], [s], [m] and [b] represent insignificant, small, medium and large A12 effect size against SMOClust. Presence/absence of the sign “-” in the effect size means that the corresponding approach was worse/better than SMOClust.
\end{footnotesize}
\end{tablenotes}
\end{threeparttable}
\end{table}

\begin{figure}[!ht]
\centering
\includegraphics[width=\textwidth]{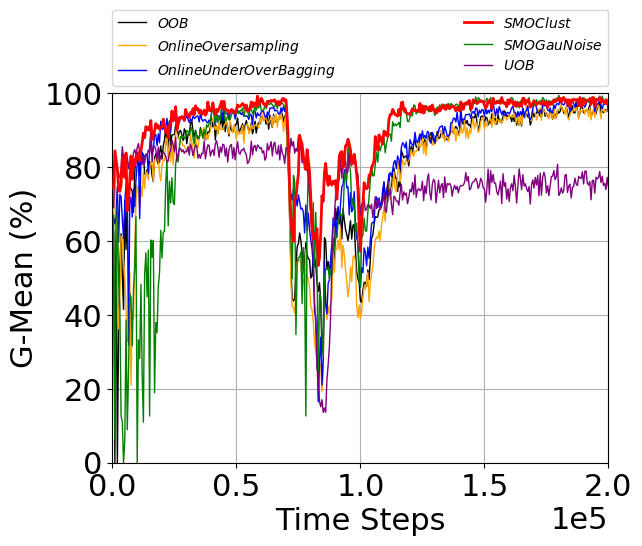}
% \vspace{-0.4cm}
\caption{Periodic Class Balanced Holdout Test G-Mean Against Time Steps in Two-Dimensional StaticIm1\_Move7}
\label{figure:StaticIm1-Move7-GMean}
\vspace{-0.7cm}
\end{figure}

Figure \ref{figure:StaticIm1-Move7-GMean} presents the approaches' predictive performance over time steps of their median run\footnote{Median run refers to the run that leads to the median of predictive performances averaged across time steps.}. \review{To maintain readability, we omitted the predictive performance of OOB\textsubscript{d}, UOB\textsubscript{d}, oOS\textsubscript{d}, oUnderOverB\textsubscript{d}\reviewII{, VFC-SMOTE, and SMOTE-OB} from Figure \ref{figure:StaticIm1-Move7-GMean}, as their performance fluctuates significantly throughout the stream. For the comparison of SMOClust against these approaches, please refer to the supplementary document.} It shows that SMOClust performed the best in most time steps. In particular, SMOClust maintained the predictive performance to have at least 50\% G-Mean on the class balanced holdout test sets during the concept drift (from 70k to 100k time steps) and recovered from the drift better than other approaches (the solid red line has a rapid recovery since 100k time steps). In contrast, other approaches usually dropped to around 0-20\% G-Mean during the drift. This case showed the superior performance achieved by SMOClust in handling severely class imbalanced drifting data streams.

\begin{figure}[!ht]
\centering
\subfigure[OOB]{\includegraphics[width=0.29\textwidth]{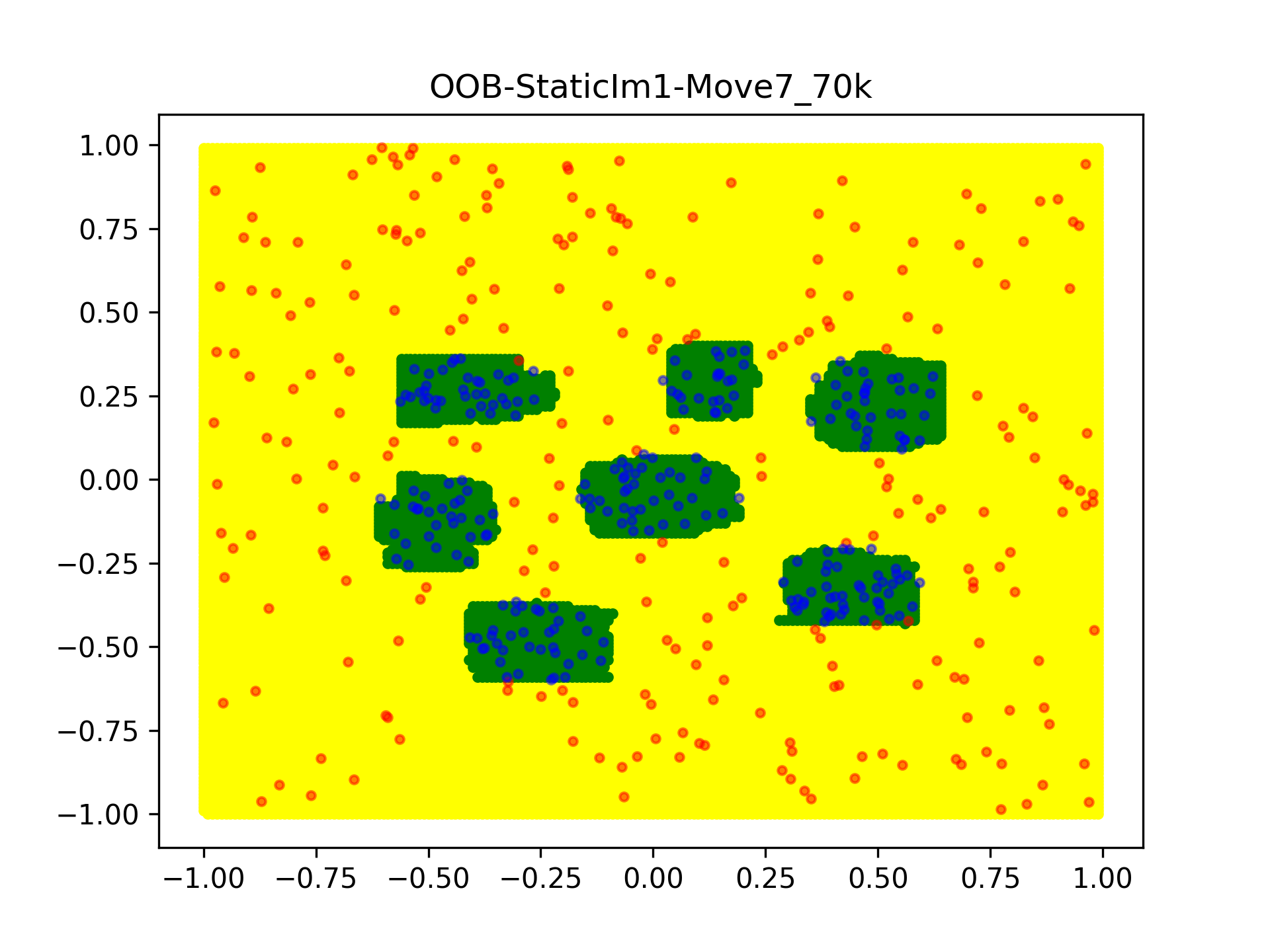} \label{figure:StaticIm1-Move7-dec_bound-OOB-70k}}
\subfigure[UOB]{\includegraphics[width=0.29\textwidth]{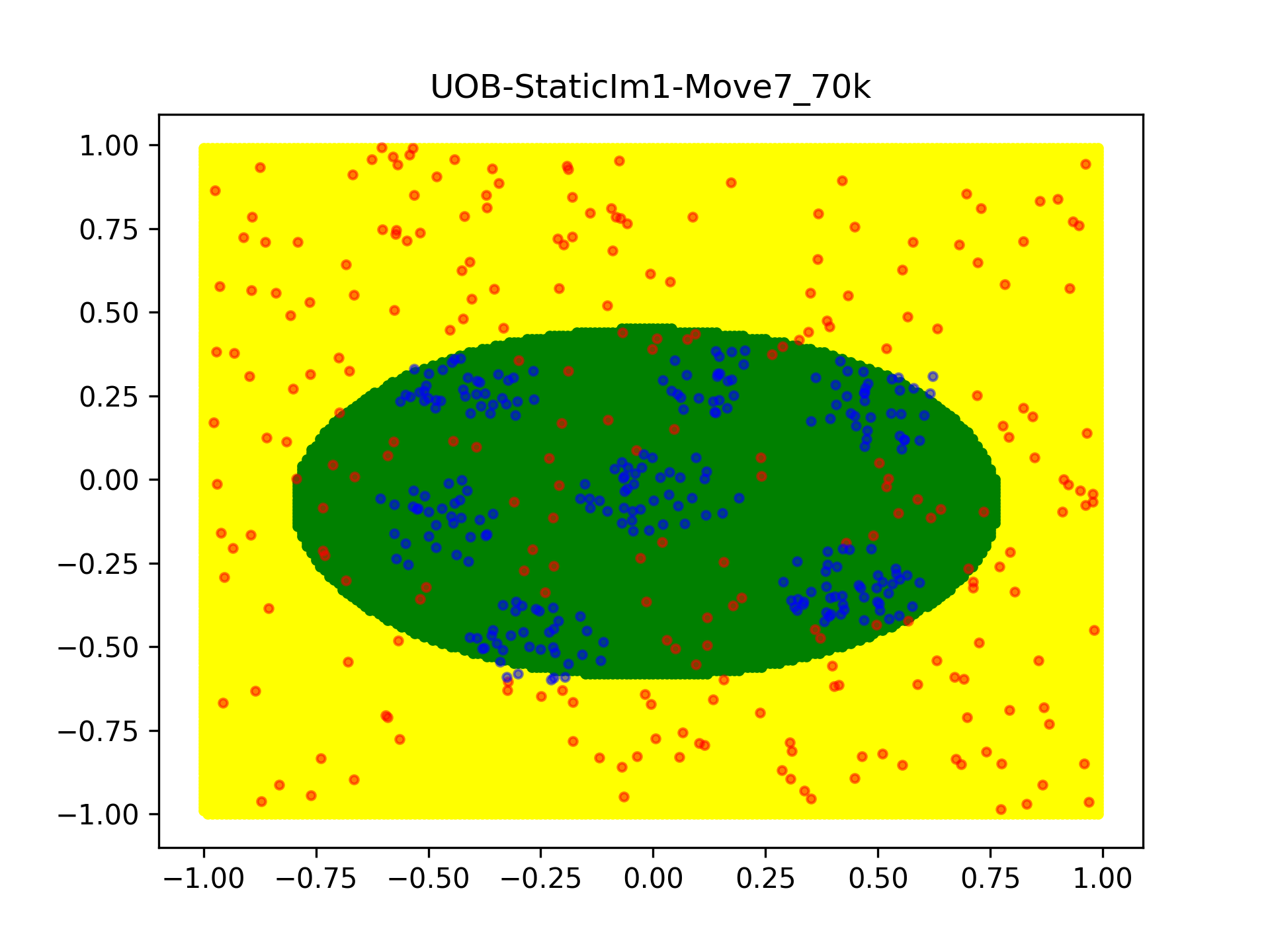} \label{figure:StaticIm1-Move7-dec_bound-UOB-70k}}
\subfigure[oOS]{\includegraphics[width=0.29\textwidth]{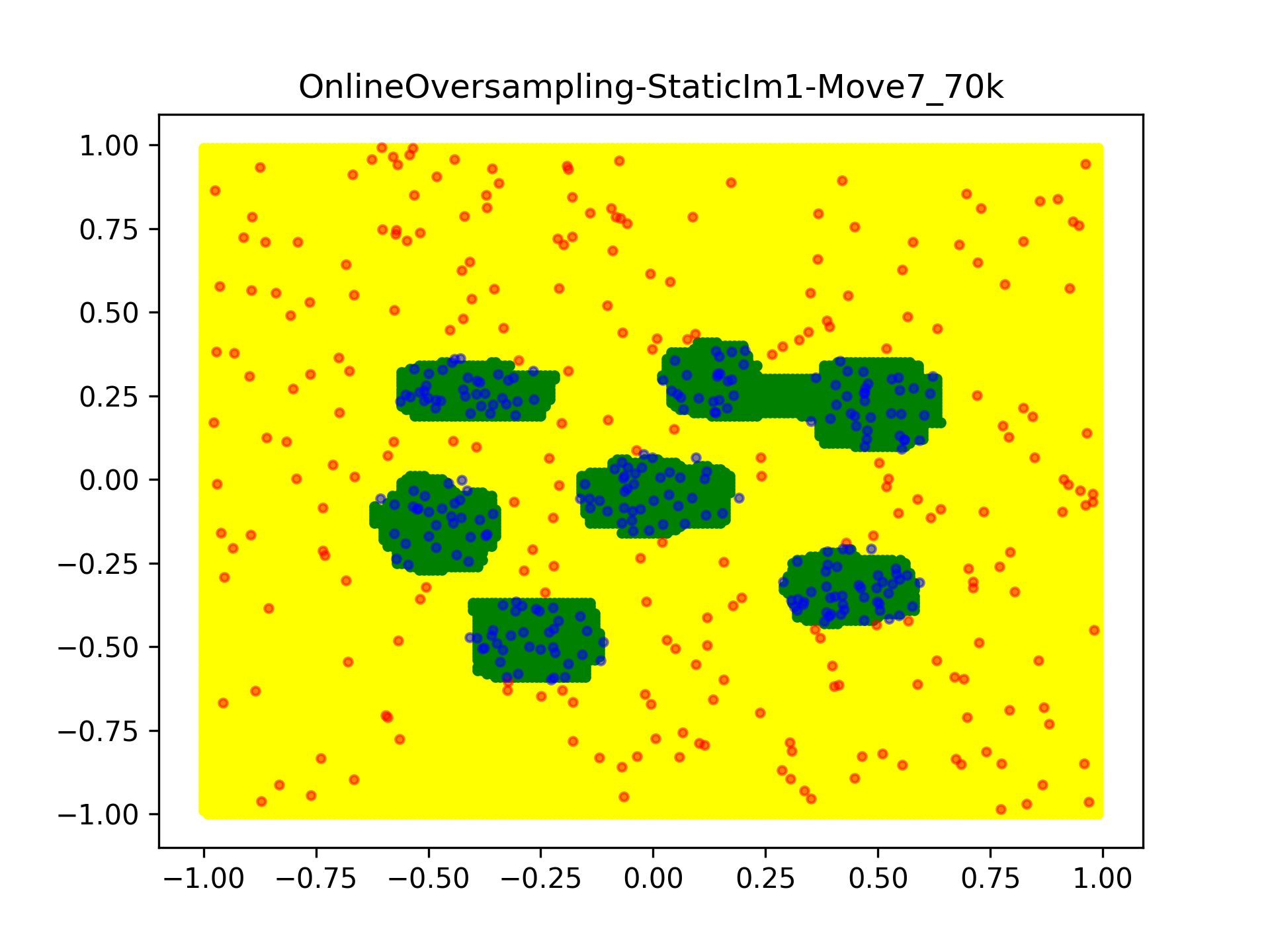} \label{figure:StaticIm1-Move7-dec_bound-OnlineOversampling-70k}}
\subfigure[oUnderOverB]{\includegraphics[width=0.29\textwidth]{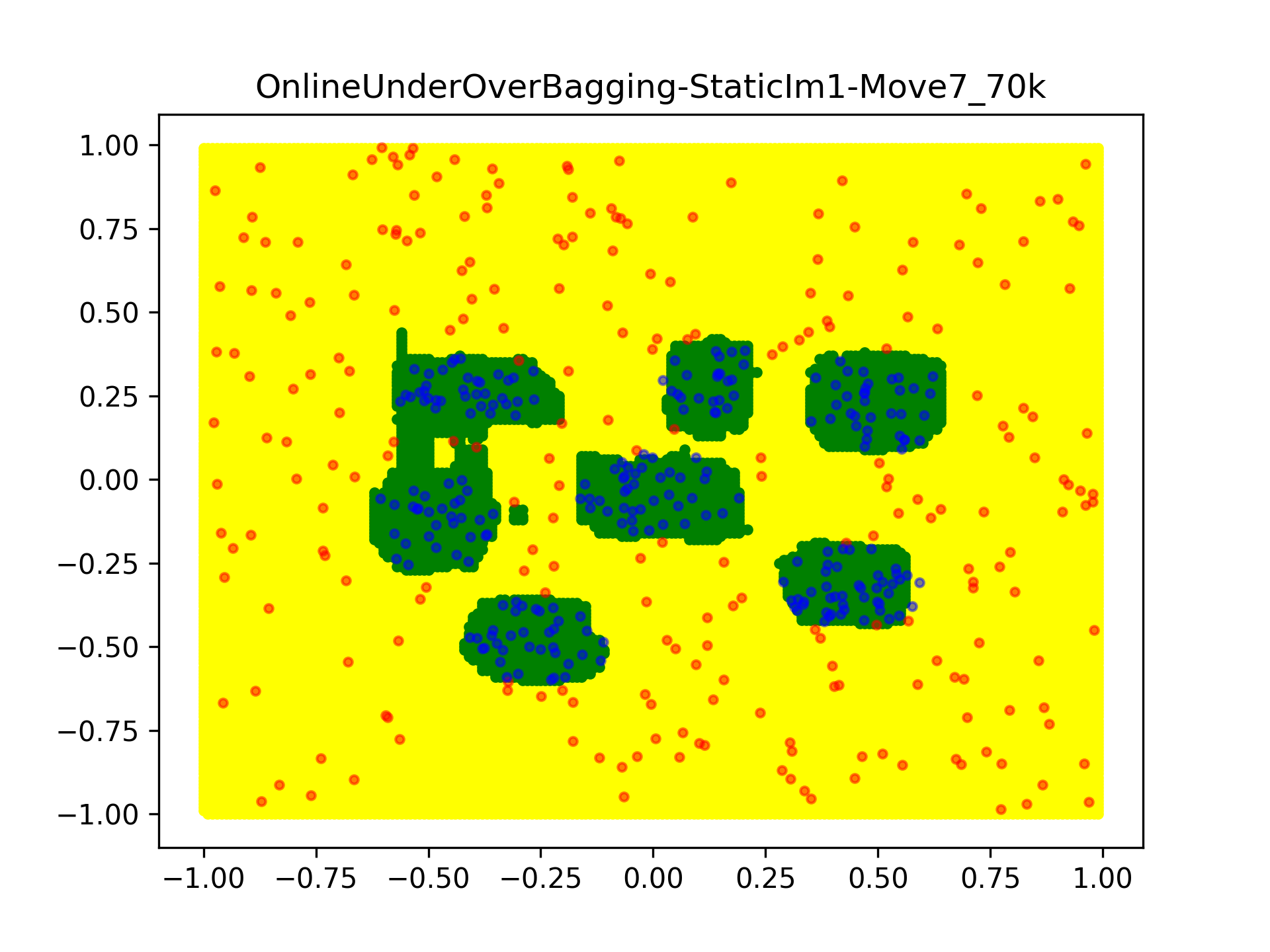} \label{figure:StaticIm1-Move7-dec_bound-OnlineUnderOverBagging-70k}}
\subfigure[OOB\textsubscript{d}]{\includegraphics[width=0.29\textwidth]{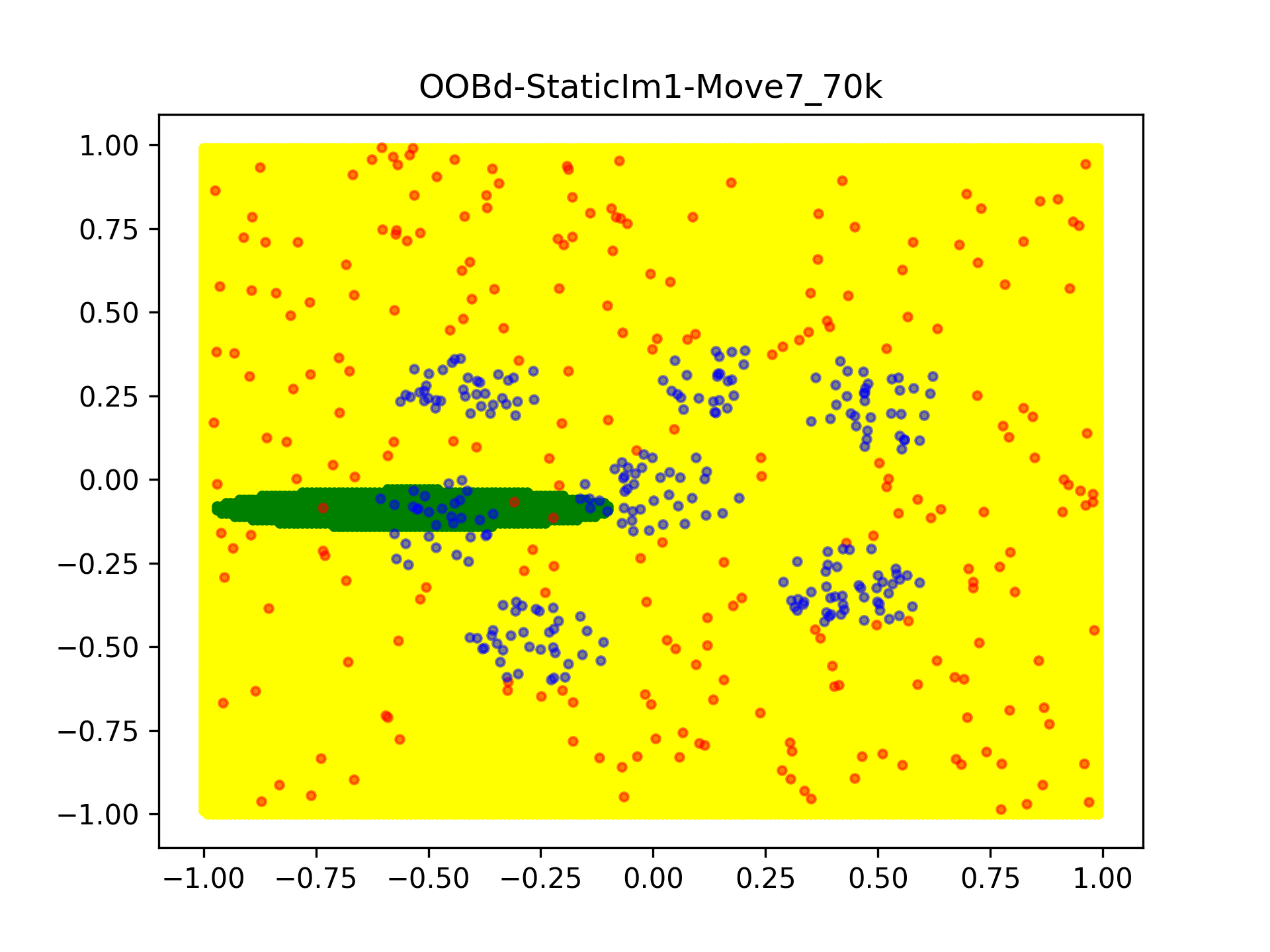} \label{figure:StaticIm1-Move7-dec_bound-OOBd-70k}}
\subfigure[UOB\textsubscript{d}]{\includegraphics[width=0.29\textwidth]{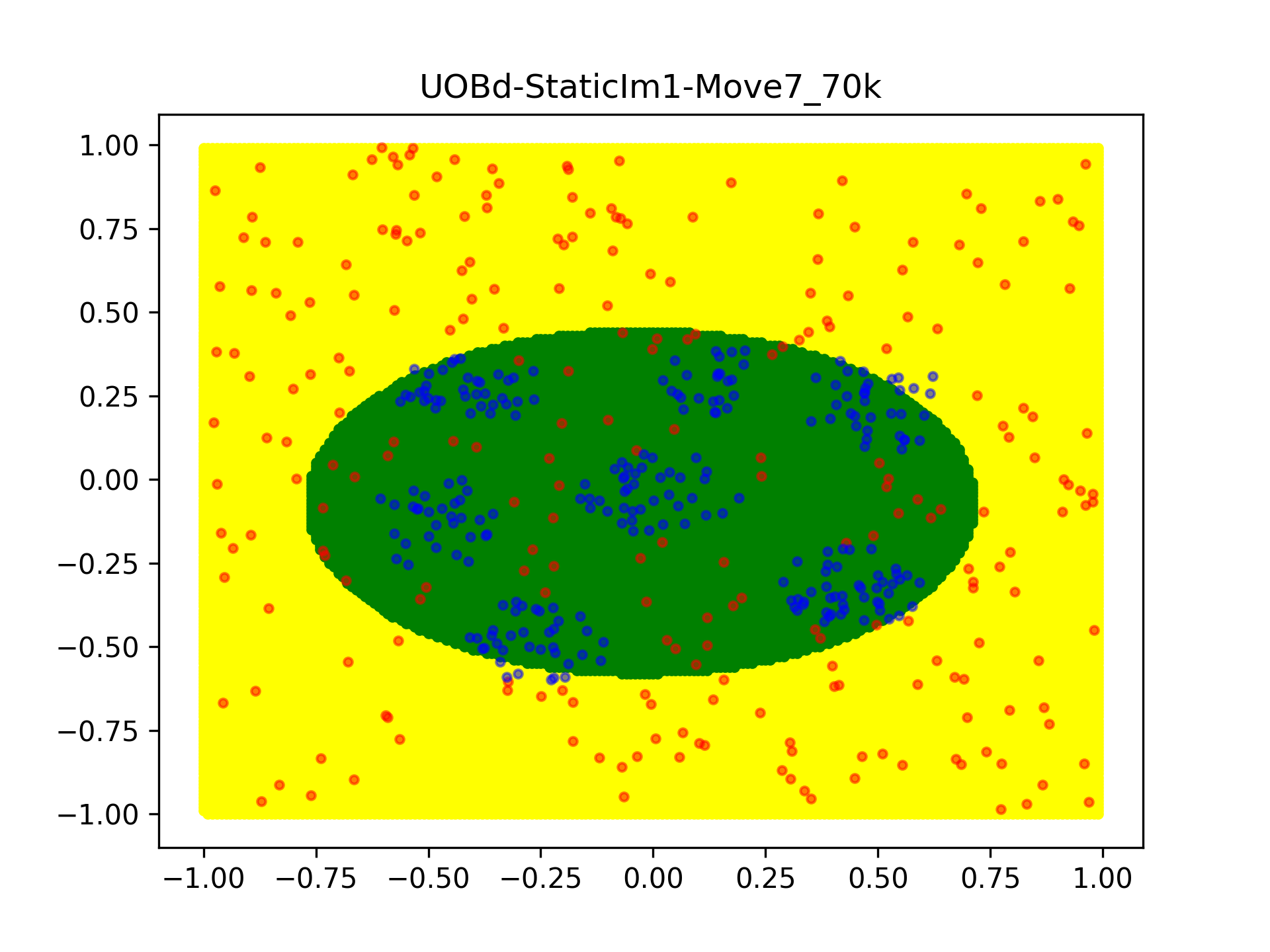} \label{figure:StaticIm1-Move7-dec_bound-UOBd-70k}}
\subfigure[oOS\textsubscript{d}]{\includegraphics[width=0.29\textwidth]{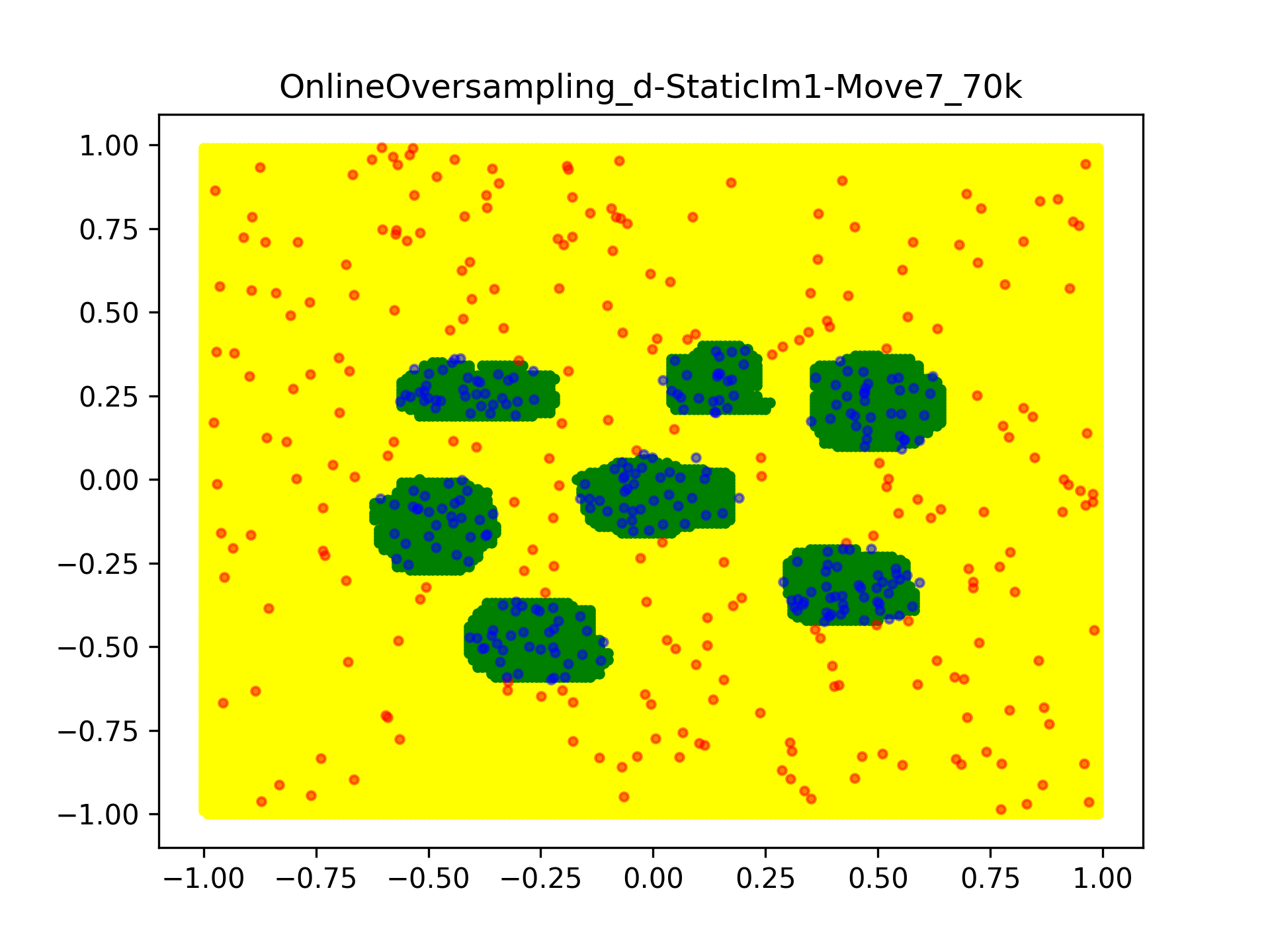} \label{figure:StaticIm1-Move7-dec_bound-OnlineOversampling_d-70k}}
\subfigure[oUnderOverB\textsubscript{d}]{\includegraphics[width=0.29\textwidth]{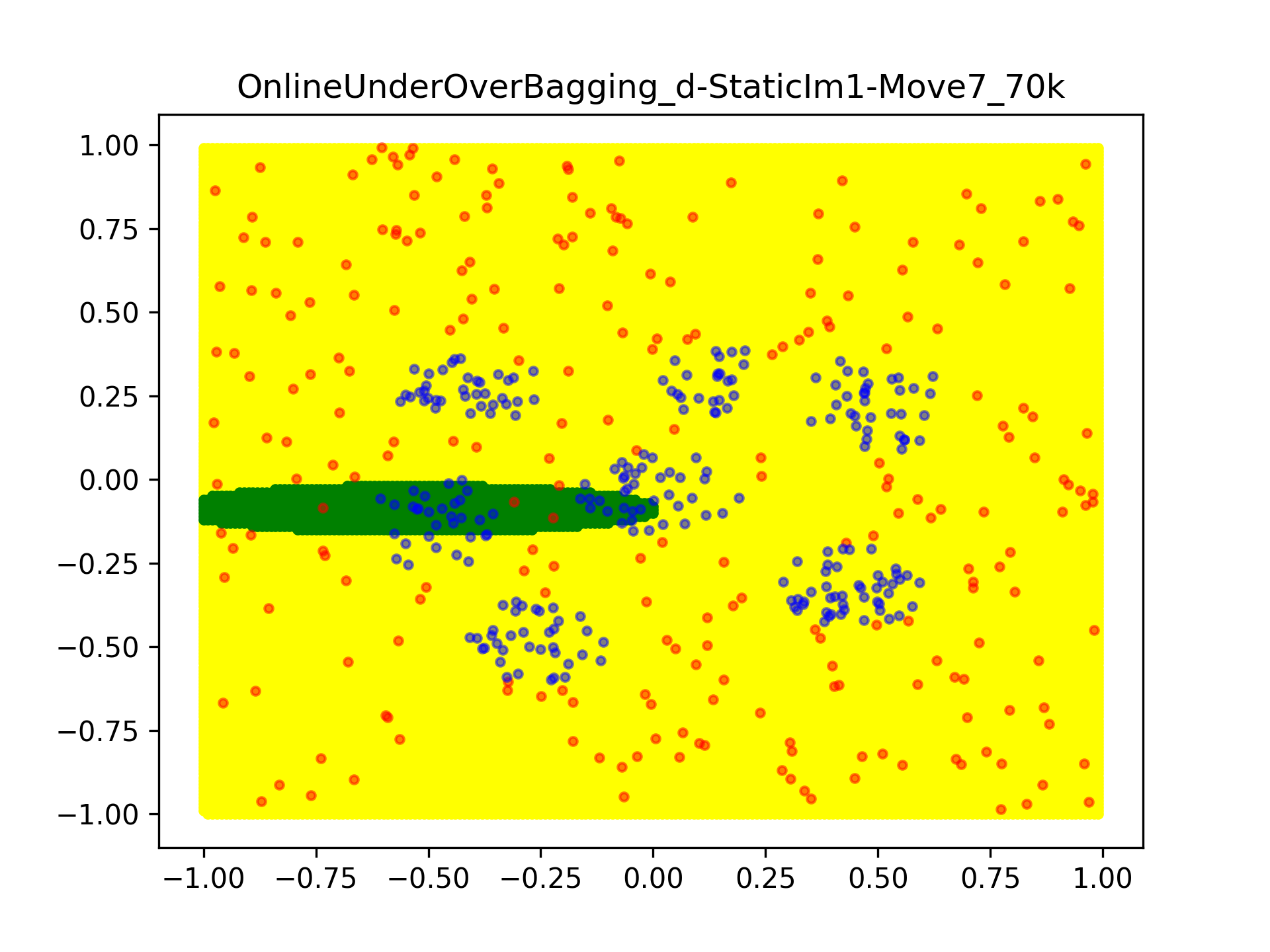} \label{figure:StaticIm1-Move7-dec_bound-OnlineUnderOverBagging_d-70k}}
\subfigure[SMOGauNoise]{\includegraphics[width=0.29\textwidth]{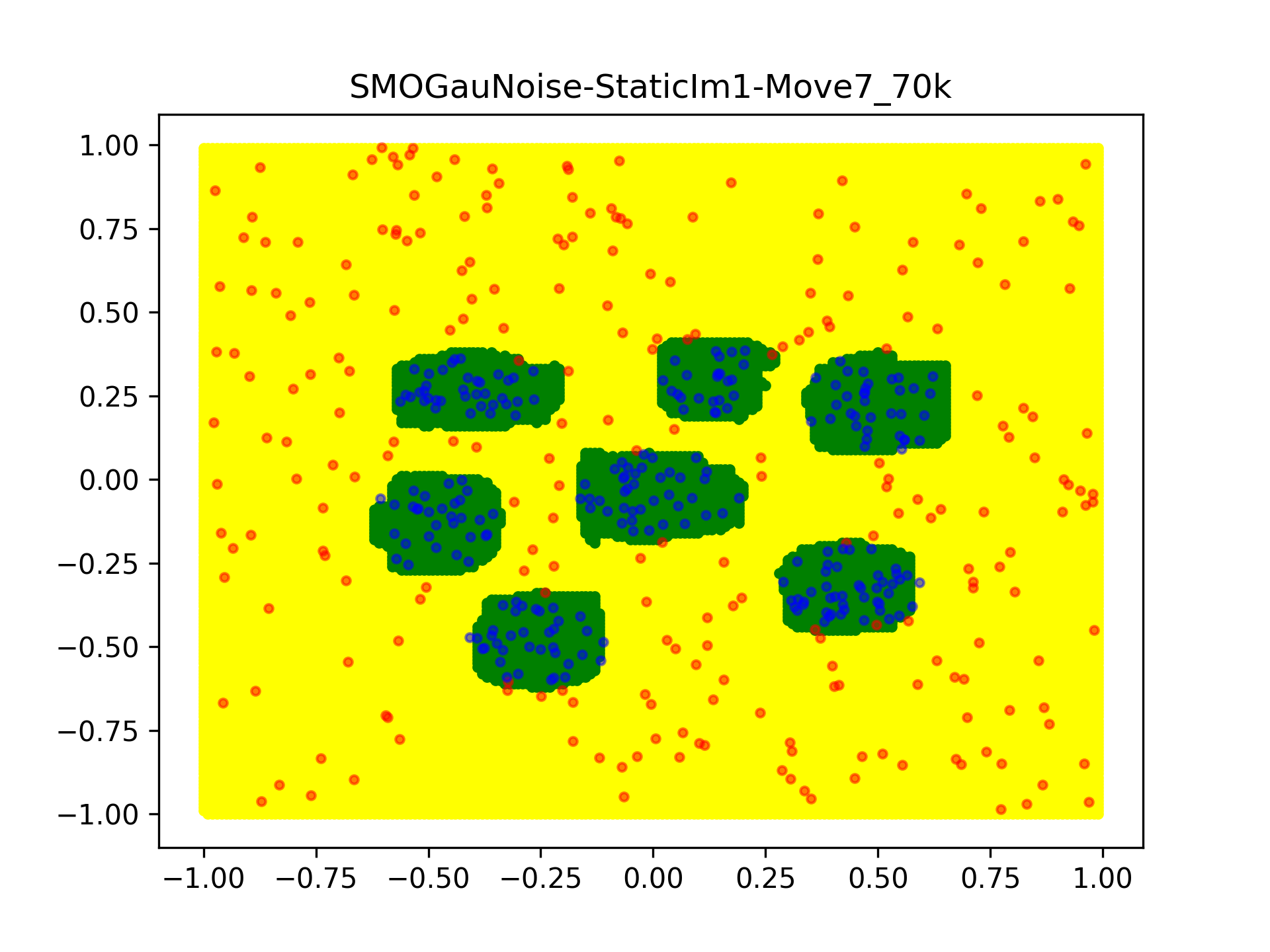} \label{figure:StaticIm1-Move7-dec_bound-SMOGauNoise-70k}}
\subfigure[\reviewII{VFC-SMOTE}]{\includegraphics[width=0.29\textwidth]{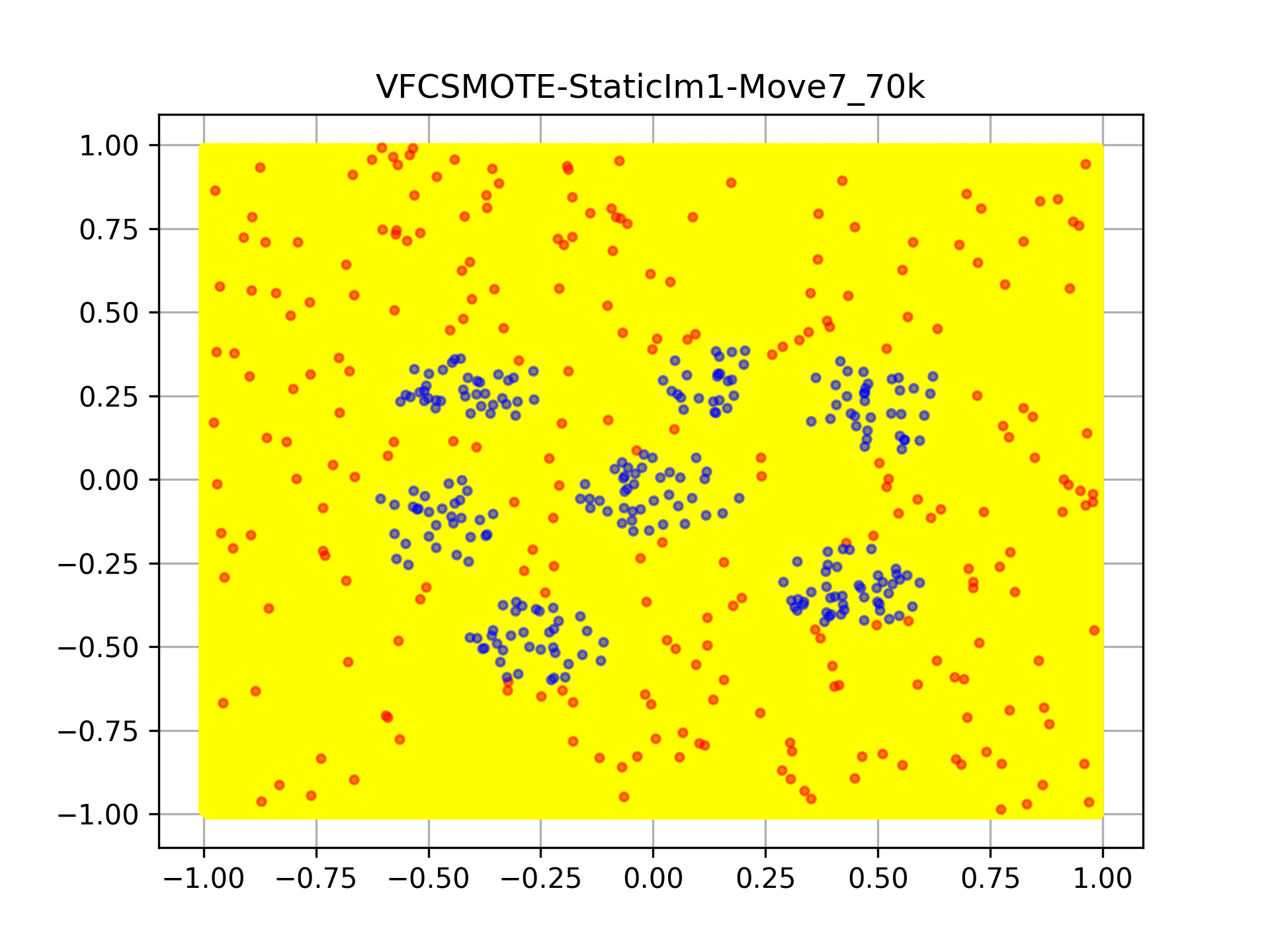} \label{figure:StaticIm1-Move7-dec_bound-VFCSMOTE-70k}}
\subfigure[\reviewII{SMOTE-OB}]{\includegraphics[width=0.29\textwidth]{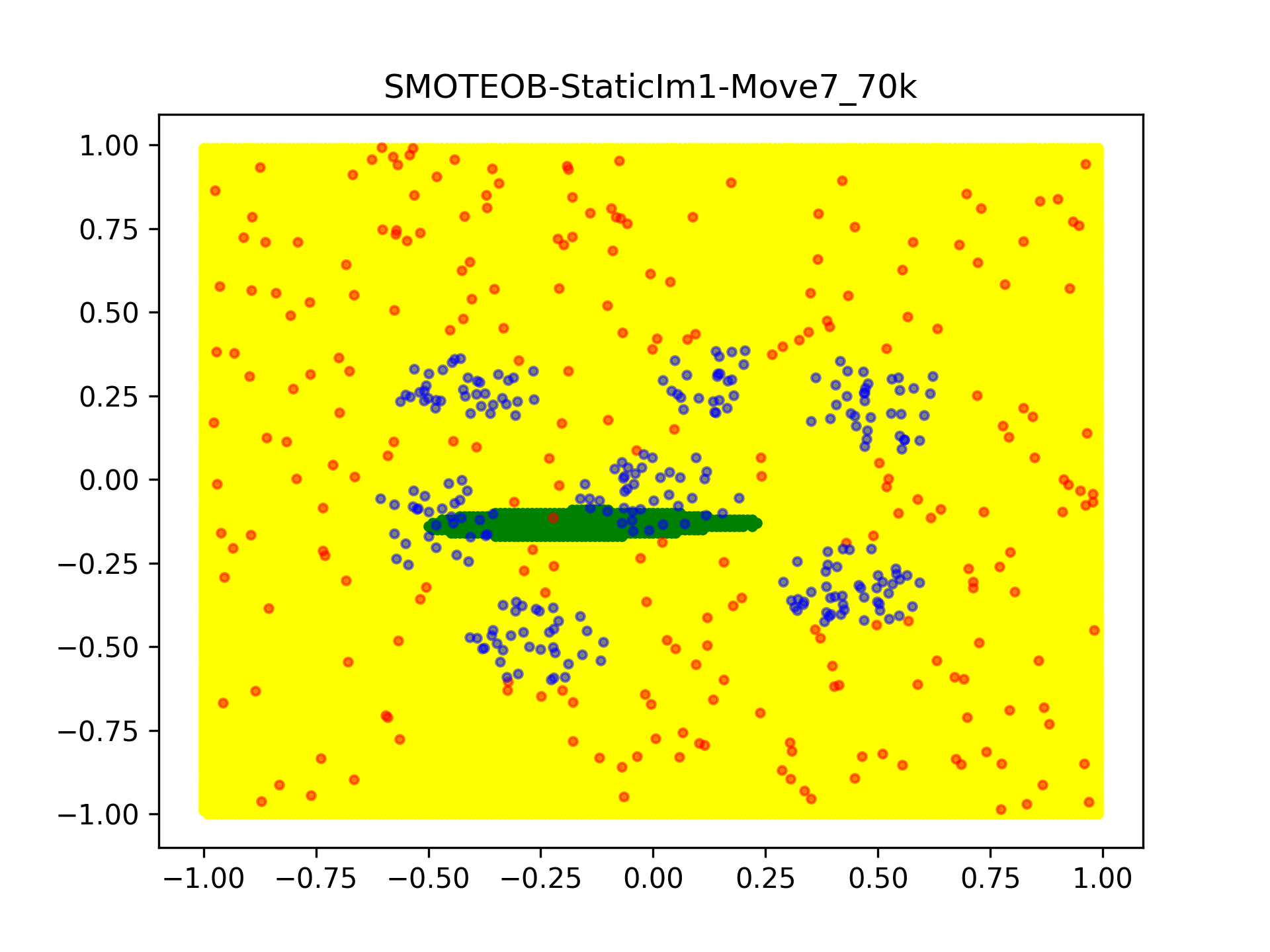} \label{figure:StaticIm1-Move7-dec_bound-SMOTEOB-70k}}
\subfigure[SMOClust]{\includegraphics[width=0.29\textwidth]{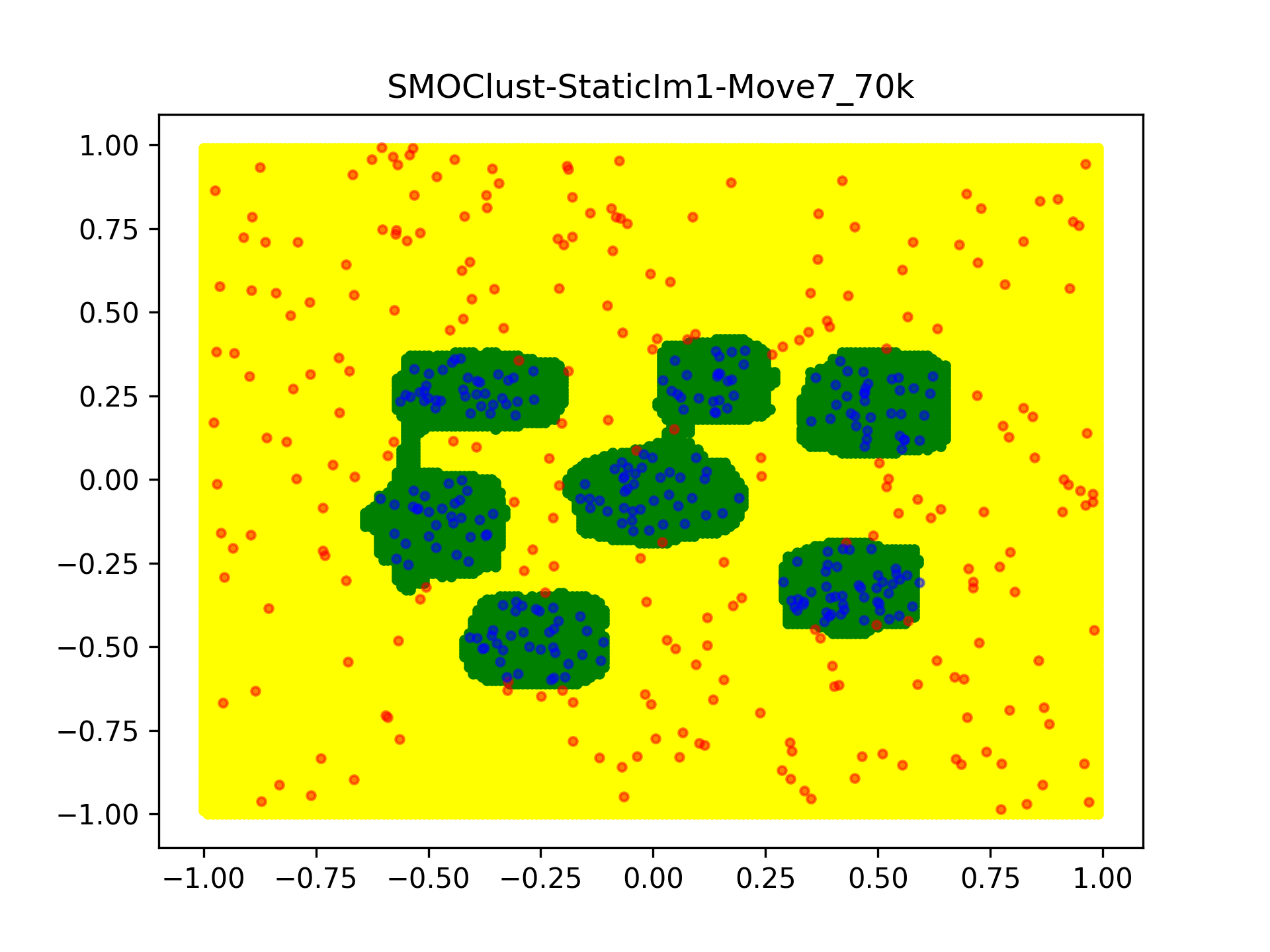} \label{figure:StaticIm1-Move7-dec_bound-SMOClust-70k}}
\caption{Decision Areas Against Class Balanced Test Set at 70k Time Steps (Before Drift) of Two-Dimensional StaticIm1\_Move7}
\label{figure:StaticIm1-Move7-dec_bound-70k}
\end{figure}

\begin{figure}[!ht]
\centering
\subfigure[OOB]{\includegraphics[width=0.29\textwidth]{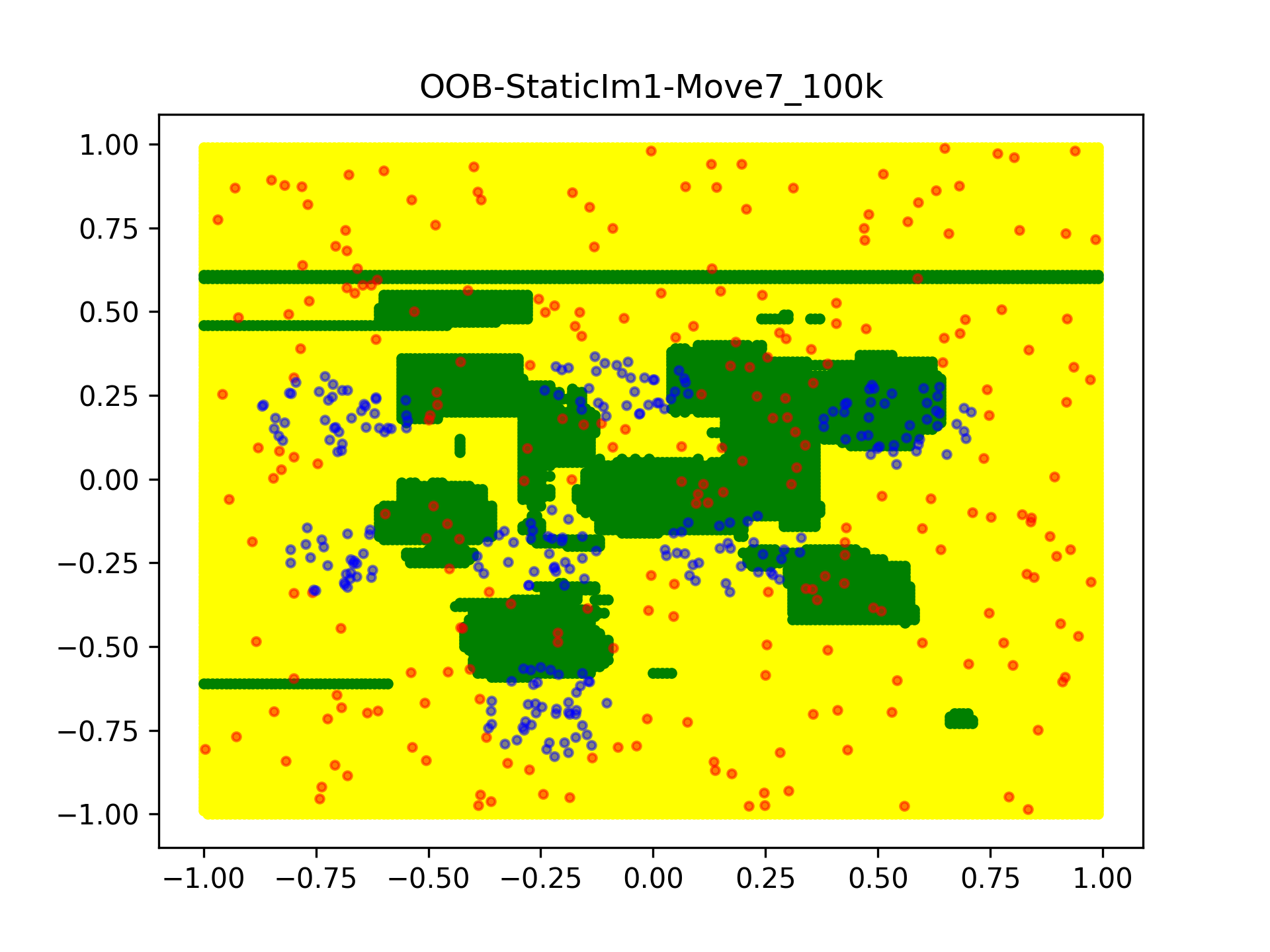} \label{figure:StaticIm1-Move7-dec_bound-OOB-100k}}
\subfigure[UOB]{\includegraphics[width=0.29\textwidth]{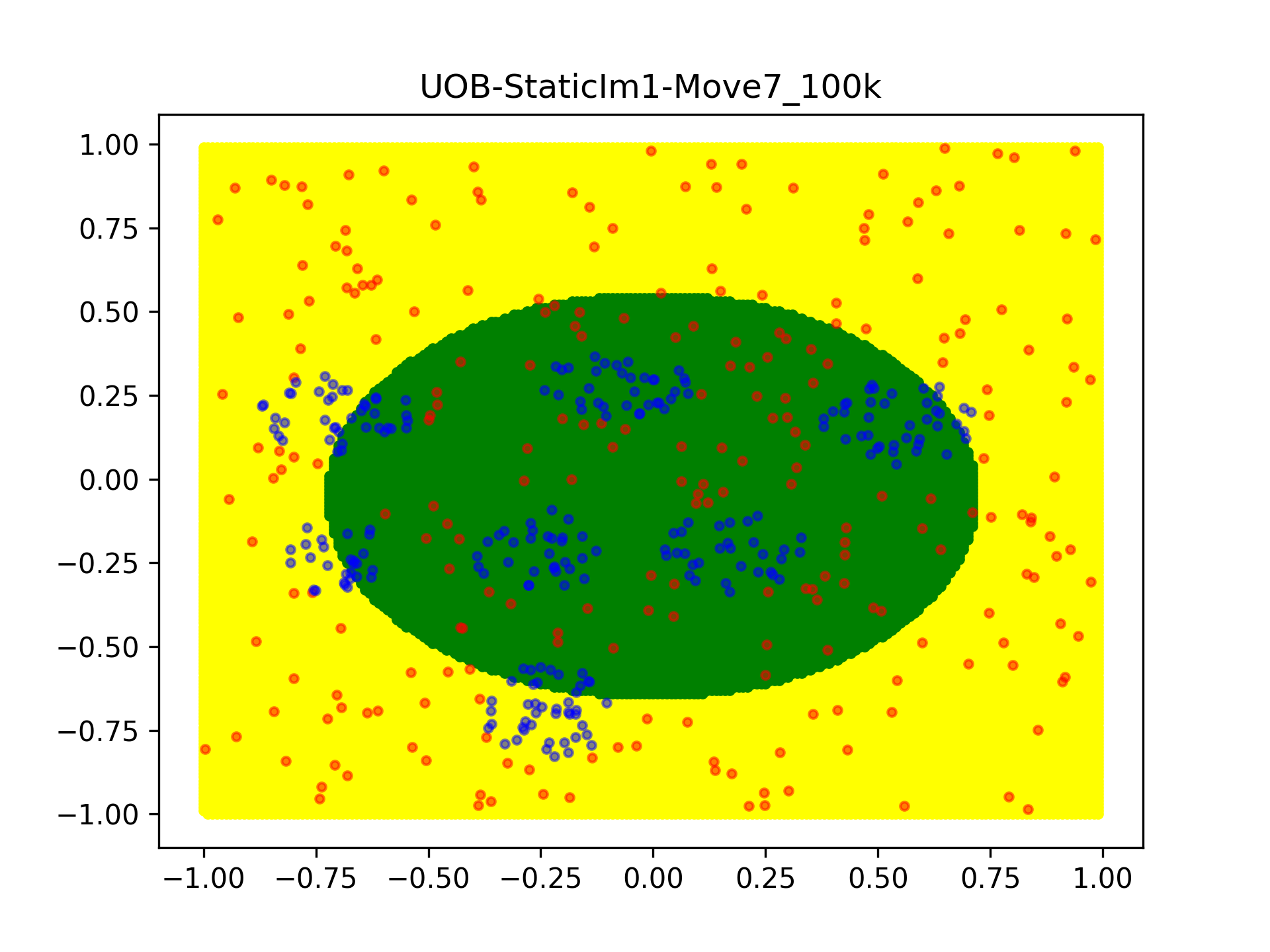} \label{figure:StaticIm1-Move7-dec_bound-UOB-100k}}
\subfigure[oOS]{\includegraphics[width=0.29\textwidth]{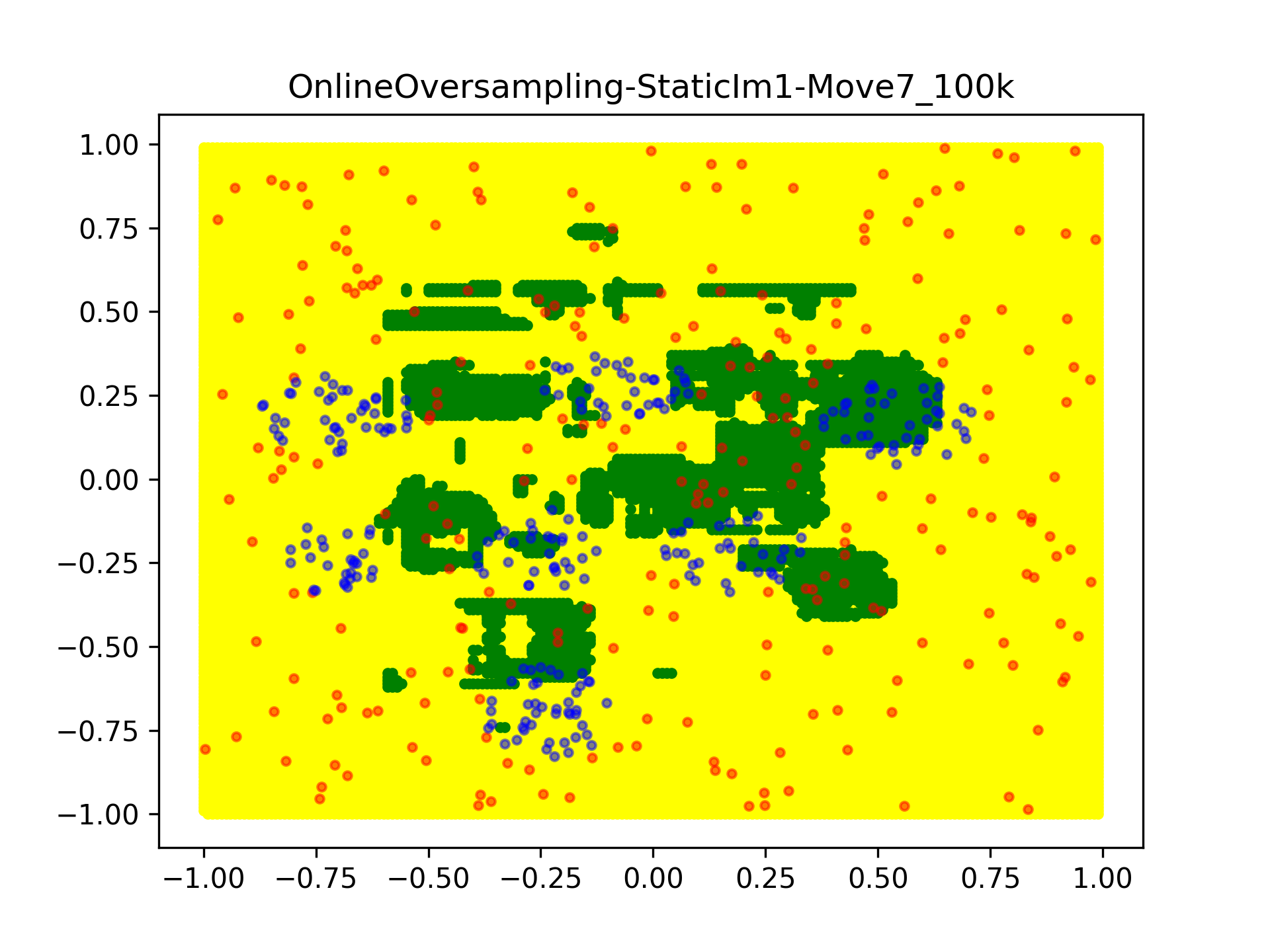} \label{figure:StaticIm1-Move7-dec_bound-OnlineOversampling-100k}}
\subfigure[oUnderOverB]{\includegraphics[width=0.29\textwidth]{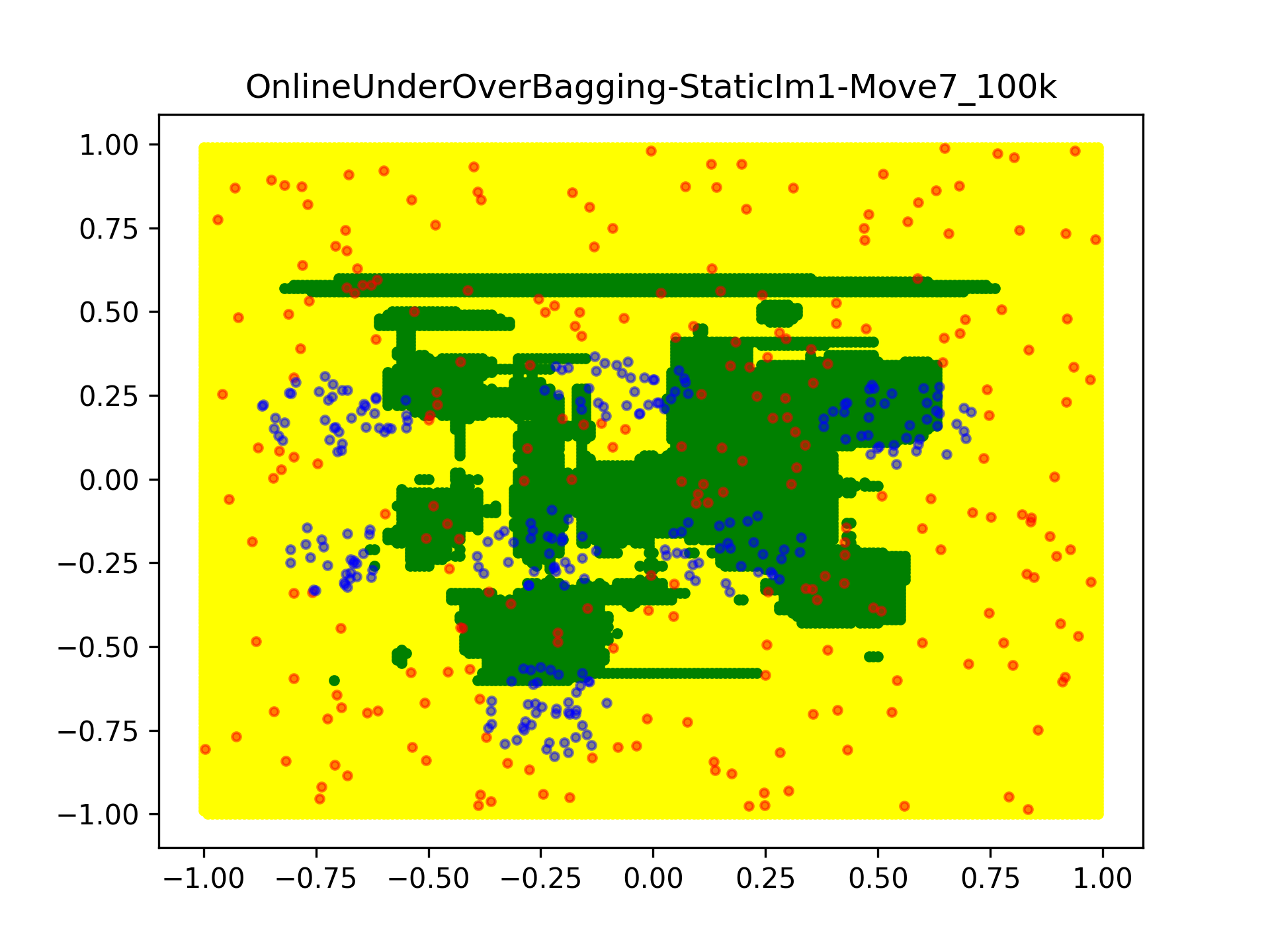} \label{figure:StaticIm1-Move7-dec_bound-OnlineUnderOverBagging-100k}}
\subfigure[OOB\textsubscript{d}]{\includegraphics[width=0.29\textwidth]{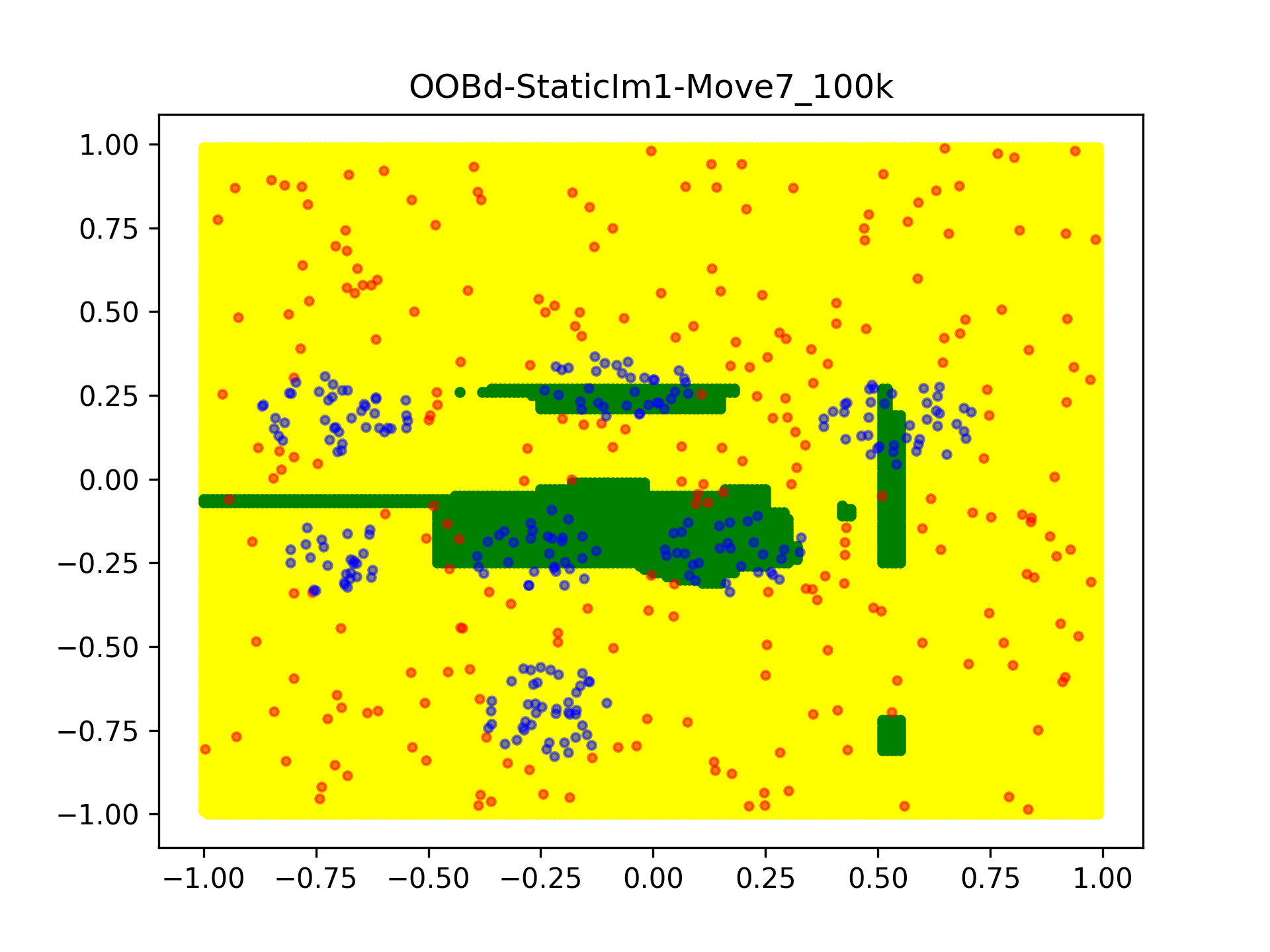} \label{figure:StaticIm1-Move7-dec_bound-OOBd-100k}}
\subfigure[UOB\textsubscript{d}]{\includegraphics[width=0.29\textwidth]{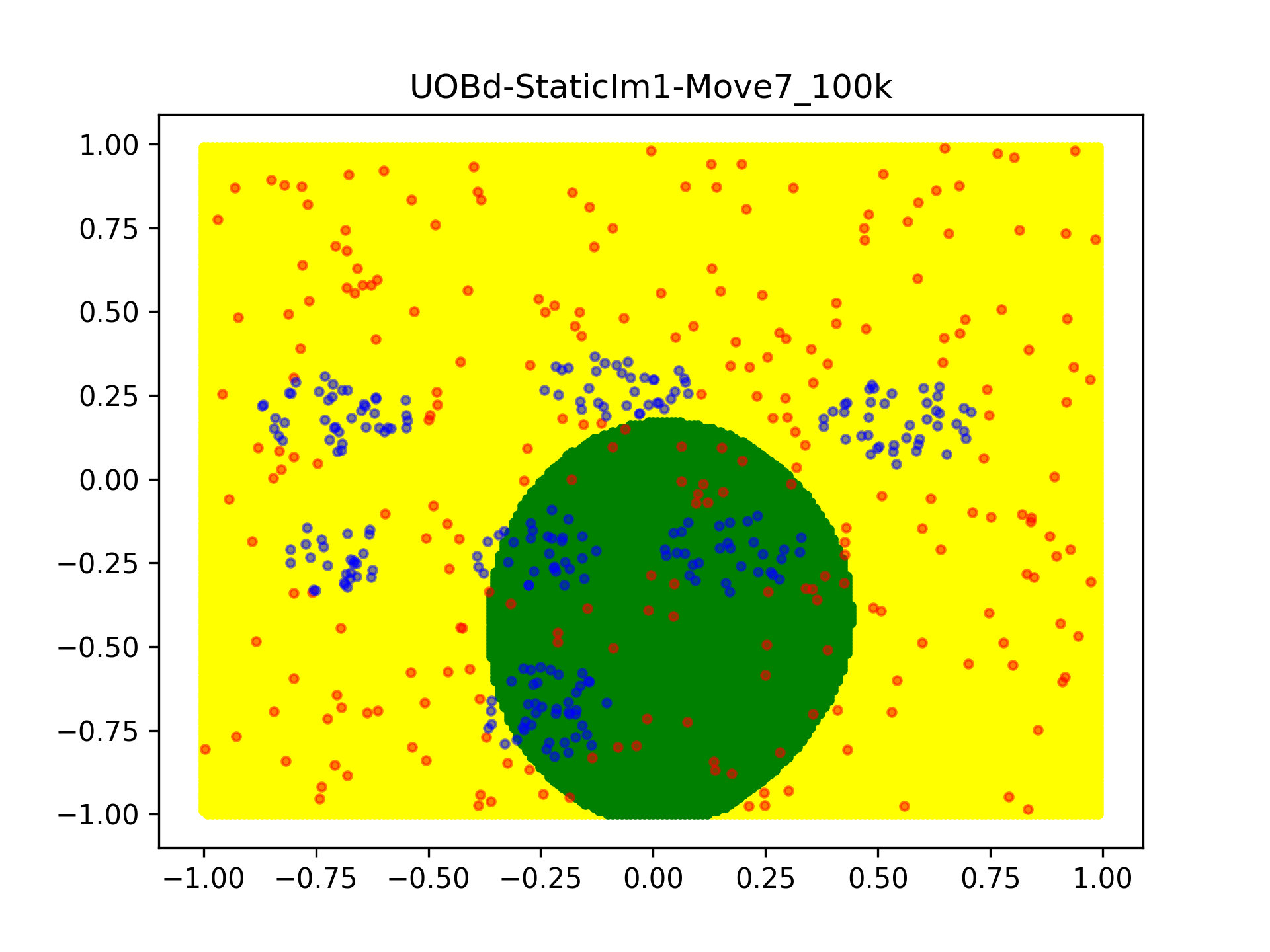} \label{figure:StaticIm1-Move7-dec_bound-UOBd-100k}}
\subfigure[oOS\textsubscript{d}]{\includegraphics[width=0.29\textwidth]{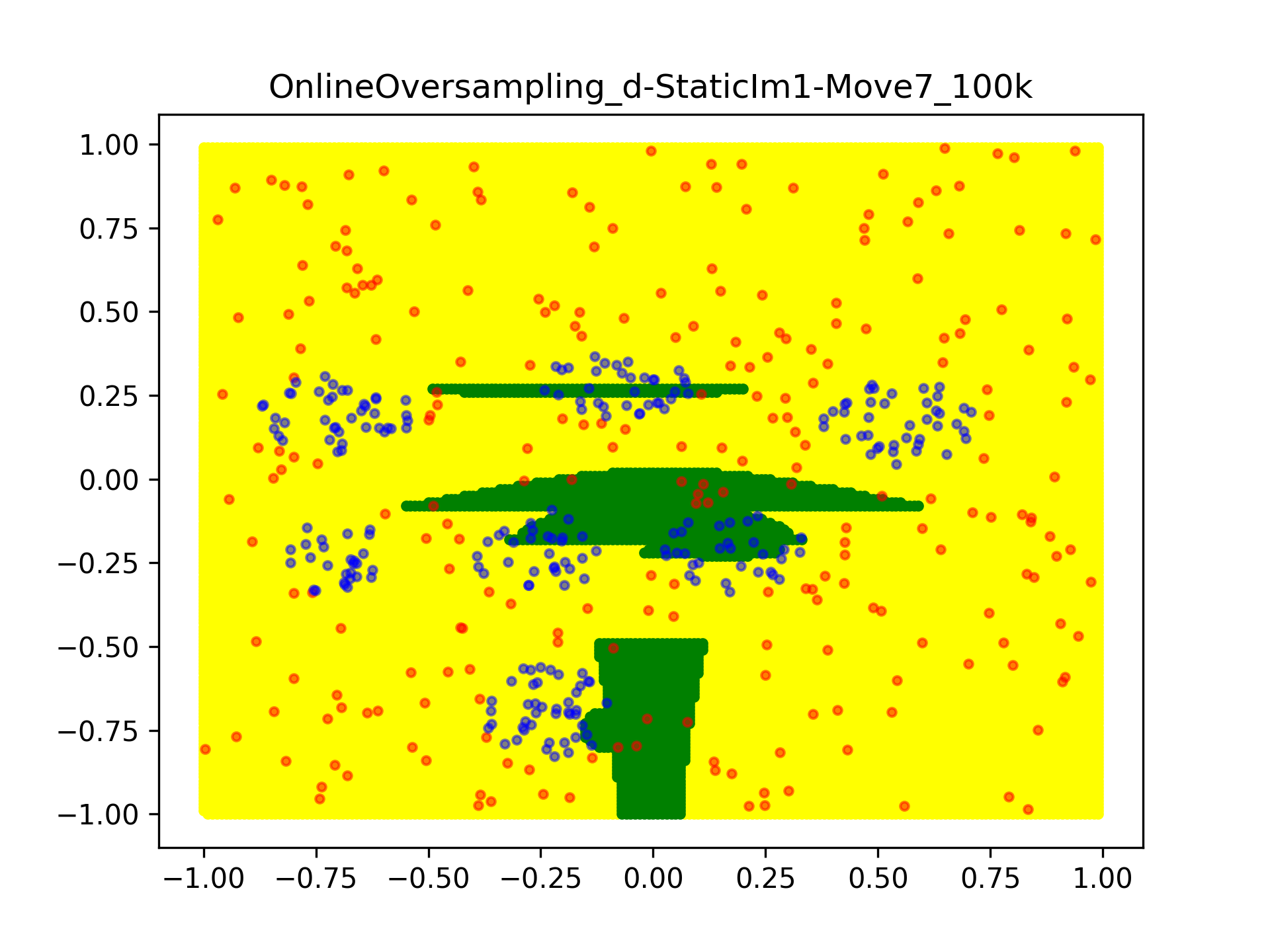} \label{figure:StaticIm1-Move7-dec_bound-OnlineOversampling_d-100k}}
\subfigure[oUnderOverB\textsubscript{d}]{\includegraphics[width=0.29\textwidth]{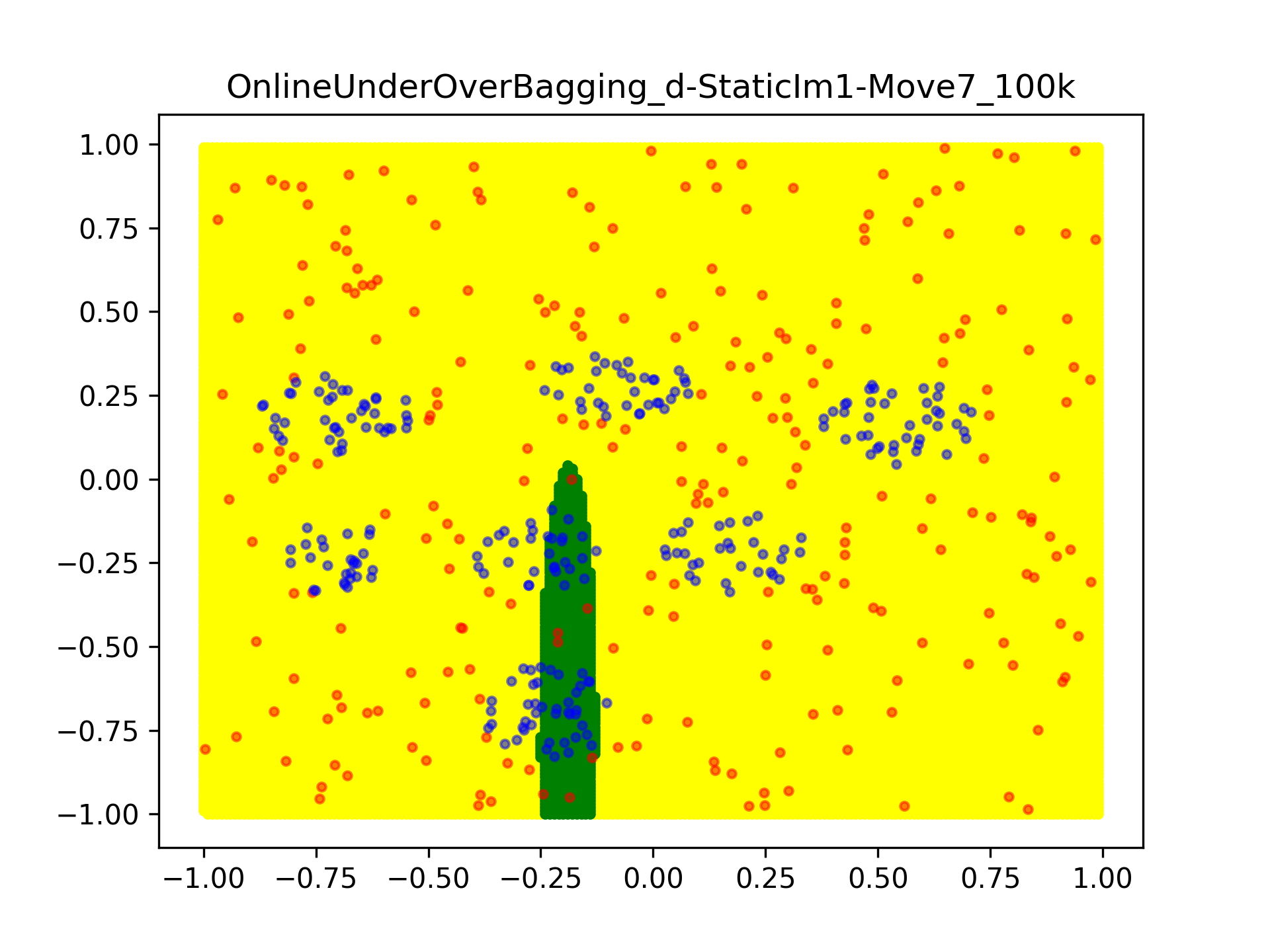} \label{figure:StaticIm1-Move7-dec_bound-OnlineUnderOverBagging_d-100k}}
\subfigure[SMOGauNoise]{\includegraphics[width=0.29\textwidth]{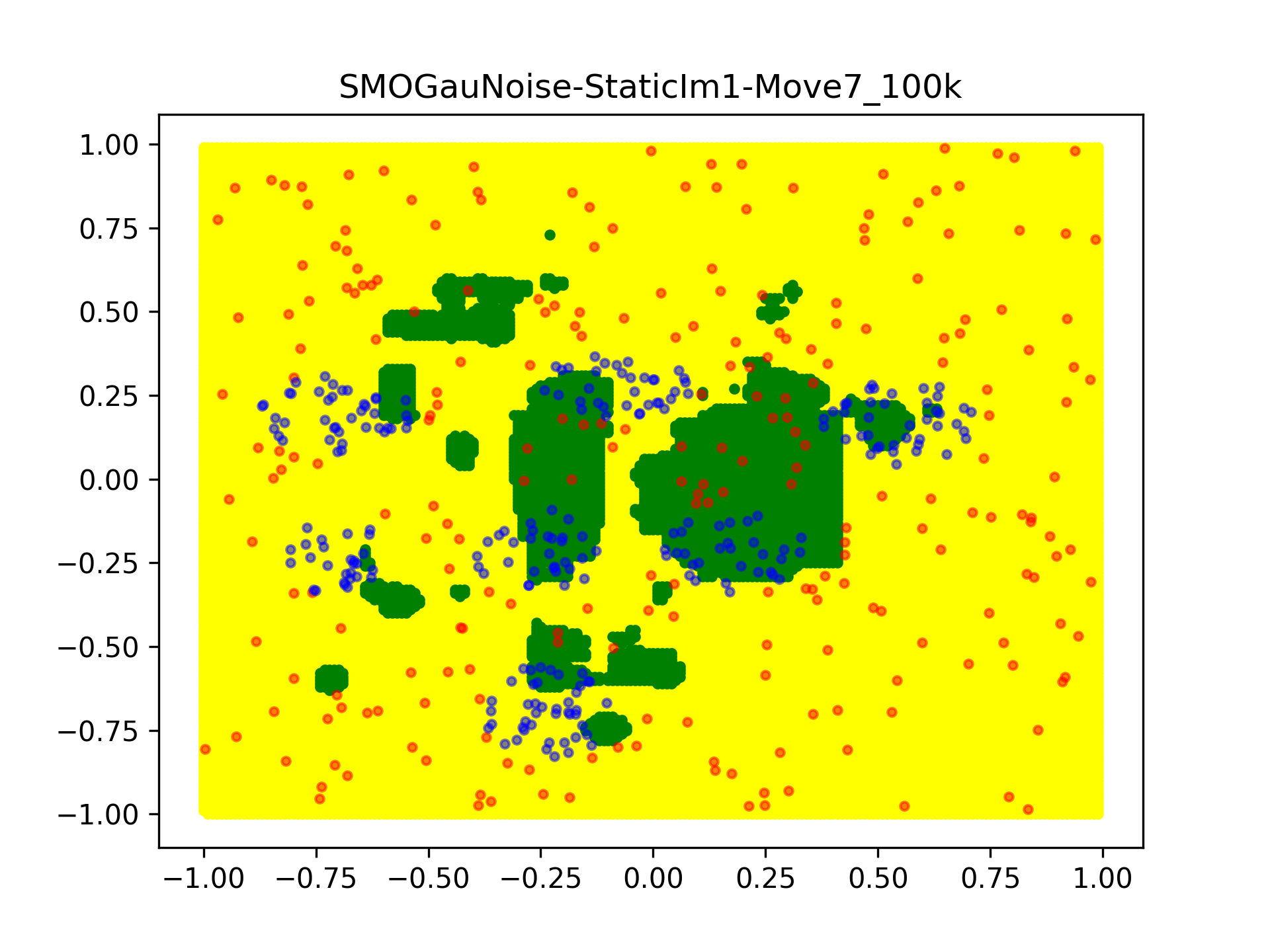} \label{figure:StaticIm1-Move7-dec_bound-SMOGauNoise-100k}}
\subfigure[\reviewII{VFC-SMOTE}]{\includegraphics[width=0.29\textwidth]{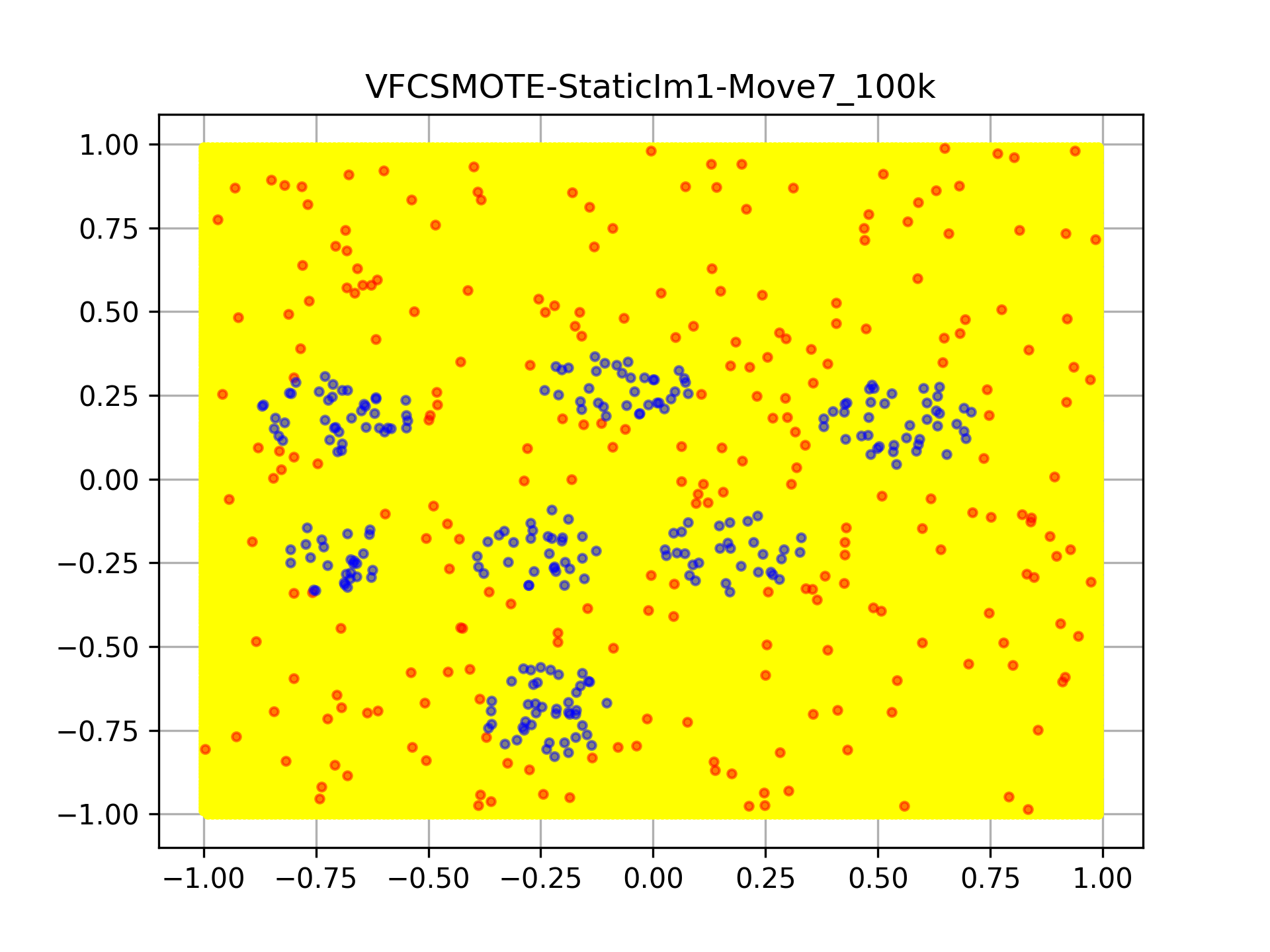} \label{figure:StaticIm1-Move7-dec_bound-VFCSMOTE-100k}}
\subfigure[\reviewII{SMOTE-OB}]{\includegraphics[width=0.29\textwidth]{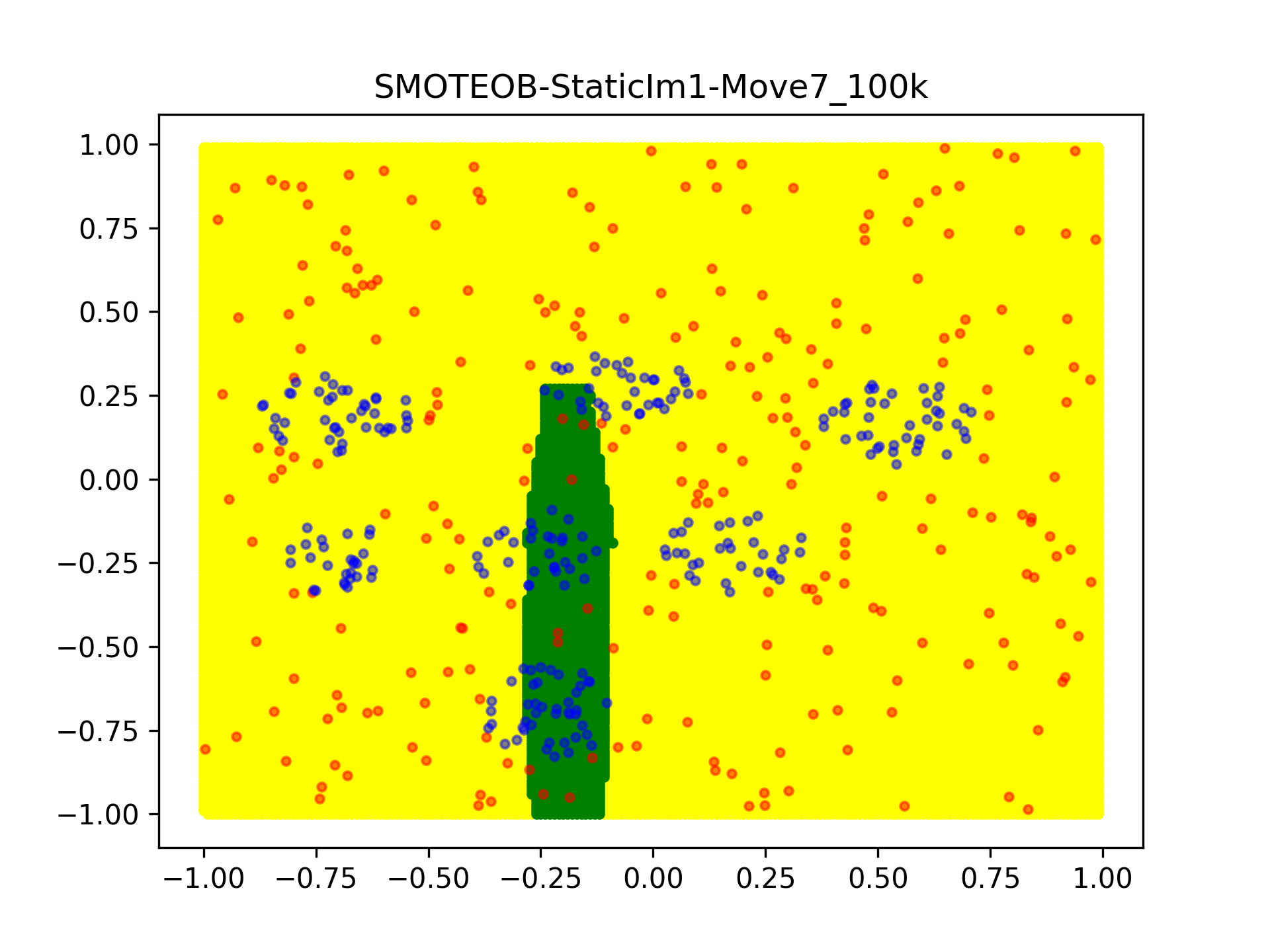} \label{figure:StaticIm1-Move7-dec_bound-SMOTEOB-100k}}
\subfigure[SMOClust]{\includegraphics[width=0.29\textwidth]{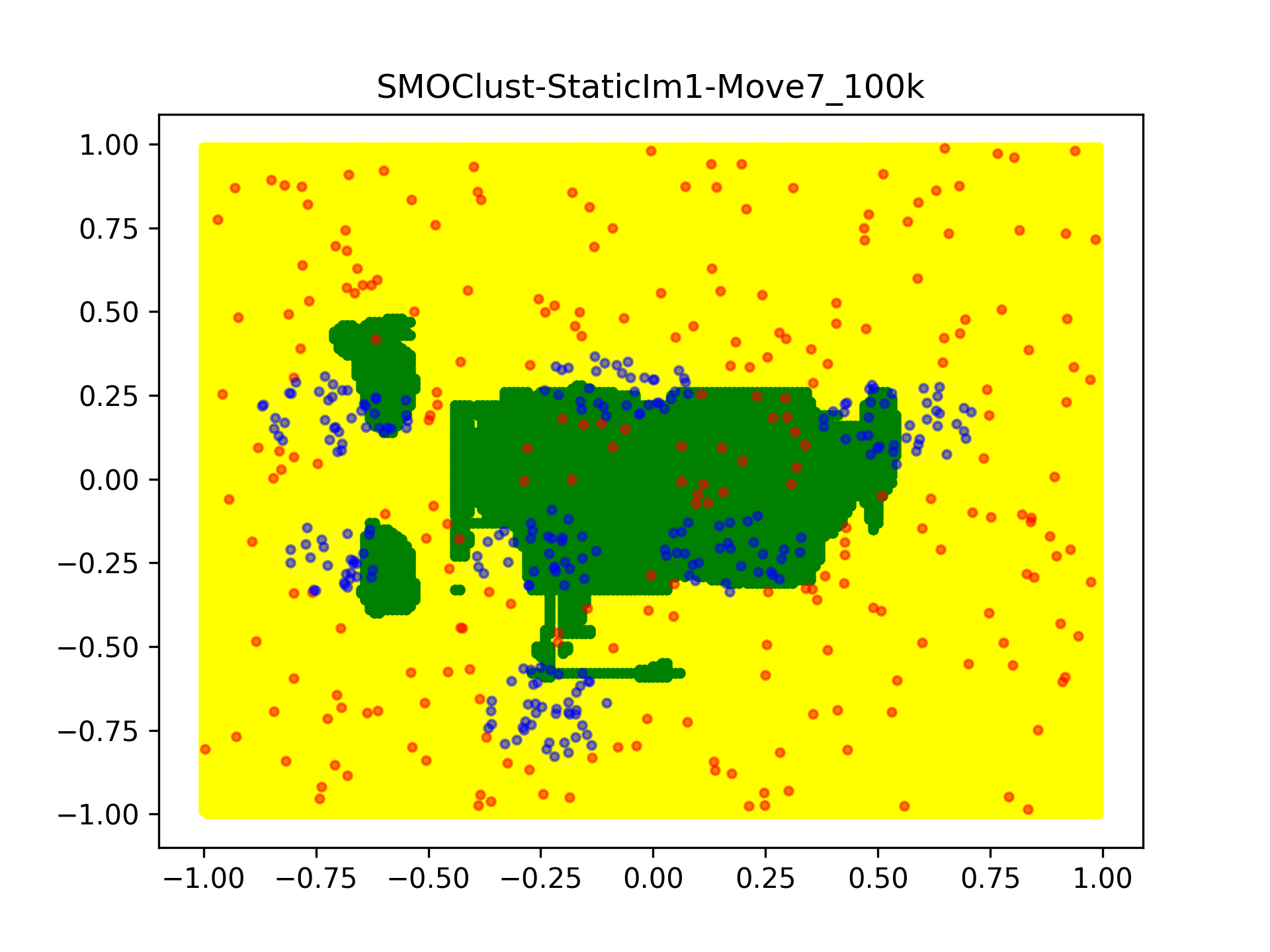} \label{figure:StaticIm1-Move7-dec_bound-SMOClust-100k}}
\caption{Decision Areas Against Class Balanced Test Set at 100k Time Steps (After Drift) of Two-Dimensional StaticIm1\_Move7}
\label{figure:StaticIm1-Move7-dec_bound-100k}
\end{figure}

\begin{figure}[!ht]
\centering
\subfigure[OOB]{\includegraphics[width=0.29\textwidth]{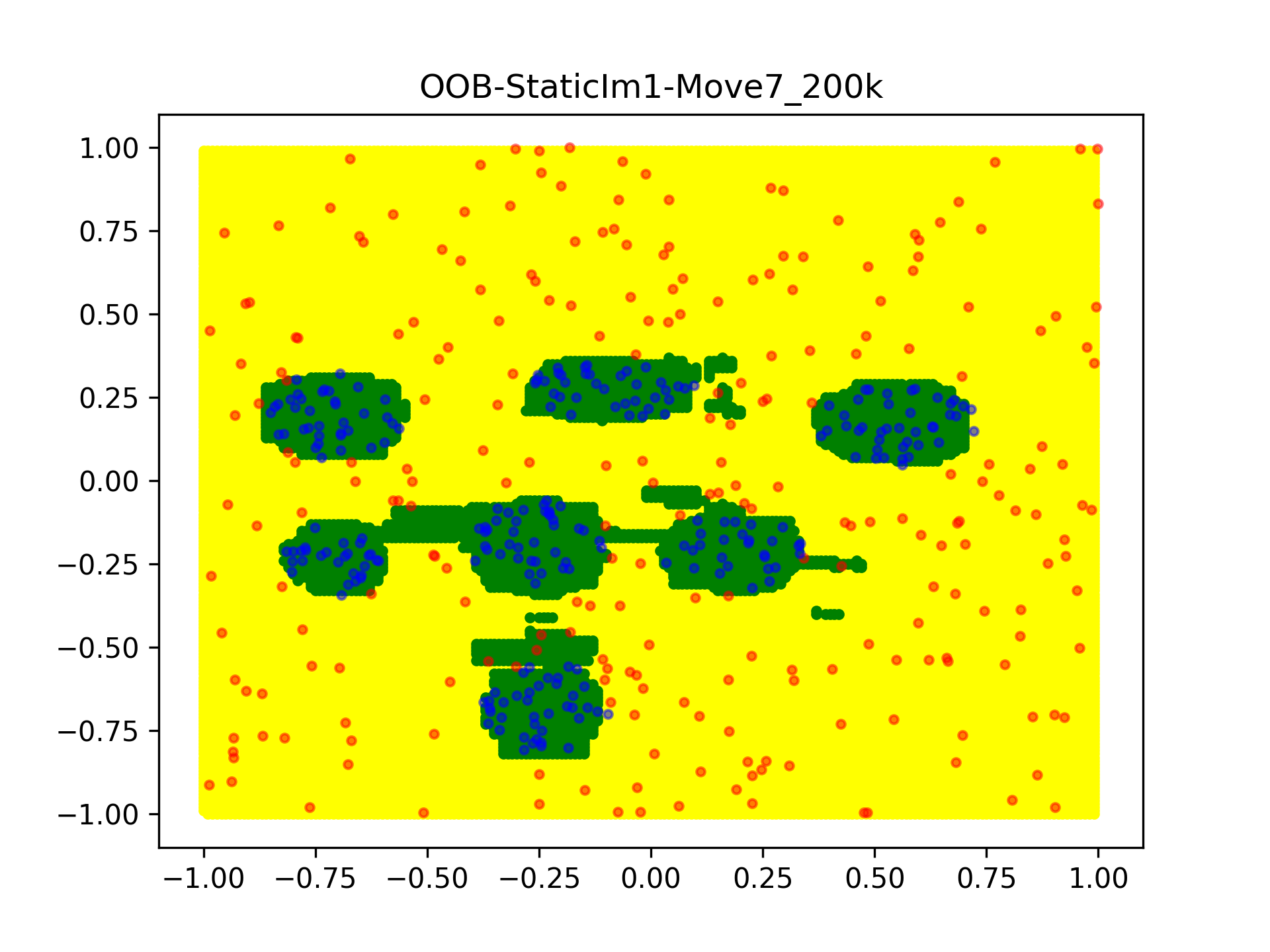} \label{figure:StaticIm1-Move7-dec_bound-OOB-200k}}
\subfigure[UOB]{\includegraphics[width=0.29\textwidth]{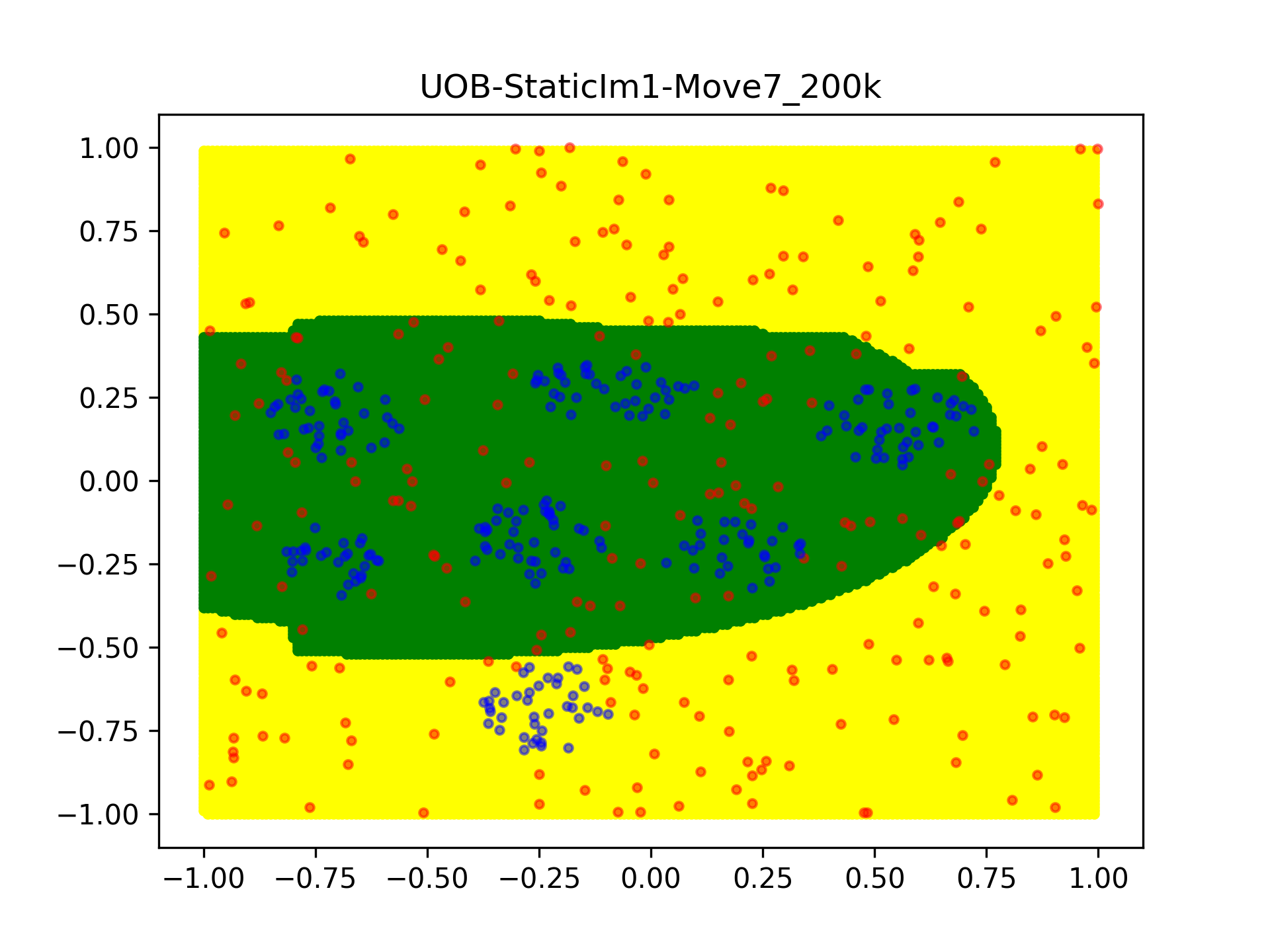} \label{figure:StaticIm1-Move7-dec_bound-UOB-200k}}
\subfigure[oOS]{\includegraphics[width=0.29\textwidth]{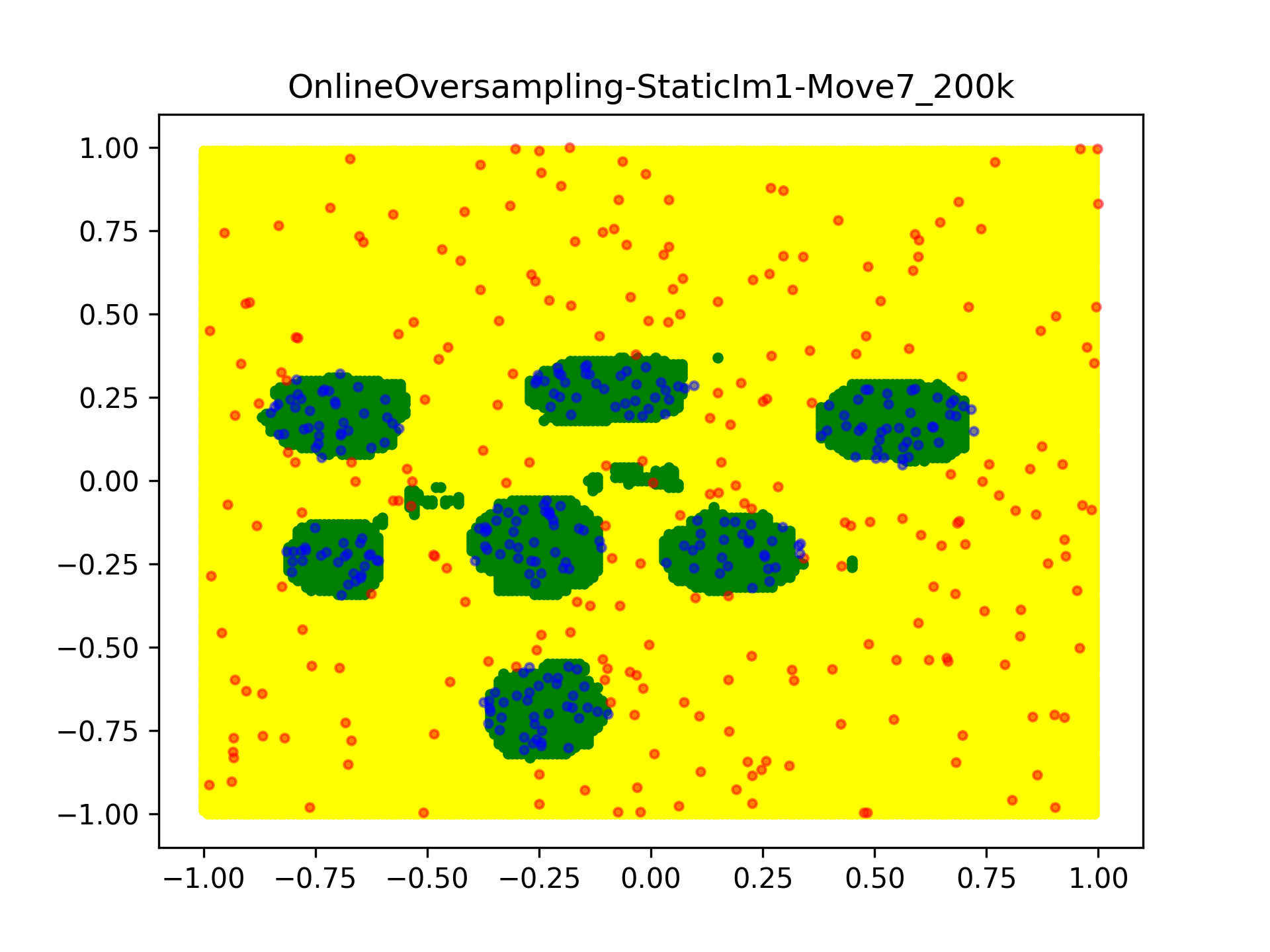} \label{figure:StaticIm1-Move7-dec_bound-OnlineOversampling-200k}}
\subfigure[oUnderOverB]{\includegraphics[width=0.29\textwidth]{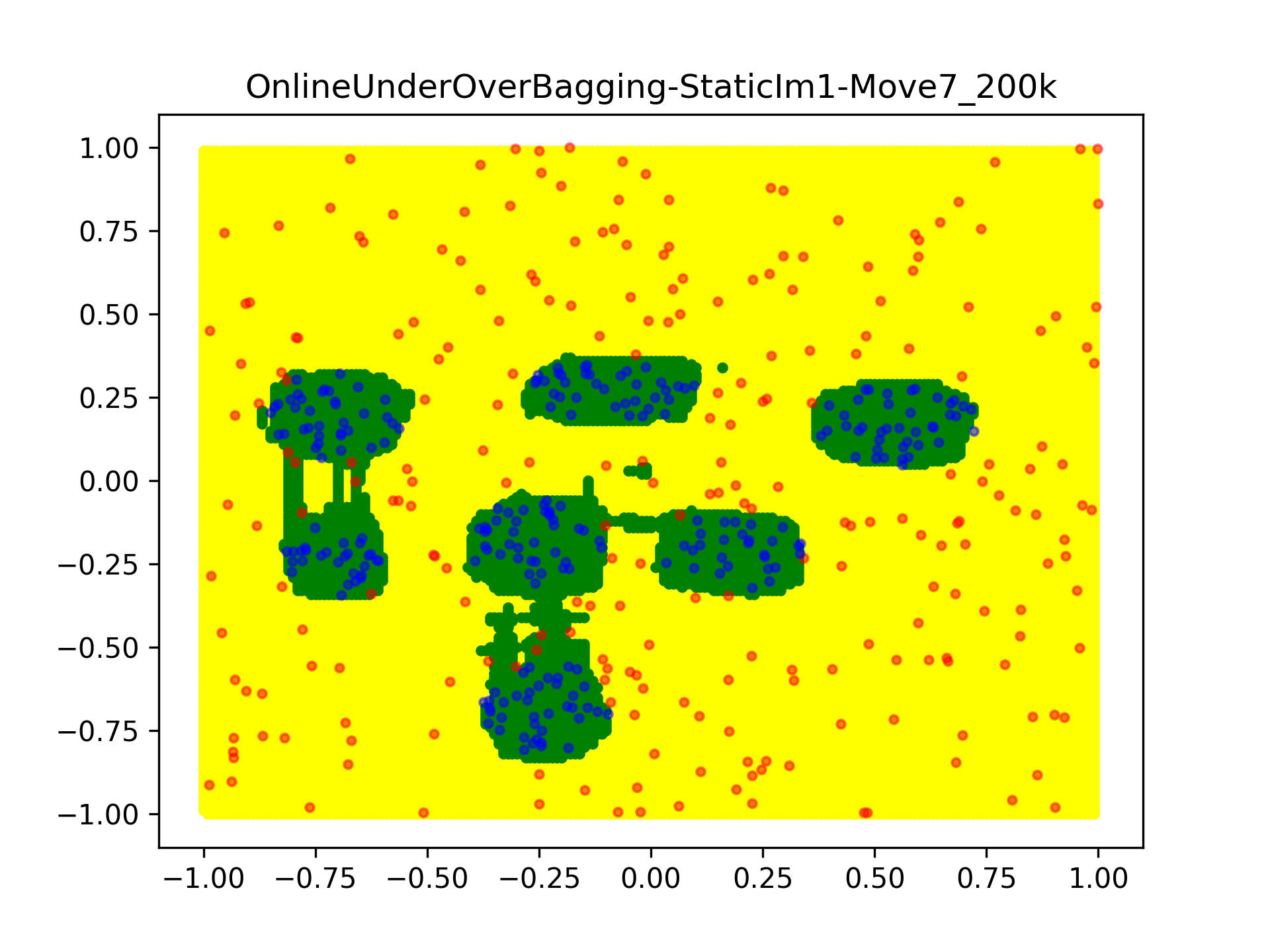} \label{figure:StaticIm1-Move7-dec_bound-OnlineUnderOverBagging-200k}}
\subfigure[OOB\textsubscript{d}]{\includegraphics[width=0.29\textwidth]{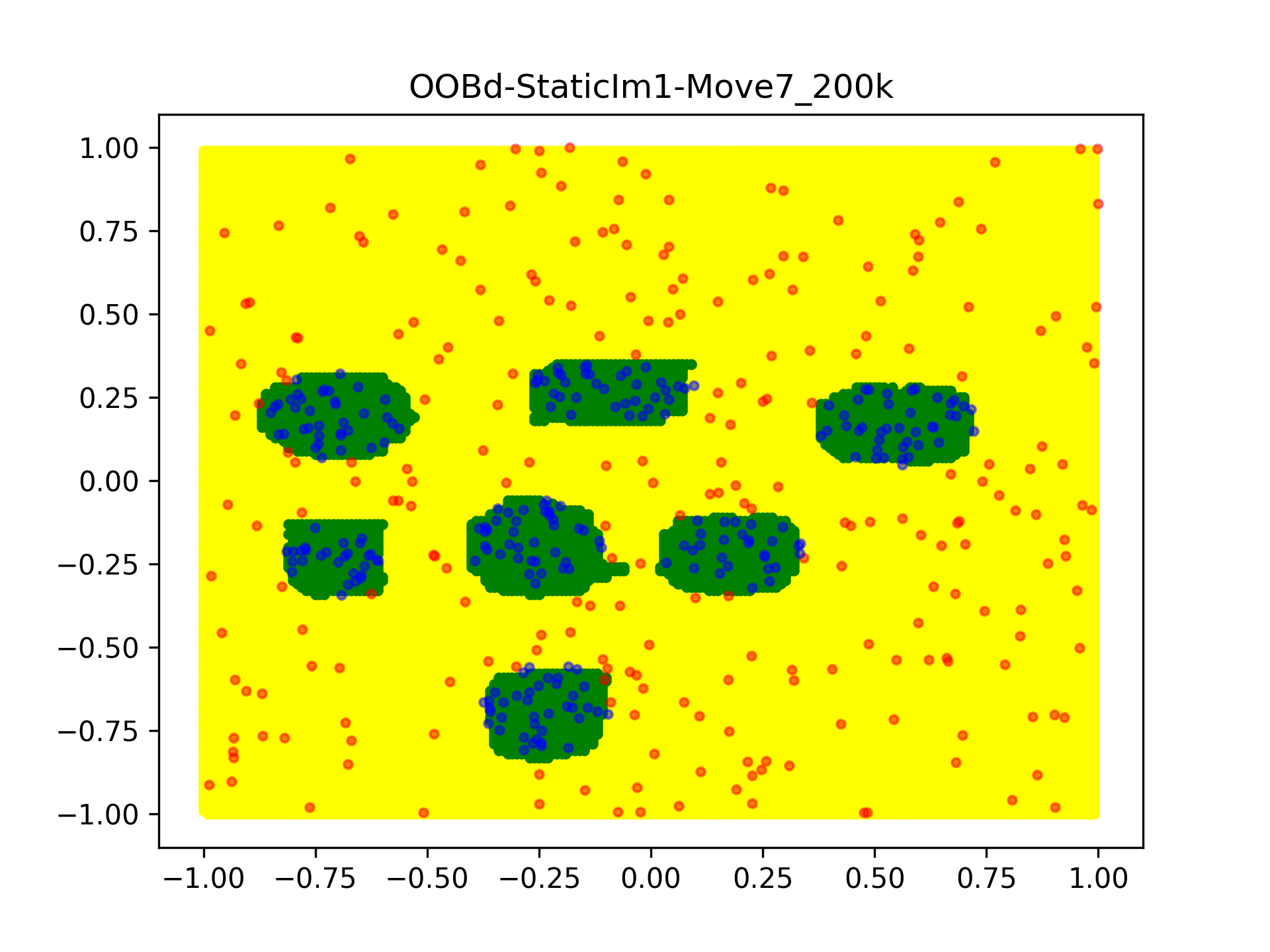} \label{figure:StaticIm1-Move7-dec_bound-OOBd-200k}}
\subfigure[UOB\textsubscript{d}]{\includegraphics[width=0.29\textwidth]{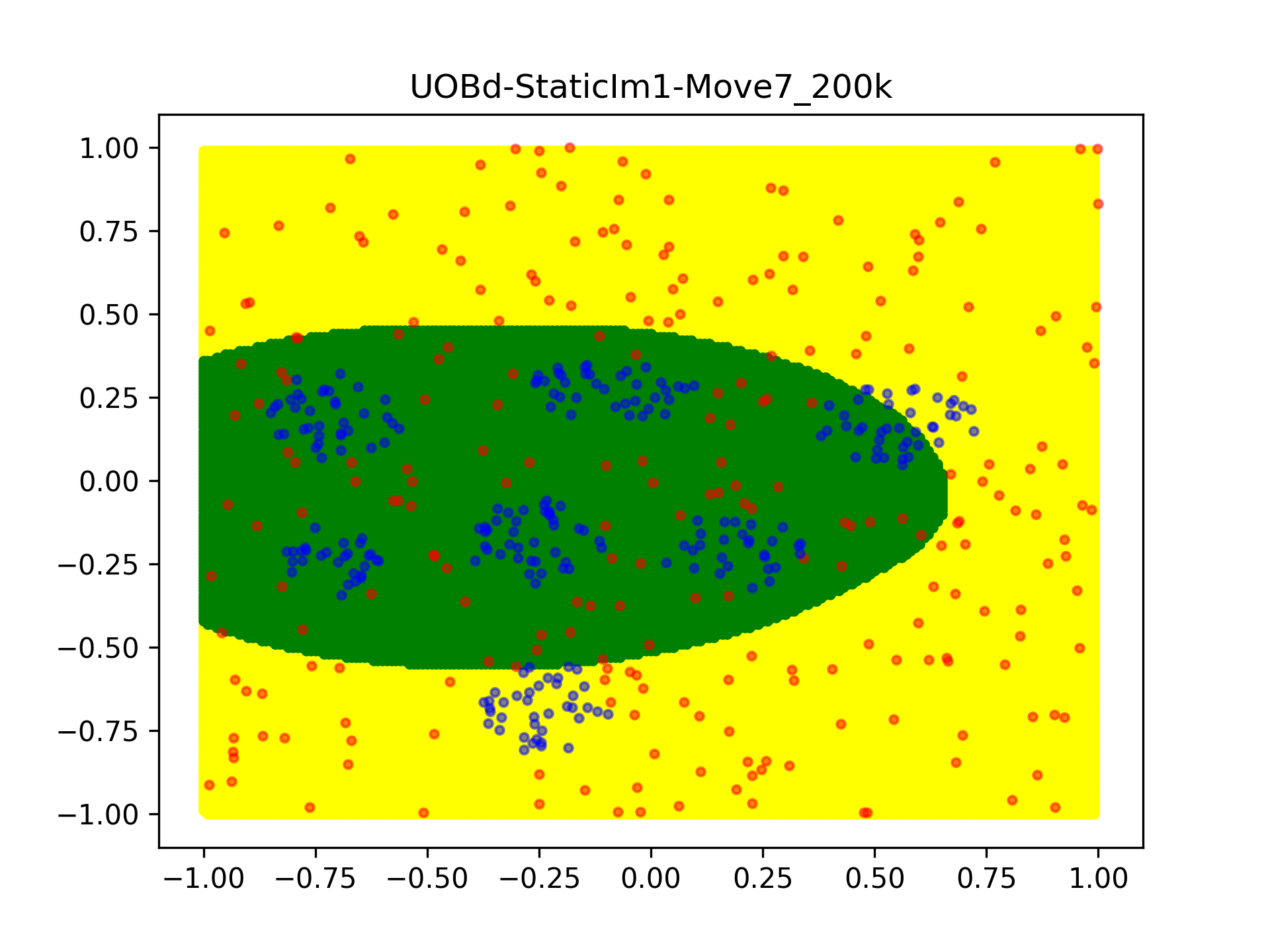} \label{figure:StaticIm1-Move7-dec_bound-UOBd-200k}}
\subfigure[oOS\textsubscript{d}]{\includegraphics[width=0.29\textwidth]{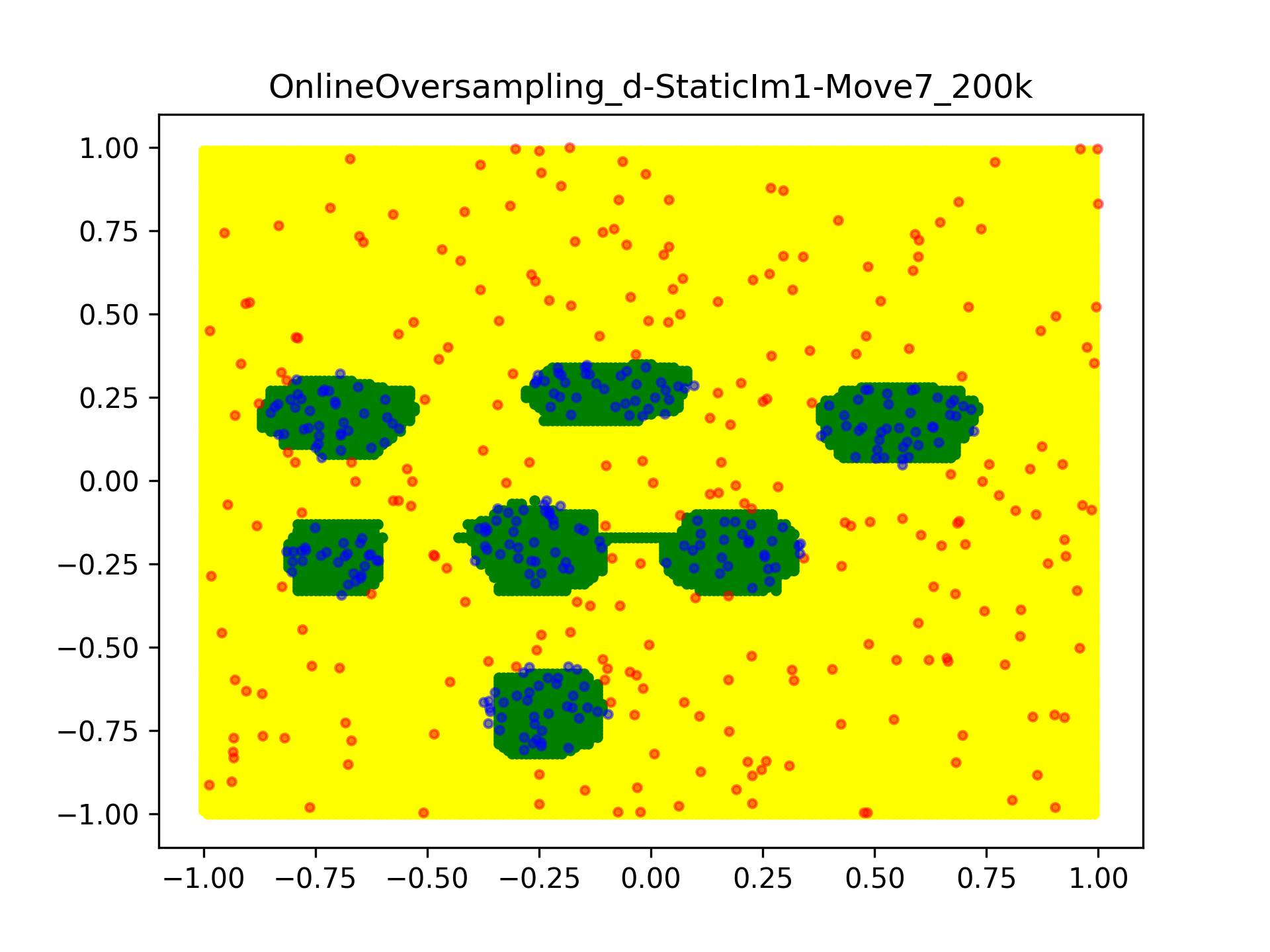} \label{figure:StaticIm1-Move7-dec_bound-OnlineOversampling_d-200k}}
\subfigure[oUnderOverB\textsubscript{d}]{\includegraphics[width=0.29\textwidth]{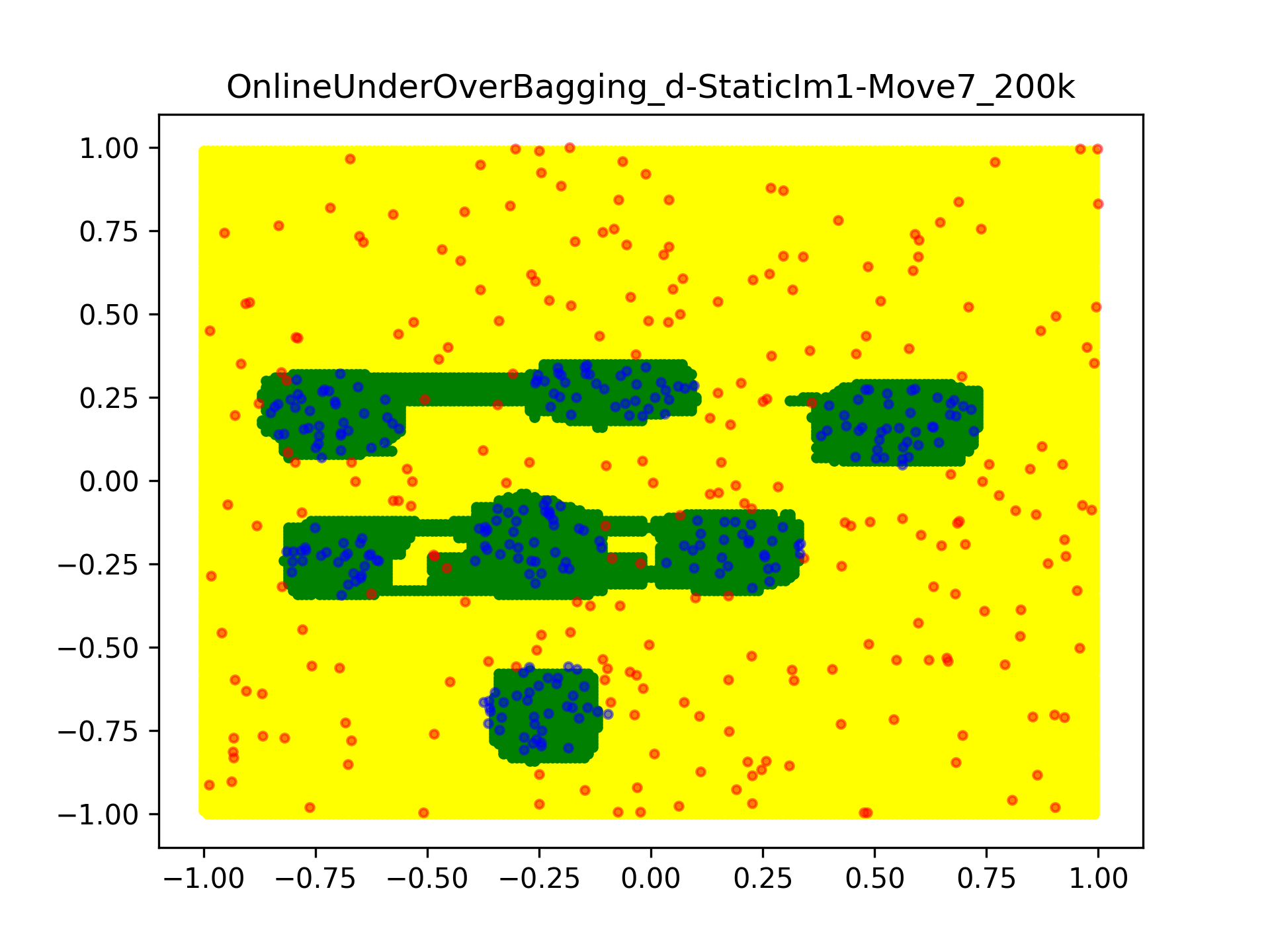} \label{figure:StaticIm1-Move7-dec_bound-OnlineUnderOverBagging_d-200k}}
\subfigure[SMOGauNoise]{\includegraphics[width=0.29\textwidth]{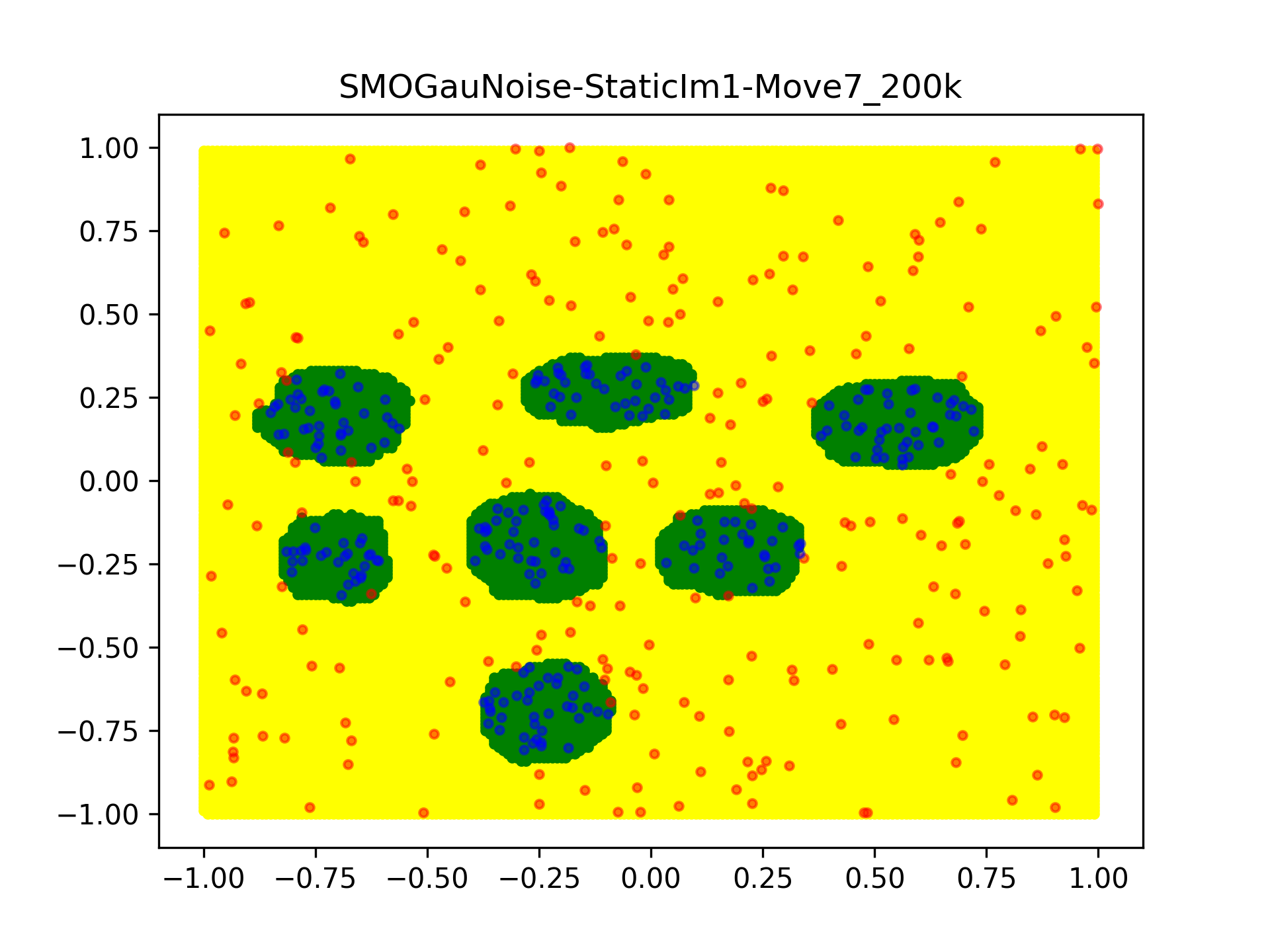} \label{figure:StaticIm1-Move7-dec_bound-SMOGauNoise-200k}}
\subfigure[\reviewII{VFC-SMOTE}]{\includegraphics[width=0.29\textwidth]{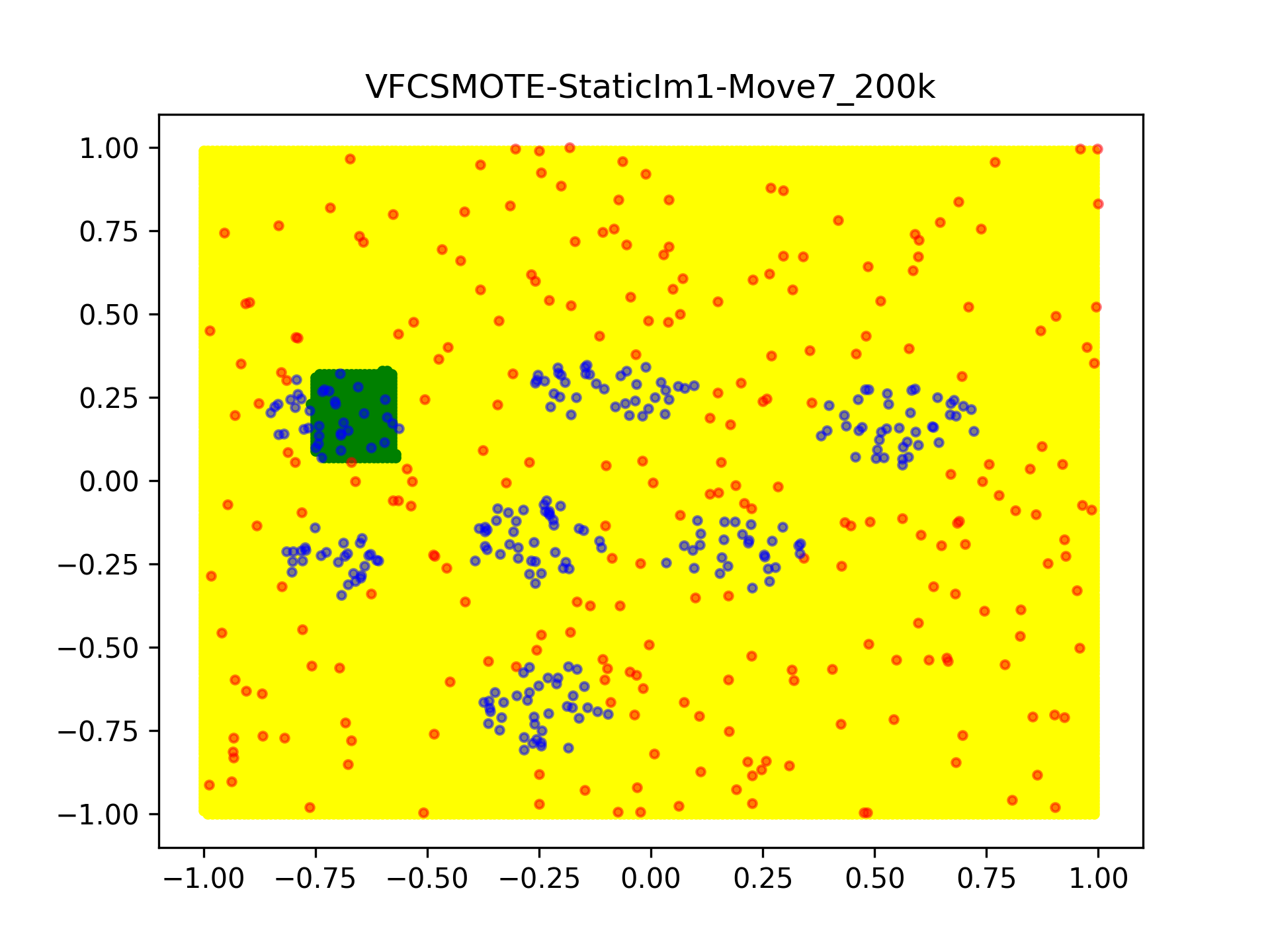} \label{figure:StaticIm1-Move7-dec_bound-VFCSMOTE-200k}}
\subfigure[\reviewII{SMOTE-OB}]{\includegraphics[width=0.29\textwidth]{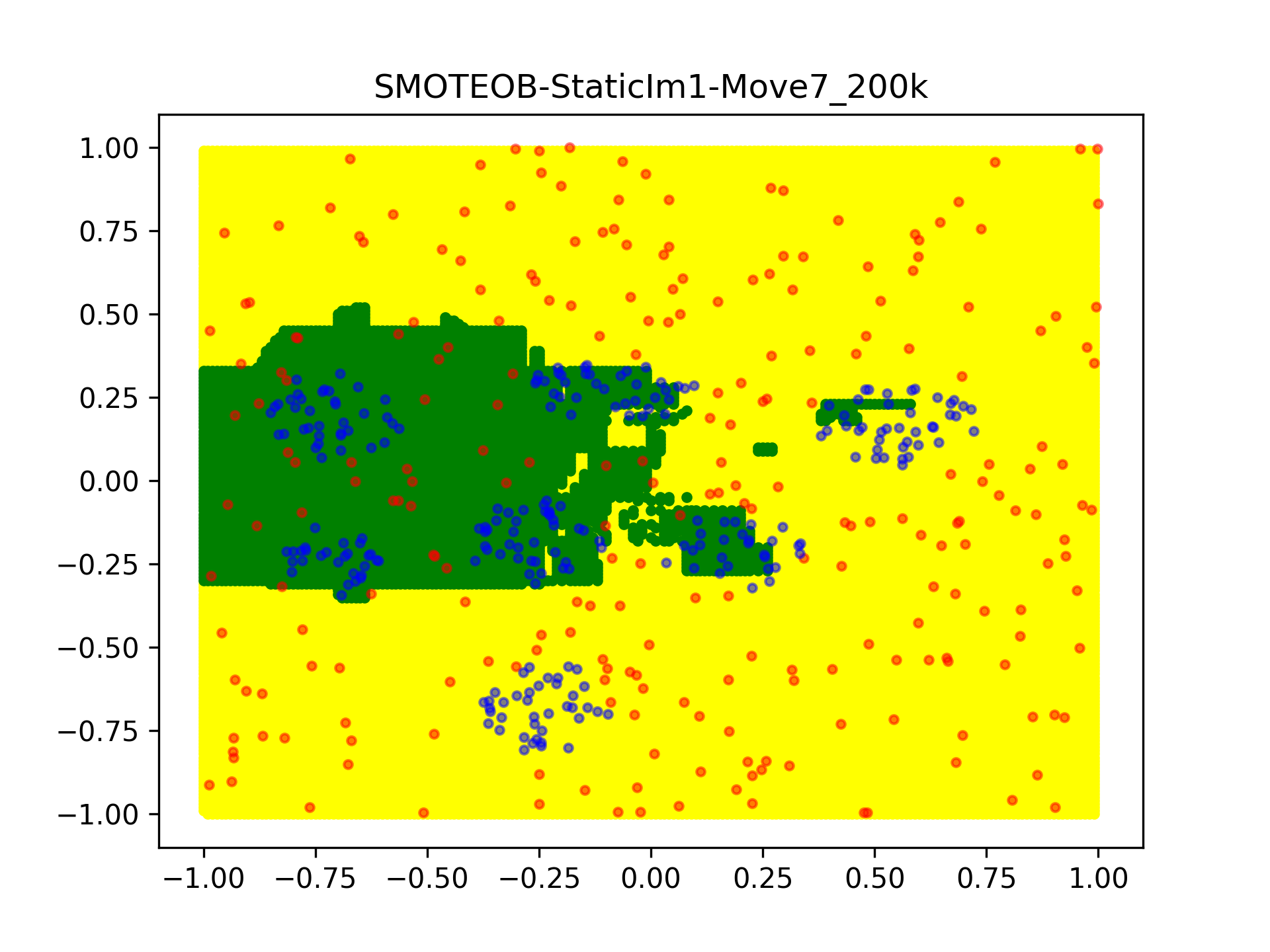} \label{figure:StaticIm1-Move7-dec_bound-SMOTEOB-200k}}
\subfigure[SMOClust]{\includegraphics[width=0.29\textwidth]{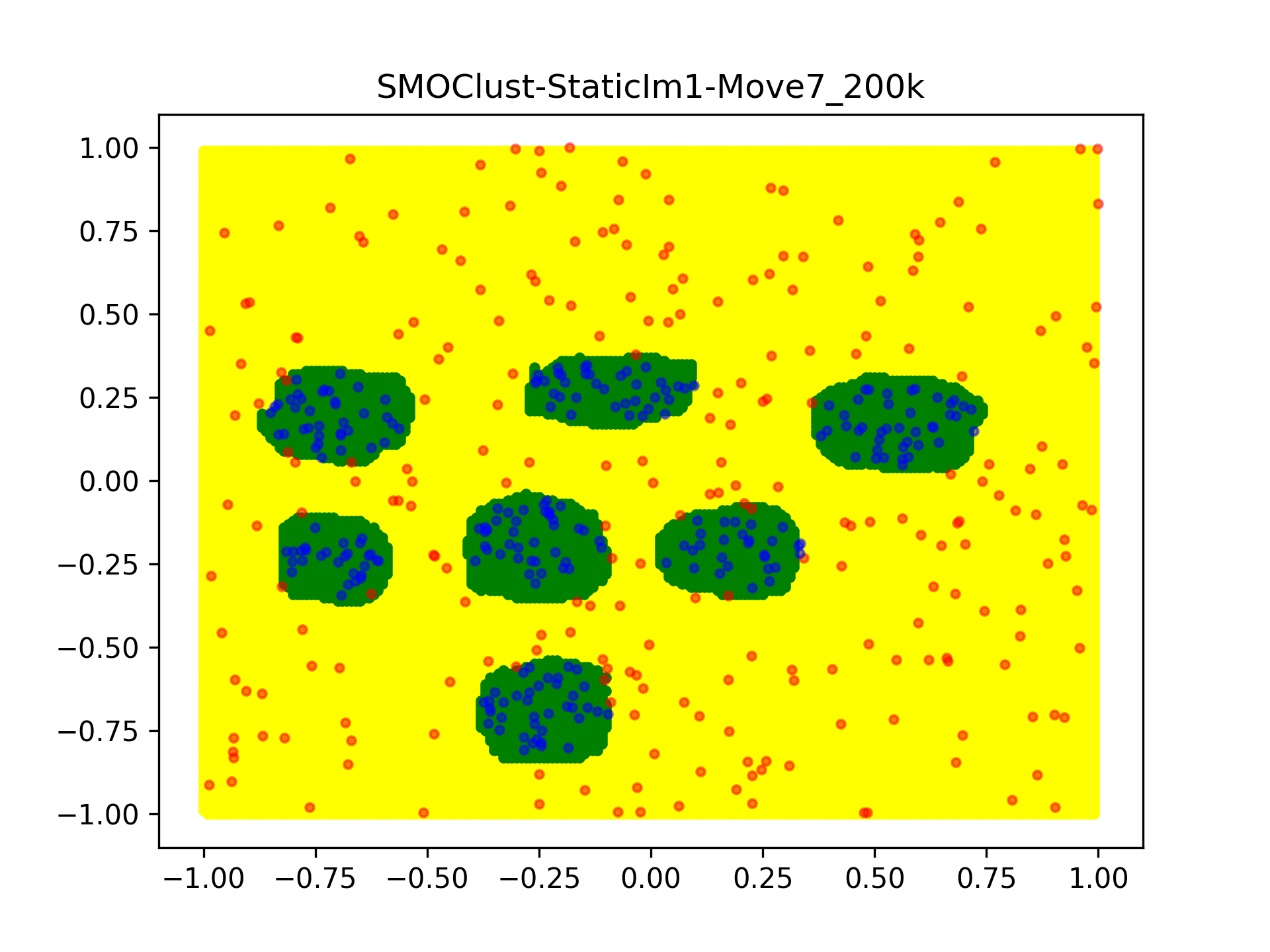} \label{figure:StaticIm1-Move7-dec_bound-SMOClust-200k}}
\caption{Decision Areas Against Class Balanced Test Set at 200k Time Steps (End of Stream) of Two-Dimensional StaticIm1\_Move7}
\label{figure:StaticIm1-Move7-dec_bound-200k}
\end{figure}

Figures \ref{figure:StaticIm1-Move7-dec_bound-70k}, \ref{figure:StaticIm1-Move7-dec_bound-100k} and \ref{figure:StaticIm1-Move7-dec_bound-200k} visualise the learnt decision areas of approaches at the time steps right before and after concept drift (70k and 100k time steps) and at the end (200k time steps) of the two-dimensional StaticIm1\_Move7 stream respectively. The yellow and green regions represent their learnt decision areas of class 0 (majority class) and class 1 (minority class) respectively, while the red and blue dots are the class 0 (majority class) and class 1 (minority class) examples in the class balanced test set, corresponding to the time steps.

First of all, we compare the learnt decision areas of the approaches at the time steps right before concept drift (at 70k time steps). Figure \ref{figure:StaticIm1-Move7-dec_bound-70k} shows that OOB, OnlineOversampling, OnlineUnderOverBagging, SMOGauNoise and SMOClust had learnt decision areas which match the corresponding class balanced test set. This explains why they performed the best before the drift (0-70k time steps, Figure \ref{figure:StaticIm1-Move7-GMean}). Figure \ref{figure:StaticIm1-Move7-dec_bound-SMOGauNoise-70k} and Figure \ref{figure:StaticIm1-Move7-dec_bound-SMOClust-70k} show that the learnt decision areas of SMOClust were similar to SMOGauNoise because they both have strategies to explore the minority class decision boundaries. The expansion by SMOClust was slightly more aggressive than SMOGauNoise, with some sub-areas linked together. Although the proposed synthetic minority oversampling strategy prioritises ``safe'' areas to generate synthetic minority class examples, the strategy of using synthetic examples to train the stream clustering methods may contribute to such aggressiveness in exploring the minority class decision boundaries.

Figures \ref{figure:StaticIm1-Move7-dec_bound-OOB-70k} and \ref{figure:StaticIm1-Move7-dec_bound-OnlineOversampling-70k} show that OOB and OnlineOversampling learnt the most compact minority class decision areas because they reuse the existing minority class examples for oversampling. Figure \ref{figure:StaticIm1-Move7-dec_bound-OnlineUnderOverBagging-70k} shows that the minority class decision areas of OnlineUnderOverBagging were slightly larger than that of OOB and OnlineOversampling. Particularly, there were two green areas linked together. This may be the result of using oversampling and undersampling together, which managed to cover the true minority class clusters while preserving some aggressiveness from undersampling. In contrast, Figures \ref{figure:StaticIm1-Move7-dec_bound-UOB-70k} and \ref{figure:StaticIm1-Move7-dec_bound-UOBd-70k} show that UOB and UOB\textsubscript{d} learnt a single cluster to aggressively cover most minority class areas, considering the small majority class areas in between as part of the minority class. This is likely to cost some predictive performance in the majority class. Thus, we can see that UOB and UOB\textsubscript{d} performed slightly worse than the other approaches before the concept drift (0-70k time steps, Figure \ref{figure:StaticIm1-Move7-GMean}). However, Figure \ref{figure:SMOClust-Avg GMean Artificial Severe} shows that UOB performed slightly better than SMOClust in the five-dimensional StaticIm1\_Move7 stream, indicating that the aggressive nature of undersampling may be an advantage in learning the minority class when the feature space is sparse and presents very few minority class examples. When the feature space is more compact, the proposed strategy in SMOClust is more advantageous.

\reviewII{Considering OOB\textsubscript{d}, OnlineUnderOverBagging\textsubscript{d}, VFC-SMOTE, and SMOTE-OB, Figures \ref{figure:StaticIm1-Move7-dec_bound-OOBd-70k}, \ref{figure:StaticIm1-Move7-dec_bound-OnlineUnderOverBagging_d-70k}, \ref{figure:StaticIm1-Move7-dec_bound-VFCSMOTE-70k}, and \ref{figure:StaticIm1-Move7-dec_bound-SMOTEOB-70k} show that their learnt minority decision areas were very small which only covered a small proportion of the true minority class areas. In the case of VFC-SMOTE, it predicted every example as majority class at 70k time steps. As previously mentioned, their predictive performance fluctuated a lot throughout the stream. So, it can be deduced that they were greatly affected by false-positive drift detections.}

Over the next paragraphs, we compare the predictive performance and the decision boundaries of SMOClust against other approaches at the time steps right after concept drift (at 100k time steps) and at the end of the data stream (at 200k time steps), to understand how SMOClust handles concept drift of moving minority class sub-clusters when the data stream is severely class imbalanced. 

Figure \ref{figure:StaticIm1-Move7-GMean} shows that the predictive performance of SMOClust fluctuated during the drift (70k-100k time steps, Figure \ref{figure:StaticIm1-Move7-GMean}). Thus, it is likely that its base learner had been reset several times due to drift detection. Yet, it was the fastest approach to recovering predictive performance from the drift. Figure \ref{figure:StaticIm1-Move7-dec_bound-100k} presents the learnt decision boundaries right after the drift. It shows that SMOClust and SMOGauNoise made the best attempt in adapting the drift. They were able to cover most minority class sub-clusters at the new position, especially SMOClust. The potential reason is that, although the base learner of SMOClust is reset upon drift detection, the stream clustering methods are not reset as they are expected to be drift adaptable. Therefore, SMOClust is more robust to incremental and gradual drifts than SMOGauNoise, explaining its rapid predictive performance recovery from the drift. 

On the other hand, Figures \ref{figure:StaticIm1-Move7-dec_bound-OOB-100k}, \ref{figure:StaticIm1-Move7-dec_bound-UOB-100k}, \ref{figure:StaticIm1-Move7-dec_bound-OnlineOversampling-100k}, and \ref{figure:StaticIm1-Move7-dec_bound-OnlineUnderOverBagging-100k} show that the learnt minority class decision areas of OOB, UOB, OnlineOversampling and OnlineUnderOverBagging mainly retained at the pre-drift position because they are not concept drift adaptable. Their concept drift adaptable counterparts \reviewII{, VFC-SMOTE and SMOTe-OB} did not handle the drift very well either. Figures \ref{figure:StaticIm1-Move7-dec_bound-OOBd-100k}, \ref{figure:StaticIm1-Move7-dec_bound-UOBd-100k}, \ref{figure:StaticIm1-Move7-dec_bound-OnlineOversampling_d-100k}, \ref{figure:StaticIm1-Move7-dec_bound-OnlineUnderOverBagging_d-100k}\reviewII{, \ref{figure:StaticIm1-Move7-dec_bound-VFCSMOTE-100k}, and \ref{figure:StaticIm1-Move7-dec_bound-SMOTEOB-100k}} show that their learnt minority class decision areas only covered a few minority class sub-clusters at the post-drift position, which is likely because their base learners had been reset for several times caused by drift detection and they do not have any strategy to deal with incremental and gradual drifts. As the result, they struggled to recover their predictive performance from the drift, as shown in Figure \ref{figure:StaticIm1-Move7-GMean}.

Lastly, we compare the learnt decision areas of the approaches at the end of the two-dimensional StaticIm1\_Move7 stream. Figure \ref{figure:StaticIm1-Move7-dec_bound-200k} shows that OOB\textsubscript{d}, SMOGauNoise and SMOClust are the best approaches in converging to the post-drift position of minority class sub-clusters. In particular, a few green areas of SMOClust and SMOGauNoise were slightly less compact than OOB\textsubscript{d}, showing that SMOClust and SMOGauNoise had slightly better generalisation than OOB\textsubscript{d}.

Figures \ref{figure:StaticIm1-Move7-dec_bound-OOB-200k}, \ref{figure:StaticIm1-Move7-dec_bound-OnlineOversampling-200k}, and \ref{figure:StaticIm1-Move7-dec_bound-OnlineUnderOverBagging-200k} show that OOB, OnlineOversampling and OnlineUnderOverBagging managed to converge to the new concept after the drift. However, they also retained a small portion of green areas which corresponds to the pre-drift position of the minority class. This shows that OOB, OnlineOversampling and OnlineUnderOverBagging can adapt to concept drift involving minority class sub-cluster movement. However, they required a longer period to adapt as they were hindered by the knowledge acquired pre-drift. Meanwhile, Figures \ref{figure:StaticIm1-Move7-dec_bound-OOBd-200k}, \ref{figure:StaticIm1-Move7-dec_bound-OnlineOversampling_d-200k}, and \ref{figure:StaticIm1-Move7-dec_bound-OnlineUnderOverBagging_d-200k} show that their concept drift adaptable counterparts adapted better, except OnlineUnderOverBagging\textsubscript{d}. While resetting base learners helps to adapt to concept drift, OnlineUnderOverBagging\textsubscript{d} partly uses undersampling in its strategy to deal with class imbalance led to some over-generalisation between the learnt minority class areas. UOB and UOB\textsubscript{d} use undersampling to deal with class imbalance, thus Figures \ref{figure:StaticIm1-Move7-dec_bound-OOB-200k} and \ref{figure:StaticIm1-Move7-dec_bound-UOBd-200k} show that they had the greatest over generalisation due to the aggressive nature of undersampling. \reviewII{VFC-SMOTE and SMOTE-OB continued to struggle, as shown in Figures \ref{figure:StaticIm1-Move7-dec_bound-VFCSMOTE-200k} and \ref{figure:StaticIm1-Move7-dec_bound-SMOTEOB-200k}, because of frequent false-positive drift detections.}

%\noindent\fbox{\begin{minipage}{\textwidth}
\begin{shadequote}
\textit{Short Summary: Through the pre-drift analysis, the ability of SMOClust in handling stationary severely class imbalanced data streams presenting several minority class sub-clusters is validated. In particular, it shows that SMOClust was able to learn and explore the true decision boundaries despite the data stream presents very few minority class examples. The post-drift analysis shows that SMOClust was more robust in adapting incremental and gradual drift involving minority class sub-clusters movement than existing approaches. Although most of the approaches converged to the new concept at the end of the data stream, SMOClust was the best and the fastest approach in recovering predictive performance from the drift. The inconsistent results between two and five-dimensional versions of this representative case indicate that SMOClust may be more advantageous in lower-dimensional data streams.}
\end{shadequote}
%\end{minipage}}

% Concluding the comparison of learnt decision areas at the time steps right before and after concept drift, and at the end of two-dimensional StaticIm1\_Move7 stream, we can see that SMOClust handled severely class imbalanced data streams similar or better than compared approaches. Also, SMOClust had the best adaptability to concept drift of minority class clusters movement among compared approaches in this severely class imbalanced data stream.

\subsubsection{Cases where SMOClust performed worse} \label{section:SMOClust-analysis-artificial data streams-worse}

This section discusses the situations where SMOClust performed worse than other approaches, particularly in cases with concept drift leading to 100\% rare minority examples. StaticIm10\_Rare100 stream was chosen from Table \ref{figure:SMOClust-Avg GMean Artificial} as the representative case to discuss the behaviour of SMOClust in detail. Following the method of analysis in Section \ref{section:SMOClust-analysis-artificial data streams-better}, we also created a two-dimensional version of StaticIm10\_Rare100 stream such that we can visualise and compare the learnt decision boundaries of the approaches to understand their behaviour.

\begin{figure}[!ht]
\centering
\includegraphics[width=\textwidth]{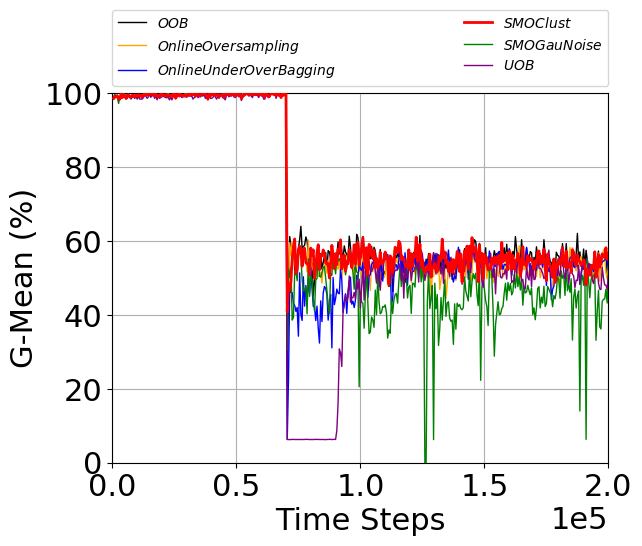}
% \vspace{-0.4cm}
\caption{Periodic Class Balanced Holdout Test G-Mean Against Time Steps in Two-Dimensional StaticIm10\_Rare100}
\label{figure:StaticIm10-Rare100-GMean}
\end{figure}

Table \ref{table:SMOClust-Average GMean 2D worse} presents the approaches' thirty runs average G-Mean on the two-dimensional StaticIm10\_Rare100 stream. It shows that SMOClust performed better than most other approaches. Figure \ref{figure:StaticIm10-Rare100-GMean}, showing the G-Mean of the approaches in their median run\footnote{Median run refers to the run that leads to the median of predictive performances averaged across time steps.} throughout the two-dimensional StaticIm10\_Rare100 stream, also supports the results on Table \ref{table:SMOClust-Average GMean 2D worse}. \review{Note that, to improve readability, we have omitted the predictive performance of OOB\textsubscript{d}, UOB\textsubscript{d}, oOS\textsubscript{d}, oUnderOverB\textsubscript{d}\reviewII{, VFC-SMOTE and SMOTE-OB} from Figure \ref{figure:StaticIm10-Rare100-GMean}, similar to Figure \ref{figure:StaticIm1-Move7-GMean}, due to their values fluctuating significantly throughout the stream. For a comparison of SMOClust against these approaches, please refer to the supplementary document.}

As these results are not consistent with the results of the five-dimensional StaticIm10\_Rare100 stream, shown in Table \ref{table:SMOClust-Friedman Ranks-GMean-artificial} and \ref{table:SMOClust-Friedman Ranks-GMean-artificial severe}, we preliminary checked if using a different set of random seeds or picking another case that involves drift leading to 100\% rare minority class examples would yield results that are consistent with Table \ref{table:SMOClust-Friedman Ranks-GMean-artificial} and \ref{table:SMOClust-Friedman Ranks-GMean-artificial severe}. Yet, it still shows that SMOClust performed similar to or better than other approaches in two-dimensional StaticIm10\_Rare100 stream. Thus, in this analysis, we focus on why SMOClust can handle concept drift leading to 100\% rare minority class examples than other approaches when the data stream has only two dimensions while attempting to deduce why it could not when the data stream has five dimensions.

\begin{table}
\footnotesize
\centering
% \caption{30 Runs Average G-Mean on Two-Dimensional Version of Representative Artificial Data Streams where SMOClust Performed Worse (A12 SMOClust vs Others)}
\caption{30 Runs Average G-Mean on Two-Dimensional Version of Representative Artificial Data Streams where SMOClust Performed Worse}
\label{table:SMOClust-Average GMean 2D worse}
\renewcommand\tabcolsep{3pt}
\begin{threeparttable}
\begin{tabular}{c|cccccc}
\hline
Stream & OOB & UOB & oOS & \makecell{oUnder- \\ OverB} & OOB\textsubscript{d} & UOB\textsubscript{d} \\
\hline
\hline
% \multicolumn{11}{c}{\cellcolor[hsb]{0.95,0.95,0.95} Set 3'} \\
% \hline
StaticIm10\_Rare100 & \cellcolor[hsb]{0.083,0.029,1} 70.61\% & \cellcolor[hsb]{0.25,0.667,1} 63.65\% & \cellcolor[hsb]{0.25,0.118,1} 69.14\% & \cellcolor[hsb]{0.25,0.213,1} 68.19\% & \cellcolor[hsb]{0.25,0.483,1} 65.49\% & \cellcolor[hsb]{0.25,0.215,1} 68.17\% \\
\hline
\hline
Stream & oOS\textsubscript{d} & \makecell{oUnder- \\ OverB\textsubscript{d}} & \makecell{SMO- \\ GauNoise} & \makecell{VFC- \\ SMOTE} & \makecell{SMOTE- \\ OB} & SMOClust \\
\hline
\hline
StaticIm10\_Rare100 & \cellcolor[hsb]{0.25,0.528,1} 65.04\% & \cellcolor[hsb]{0.25,0.535,1} 64.98\% & \cellcolor[hsb]{0.25,0.576,1} 64.56\% & \cellcolor[hsb]{0.25,1.0,1} 54.64\% & \cellcolor[hsb]{0.25,1.0,1} 58.98\% & 70.32\% \\
\hline
\end{tabular}
\begin{tablenotes}
\begin{footnotesize}
\item[-] Based on the average G-Mean, cells are highlighted in lime / light grey when SMOClust performed better than the corresponding approach and cells are highlighted in orange / dark grey cells when SMOClust performed worse than the corresponding approach. The colour intensity scales with the absolute difference of average G-Mean between the SMOClust and the approach of the column and the intensity reaches the maximum when such difference is $\geq 10\%$.
% \item[-] Symbols [*], [s], [m] and [b] represent insignificant, small, medium and large A12 effect size against SMOClust. Presence/absence of the sign “-” in the effect size means that the corresponding approach was worse/better than SMOClust.
\end{footnotesize}
\end{tablenotes}
\end{threeparttable}
\end{table}

\begin{figure}[!ht]
\centering
\subfigure[OOB]{\includegraphics[width=0.29\textwidth]{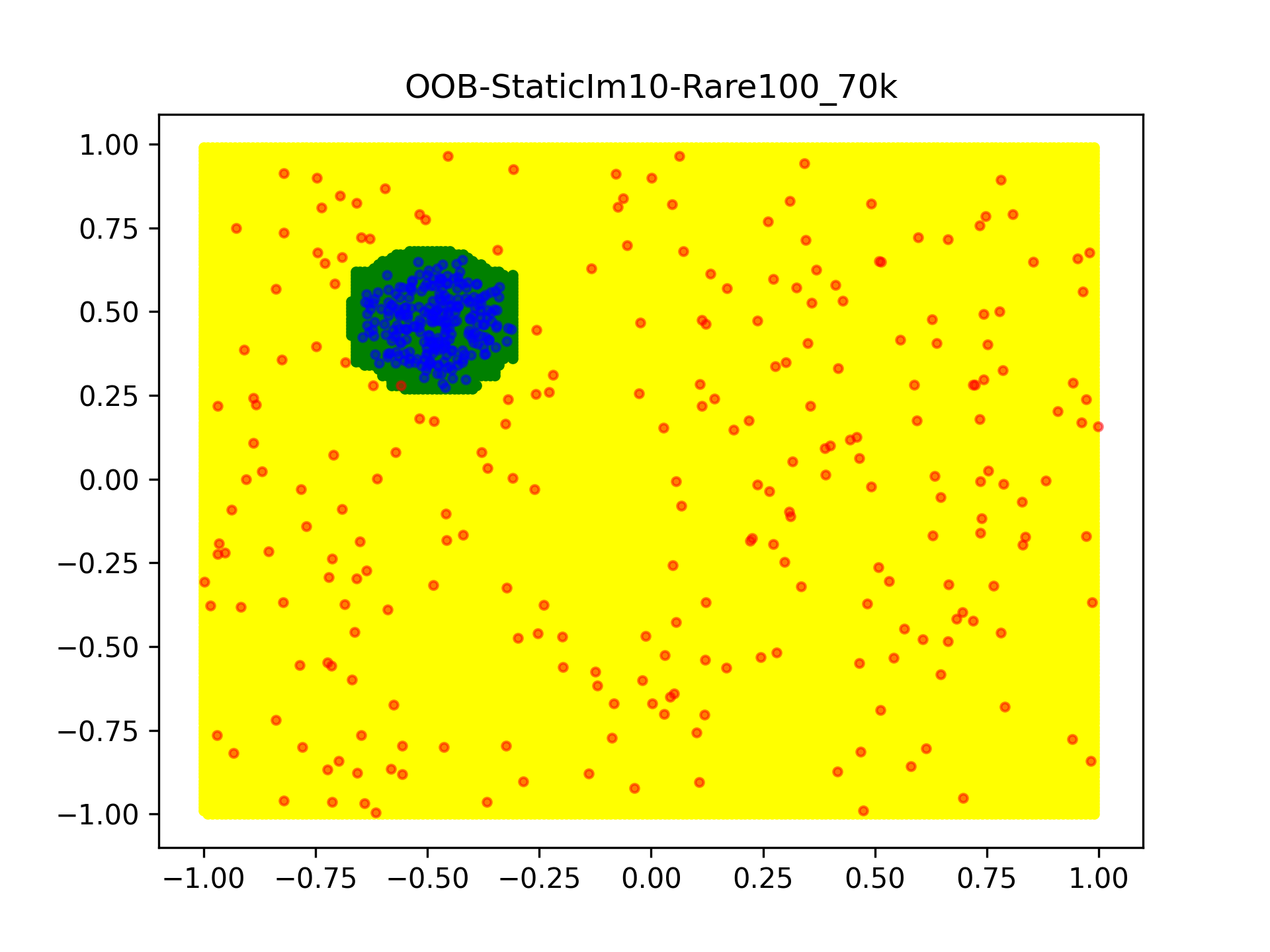} \label{figure:StaticIm10-Rare100-dec_bound-OOB-70k}}
\subfigure[UOB]{\includegraphics[width=0.29\textwidth]{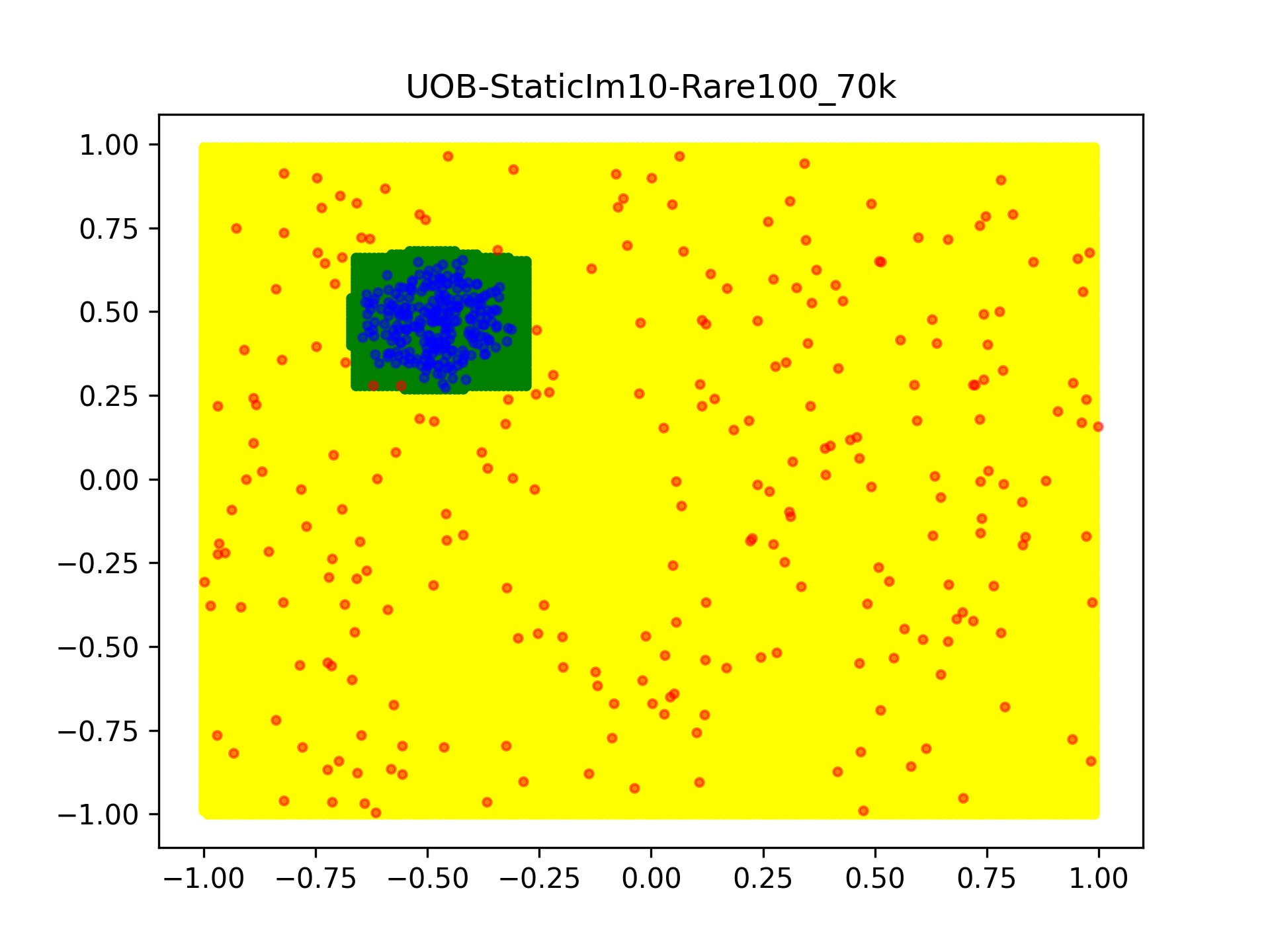} \label{figure:StaticIm10-Rare100-dec_bound-UOB-70k}}
\subfigure[oOS]{\includegraphics[width=0.29\textwidth]{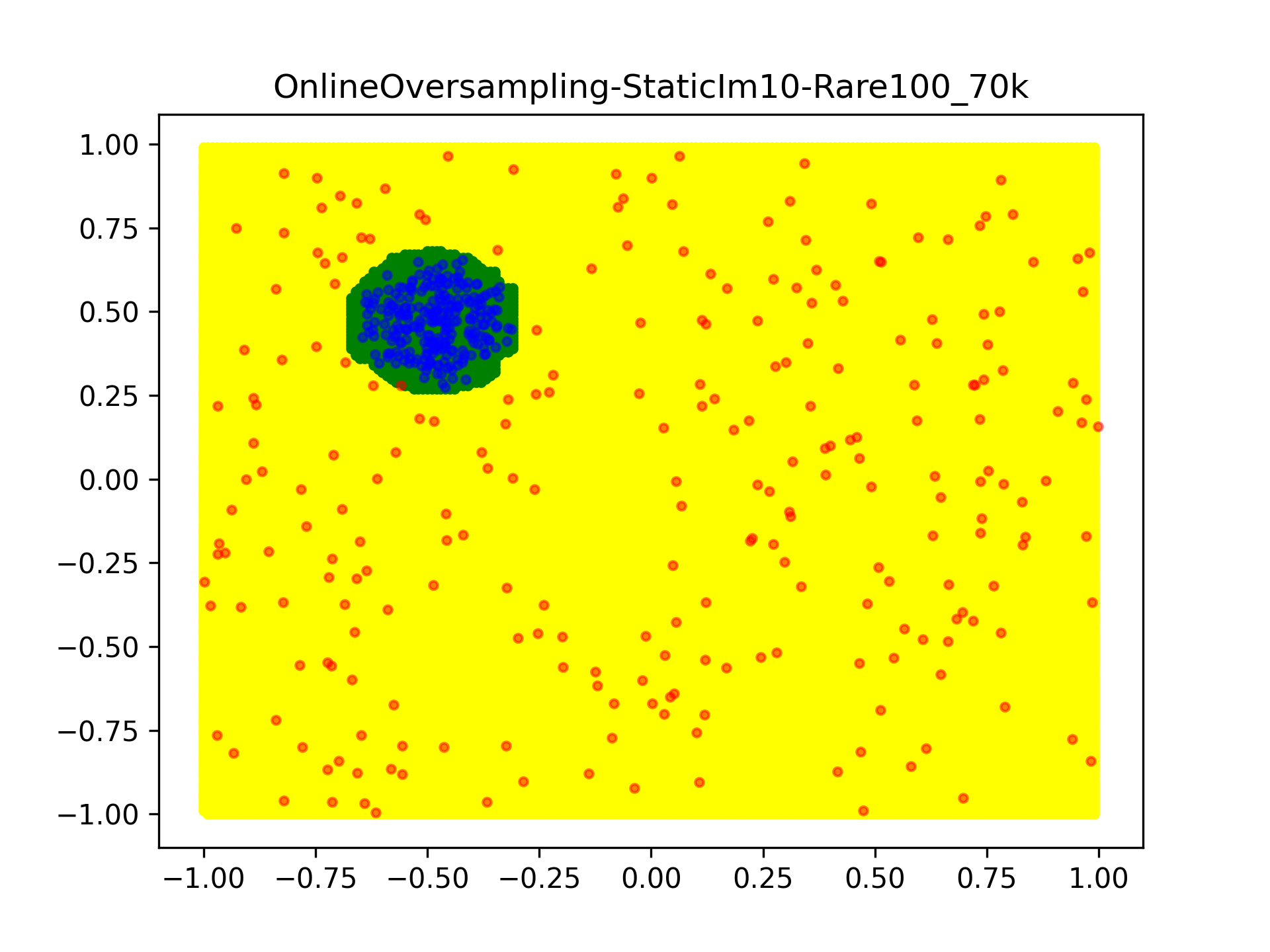} \label{figure:StaticIm10-Rare100-dec_bound-OnlineOversampling-70k}}
\subfigure[oUnderOverB]{\includegraphics[width=0.29\textwidth]{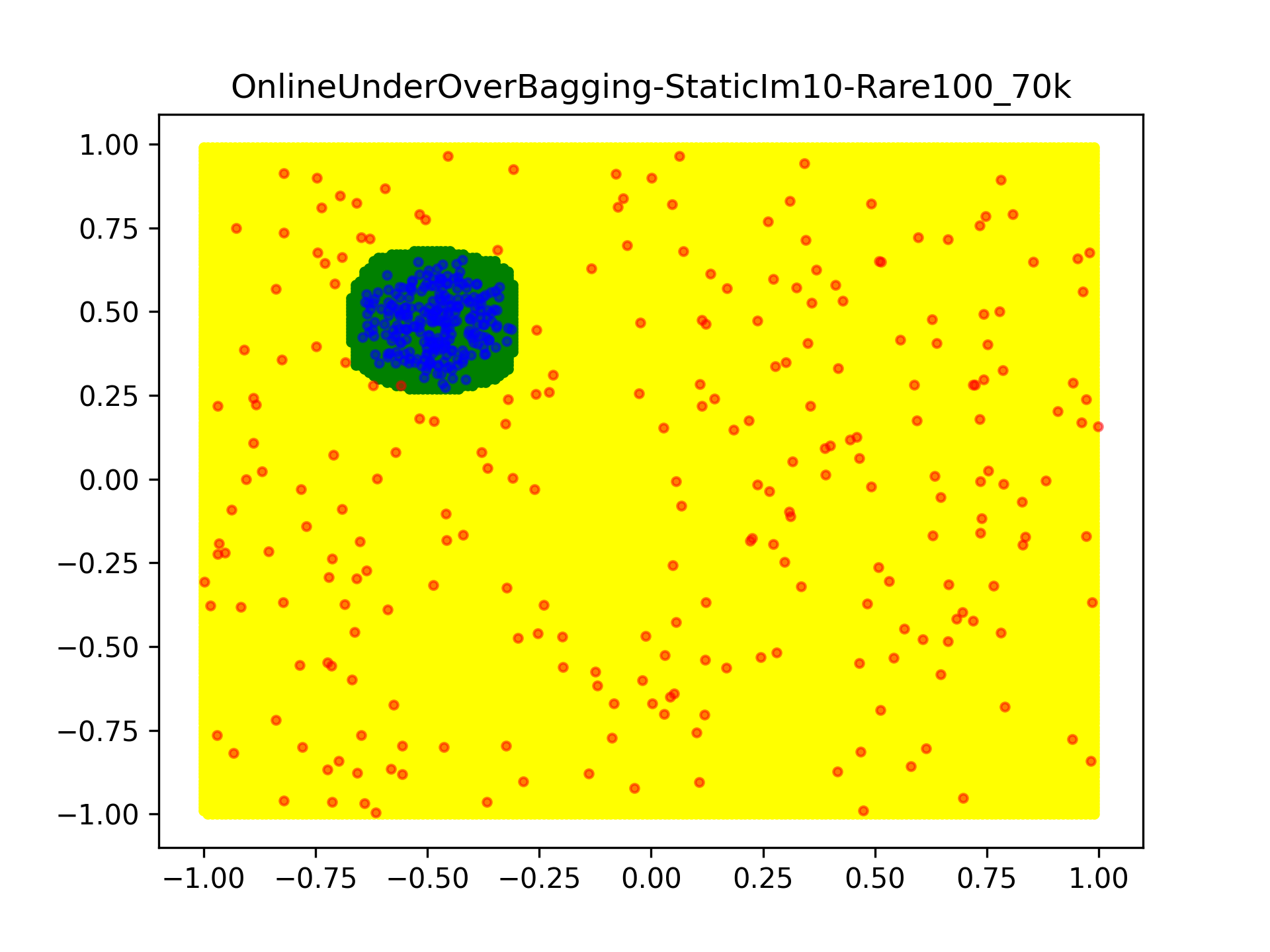} \label{figure:StaticIm10-Rare100-dec_bound-OnlineUnderOverBagging-70k}}
\subfigure[OOB\textsubscript{d}]{\includegraphics[width=0.29\textwidth]{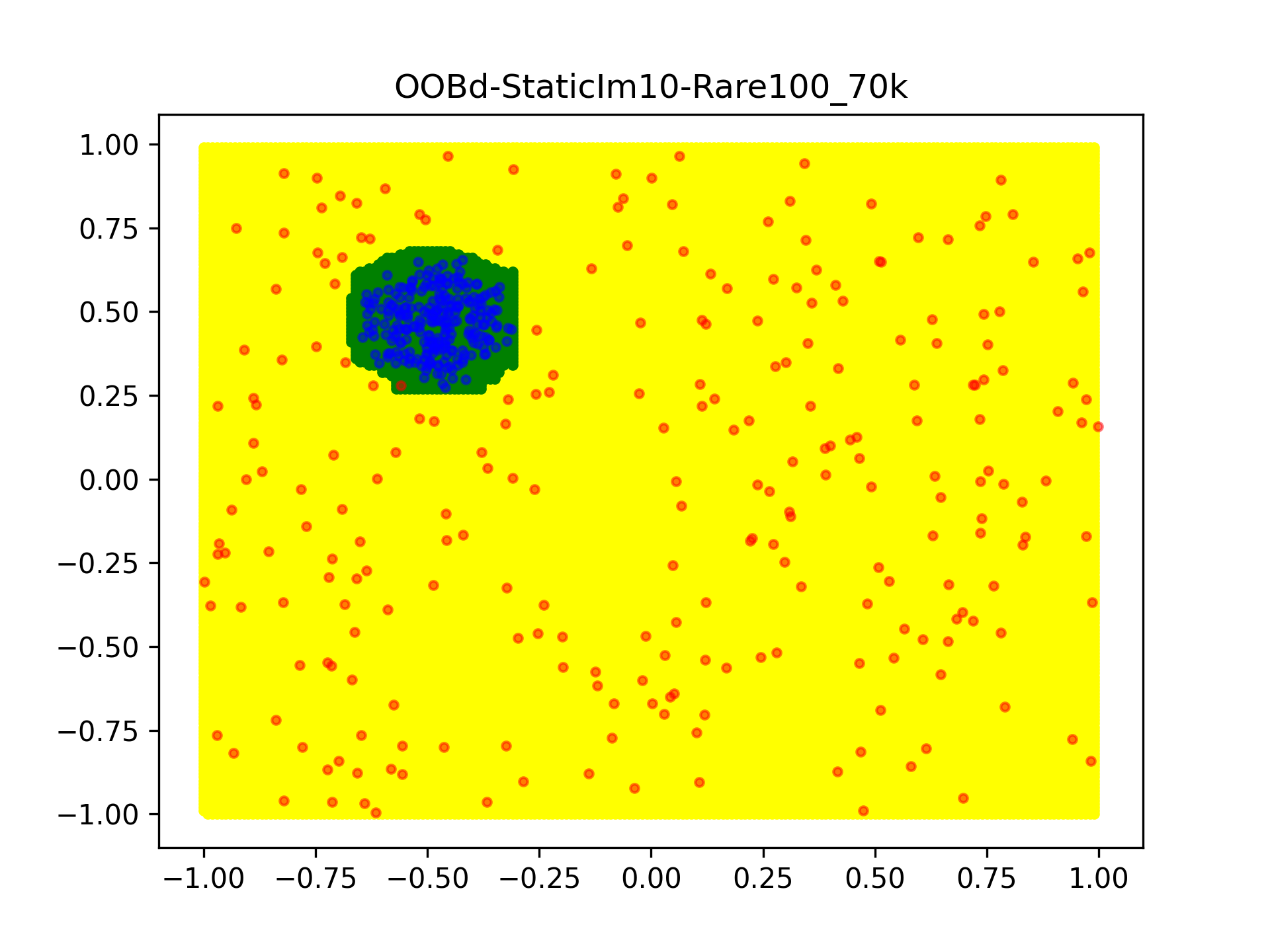} \label{figure:StaticIm10-Rare100-dec_bound-OOBd-70k}}
\subfigure[UOB\textsubscript{d}]{\includegraphics[width=0.29\textwidth]{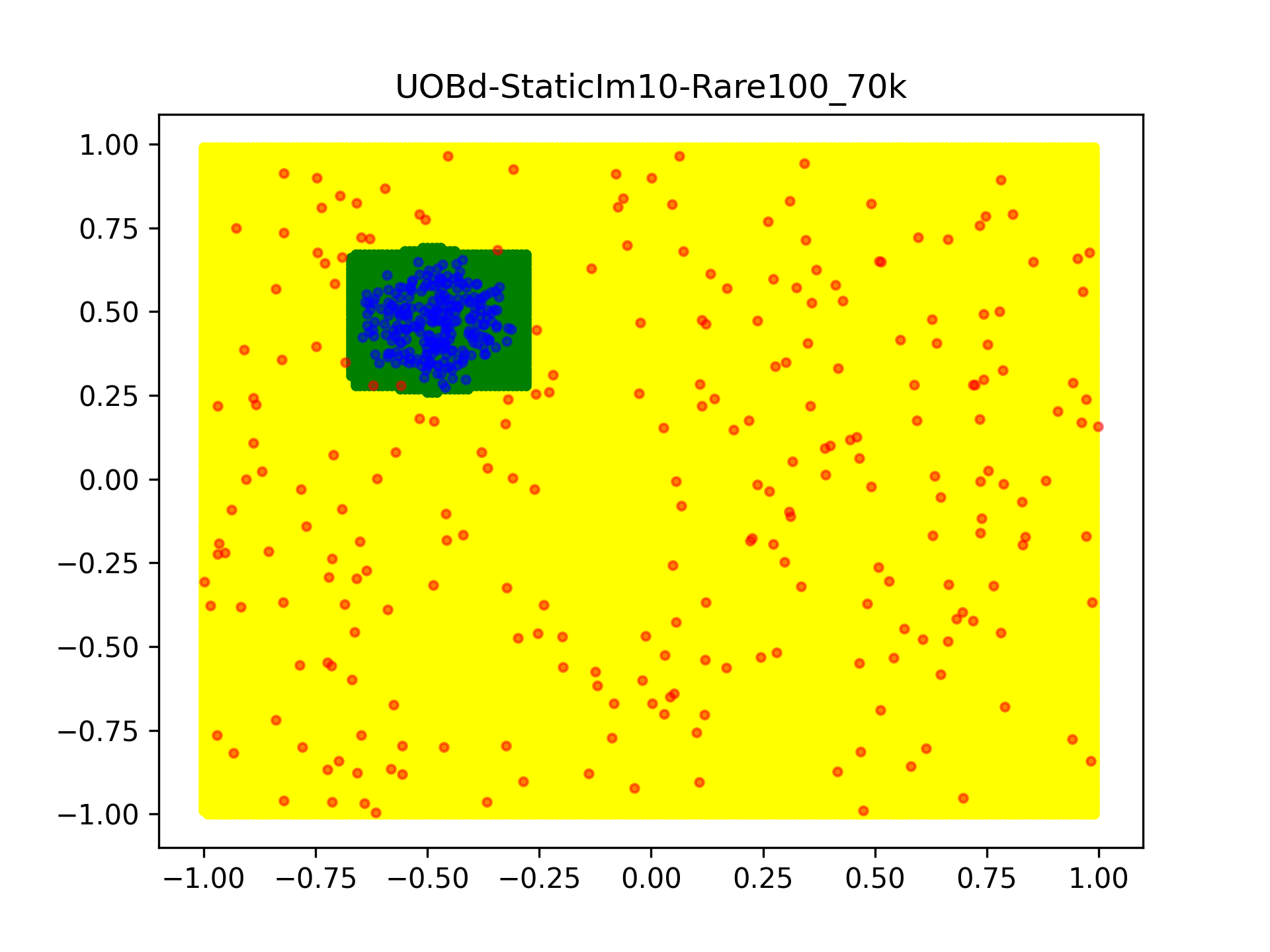} \label{figure:StaticIm10-Rare100-dec_bound-UOBd-70k}}
\subfigure[oOS\textsubscript{d}]{\includegraphics[width=0.29\textwidth]{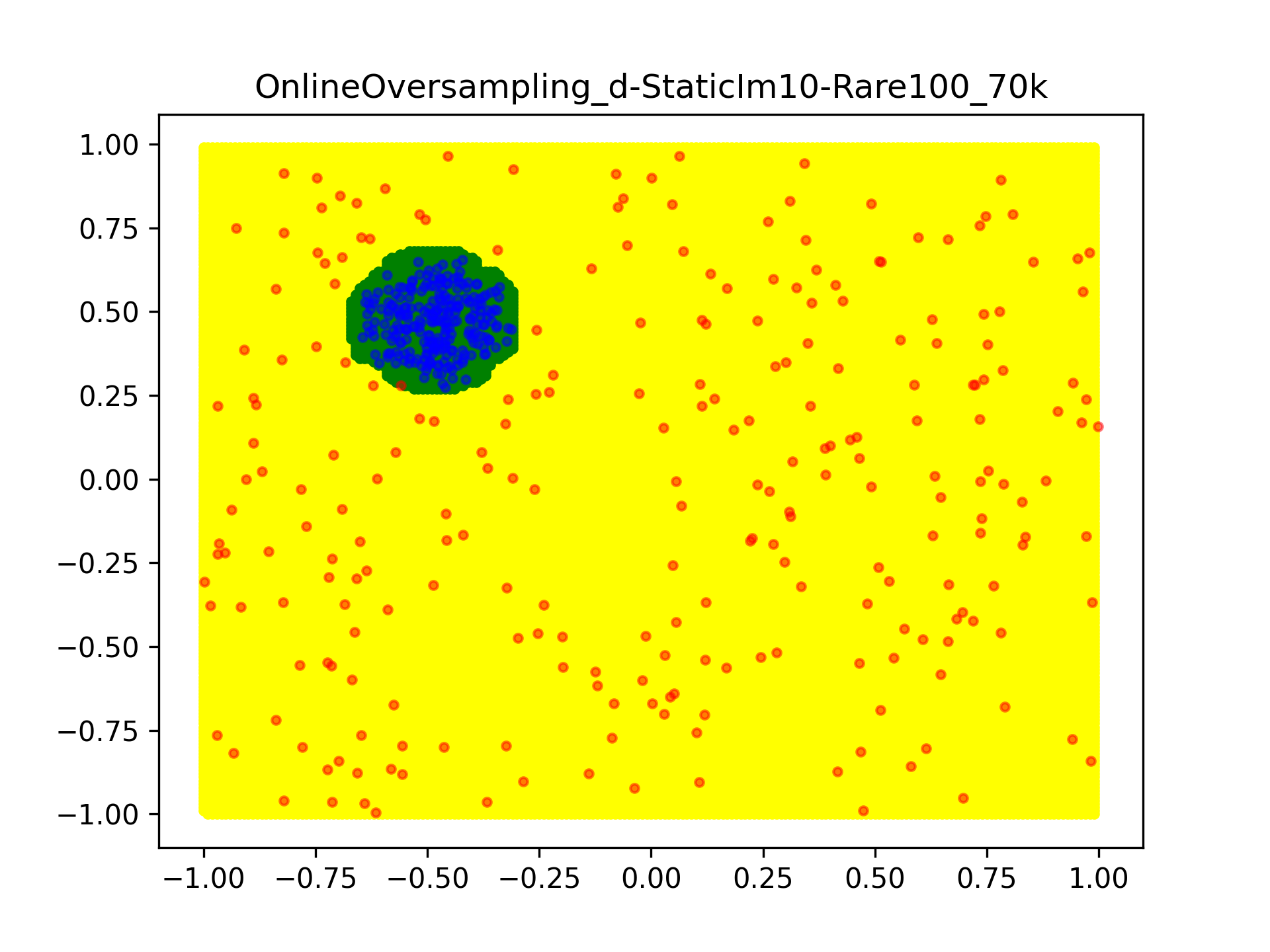} \label{figure:StaticIm10-Rare100-dec_bound-OnlineOversampling_d-70k}}
\subfigure[oUnderOverB\textsubscript{d}]{\includegraphics[width=0.29\textwidth]{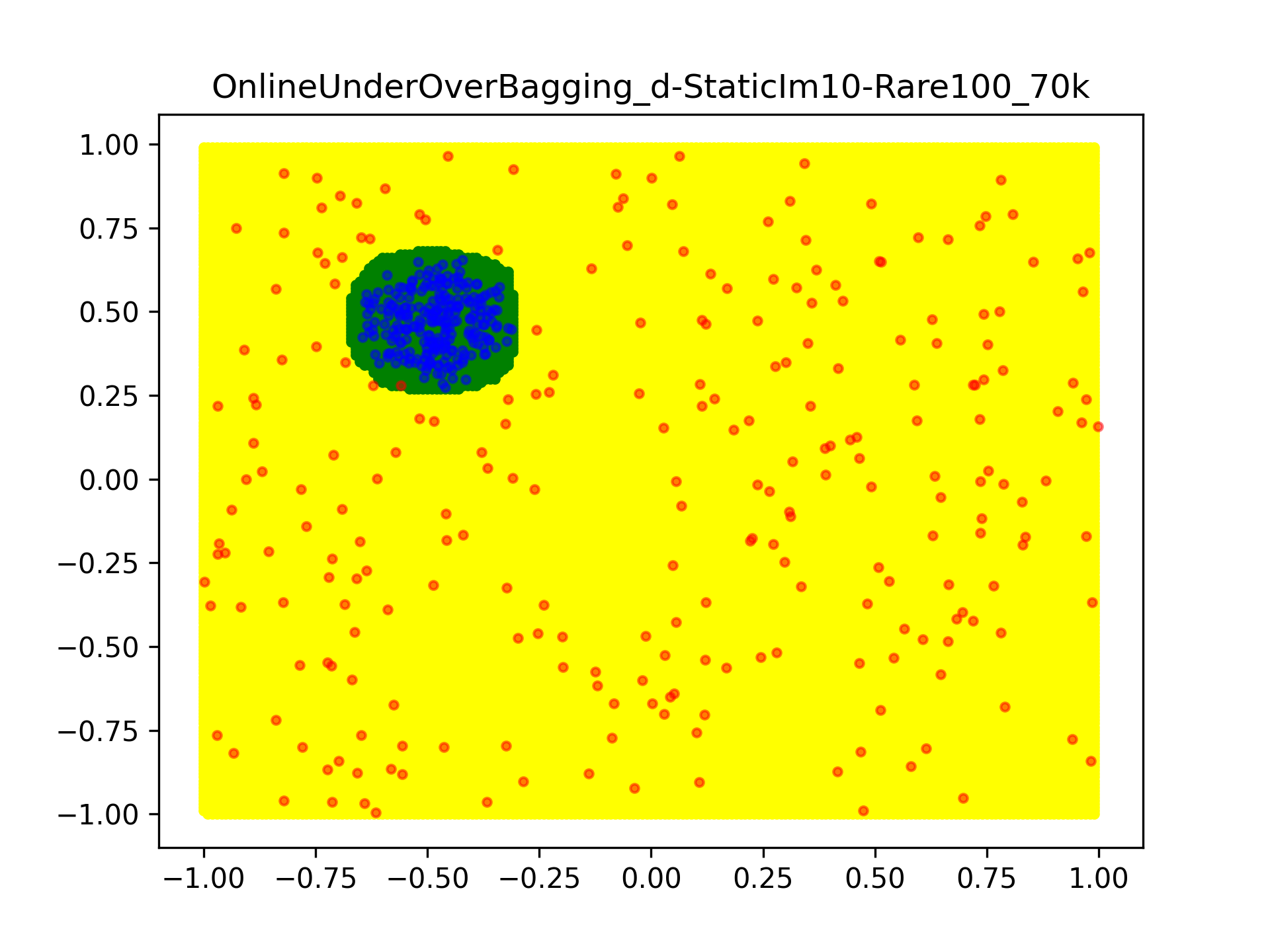} \label{figure:StaticIm10-Rare100-dec_bound-OnlineUnderOverBagging_d-70k}}
\subfigure[SMOGauNoise]{\includegraphics[width=0.29\textwidth]{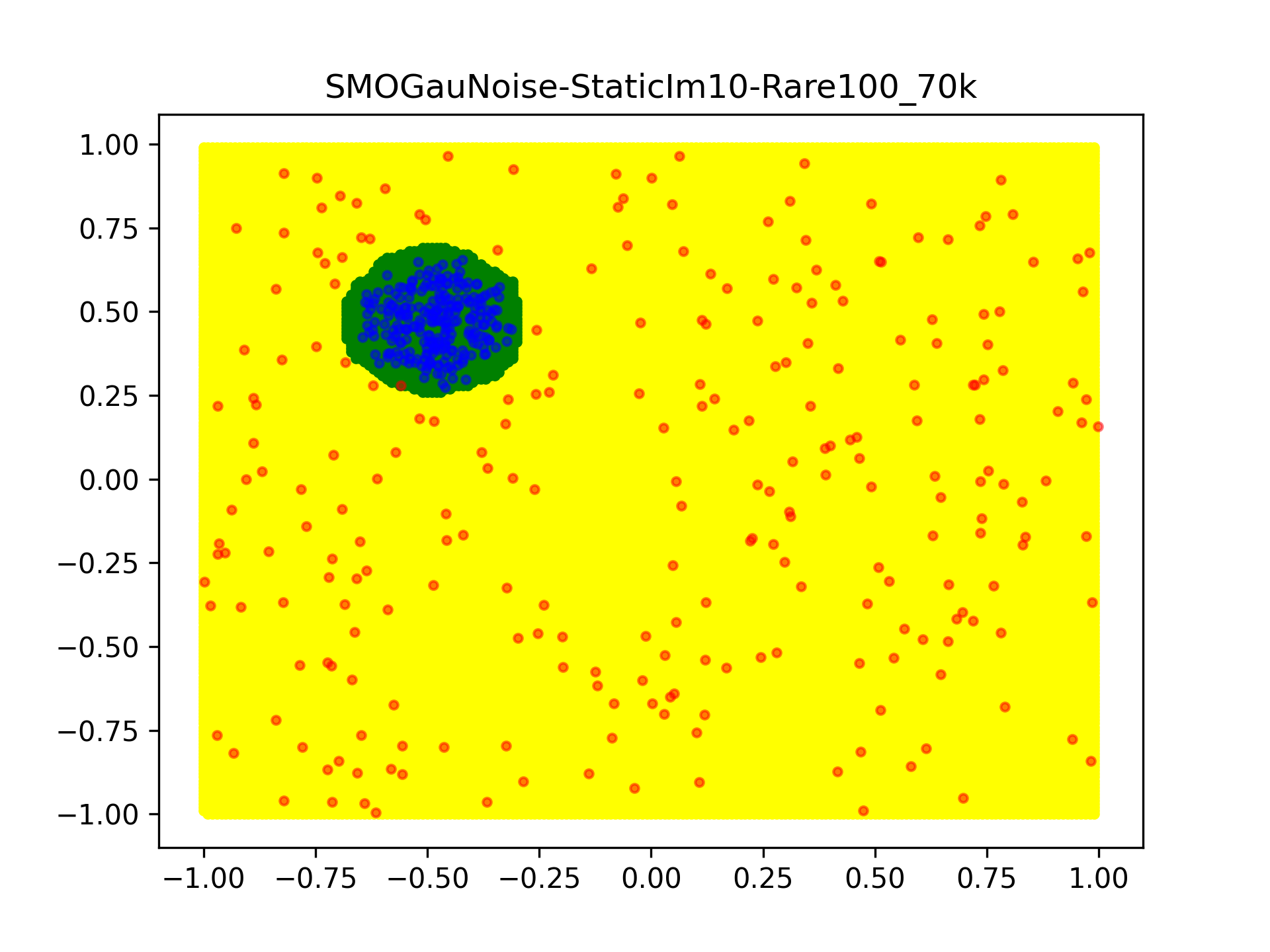} \label{figure:StaticIm10-Rare100-dec_bound-SMOGauNoise-70k}}
\subfigure[VFC-SMOTE]{\includegraphics[width=0.29\textwidth]{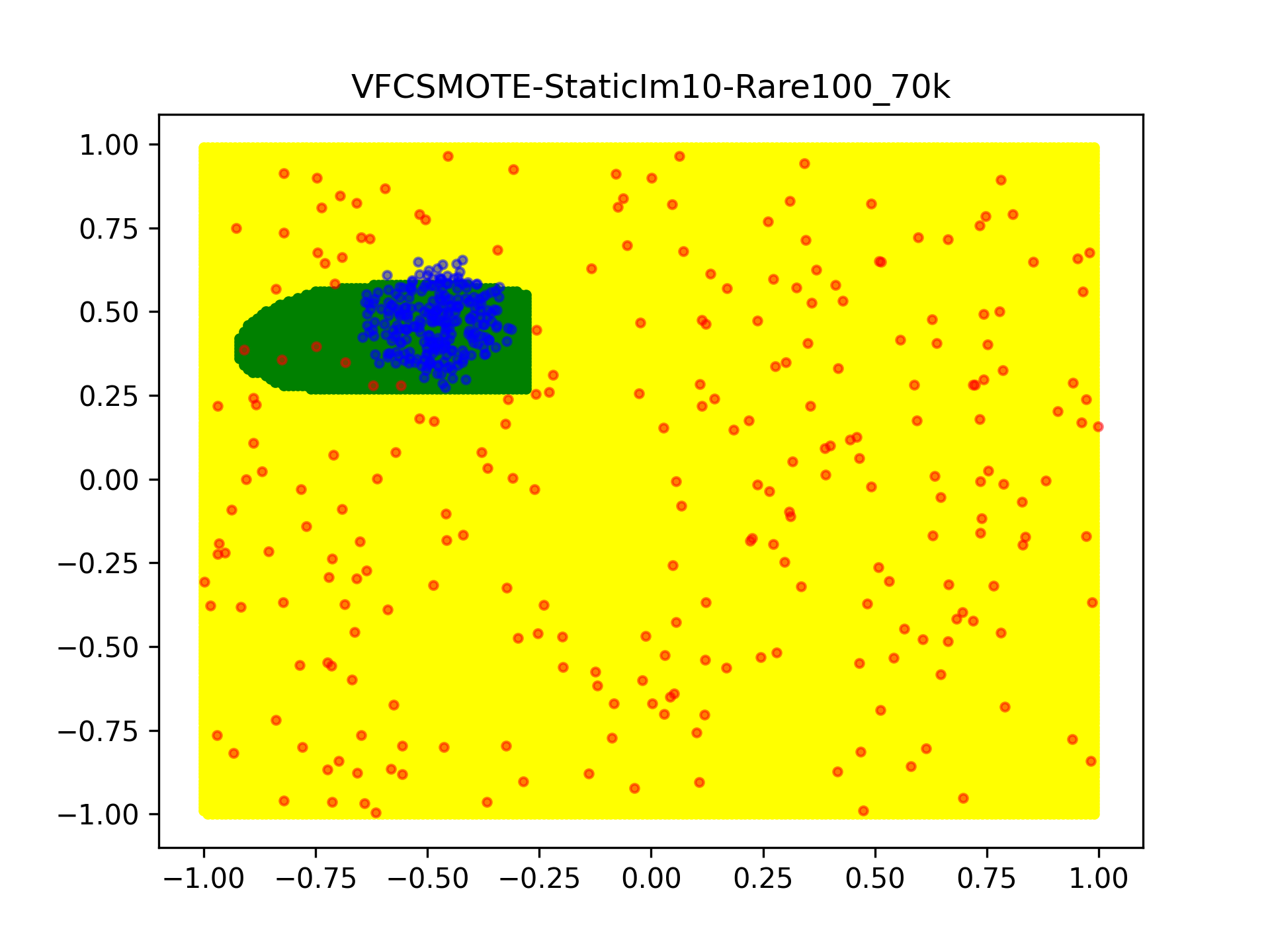} \label{figure:StaticIm10-Rare100-dec_bound-VFCSMOTE-70k}}
\subfigure[SMOTE-OB]{\includegraphics[width=0.29\textwidth]{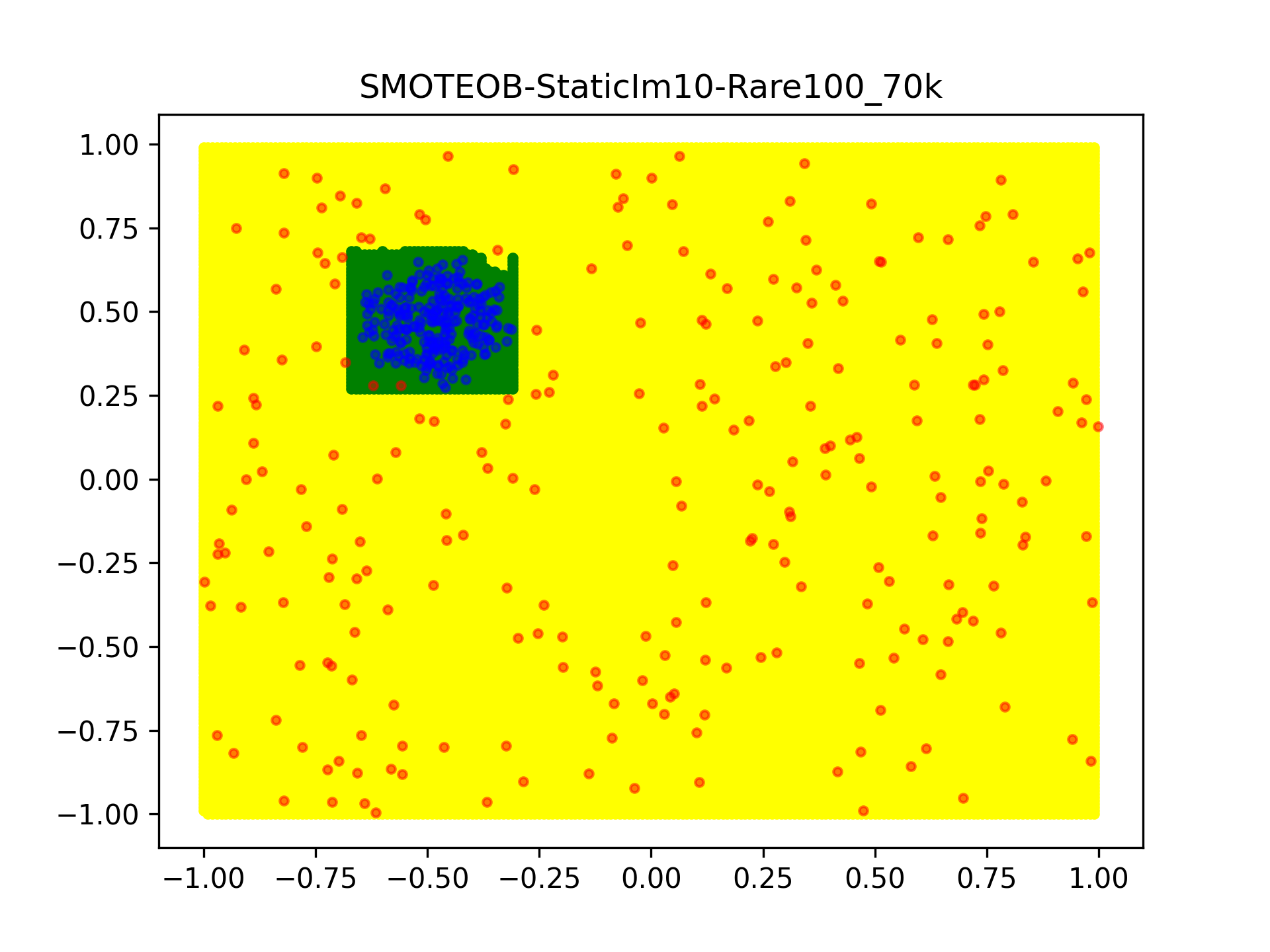} \label{figure:StaticIm10-Rare100-dec_bound-SMOTEOB-70k}}
\subfigure[SMOClust]{\includegraphics[width=0.29\textwidth]{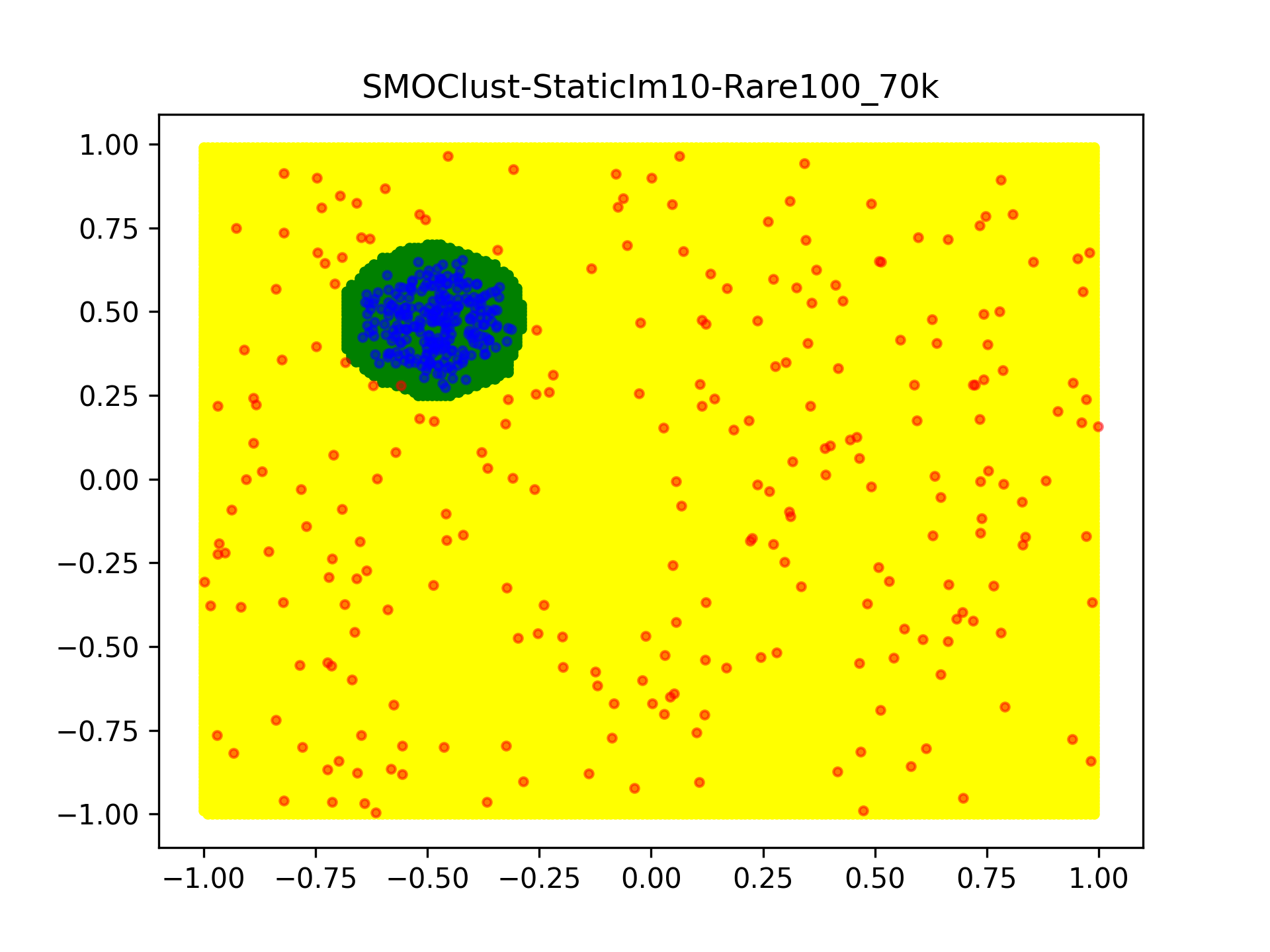} \label{figure:StaticIm10-Rare100-dec_bound-SMOClust-70k}}
\caption{Decision Areas Against Class Balanced Test Set at 70k Time Steps (Before Drift) of Two-Dimensional StaticIm10\_Rare100}
\label{figure:StaticIm10-Rare100-dec_bound-70k}
\end{figure}

\begin{figure}[!ht]
\centering
\subfigure[OOB]{\includegraphics[width=0.29\textwidth]{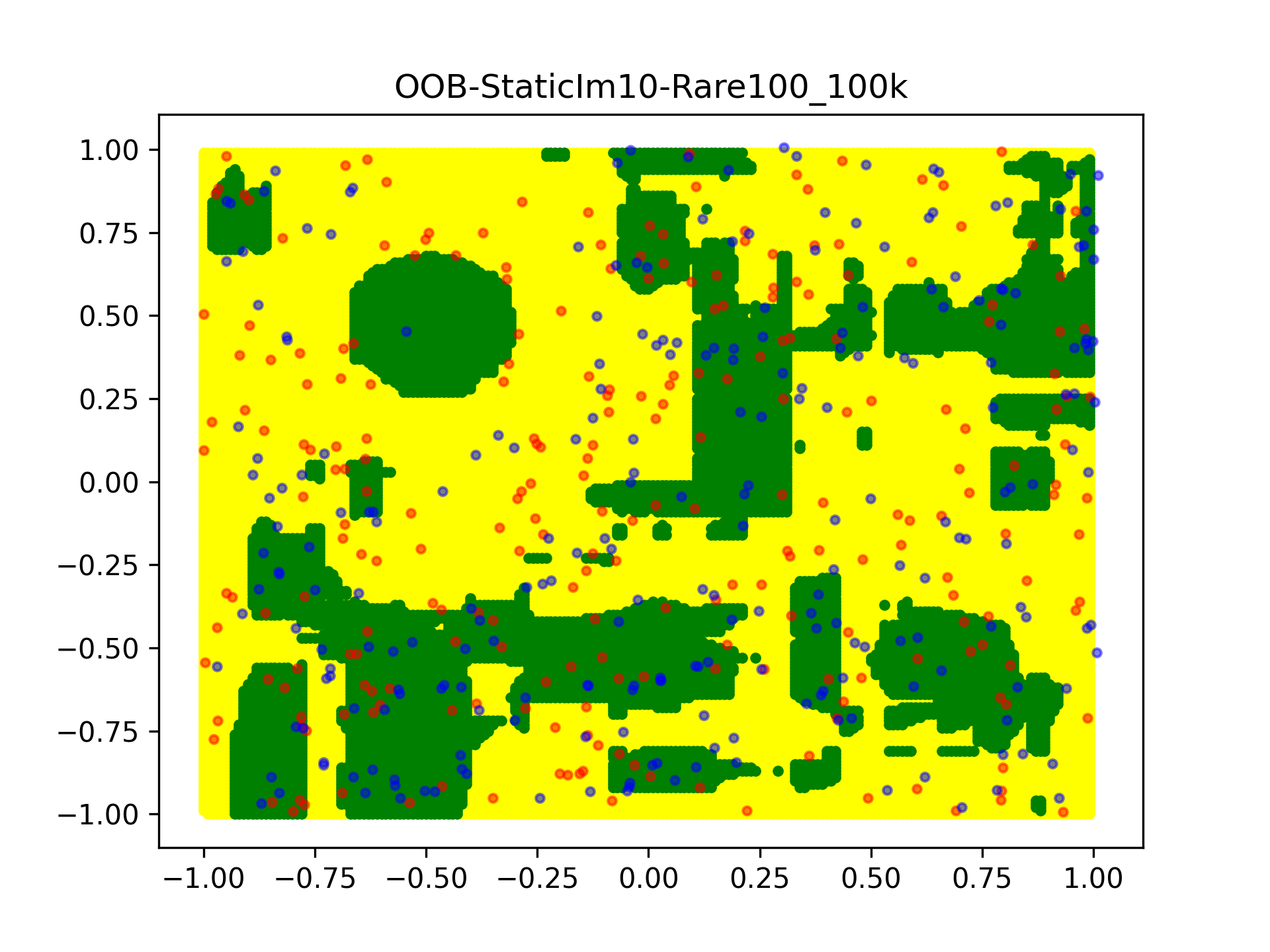} \label{figure:StaticIm10-Rare100-dec_bound-OOB-100k}}
\subfigure[UOB]{\includegraphics[width=0.29\textwidth]{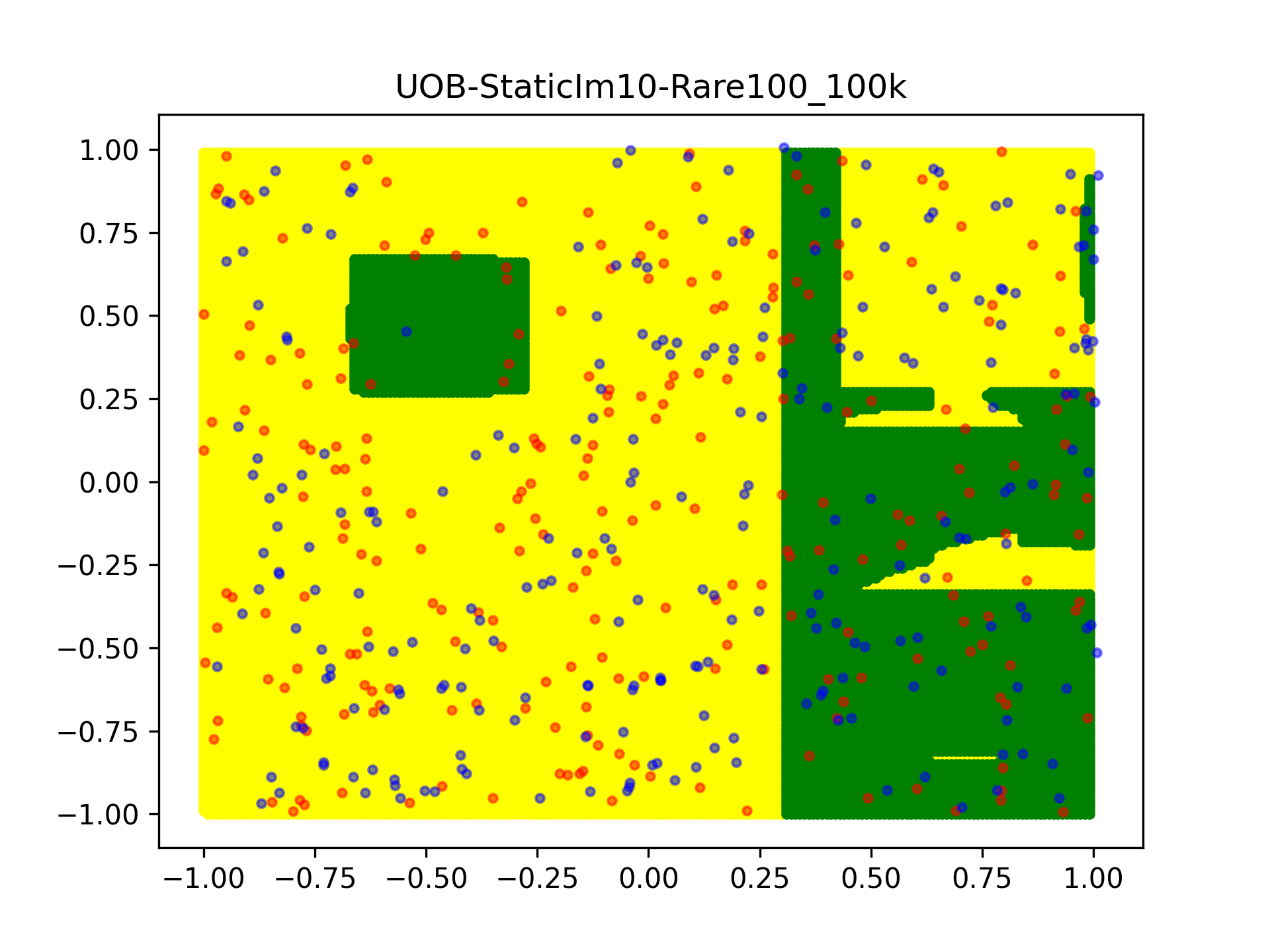} \label{figure:StaticIm10-Rare100-dec_bound-UOB-100k}}
\subfigure[oOS]{\includegraphics[width=0.29\textwidth]{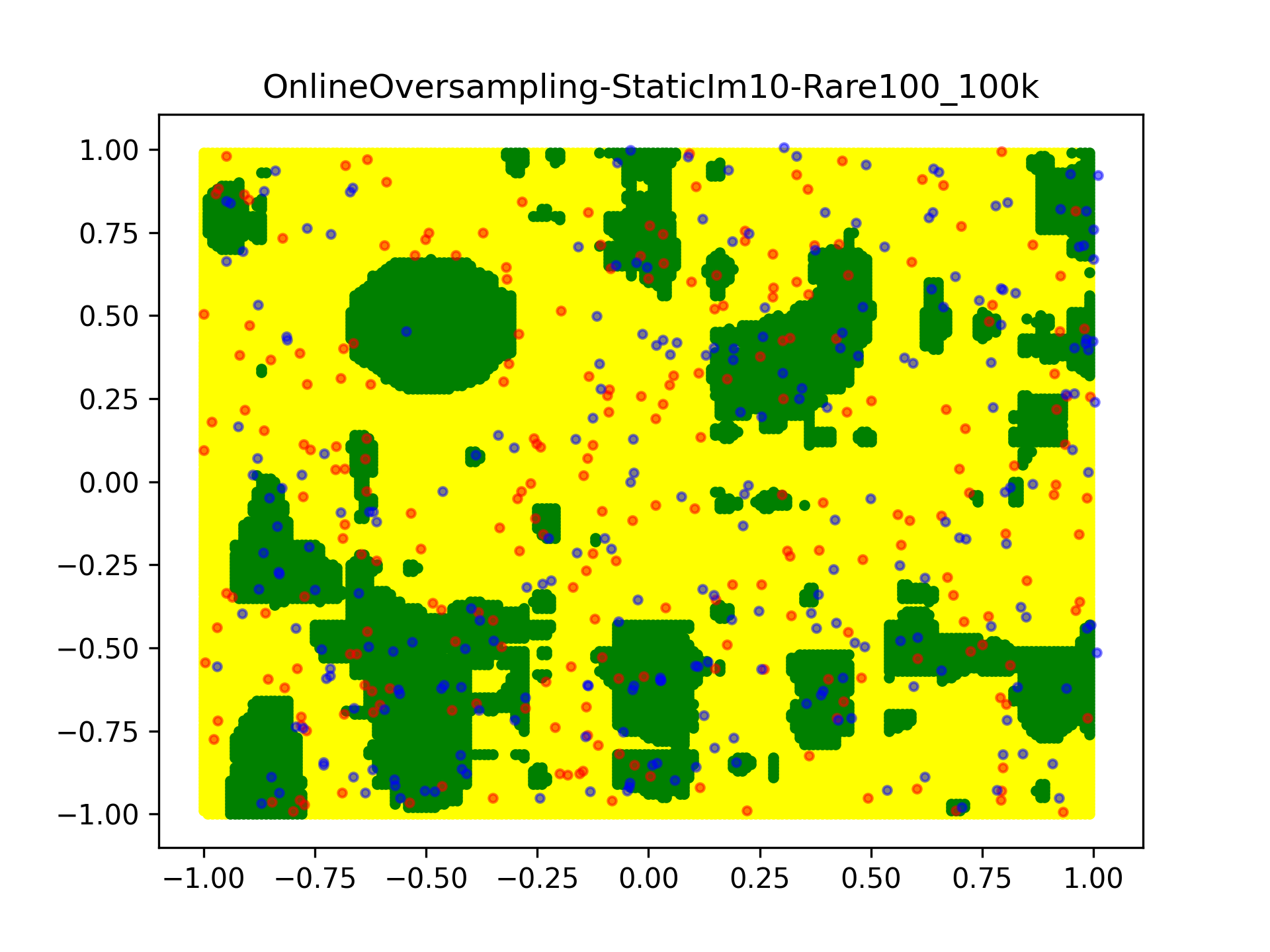} \label{figure:StaticIm10-Rare100-dec_bound-OnlineOversampling-100k}}
\subfigure[oUnderOverB]{\includegraphics[width=0.29\textwidth]{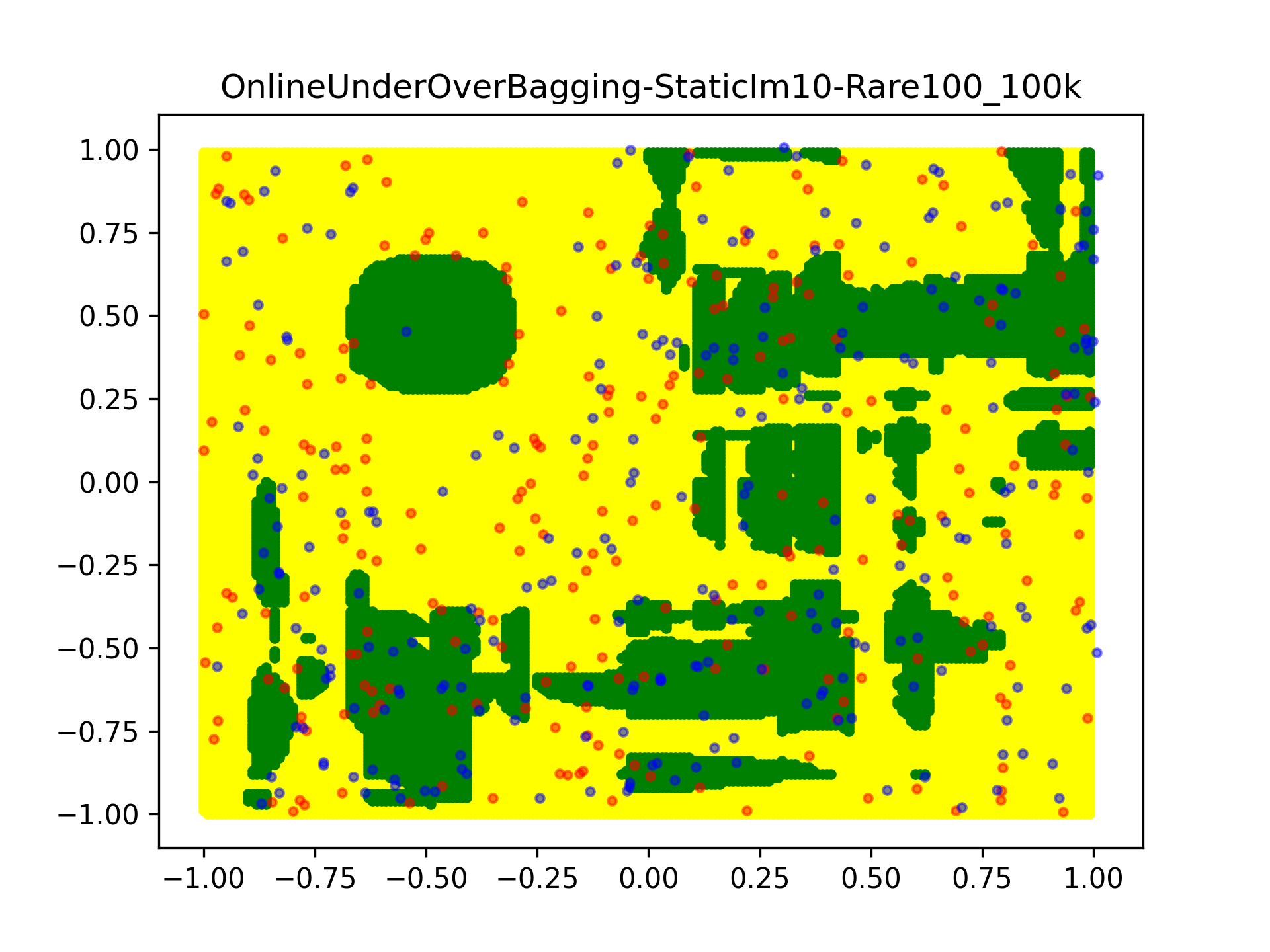} \label{figure:StaticIm10-Rare100-dec_bound-OnlineUnderOverBagging-100k}}
\subfigure[OOB\textsubscript{d}]{\includegraphics[width=0.29\textwidth]{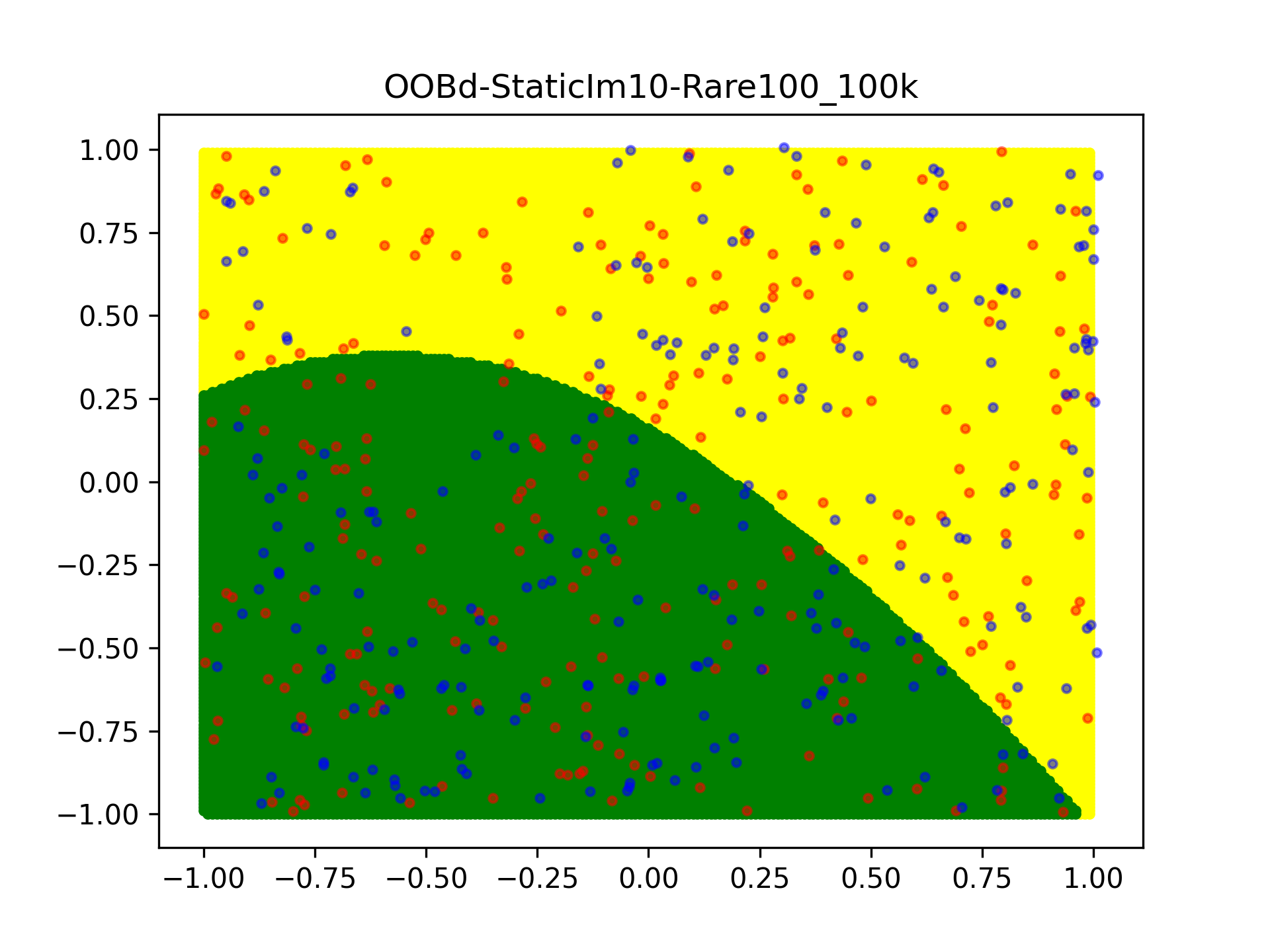} \label{figure:StaticIm10-Rare100-dec_bound-OOBd-100k}}
\subfigure[UOB\textsubscript{d}]{\includegraphics[width=0.29\textwidth]{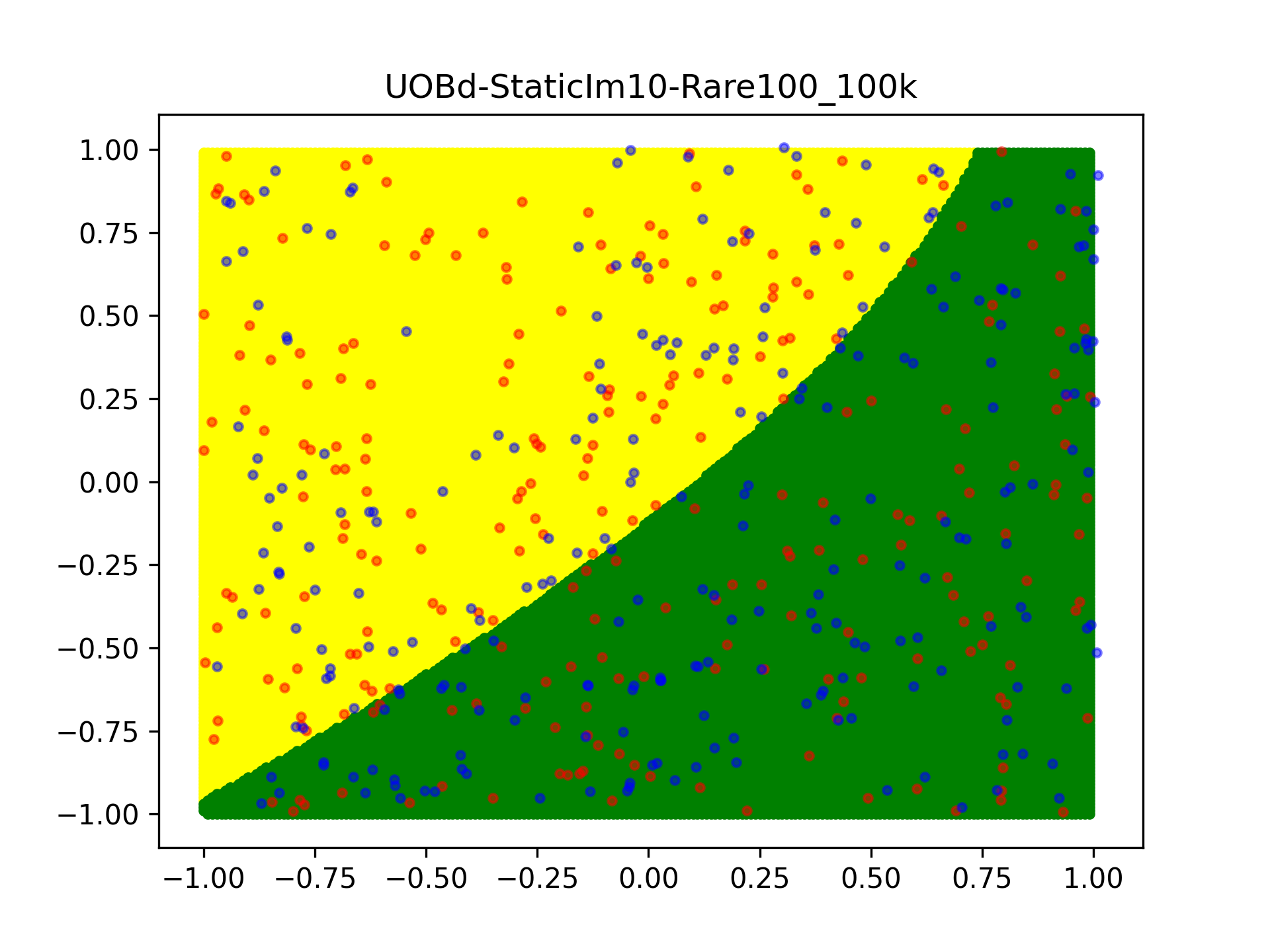} \label{figure:StaticIm10-Rare100-dec_bound-UOBd-100k}}
\subfigure[oOS\textsubscript{d}]{\includegraphics[width=0.29\textwidth]{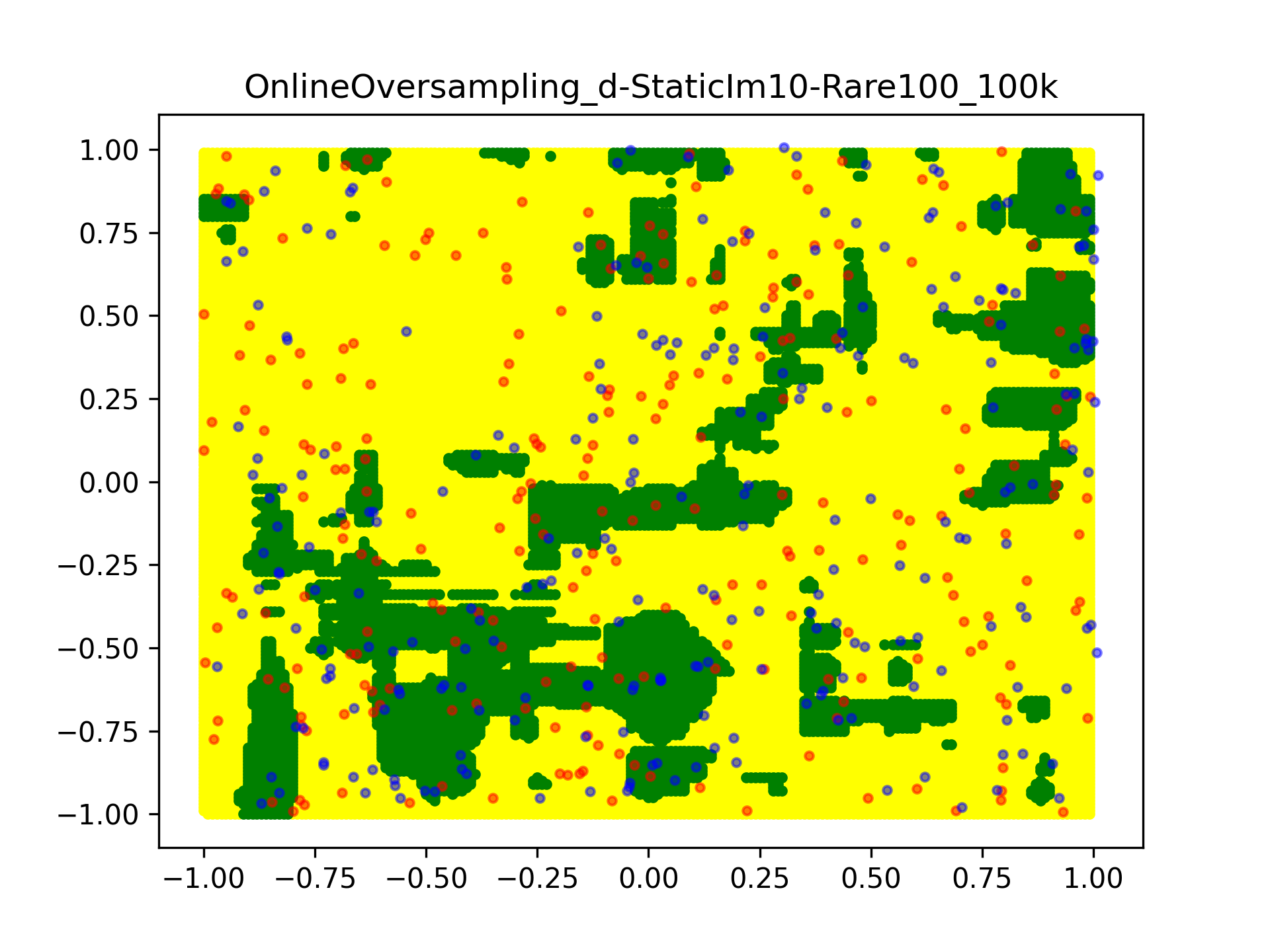} \label{figure:StaticIm10-Rare100-dec_bound-OnlineOversampling_d-100k}}
\subfigure[oUnderOverB\textsubscript{d}]{\includegraphics[width=0.29\textwidth]{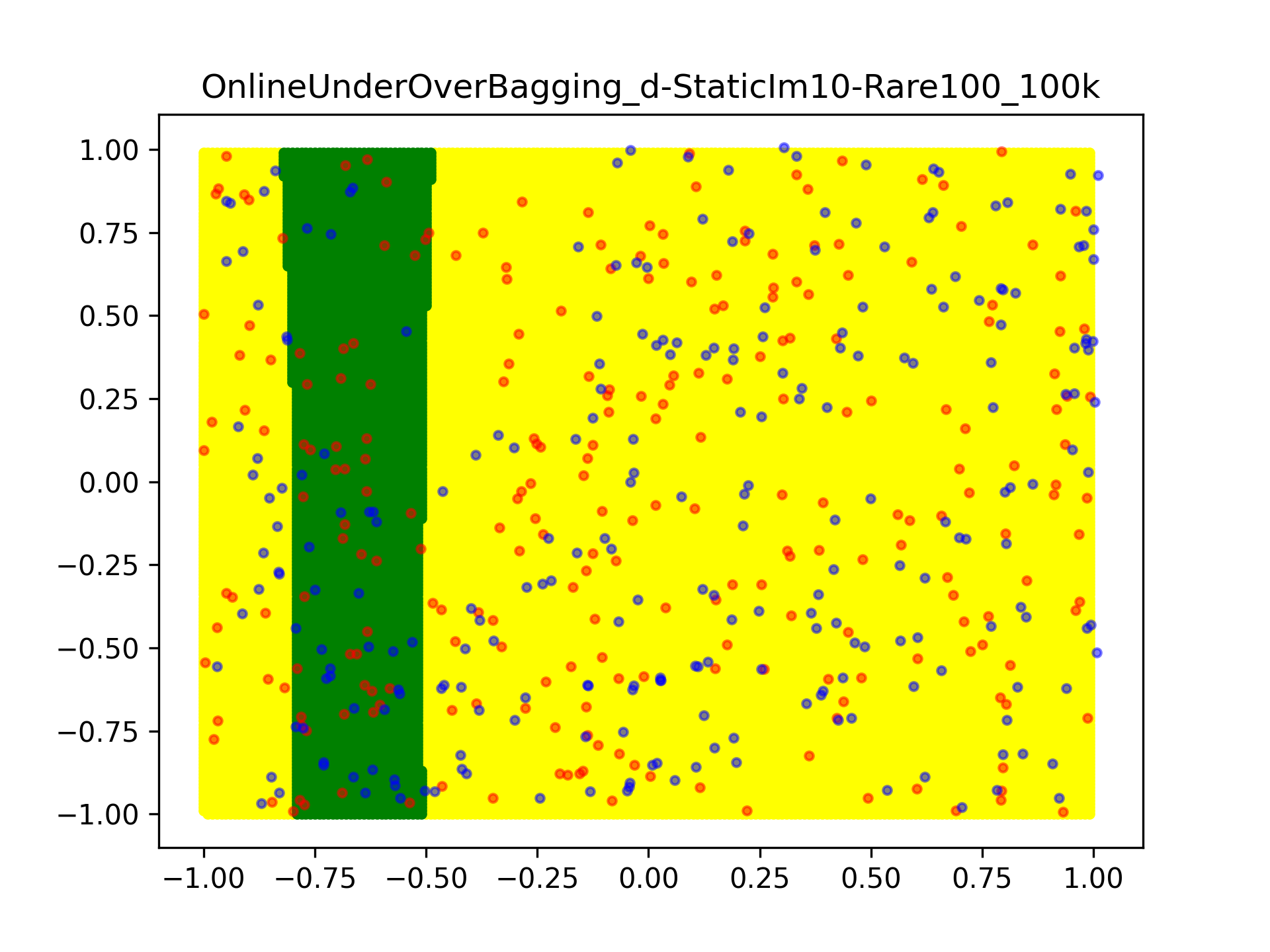} \label{figure:StaticIm10-Rare100-dec_bound-OnlineUnderOverBagging_d-100k}}
\subfigure[SMOGauNoise]{\includegraphics[width=0.29\textwidth]{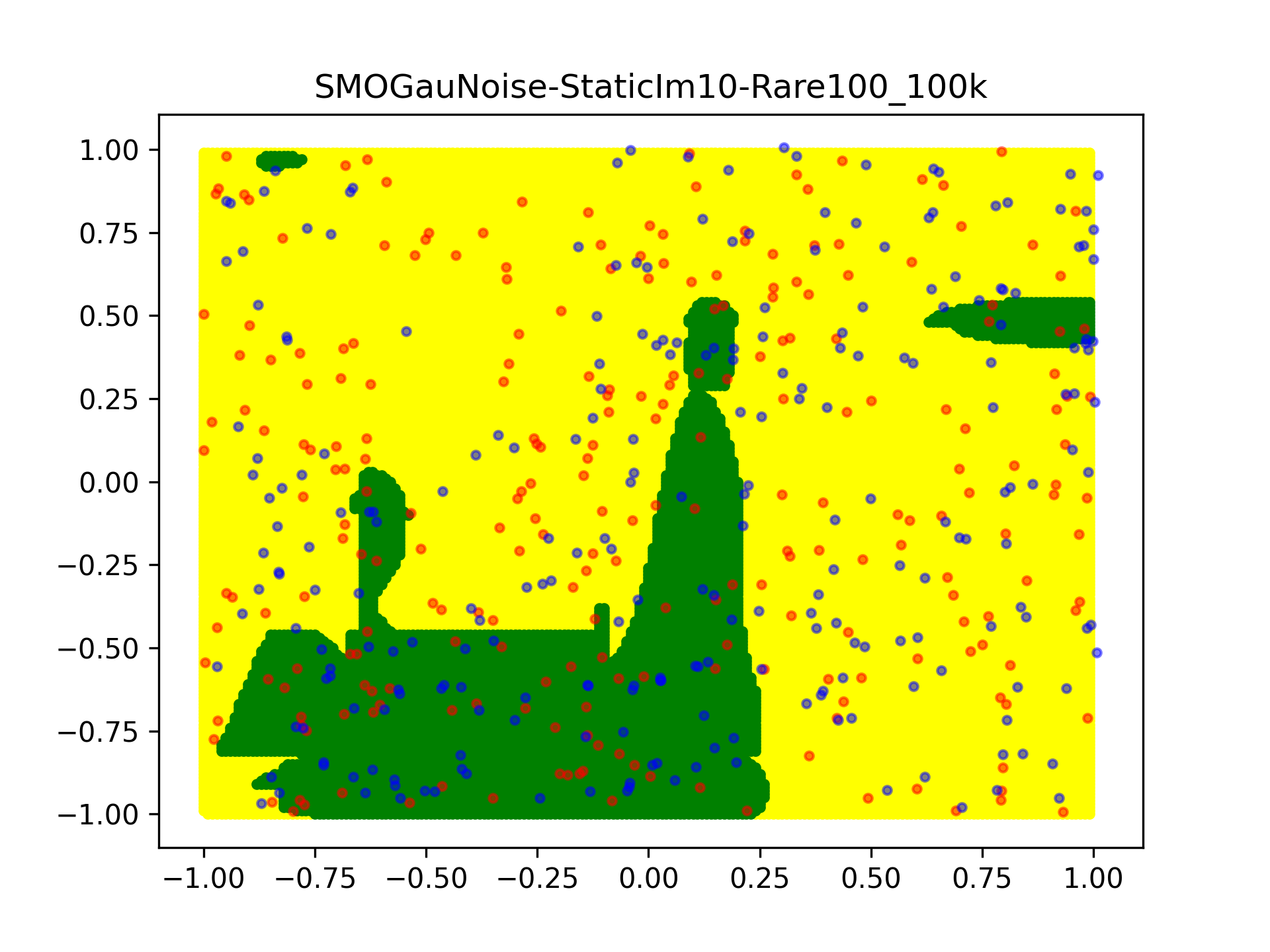} \label{figure:StaticIm10-Rare100-dec_bound-SMOGauNoise-100k}}
\subfigure[VFC-SMOTE]{\includegraphics[width=0.29\textwidth]{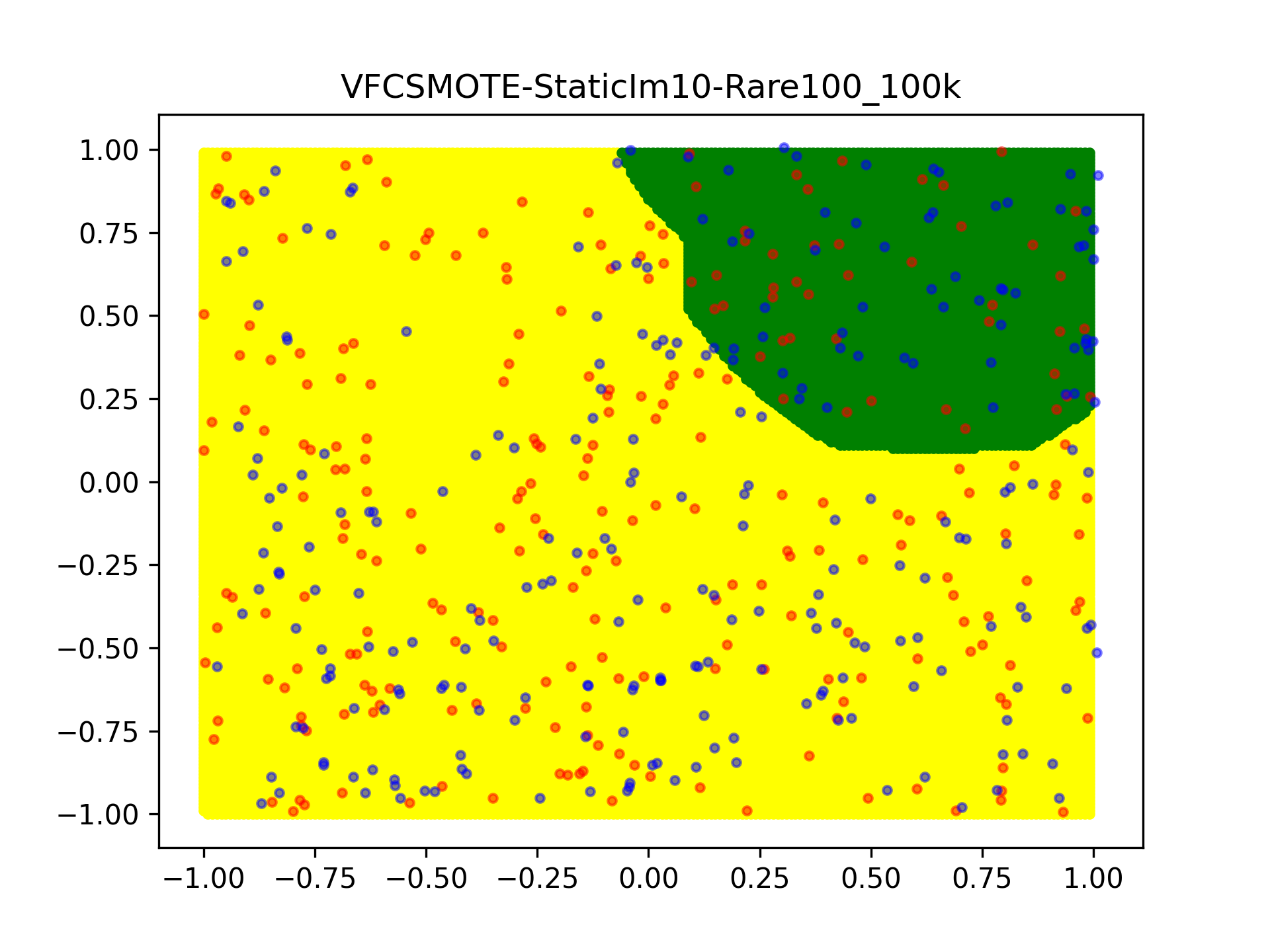} \label{figure:StaticIm10-Rare100-dec_bound-VFCSMOTE-100k}}
\subfigure[SMOTE-OB]{\includegraphics[width=0.29\textwidth]{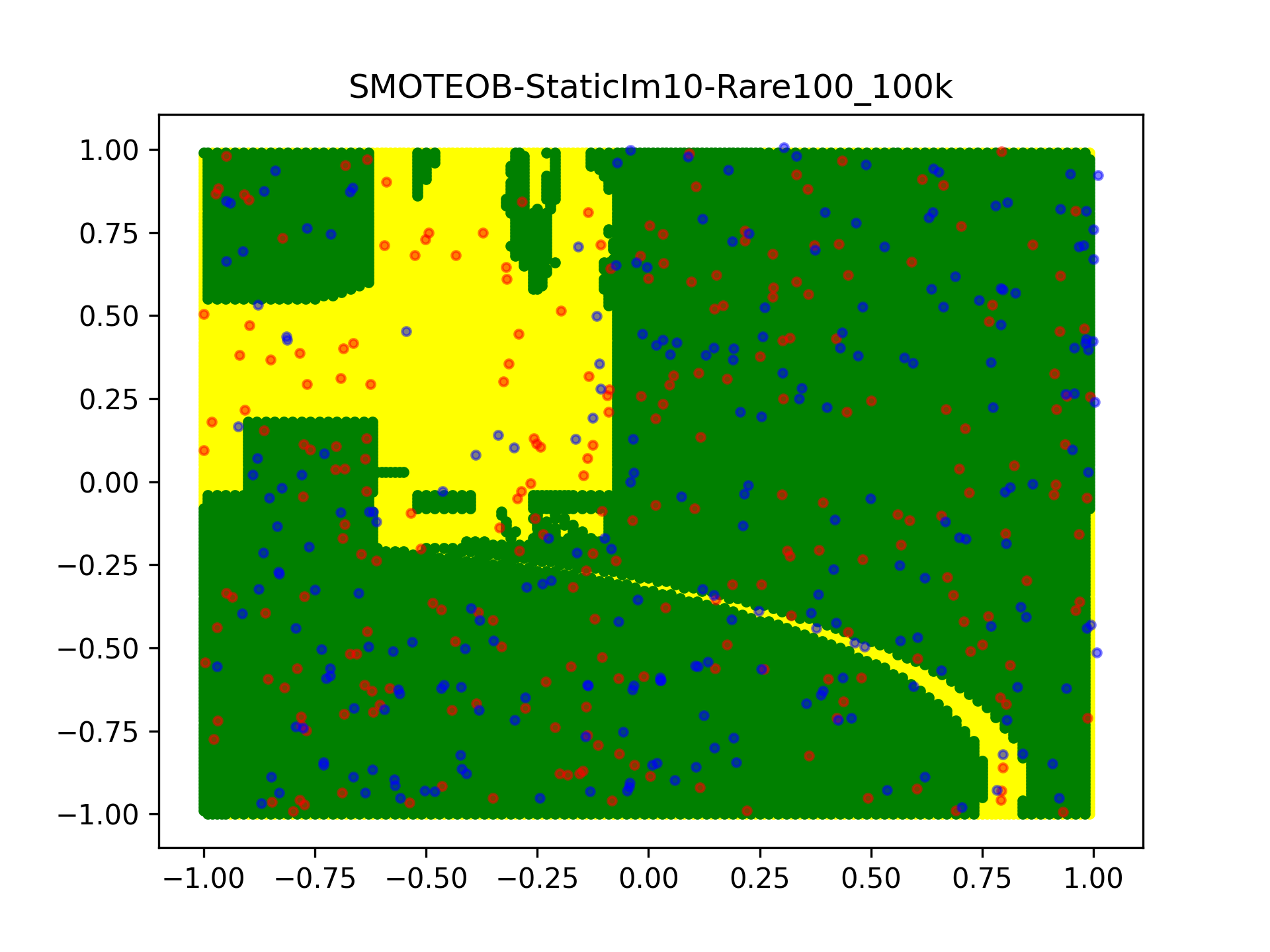} \label{figure:StaticIm10-Rare100-dec_bound-SMOTEOB-100k}}
\subfigure[SMOClust]{\includegraphics[width=0.29\textwidth]{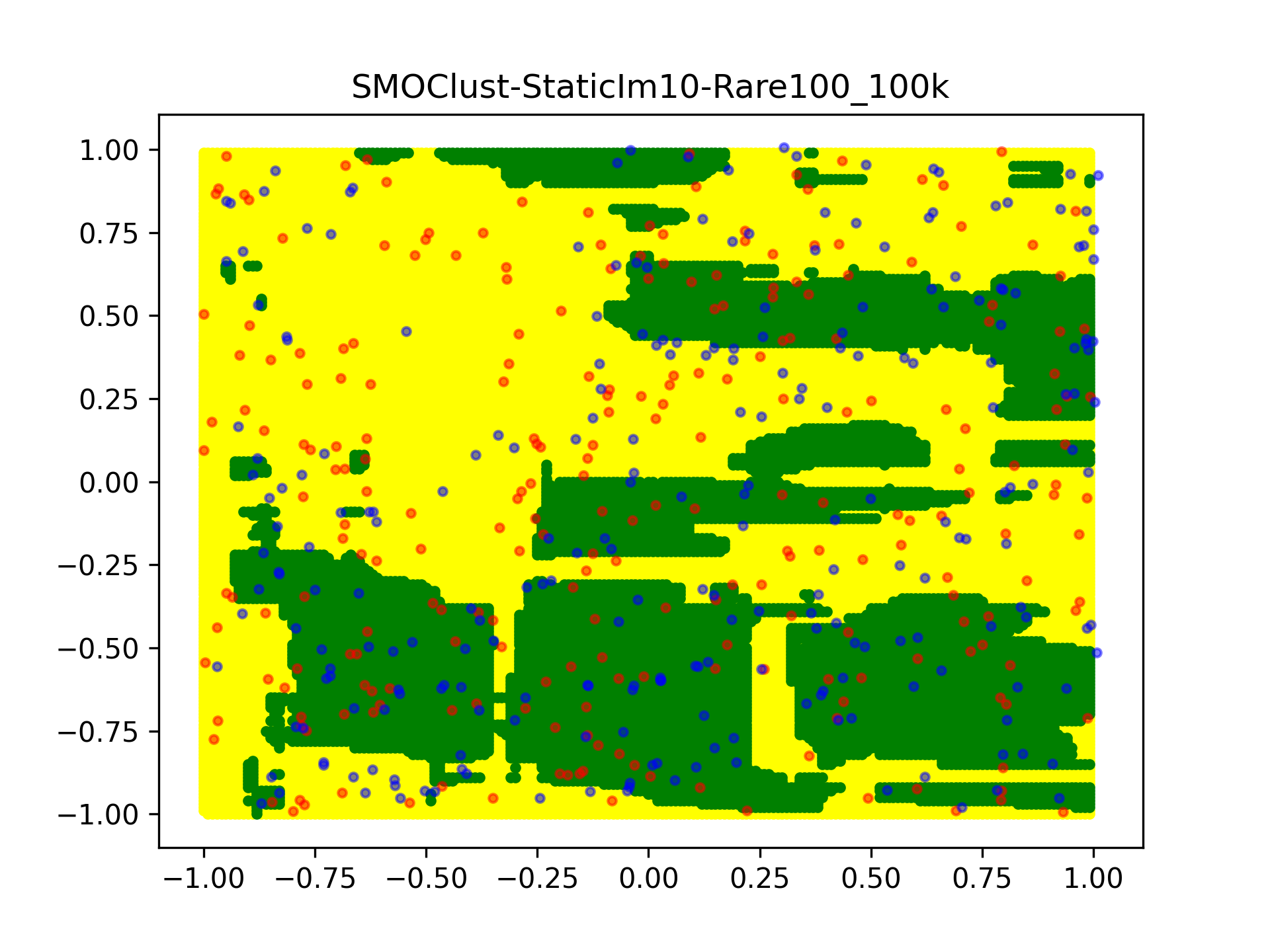} \label{figure:StaticIm10-Rare100-dec_bound-SMOClust-100k}}
\caption{Decision Areas Against Class Balanced Test Set at 100k Time Steps (After Drift) of Two-Dimensional StaticIm10\_Rare100}
\label{figure:StaticIm10-Rare100-dec_bound-100k}
\end{figure}

\begin{figure}[!ht]
\centering
\subfigure[OOB]{\includegraphics[width=0.29\textwidth]{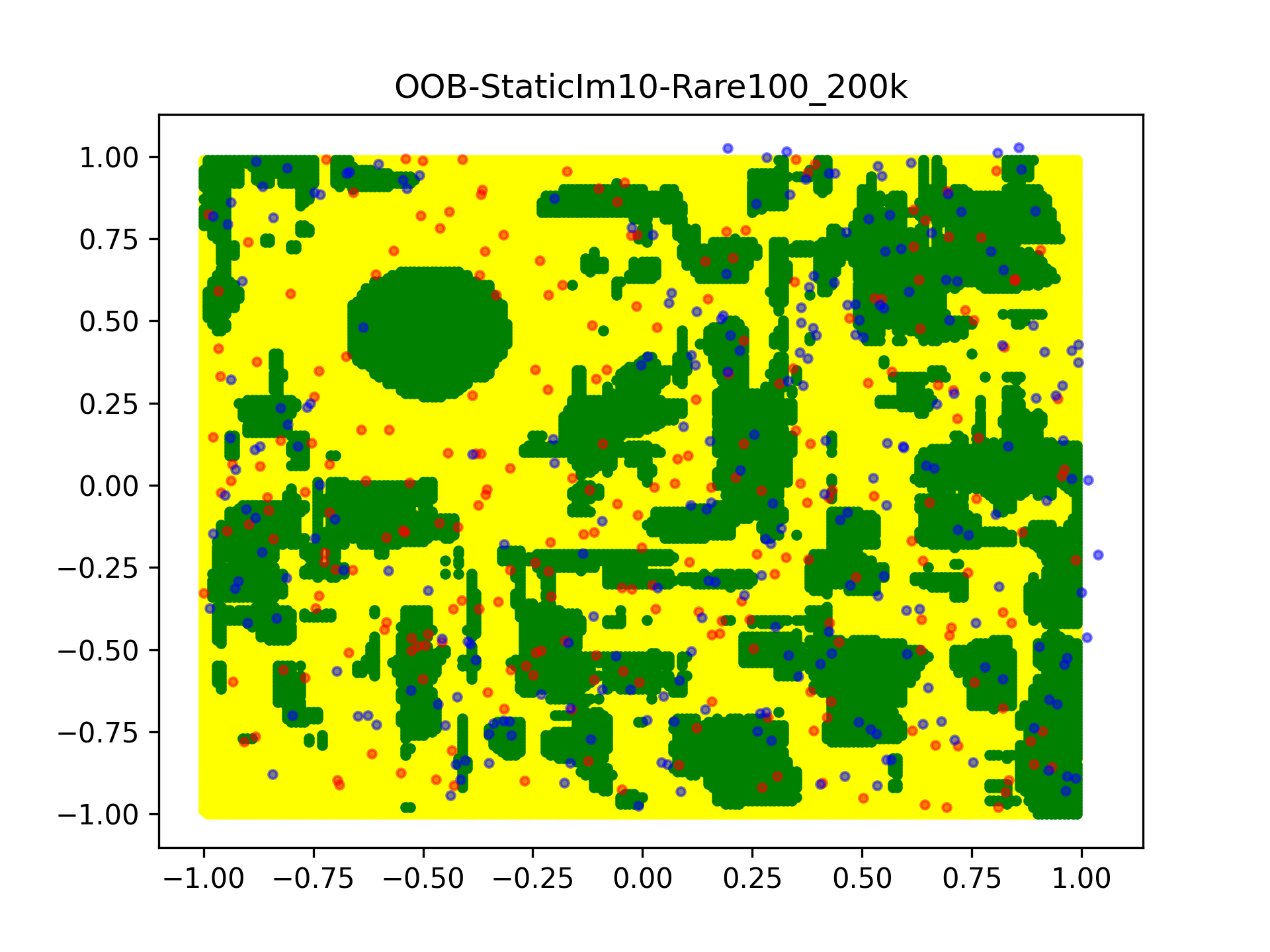} \label{figure:StaticIm10-Rare100-dec_bound-OOB-200k}}
\subfigure[UOB]{\includegraphics[width=0.29\textwidth]{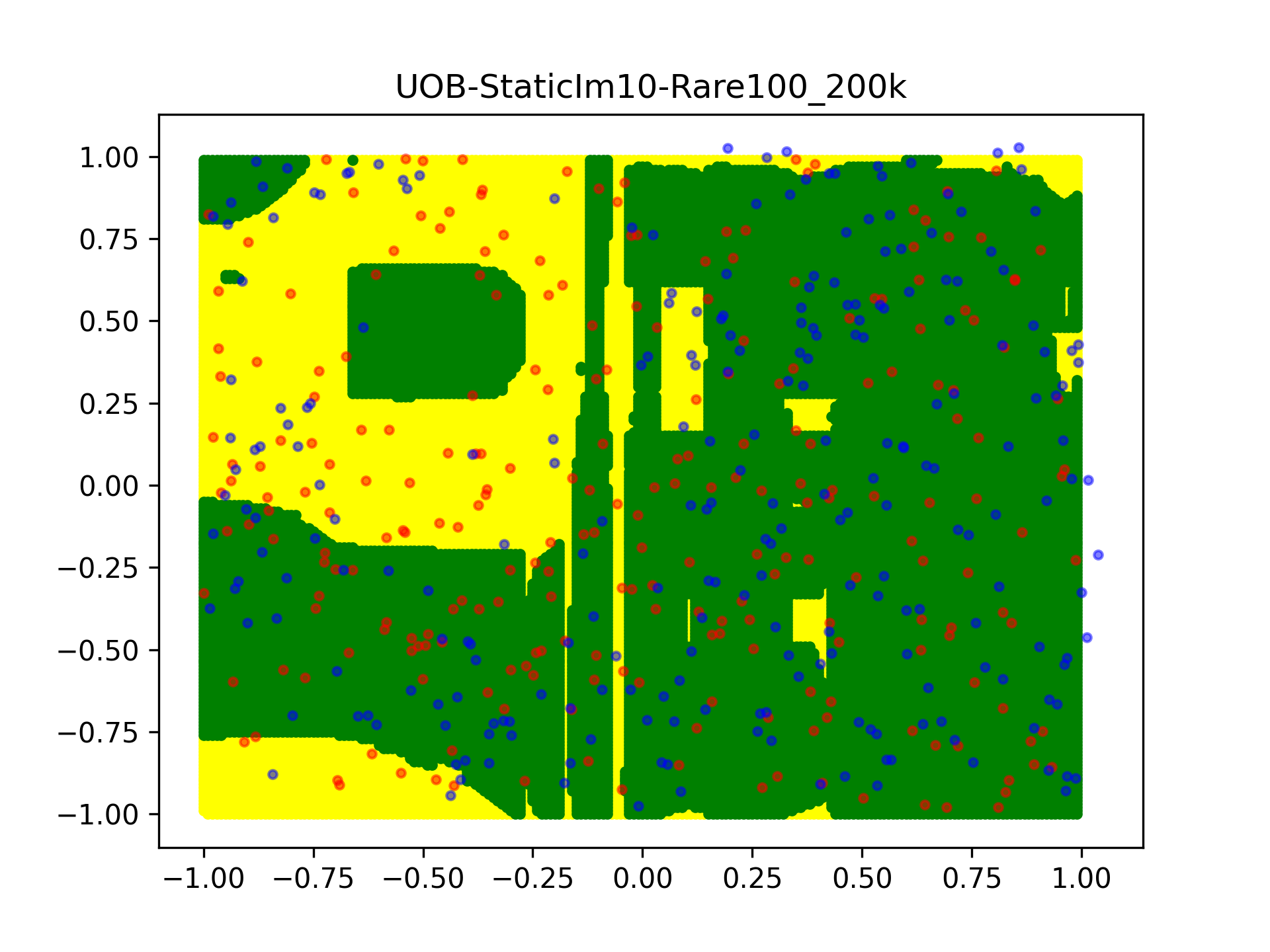} \label{figure:StaticIm10-Rare100-dec_bound-UOB-200k}}
\subfigure[oOS]{\includegraphics[width=0.29\textwidth]{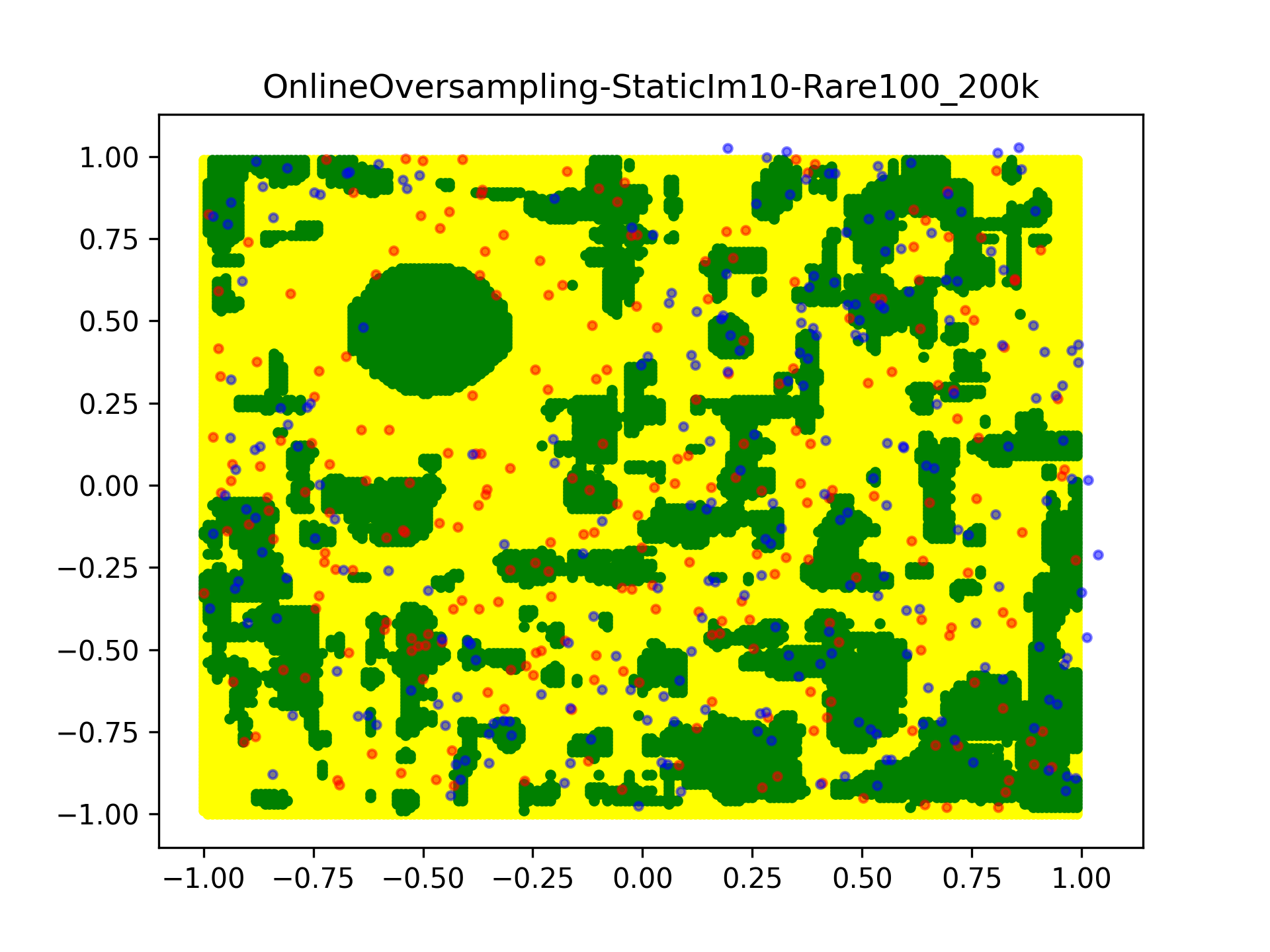} \label{figure:StaticIm10-Rare100-dec_bound-OnlineOversampling-200k}}
\subfigure[oUnderOverB]{\includegraphics[width=0.29\textwidth]{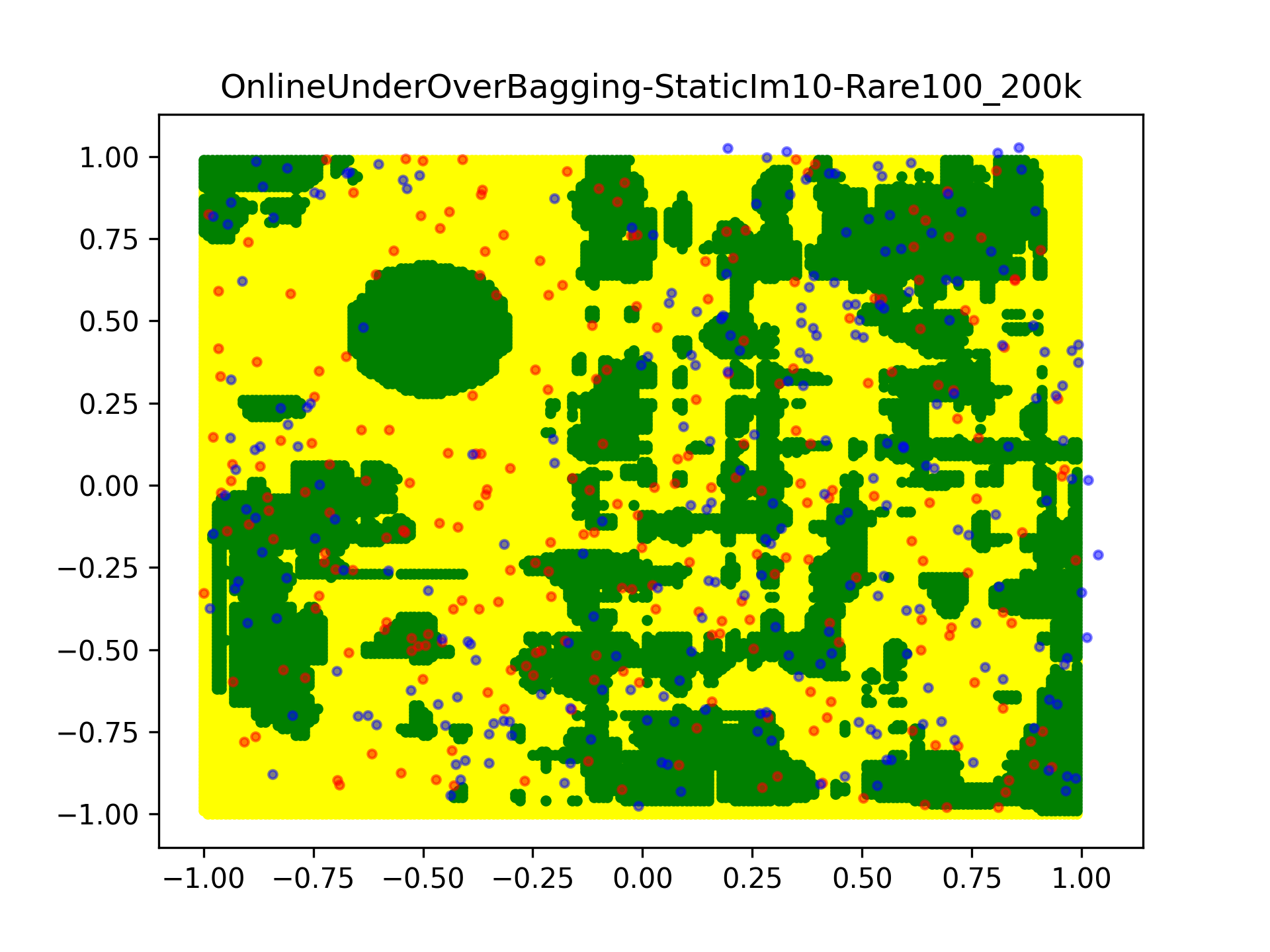} \label{figure:StaticIm10-Rare100-dec_bound-OnlineUnderOverBagging-200k}}
\subfigure[OOB\textsubscript{d}]{\includegraphics[width=0.29\textwidth]{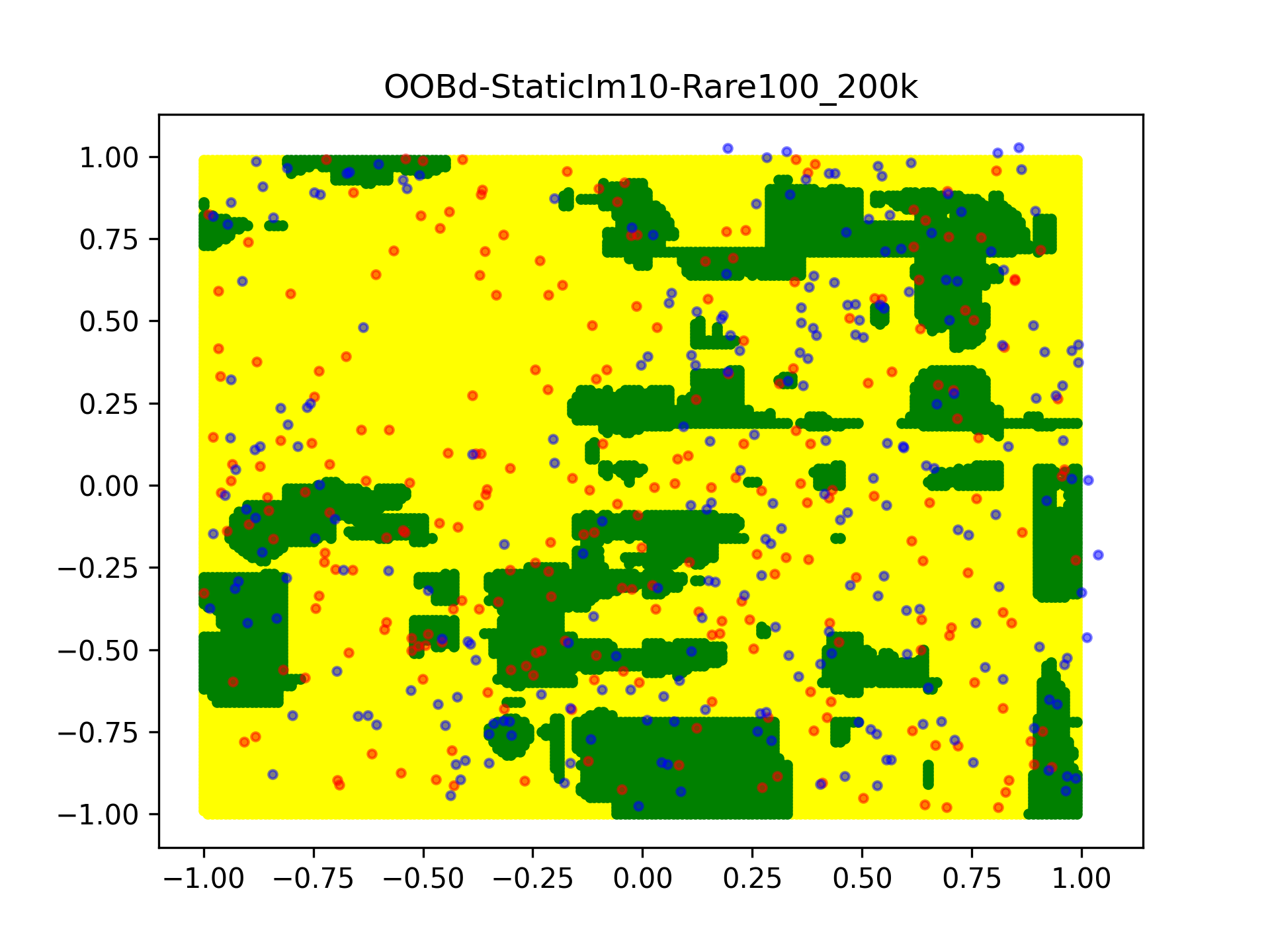} \label{figure:StaticIm10-Rare100-dec_bound-OOBd-200k}}
\subfigure[UOB\textsubscript{d}]{\includegraphics[width=0.29\textwidth]{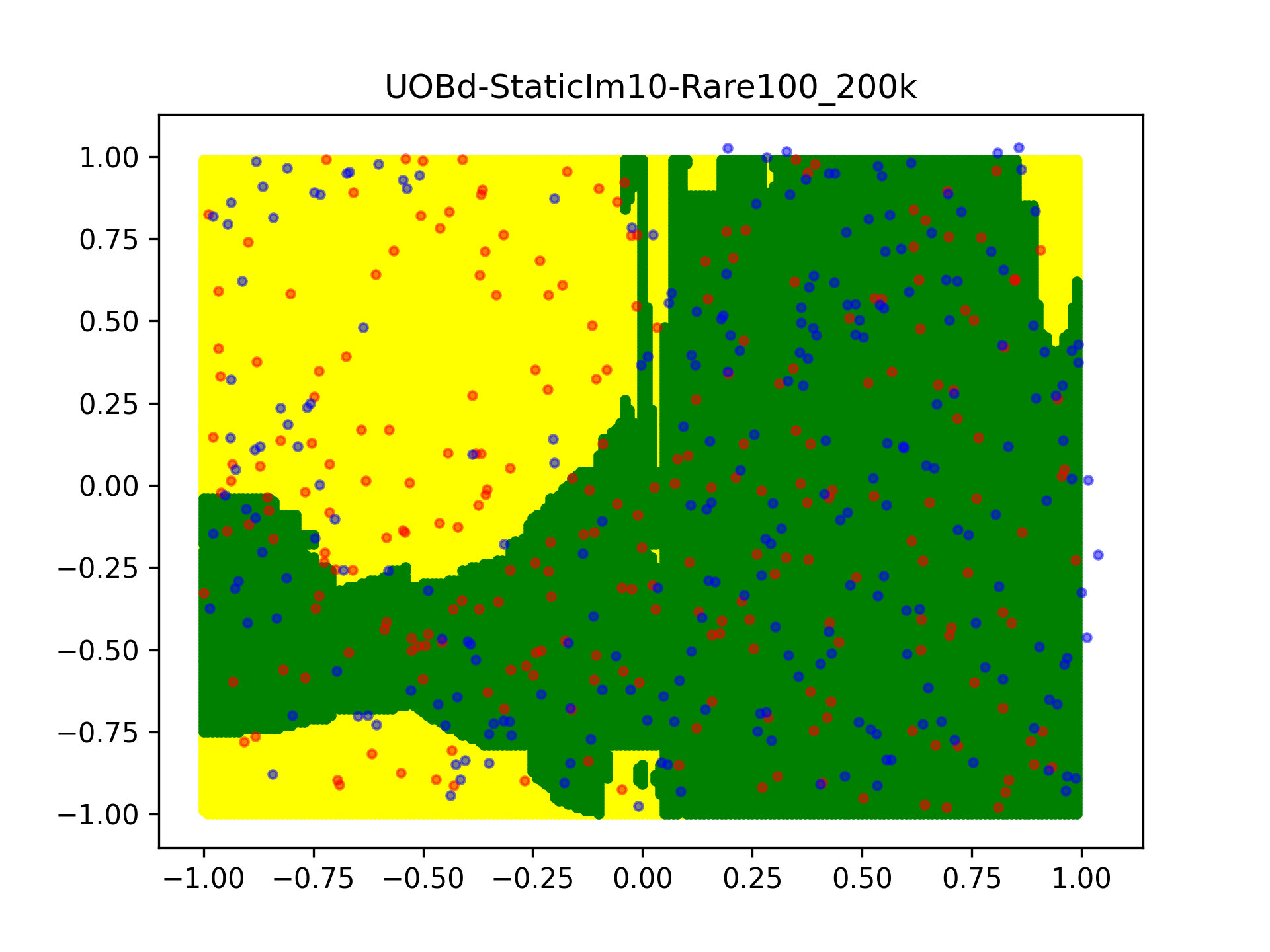} \label{figure:StaticIm10-Rare100-dec_bound-UOBd-200k}}
\subfigure[oOS\textsubscript{d}]{\includegraphics[width=0.29\textwidth]{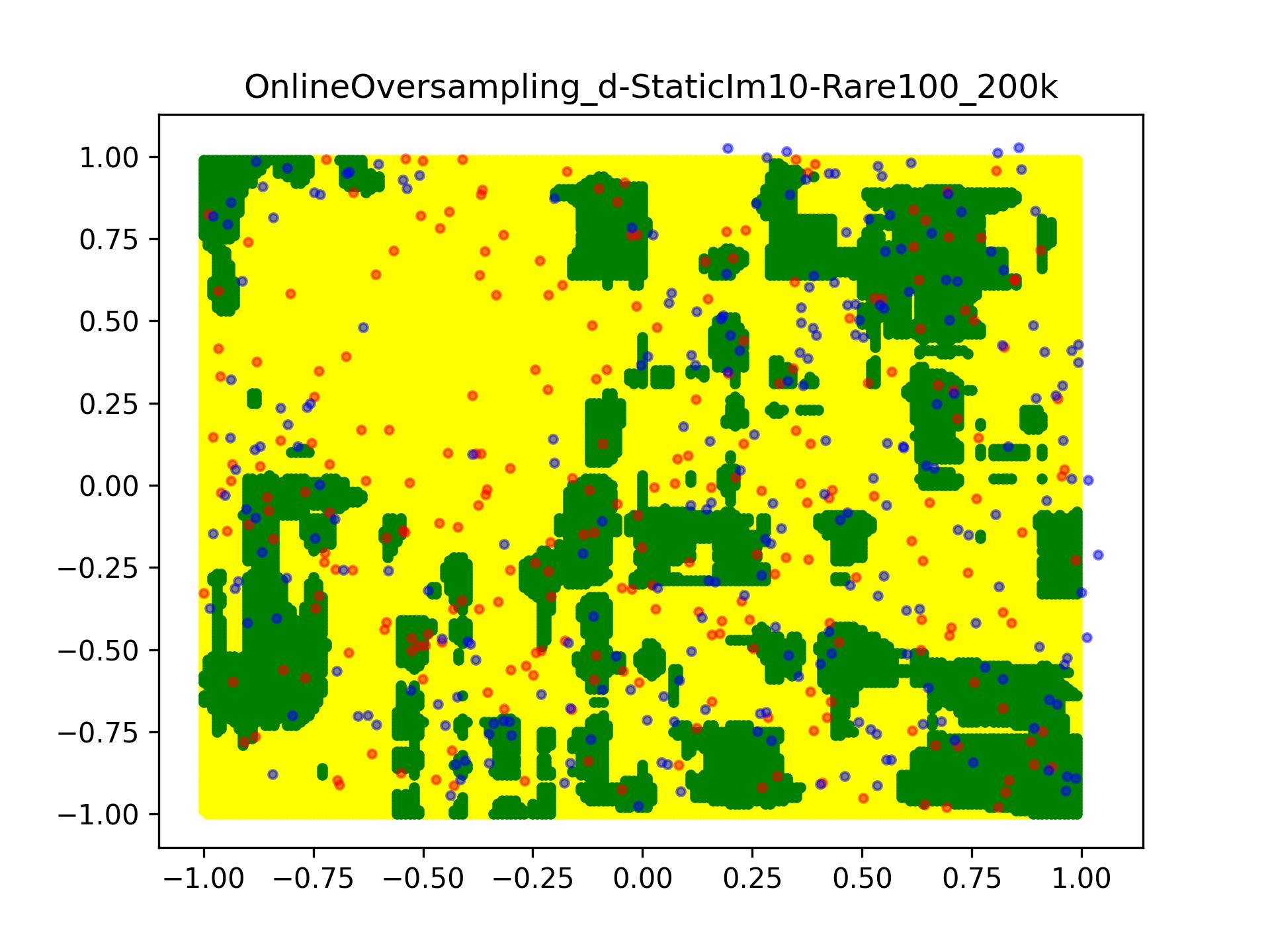} \label{figure:StaticIm10-Rare100-dec_bound-OnlineOversampling_d-200k}}
\subfigure[oUnderOverB\textsubscript{d}]{\includegraphics[width=0.29\textwidth]{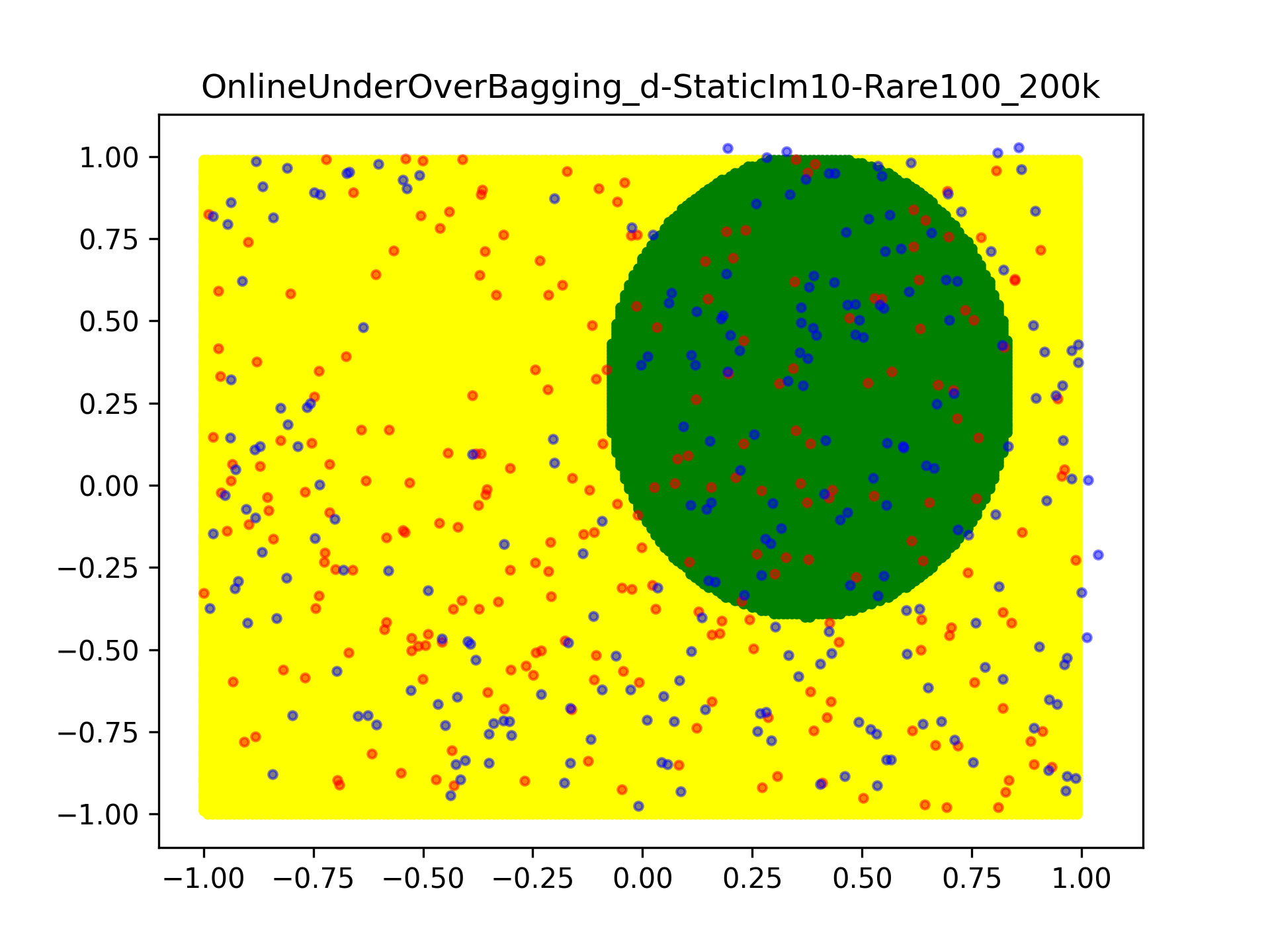} \label{figure:StaticIm10-Rare100-dec_bound-OnlineUnderOverBagging_d-200k}}
\subfigure[SMOGauNoise]{\includegraphics[width=0.29\textwidth]{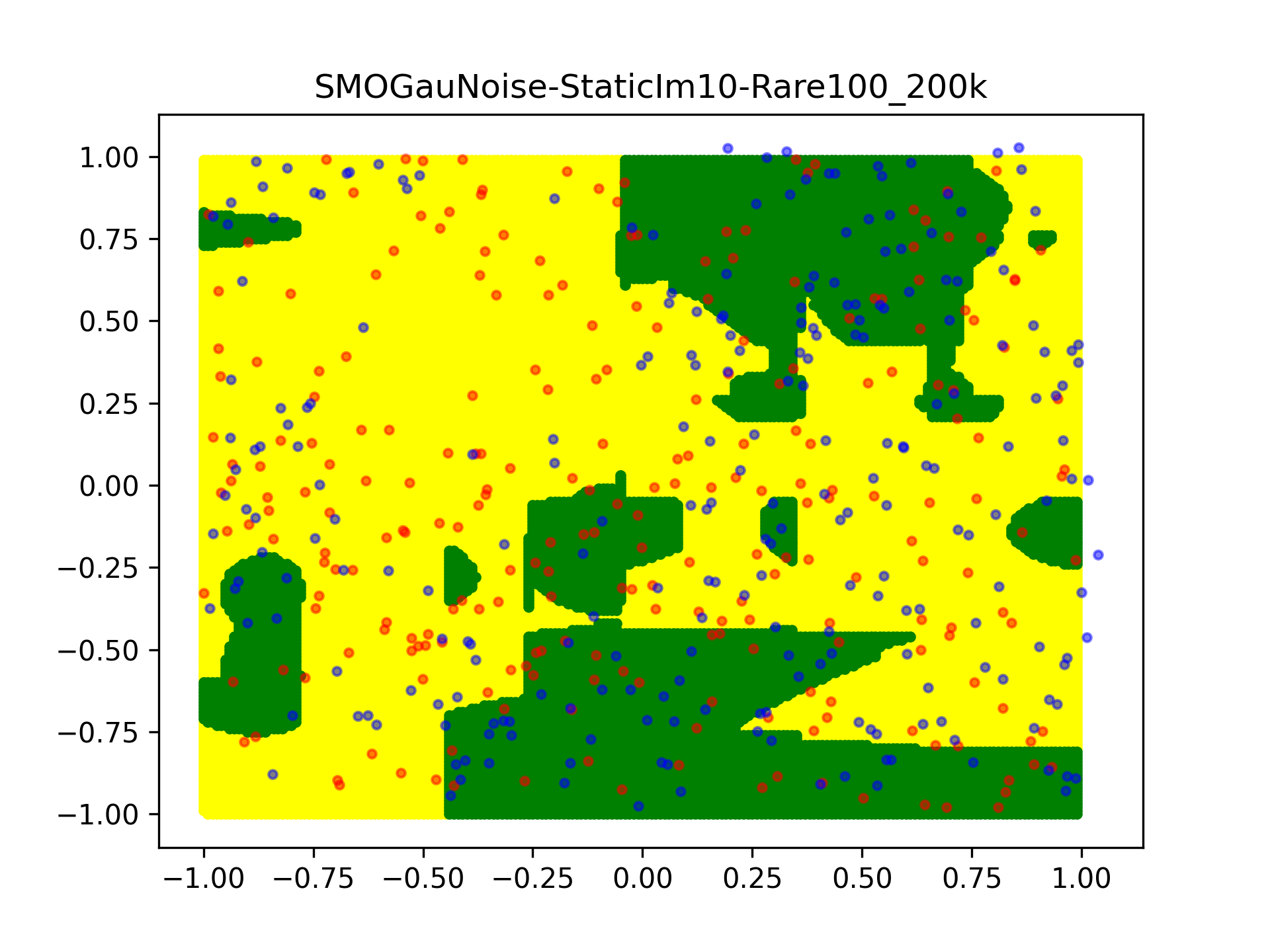} \label{figure:StaticIm10-Rare100-dec_bound-SMOGauNoise-200k}}
\subfigure[VFC-SMOTE]{\includegraphics[width=0.29\textwidth]{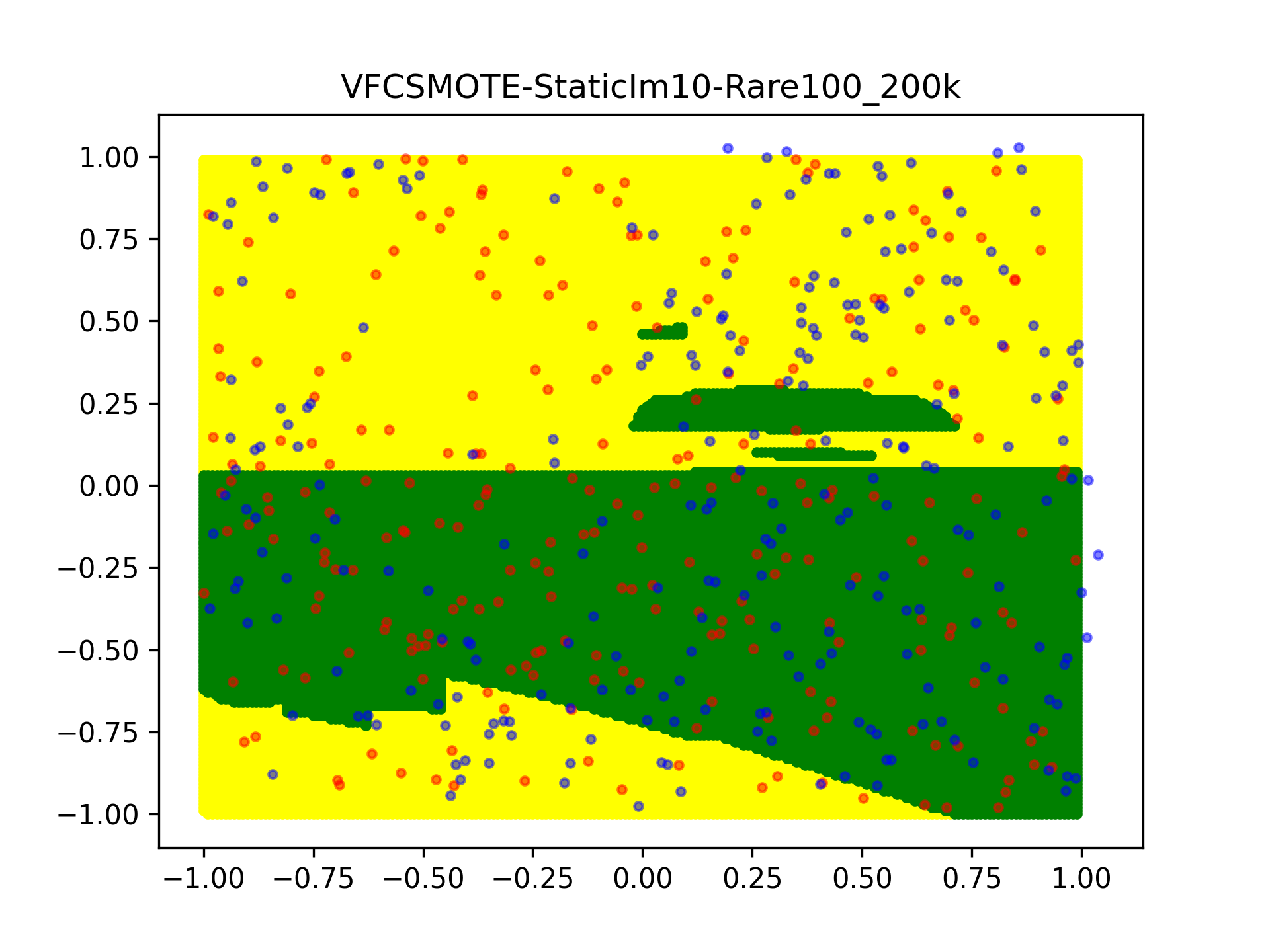} \label{figure:StaticIm10-Rare100-dec_bound-VFCSMOTE-200k}}
\subfigure[SMOTE-OB]{\includegraphics[width=0.29\textwidth]{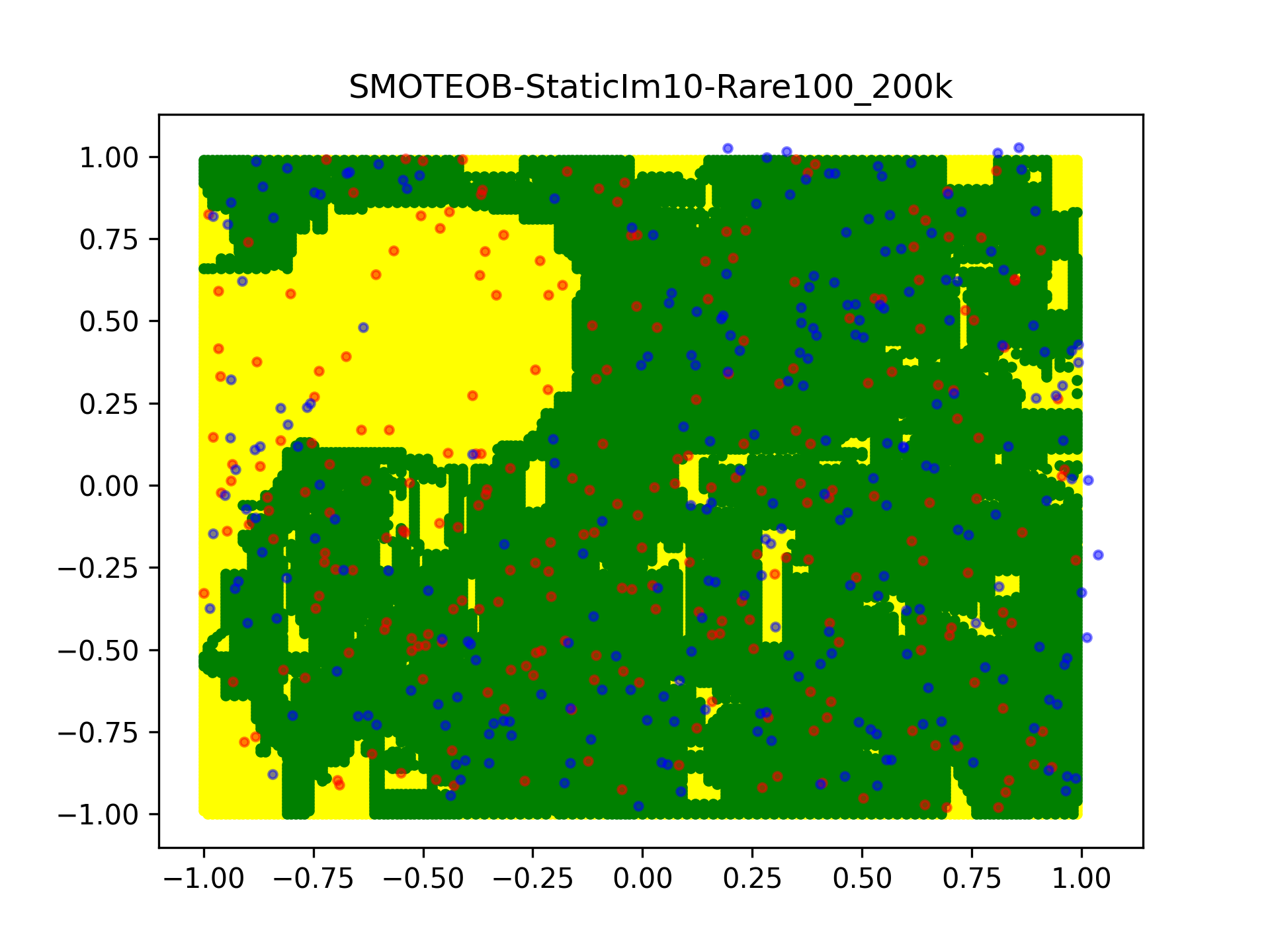} \label{figure:StaticIm10-Rare100-dec_bound-SMOTEOB-200k}}
\subfigure[SMOClust]{\includegraphics[width=0.29\textwidth]{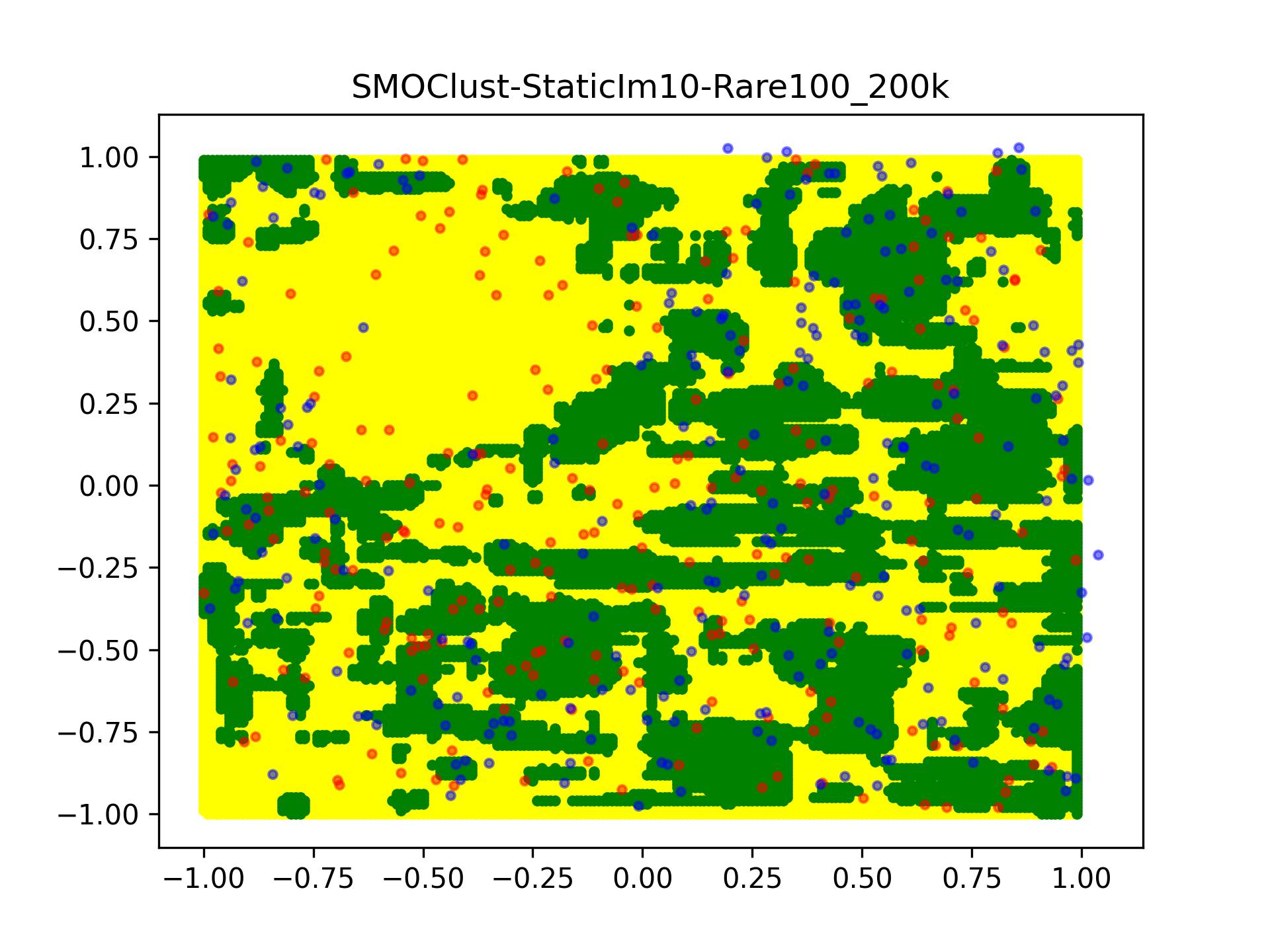} \label{figure:StaticIm10-Rare100-dec_bound-SMOClust-200k}}
\caption{Decision Areas Against Class Balanced Test Set at 200k Time Steps (End of Stream) of Two-Dimensional StaticIm10\_Rare100}
\label{figure:StaticIm10-Rare100-dec_bound-200k}
\end{figure}

Figures \ref{figure:StaticIm10-Rare100-dec_bound-70k}, \ref{figure:StaticIm10-Rare100-dec_bound-100k} and \ref{figure:StaticIm10-Rare100-dec_bound-200k} visualise the learnt decision areas of the approaches at the time steps right before and after concept drift (70k and 100k time steps) and at the end (200k time steps) of the two-dimensional StaticIm10\_Rare100 stream respectively. The yellow and green regions represent their learnt decision areas of class 0 (majority class) and class 1 (minority class) respectively, while the red and blue dots are the class 0 (majority class) and class 1 (minority class) examples in the class balanced test set which corresponds to the time steps.

Figure \ref{figure:StaticIm10-Rare100-GMean} shows that all approaches performed very well during the pre-drift period (0-70k time steps). Figure \ref{figure:StaticIm10-Rare100-dec_bound-70k} reveals that it is because they learnt the decision boundary of the pre-drift concept very well, as the minority class was just a single cluster. While most approaches learnt an oval shape decision boundary, UOB, UOB\textsubscript{d} \reviewII{and SMOTE-OB} learnt a rectangular shape, which could be due to the use of undersampling. \reviewII{VFC-SMOTE learnt a peculiar shape decision boundary which would cause more frequent false-positive drift detections. These may have been due to minority class examples generated by VFC-SMOTE with considerable amount of noise. Meanwhile, SMOTE-OB adopts the same strategy as VFC-SMOTE for generating synthetic minority class examples but simultaneously incorporating undersampling to address class imbalance. This integration of undersampling might explain why SMOTE-OB more successfully circumvented the issue encountered by VFC-SMOTE.}

% As the approaches' learnt decision areas did not have any significant visual difference at the time steps right before concept drift (at 70k time steps), we further investigate why SMOClust was one of the best performing approach in two-dimensional StaticIm10\_Rare100 stream by comparing the approaches' decision areas at the time steps right after the drift (at 100k time steps) and at the end of the stream (at 200k time steps) respectively.

Figure \ref{figure:StaticIm10-Rare100-GMean} shows that the predictive performance of the approaches dropped to below 60\% G-Mean and started to differ since the concept drift began (70k time steps). While most approaches' predictive performance fluctuated with large magnitude, SMOClust's predictive performance was relatively steady, bouncing between 50\%-60\% G-Mean. UOB performed poorly since the drift began at 70k time steps until the drift was close to finishing at 100k time steps, indicating that undersampling struggled in dealing with this drift without the help of a concept drift detector.

Figure \ref{figure:StaticIm10-Rare100-dec_bound-100k} presents the learnt decision boundaries of the approaches right after the drift (100k time steps). It shows that OOB, OnlineOversampling, OnlineUnderOverBagging, OnlineOversampling\textsubscript{d}, SMOGauNoise\reviewII{, SMOTE-OB} and SMOClust learnt very complex decision areas, indicating that they made great efforts to learn all the areas that spawn rare minority class examples belonging to the post-drift concept. However, only approaches with a concept drift detector were able to forget the old area of the minority class at the top left corner. This shows that, although this drift was gradual, concept drift detection was important in helping the system to forget irrelevant past knowledge. In contrast, approaches without a drift detector retained the oval minority class cluster at the top left corner which belongs to the pre-drift concept. Most of them struggled to perform well since the drift started at 70k time steps, as shown in Figure \ref{figure:StaticIm10-Rare100-GMean}. OOB was an exception in terms of predictive performance. However, the fact that it retained the knowledge about the pre-drift minority class areas makes it disadvantageous in dealing with other types of drift, as discussed in Section \ref{section:SMOClust-analysis-artificial data streams-better}.

Comparing the learnt decision areas of SMOClust against other approaches with drift detector (OOB\textsubscript{d}, UOB\textsubscript{d}, OnlineOversampling\textsubscript{d}, OnlineUnderOverBagging\textsubscript{d}\reviewII{, VFC-SMOTE, SMOTE-OB} and SMOGauNoise), it can be observed that the learnt minority class areas of SMOClust were complex and covered the feature space spawning minority class examples the most. While OnlineOversampling\textsubscript{d}'s, SMOGauNoise's \reviewII{and SMOTE-OB's} were also complex \reviewII{(see Figures \ref{figure:StaticIm10-Rare100-dec_bound-OnlineOversampling_d-100k}, \ref{figure:StaticIm10-Rare100-dec_bound-SMOGauNoise-100k} and \ref{figure:StaticIm10-Rare100-dec_bound-SMOTEOB-100k})}, they either did not cover the feature space spawning minority class examples as much as SMOClust's did or exhibited over-generalisation. The fact that OnlineOversampling\textsubscript{d} only reuses the recently seen minority class example for oversampling likely leads to overfitting to such most recent area. SMOGauNoise also has a strategy to explore the decision boundaries of the minority class, but such strategy only explores the area around the recently seen minority class example. This could be disadvantageous when false-positive drift detections were triggered, resetting the base learner. \reviewII{SMOTE-OB's over-generalisation could be explained by the use of undersampling and noisy minority class examples generated.} SMOClust, on the other hand, does not have this disadvantage because the stream clustering methods are not reset upon drift detection. This makes it more robust to false-positive drift detections than other approaches. As the drift was gradual, OOB\textsubscript{d}, UOB\textsubscript{d} and OnlineUnderOverBagging\textsubscript{d} likely also suffered from multiple drift detection, as Figures \ref{figure:StaticIm10-Rare100-dec_bound-OOBd-100k}, \ref{figure:StaticIm10-Rare100-dec_bound-UOBd-100k} and \ref{figure:StaticIm10-Rare100-dec_bound-OnlineUnderOverBagging_d-100k} show that the learnt a simple decision boundary right after the drift.

Figure \ref{figure:StaticIm10-Rare100-dec_bound-200k} presents the learnt decision boundaries of the approaches at the end of the two-dimensional StaticIm10\_Rare100 stream (at 200k time steps). While most approaches continued to further improve their learnt decision boundaries since the drift had finished, Figures \ref{figure:StaticIm10-Rare100-dec_bound-OnlineUnderOverBagging_d-200k} and \ref{figure:StaticIm10-Rare100-dec_bound-SMOClust-200k} show that OnlineUnderOverBagging\textsubscript{d} and SMOGauNoise did not improve as much as other approaches, meaning that they suffered from false-positive drift detections during the post-drift period. \reviewII{Besides, UOB, UOB\textsubscript{d}, and SMOTE-OB exhibited an extensive and predominantly continuous decision area for the minority class, demonstrating the aggressiveness of undersampling. However, in the case of SMOTE-OB, the approach's synthetic minority class generation strategy exacerbates this aggressiveness.}

From this analysis, it has been shown that SMOClust managed to forget the pre-drift concept and adapt to drift leading to 100\% rare minority class examples and more robust to false-positive drift detections than other approaches in the two-dimensional StaticIm10\_Rare100 stream. However, the experiment with the five-dimensional StaticIm10\_Rare100 stream presents different results (Figure \ref{figure:SMOClust-Avg GMean Artificial}). It shows that SMOClust only performed better than OnlineOversampling\textsubscript{d} but worse or similar to most other approaches. One potential reason is the fact that two-dimensional space is more compact than five-dimensional space, the rare minority class examples have a lot less space to randomly spawn, which means they are likely to spawn at the locations that had already been learnt and covered by SMOClust using micro-clusters. Therefore, SMOClust can predict their class label correctly. However, five-dimensional space is sparser than two-dimensional space, meaning that new rare minority class examples are less likely to spawn at previous locations. Therefore, SMOClust struggled to make correct predictions to new rare minority class examples.  Another potential reason is that the stream clustering method may be less effective in data streams with more dimensions. For example, it may create some minority class micro-clusters that overlap with the majority class region because of the sparsity of the feature space. Therefore, the aforementioned advantage of SMOClust in dealing with drift could not be manifested. Anyhow, future work is needed to further confirm whether SMOClust tends to perform better in data streams with fewer dimensions.

%\noindent\fbox{\begin{minipage}{\textwidth}
\begin{shadequote}
\textit{Short Summary: This analysis shows that SMOClust managed to adapt to concept drift leading to 100\% rare minority class examples and was robust to multiple drift detection during gradual drift as well as false-positive drift detections when the data stream has only two dimensions. However, the experiments with the corresponding five-dimensional stream present a different set of results, as the stream clustering methods used by SMOClust might not perform well when the data stream has more dimensions.}
\end{shadequote}
%\end{minipage}}

\subsubsection{\review{Results with Two-Dimensional Artificial Data Streams}} \label{section:SMOClust-analysis-artificial data streams-2d}

\review{To investigate whether SMOClust performs better in lower-dimensional data streams, we performed additional experiments on the same artificial data streams presented in Section \ref{section:SMOClust-data streams}, but with only two input features. We also created a randomised two-dimensional data stream for the purpose of hyper-parameter tuning, following the procedure described in Section \ref{section:SMOClust-analysis-artificial data streams}.}

\review{Figure \ref{figure:SMOClust-Avg GMean Artificial, 2D} presents the difference in average G-Mean (based on thirty runs) between compared approaches and SMOClust on two-dimensional artificial data streams in the form of a heat-map. Green cells indicate results favourable to SMOClust, whereas red cells indicate results favourable to the compared approach. For a comprehensive table of the predictive performance of the approaches, please refer to the supplementary document. Compared to Figure \ref{figure:SMOClust-Avg GMean Artificial}, there are fewer red cells in this figure, indicating that SMOClust generally performed better in the lower-dimensional version of the same set of data streams. In particular, the sections of the heat-map corresponding to StaticIm30 and StaticIm10 data streams, which were mostly reddish in Figure \ref{figure:SMOClust-Avg GMean Artificial}, are mostly greenish in Figure \ref{figure:SMOClust-Avg GMean Artificial, 2D}.}

\begin{figure}[p]
\centering
\includegraphics[width=\textwidth]{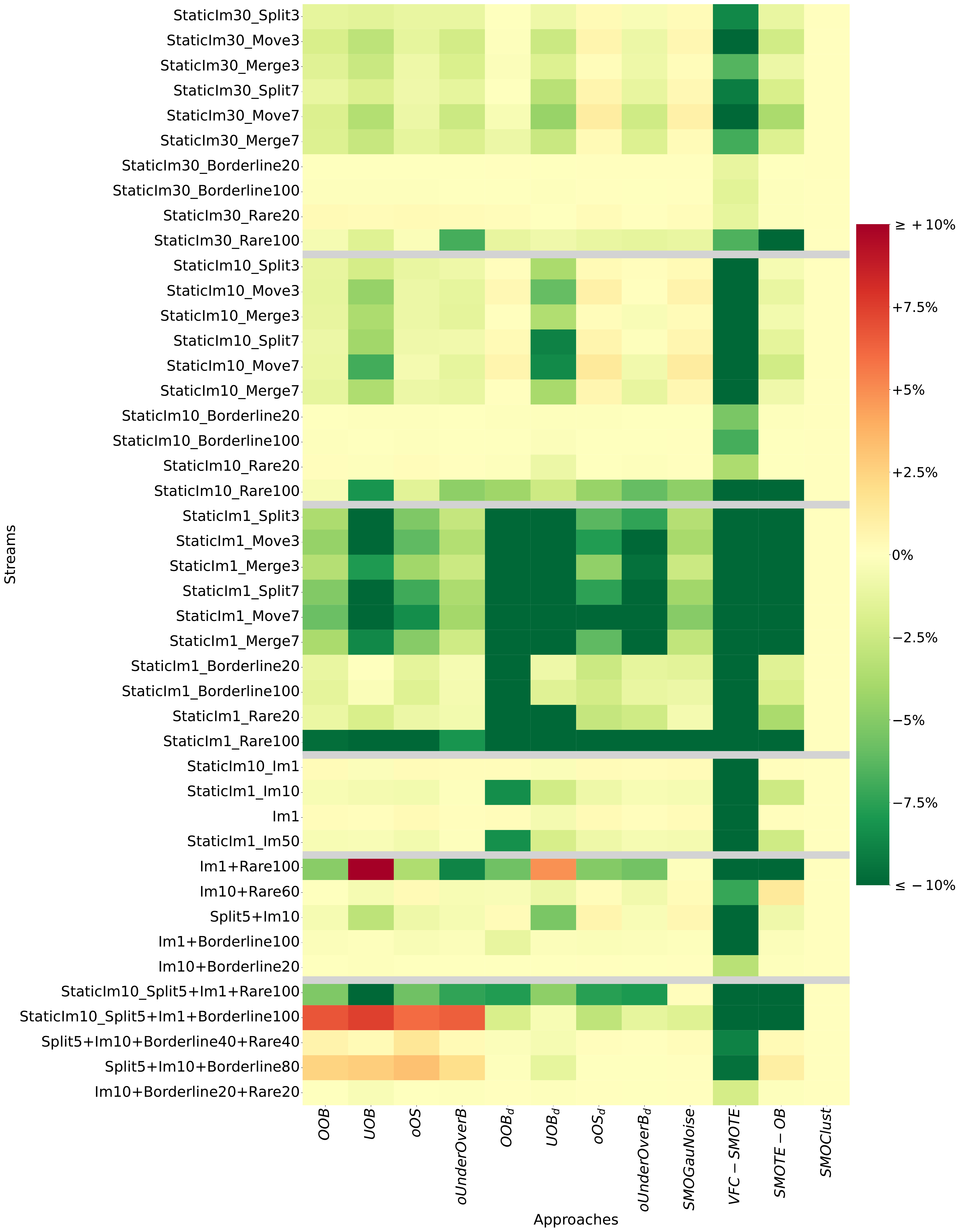}
\caption{Difference in Average G-Mean Against SMOClust on Two-Dimensional Class Imbalanced Artificial Data Streams Based on 30 Runs (Green cells indicate SMOClust performed better; Red cells indicate SMOClust performed worse; Grey horizontal lines separate different groups of data streams, i.e., StaticIm\{30/10/1\}, Imbalance Ratio Drift, Double Factor, and Complex Factor)}
\label{figure:SMOClust-Avg GMean Artificial, 2D}
\end{figure}

\review{Figure \ref{figure:SMOClust-Avg GMean Artificial, 2D} also confirms the trend shown in Figure \ref{figure:SMOClust-Avg GMean Artificial}, showing that SMOClust tends to outperform other approaches in severely class-imbalanced data streams. To further validate this trend in lower-dimensional data streams, we performed further experiments on the same set of single factor drift artificial data streams, but with enforced extremely severe class imbalance ratios (minority class ratio $0.3\%$ to $5\%$, as summarised in Table \ref{table:SMOClust-severe imbalance artificial stream details}). The results are presented in Figure \ref{figure:SMOClust-Avg GMean Artificial Severe, 2D} in the form of a heat-map, using the same colour scheme as Figure \ref{figure:SMOClust-Avg GMean Artificial, 2D}. Similarly, please refer to the supplementary document for a comprehensive table of the predictive performance of the approaches.}

\begin{figure}[p]
\centering
\includegraphics[width=\textwidth]{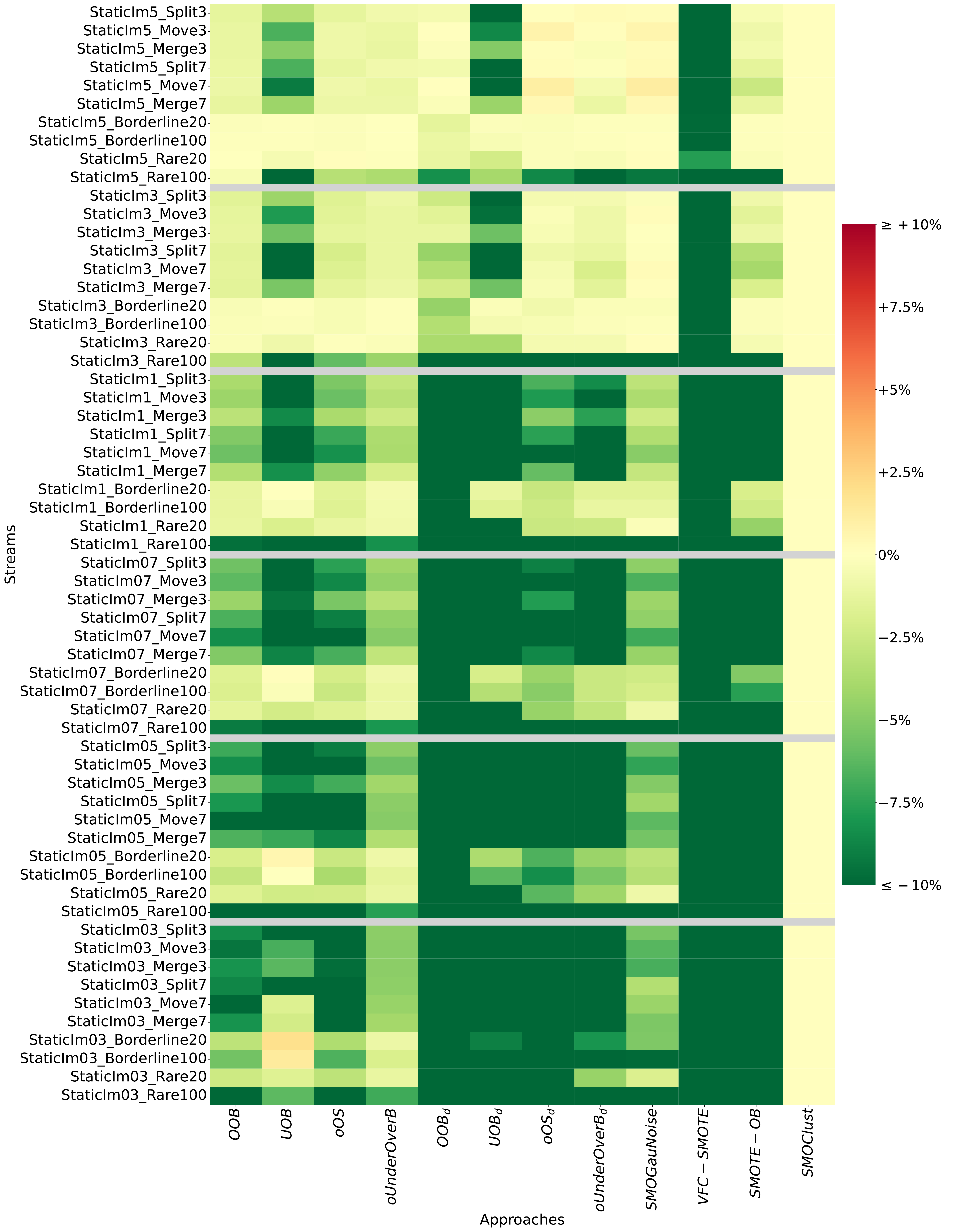}
\caption{Difference in Average G-Mean Against SMOClust on Two-Dimensional Severely Class Imbalanced Artificial Data Streams Based on 30 Runs (Green cells indicate SMOClust performed better; Red cells indicate SMOClust performed worse; Grey horizontal lines separate different groups of data streams, i.e., StaticIm\{5/3/1/07/05/03\}}
\label{figure:SMOClust-Avg GMean Artificial Severe, 2D}
\end{figure}

\review{Figure \ref{figure:SMOClust-Avg GMean Artificial Severe, 2D} presents more solid green cells than Figure \ref{figure:SMOClust-Avg GMean Artificial, 2D}, indicating that SMOClust performed better than other approaches in extremely severe class-imbalanced data streams, even in the lower-dimensional case. Additionally, the fact that Figure \ref{figure:SMOClust-Avg GMean Artificial Severe, 2D} has more green cells than Figure \ref{figure:SMOClust-Avg GMean Artificial Severe} supports the conclusion that SMOClust tends to perform better in lower-dimensional data streams.}

\subsection{Results with Real-world Data Streams} \label{section:SMOClust-analysis-real-world data streams}

This section presents the analysis done to compare the predictive performance of SMOClust against nine existing approaches in real-world data streams. Experiments with real-world data streams allow us to obtain a general idea of SMOClust's predictive performance in practical applications, where the class imbalance ratio, the position and the type of the concept drifts are unknown. Table \ref{table:SMOClust-Friedman Ranks-Preq. GMean-real world} presents the Friedman rankings of approaches' G-Mean on real-world data streams group by factors.

\begin{table}[!ht]
\footnotesize
\centering
\caption[]{Statistical (Friedman) Ranking of prequential G-Mean on Real-World Streams Grouped by Factors}
\label{table:SMOClust-Friedman Ranks-Preq. GMean-real world}
\renewcommand\tabcolsep{1.5pt}
\begin{threeparttable}
\begin{tabular}{c|ccccccccccc|c}
\hline
Groups & OOB & UOB & oOS & \makecell{oUnder- \\ OverB} & OOB\textsubscript{d} & UOB\textsubscript{d} & oOS\textsubscript{d} & \makecell{oUnder- \\ OverB\textsubscript{d}} & \makecell{SMO- \\ Gau- \\ Noise} & \makecell{VFC- \\ SMO- \\ TE} & \makecell{SMO- \\ TE- \\ OB} & \makecell{SMO- \\ Clust} \\
\hline
\hline
% Airlines & \\
\makecell{Luxum- \\ bourg} & \underline{5.43} & \underline{8.83} & \cellcolor{lime} 2.03 & \underline{7.33} & \underline{5.43} & \underline{10.27} & \cellcolor{lime} 2.03 & \underline{7.33} & \cellcolor{lime} 2.97 & 11.87 & \underline{6.23} & \underline{8.23} \\
NOAA & \cellcolor{lime} 1.80 & \underline{5.80} & \underline{5.00} & \cellcolor{lime} 1.93 & \underline{6.10} & 8.27 & 10.73 & \underline{6.97} & 12.00 & 9.67 & \cellcolor{lime} 3.13 & \underline{6.60} \\
Ozone & \cellcolor{lime} 3.27 & \cellcolor{lime} 2.68 & \underline{9.25} & \cellcolor{lime} 3.93 & \underline{6.10} & \cellcolor{lime} 2.82 & \underline{9.02} & \underline{7.53} & \underline{10.70} & 12.00 & \cellcolor{lime} 2.30 & \underline{8.40} \\
\makecell{PAKDD- \\ 2009} & 5.90 & \cellcolor{lime} 2.43 & \cellcolor{lime} 1.20 & 6.43 & 8.00 & 6.97 & \cellcolor{lime} 2.37 & 7.67 & \underline{10.70} & \underline{10.30} & \cellcolor{lime} 4.03 & \underline{12.00} \\
Covtype & \cellcolor{lime} 2.79 & 6.60 & 9.34 & \cellcolor{lime} 3.33 & \underline{5.44} & \underline{5.23} & 11.56 & 6.40 & 9.95 & 9.89 & \cellcolor{lime} 2.81 & \underline{4.66} \\
INSECTS & 5.59 & 9.39 & 7.26 & 6.28 & \cellcolor{lime} 1.23 & 8.58 & 3.75 & \cellcolor{lime} 2.54 & 5.91 & \underline{11.29} & 4.77 & \underline{11.41} \\
Amazon & \cellcolor{lime} 1.93 & \underline{8.43} & 4.23 & \underline{11.93} & 4.57 & \cellcolor{lime} 3.43 & 6.50 & 6.93 & \underline{7.93} & \underline{10.47} & \cellcolor{lime} 1.07 & \underline{10.57} \\
% Elec & \cellcolor{lime} 4.50 & \underline{10.73} & 8.07 & 7.17 & \cellcolor{lime} 1.70 & \cellcolor{lime} 2.43 & \cellcolor{lime} 2.17 & 7.40 & \underline{9.97} & 7.67 & \cellcolor{lime} 4.20 & \underline{12.00} \\
Twitter & \cellcolor{lime} 1.90 & 5.37 & 8.43 & \cellcolor{lime} 3.87 & \cellcolor{lime} 2.57 & \cellcolor{lime} 4.07 & 8.50 & 4.97 & \underline{10.53} & \underline{10.47} & 5.33 & \underline{12.00} \\
\hline
\hline
All & \cellcolor{lime} 3.7593 & 7.0417 & 7.325 & 5.0074 & \cellcolor{lime} 4.2778 & 6.4083 & \underline{7.712} & 5.4963 & \underline{8.5574} & 10.5778 & \cellcolor{lime} 3.6444 & \underline{8.1926} \\
% All & \cellcolor{lime} 3.80 & 7.24 & 7.36 & 5.12 & \cellcolor{lime} 4.14 & 6.20 & 7.42 & 5.60 & \underline{8.63} & 10.42 & \cellcolor{lime} 3.67 & \underline{8.39} \\
\hline
\end{tabular}
\begin{tablenotes}
\begin{footnotesize}
\item[-] The p-values of Friedman tests are all $\leq$2.2E-16.
\item[-] Highlighted ranks denote significant superior performance.
\item[-] Underlined ranks denote the corresponding approach's performance have no statistical significance with SMOClust.
\end{footnotesize}
\end{tablenotes}
\end{threeparttable}
\end{table}

\begin{figure}[!ht]
\centering
\includegraphics[width=\textwidth]{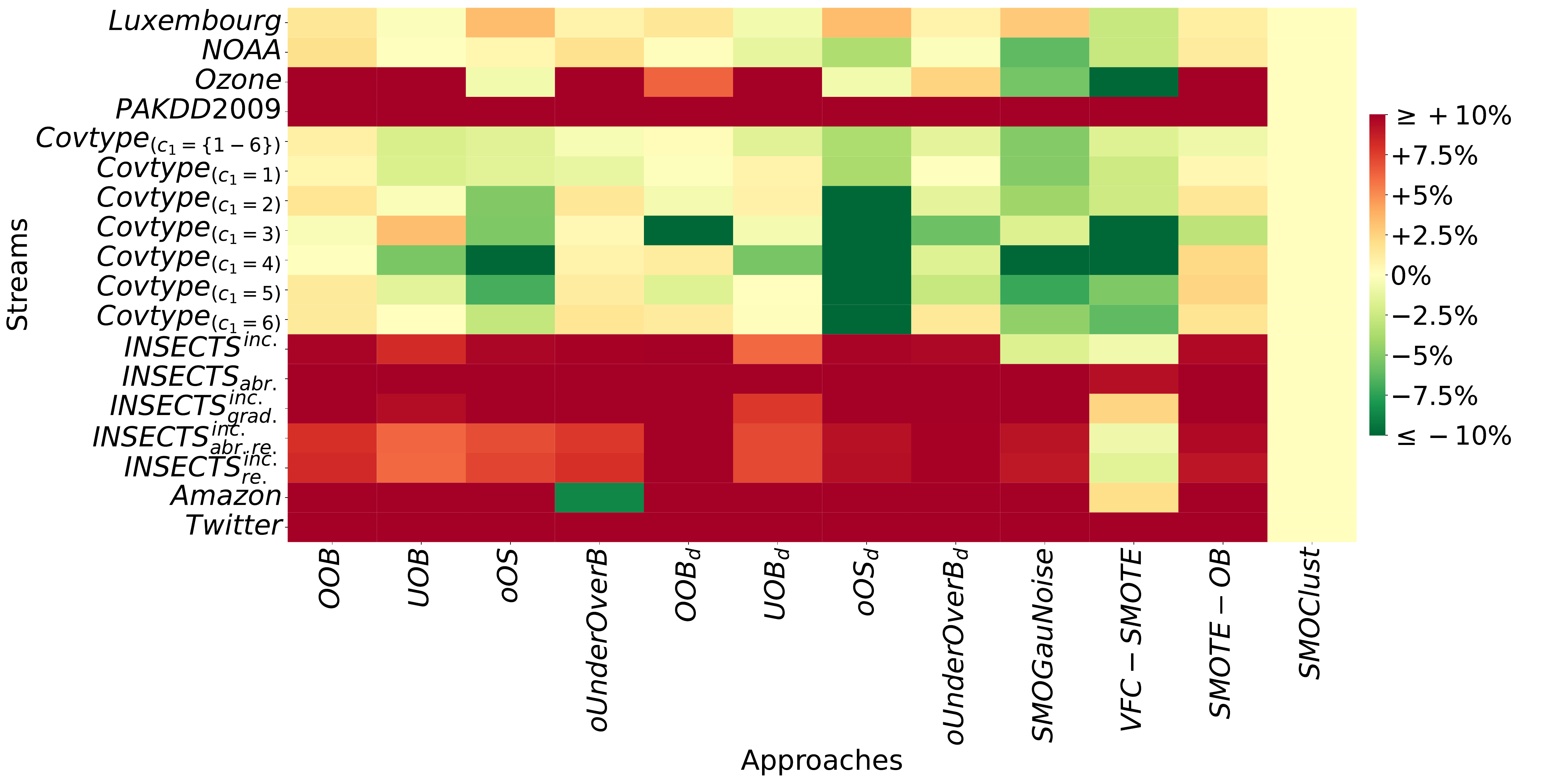}
\caption{Difference in Average G-Mean Against SMOClust on Real-World Data Streams Based on 30 Runs (Green cells indicate SMOClust performed better; Red cells indicate SMOClust performed worse)}
\label{figure:SMOClust-Avg Preq. GMean Real-World}
\end{figure}

Table \ref{table:SMOClust-Friedman Ranks-Preq. GMean-real world} shows that the overall top-ranked approaches on real-world data streams are OOB, OOB\textsubscript{d} and \reviewII{SMOTE-OB} whereas SMOClust usually achieved low rankings. SMOClust only achieved a relatively better ranking in Covtype streams than in other streams. Considering all real-world data streams, SMOClust performed similarly to \reviewII{OnlineOversampling\textsubscript{d} and SMOGauNoise}. Following the analysis method in Section \ref{section:SMOClust-analysis-artificial data streams}, we also compared the thirty runs average prequential G-Mean of the approaches on each real-world data stream
% and reported the A12 effect size comparison results
in Figure \ref{figure:SMOClust-Avg Preq. GMean Real-World} to further evaluate the predictive performance of SMOClust in real-world data streams.

% \caption[]{30 Runs Average Prequential G-Mean on Real-World Data Streams (A12 SMOClust vs Others)}

Figure \ref{figure:SMOClust-Avg Preq. GMean Real-World} shows that SMOClust usually performed similar or better than other approaches in NOAA and Covtype streams while it performed worse than other approaches in Ozone, PAKDD2009\reviewII{, INSECTS, Amazon, and Twitter} streams. Recalling the discussion in Section \ref{section:SMOClust-data streams} on estimated characteristics of real-world streams, NOAA and Covtype streams mainly consist of safe and borderline minority class examples with different movements of minority class clusters and the minority class ratios throughout Covtype streams are usually very low (except Covtype\textsubscript{(c\textsubscript{1}=\{1-6\})} and Covtype\textsubscript{(c\textsubscript{1}=1)}). As discussed in Section \ref{section:SMOClust-analysis-artificial data streams}, these are the characteristics of a data stream that SMOClust is likely to perform similar or better than other approaches, especially when the class imbalance ratio is severe, such as Covtype\textsubscript{(c\textsubscript{1}=3)} stream. Thus, we can see from Figure \ref{figure:SMOClust-Avg Preq. GMean Real-World} that the rows of NOAA and Covtype streams mainly consist of saturated green cells and pale red cells.

On the other hand, Table \ref{table:SMOClust-real world stream characteristics} shows that Ozone, PAKDD2009\reviewII{, INSECTS, Amazon, and Twitter} streams consist of large proportions of rare and outlier minority class examples. Based on the discussion in Section \ref{section:SMOClust-analysis-artificial data streams-worse}, SMOClust could not handle rare and outlier minority class examples very well, except when the dimensionality of the data stream was low or compact. Thus, it is not surprising to see a lot of red cells on these data streams.

%\noindent\fbox{\begin{minipage}{\textwidth}
\begin{shadequote}
\textit{To summarise the result of experiments with real-world data streams, the advantage of the proposed synthetic minority class oversampling strategy in SMOClust is manifested in severely class imbalanced data streams with high proportions of safe and borderline minority class examples with concept drifts of different movements of minority class sub-clusters. On the downside, SMOClust could not handle rare and outlier minority class examples very well. These findings are consistent with the result of experiments with artificial data streams.}
\end{shadequote}%\end{minipage}}

\section{Conclusion} \label{section:conclusion}

The main contribution of this work is the proposed stream clustering based synthetic minority oversampling approach, called SMOClust (RQ1). This method helps the learning system to strategically explore different decision areas of the minority class and to be robust to false-positive drift detections (RQ1). To evaluate the predictive performance and the characteristics of SMOClust, experiments with artificial data streams concerning different types of concept drift difficulties were performed. The results show that SMOClust performed particularly well in severely class imbalanced data streams with high proportions of safe and borderline minority class examples (RQ2). It also handles concept drifts of different movements of minority class clusters better than other existing approaches (RQ2). However, when the data stream presents high proportions of rare and outlier minority class examples, SMOClust becomes disadvantageous (RQ3).

To further understand the reason behind the experiment results on artificial data streams, additional experiments with representative two-dimensional artificial data streams were performed. However, it shows that SMOClust managed to handle rare minority class examples better than other approaches in these two-dimensional cases. This indicates that the reason why SMOClust could not handle rare cases very well on the corresponding five-dimensional stream was likely because of the stream clustering methods did not perform well in higher-dimensional space. In other words, SMOClust may be more advantageous when the dimensionality of the data stream is not high.

Lastly, we validated the performance of SMOClust on different real-world data streams. To facilitate the analysis of the experiment results of this part of the study, we estimated the characteristics of the real-world data streams, following the procedure adopted by \cite{DataDifficulty}. Based on the estimated characteristics and the experiment results, we concluded that the SMOClust behaved similarly to the experiments with artificial data streams (RQ3).

\review{As for future work, an investigation of new strategies to better handle large proportions of rare and outlier minority class examples is one potential direction. For example, strategies to generate synthetic minority examples for oversampling in a more diverse manner without introducing a significant amount of noise or creating artificial concept drifts could be proposed. Additionally, extending the idea of SMOClust to deal with multi-class classification tasks could also be an area to investigate in the future. Furthermore, the proposed synthetic minority oversampling strategy in this work could be adapted for use with other complex data stream learning systems easily as it is a drift adaptable data-level method to address class imbalance in data stream learning. For example, it could be incorporated into an explicit drift handling approach which exploits relevant past knowledge to handle concept drifts \cite{DP,CDCMS} or an ensemble approach which evolves themselves to adapt to concept drifts \cite{DWM,OAUE}. \reviewII{Apart from these, a comprehensive study to compare SMOClust against more approaches for learning drifting class imbalanced data streams (e.g., CSARF \cite{CSARF}, ROSE \cite{ROSE} etc.) and with more data sets could also be a potential future work.}} %However, the interaction between these systems and the proposed strategy is very likely complicated. Hence, further investigation is needed in the future.}

\backmatter

% \bmhead{Supplementary information}

% If your article has accompanying supplementary file/s please state so here. 

% Authors reporting data from electrophoretic gels and blots should supply the full unprocessed scans for key as part of their Supplementary information. This may be requested by the editorial team/s if it is missing.

% Please refer to Journal-level guidance for any specific requirements.

% \bmhead{Acknowledgments}

% Acknowledgments are not compulsory. Where included they should be brief. Grant or contribution numbers may be acknowledged.

% Please refer to Journal-level guidance for any specific requirements.

\section*{Declarations}

% Some journals require declarations to be submitted in a standardised format. Please check the Instructions for Authors of the journal to which you are submitting to see if you need to complete this section. If yes, your manuscript must contain the following sections under the heading `Declarations':

\begin{itemize}
\item \textbf{Funding:} This work was partly supported by EPSRC Grant No.~EP/R006660/2. This work was conducted while Chun Wai Chiu was a PhD student at the School of Computer Science, University of Birmingham, UK, under a School Scholarship provided in support of this grant.
\item \textbf{Conflict of interest:} The authors declare that they have no conflict of interest.
\item \textbf{Ethics approval:} Not Applicable.
\item \textbf{Consent to participate:} Not Applicable.
\item \textbf{Consent for publication:} Not Applicable.
\item \textbf{Availability of data and materials:} The data sets used in the experiments are available at: https://github.com/michaelchiucw/SMOClust
\item \textbf{Code availability:} The implementation of the proposed approach is available at: https://github.com/michaelchiucw/SMOClust
\item \textbf{Authors' contributions:} All authors contributed to the study conception, design, experiment results analysis, drafting and editing the manuscript. Chun Wai Chiu implemented the proposed approach and performed the experiments. 
\end{itemize}

\bibliography{sn-bibliography}% common bib file


\end{document}